\documentclass{article}

\PassOptionsToPackage{numbers, compress}{natbib}

\usepackage[preprint]{arxiv}

\usepackage[algo2e,ruled,vlined,linesnumbered]{algorithm2e}
\SetKwInput{KwInput}{Input}                
\SetKwInput{KwOutput}{Output}
\usepackage{algorithm}
\usepackage{algpseudocode}
\usepackage[utf8]{inputenc} 
\usepackage[T1]{fontenc}    
\usepackage{hyperref}       
\usepackage{url}            
\usepackage{booktabs}       
\usepackage{amsfonts}       
\usepackage{nicefrac}       
\usepackage{microtype}      
\usepackage{xcolor}         
\usepackage{float}  
\usepackage{amsmath}
\usepackage[font=small]{caption}
\usepackage{enumitem}
\usepackage{graphicx}
\usepackage[titletoc]{appendix}

\title{The Slingshot Mechanism: An Empirical Study of Adaptive Optimizers and the \emph{Grokking Phenomenon}}

\author{
  Vimal Thilak\\
  \texttt{vthilak@apple.com} \\
  \And
  Etai Littwin\\
  \texttt{elittwin@apple.com} \\
  \And
  Shuangfei Zhai\\
  \texttt{szhai@apple.com} \\
  \And
  Omid Saremi\\
  \texttt{osaremi@apple.com} \\
  \And
  Roni Paiss \\
  \texttt{rpaiss@apple.com} \\
  \And
  Joshua Susskind\\
  \texttt{jsusskind@apple.com} \\
}

\begin{document}
\maketitle

\begin{abstract}
    The \emph{grokking phenomenon} as reported by Power et al.~\cite{power2021grokking} refers to a regime where a long period of overfitting is followed by a seemingly sudden transition to perfect generalization. In this paper, we attempt to reveal the underpinnings of Grokking via a series of empirical studies. Specifically, we uncover an optimization anomaly plaguing adaptive optimizers at extremely late stages of training, referred to as the \emph{Slingshot Mechanism}. A prominent artifact of the Slingshot Mechanism can be measured by the cyclic phase transitions between stable and unstable training regimes, and can be easily monitored by the cyclic behavior of the norm of the last layers weights. We empirically observe that without explicit regularization, Grokking as reported in \cite{power2021grokking} almost exclusively happens at the onset of \emph{Slingshots}, and is absent without it. 
    While common and easily reproduced in more general settings, the Slingshot Mechanism does not follow from any known optimization theories that we are aware of, and can be easily overlooked without an in depth examination. Our work points to a surprising and useful inductive bias of adaptive gradient optimizers at late stages of training, calling for a revised theoretical analysis of their origin.
\end{abstract}

\begin{figure*}[h]
\centering
  \includegraphics[width=0.9\linewidth]{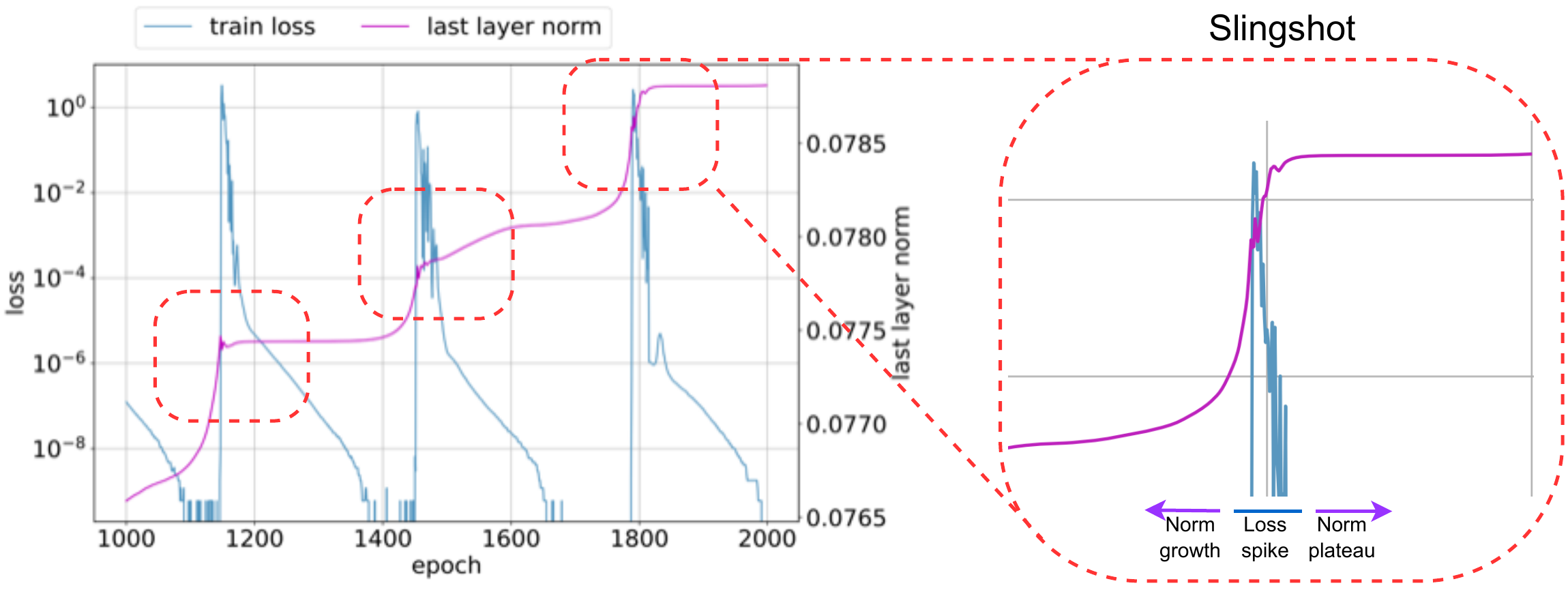} 
  \caption{Slingshot Effects are observed with a fully-connected ReLU network (FCN). The FCN is trained with 200 randomly chosen CIFAR-10 samples with Adam. Multiple Slingshot Effects occur in a cyclic fashion as indicated by the dotted red boxes. Each Slingshot Effect is characterized by a period of rapid growth of the last layer weights, an ensuing training loss spike, and a norm plateau.}
   \label{fig:slingshot_intro}
\end{figure*}

\section{Introduction}

Recently, the \emph{grokking phenomenon} was proposed by ~\cite{power2021grokking}, in the context of studying the optimization and generalization aspects in small, algorithmically generated datasets. Specifically, \emph{grokking} refers to a sudden transition from chance level validation accuracy to perfect generalization, long past the point of perfect training accuracy, i.e., \emph{Terminal Phase of Training} (TPT). This curious behavior contradicts the common belief of early stopping in the overfitting regimes, and calls for further understandings of the generalization behavior of deep neural networks. 

In the literature, it has been suggested that in some scenarios, marginal improvements in validation accuracy appears in TPT, which seem to directly support \emph{grokking}. For example, it has been shown in~\cite{soudry2018implicit} that gradient descent on logistic regression problems converges to the maximum margin solution, a result that has been since extended to cover a wider setting~\cite{lyu2019gradient, wang2021implicit}. A key finding in~\cite{soudry2018implicit} shows that when training on linearly separable data with gradient descent using logistic regression, the classifier's margin slowly improves at a rate of $\mathcal{O}(\frac{1}{\log t})$, while the weight norm of the predictor layer grows at a rate of $\mathcal{O}(t)$, where $t$ is the number of training steps. While specified for gradient descent, Wang et al.~\cite{wang2021implicit} showed that similar results also hold for adaptive optimizers (such
as Adam and RMSProp). Taking these results into consideration, one could reasonably hypothesise that deep nonlinear networks could benefit from longer training time, even after achieving zero errors on the training set. \\ 

In this paper, we provide in depth empirical analyses to the mechanism behind \emph{grokking}. We find that the phenomenology of \emph{grokking} differs from those predicted by~\cite{soudry2018implicit} in several key aspects. To be concrete, we find that \emph{grokking} occurs during the onset of another intriguing phenomenon directly related to adaptive gradient methods (see Algorithm \ref{alg:adaptive} for a generic description of adaptive gradient methods).
In particular, leveraging the basic setup in~\cite{power2021grokking}, we make the following observations:

  1. During the TPT, training exhibits a cyclic behaviour between stable and unstable regimes. A prominent artifact of this behaviour can be seen in the norm of a model's last layer weights, which exhibits a cyclical behavior with distinct, sharp phase transitions that alternate between rapid growth and plateaus over the course of training.
  
  2. The norm grows rapidly sometime after the model has perfect classification accuracy on training data. A sharp phase transition then occurs when the model missclassifies training samples.  This phase change is accompanied by a sudden spike in training loss, and a plateau in the norm growth of the final classification layer. 
  
  3. The features (pre-classification layer) show rapid evolution as the weight norm transitions from rapid growth to a growth plateau, and change relatively little at the norm growth phase. 

    4. Phase transitions between norm growth and norm plateau phases are typically accompanied by a sudden bump in generalization as measured by classification accuracy on a validation set, as observed in a dramatic fashion in ~\cite{power2021grokking}.
    
    5. It is empirically observed that grokking as reported in
    \cite{power2021grokking} almost exclusively happens at the onset of \emph{Slingshots}, and is absent without it. 

We denote the observations above as the \emph{Slingshot Effect}, which is defined to be the full cycle starting from the norm growth phase, and ending in the norm plateau phase. And empirically, a single training run typically exhibits multiple Slingshot Effects. Moreover, while \emph{grokking} as described in~\cite{power2021grokking} might be data dependent, we find that the Slingshot Mechanism is pervasive, and can be easily reproduced in multiple scenarios, encompassing a variety of models (Transformers and MLPs) and datasets (both vision, algorithmic and synthetic datasets). Since we only observe Slingshot Effects when training classification models with adaptive optimizers, our work can be seen as empirically characterizing an implicit bias of such optimizers. Finally, while our observations and conclusions hold for most variants of adaptive gradient methods, we focus on Adam in the main paper, and relegate all experiments with additional optimizers to the appendix. 

\begin{algorithm}[H]
\footnotesize
\KwInput{ $X_1 \in \mathcal{F}$, step size $\mu$, sequence of functions $\{\phi_t,\psi_t\}_{t=1}^T$, $\epsilon \in \mathbb{R}^+$}
\KwOutput{Fitted $\alpha$.}
\For{$t=1...,T$}
{
$g_t = \nabla f_t(x_t)$.

$m_t = \phi_t(g_1,...,g_t) ~ \text{and} ~ V_t = \psi_t(g_1,...,g_t)$.

$x_{t+1} = x_t - \frac{\mu m_t}{\sqrt{V_t^2} + \epsilon}$
}

\caption{Generic Adaptive Gradient Method}
\label{alg:adaptive}
\end{algorithm}

\subsection{Implications of Our Findings}

The findings in this paper have both theoretical and practical implications that go beyond characterizing Grokking.
A prominent feature of the Slingshot Mechanism is the repeating phase shifts between stable and unstable training regimes, where the unstable phase is characterized by extremely large gradients, and spiking training loss. Furthermore, we find that learning at late stages of training have a cyclic property, where non trivial feature adaptation only takes place at the onset of a phase shift. From a theoretical perspective, this is contradictory to common assumptions made in the literature of convergence of adaptive optimizers, which typically require $L$ smooth cost functions, and bounded stochastic gradients, either in the $L_2$ or $L_{\infty}$ norm,  decreasing step sizes and stable convergence \cite{Zhang2020WhyGC,AllenZhu2019ACT,Barakat2021ConvergenceAD}. 
From the apparent generalization benefits of Slingshot Effects, we cast doubt on the ability of current working theories to explain the Slingshot Mechanism.\\
Practically, our work presents additional evidence for the growing body of work indicating the importance of the TPT stage of training for optimal performance \cite{Hoffer2017TrainLG,power2021grokking,Papyan2020PrevalenceON}. 

In an era where the sheer size of models are quickly becoming out of reach for most practitioners, our work suggest focusing on improved methods to prevent excessive norm growth either implicitly through Slingshot Effects or through other forms of explicit regularization or normalization.

\section{Related Work} 

The Slingshot Mechanism we uncover here is reminiscent of the \emph{catapult mechanism} described in al.~\cite{lewkowycz2020large}. Lewkowycz et al.~\cite{lewkowycz2020large} show that loss of a model trained via gradient descent with an appropriately large learning rate shows a non-monotonic behavior \textemdash the loss initially increases and starts decreasing once the model "catapults" to a region of lower curvature \textemdash early in training. However, the catapult phenomenon differs from Slingshot Effects in several key aspects.  The \emph{catapult mechanism} is observed with vanilla or stochastic gradient descent unlike the Slingshot Mechanism that is seen with adaptive optimizers including Adam~\cite{kingma2014adam} and RMSProp~\cite{tieleman2012lecture}. Furthermore, the \emph{catapult phenomenon} relates to a large initial learning rate, and does not exhibit a repeating cyclic behavior.  More intriguingly, Slingshot Effects only emerge late in training, typically long after the model reaches perfect accuracy on the training data.

Cohen et al.~\cite{cohen2021gradient} describe a "progressive sharpening" phenomenon in which the maximum eigenvalue of the loss Hessian increases and reaches a value that is at equal to or slightly larger than $2 / \eta $ where $\eta$ is the learning rate. This "progressive sharpening" phenomenon leads to model to enter a regime Cohen et al.~\cite{cohen2021gradient} call \emph{Edge of Stability} where-in the model shows non-monotonic training loss behavior over short time spans. \emph{Edge of Stability} is similar to the Slingshot Mechanism in that it is shown to occur later on in training. However, \emph{Edge of Stability} is shown for full-batch gradient descent while we observe Slingshot Mechanism with adaptive optimizers, primarily Adam~\cite{kingma2014adam} or AdamW~\cite{loshchilov2017decoupled}.

As noted above, the Slingshot Mechanism emerges late in training, typically longer after the model reaches perfect accuracy and has low loss on training data. The benefits of continuing to training a model in this regime has been theoretically studied in several works including~\cite{soudry2018implicit, lyu2019gradient}. Soudry et al.~\cite{soudry2018implicit} show that training a linear model on separable data with gradient using the logistic loss function leads to a max-margin solution. Furthermore Soudry et al.~\cite{soudry2018implicit} prove that the loss decreases at a rate of $O(\frac{1}{t})$ while the margin increases much slower $O(\frac{1}{\log t})$, where $t$ is the number of training steps. Soudry et al.~\cite{soudry2018implicit} also note that the weight norm of the predictor layer increases at a logarithmic rate, i.e., $O({\log(t)})$. Lyu and Li~\cite{lyu2019gradient} generalize the above results to homogeneous neural networks trained with exponential-type loss function and show that loss decreases at a rate of $O(1 / t(\log(t))^{2 - 2/L})$. This is, where $L$ is defined as the order of the homogenous neural network. Although these results indeed prove the benefits of training models, their analyses are limited to gradient descent. Moreover, the analyses developed by Soudry et al~\cite{soudry2018implicit} do not predict any phenomenon that resembles the Slingshot Mechanism. Wang et al.~\cite{wang2021implicit} show that homogenous neural networks trained RMSProp~\cite{tieleman2012lecture} or Adam without momentum~\cite{wang2021implicit} do converge in direction to the max-margin solution. However, none of these papers can explain the Slingshot Mechanism and specifically the cyclical behavior of the norm of the last layer weights.

\section{The Slingshot Mechanism}
\label{sec:slingshot_mechanism}
\subsection{Experimental Setup}
We use the training setup studied by Power et al.~\cite{power2021grokking} in the main paper as a working example to illustrate the Slingshot Mechanism. 
In this setup, we train decoder-only Transformers~\cite{vaswani2017attention} on a modular division dataset~\cite{power2021grokking} of the form $a \div b = c$, where $a$, $b$ and $c$ are discrete symbols and $\div$ refers to division modulo $p$ for some prime number $p$, split into training and validation sets. The task consists of calculating $c$ given $a$ and $b$. The algorithmic operations and details of the datasets considered in our experiments are described in Appendix~\ref{appendix:xformers_setup}. The Transformer consists of 2 layers, of width 128 and 4 attention heads with approximately 450K trainable parameters and is optimized by Adam ~\cite{kingma2014adam, loshchilov2017decoupled}. For these experiments we set learning rate to 0.001, weight decay to 0, $\beta_{1} = 0.9$, $\beta_{2} = 0.98$, $\epsilon=10^{-08}$, linear learning rate warmup for the first 10 steps and minibatch size to 512 which are in line with the hyperparameters considered in~\cite{power2021grokking}.
 
\begin{figure*}[t]
\centering
  \begin{tabular}{ccc}
      \includegraphics[width=0.32\linewidth]{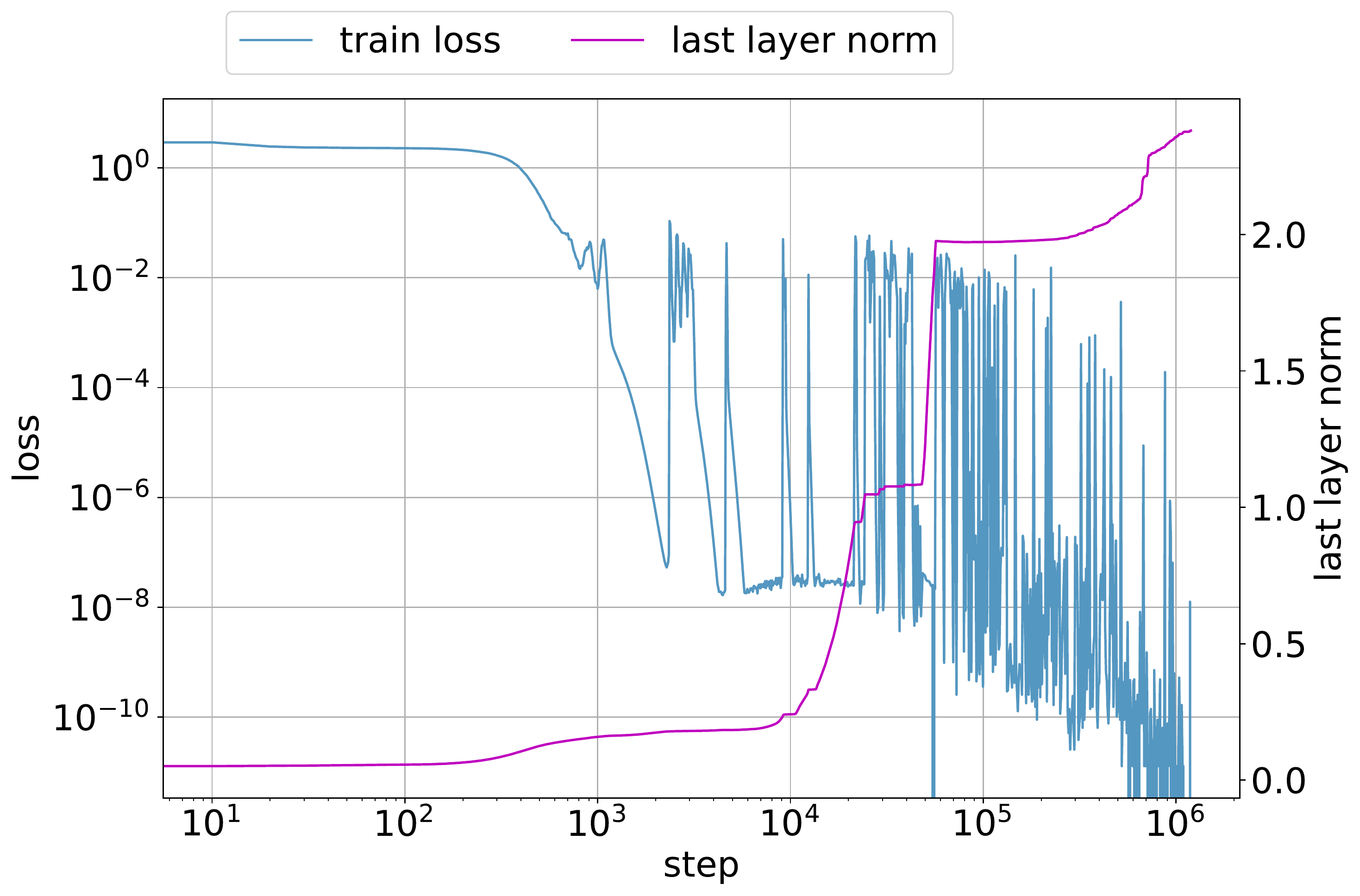} & 
      \includegraphics[width=0.32\linewidth]{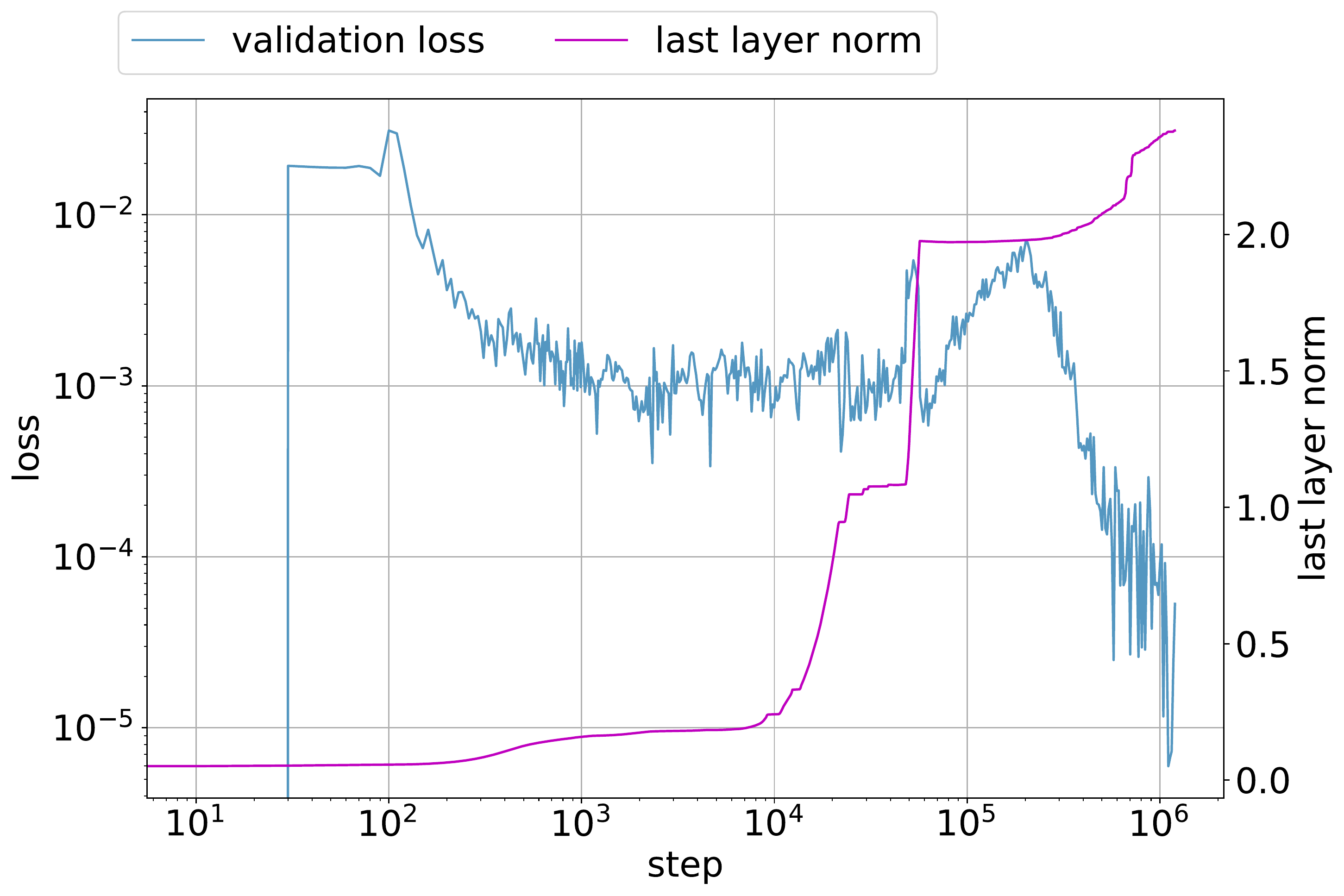} &
      
      \includegraphics[width=0.32\linewidth]{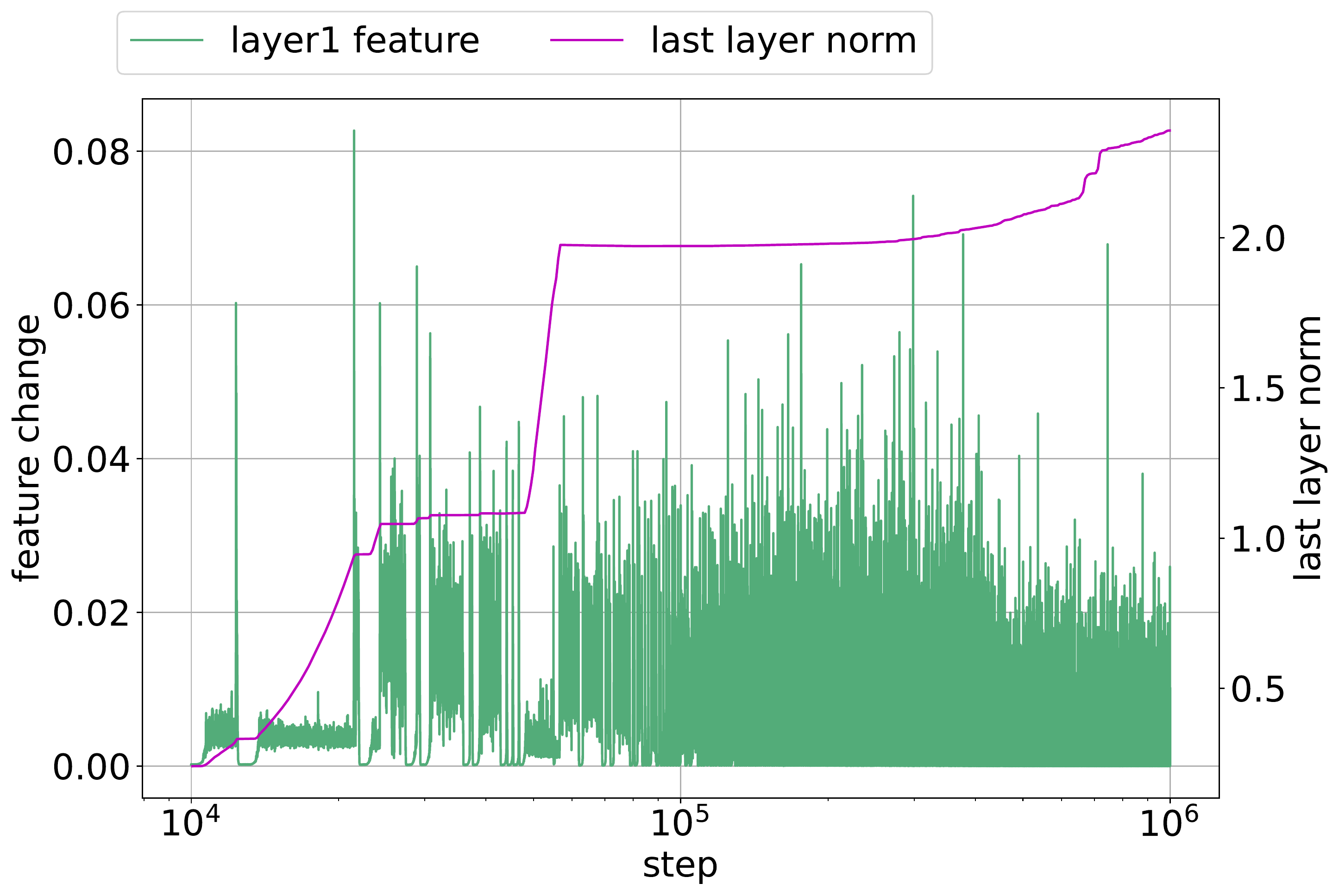} \\
      (a)  & (c) & (e) \\
     \includegraphics[width=0.32\linewidth]{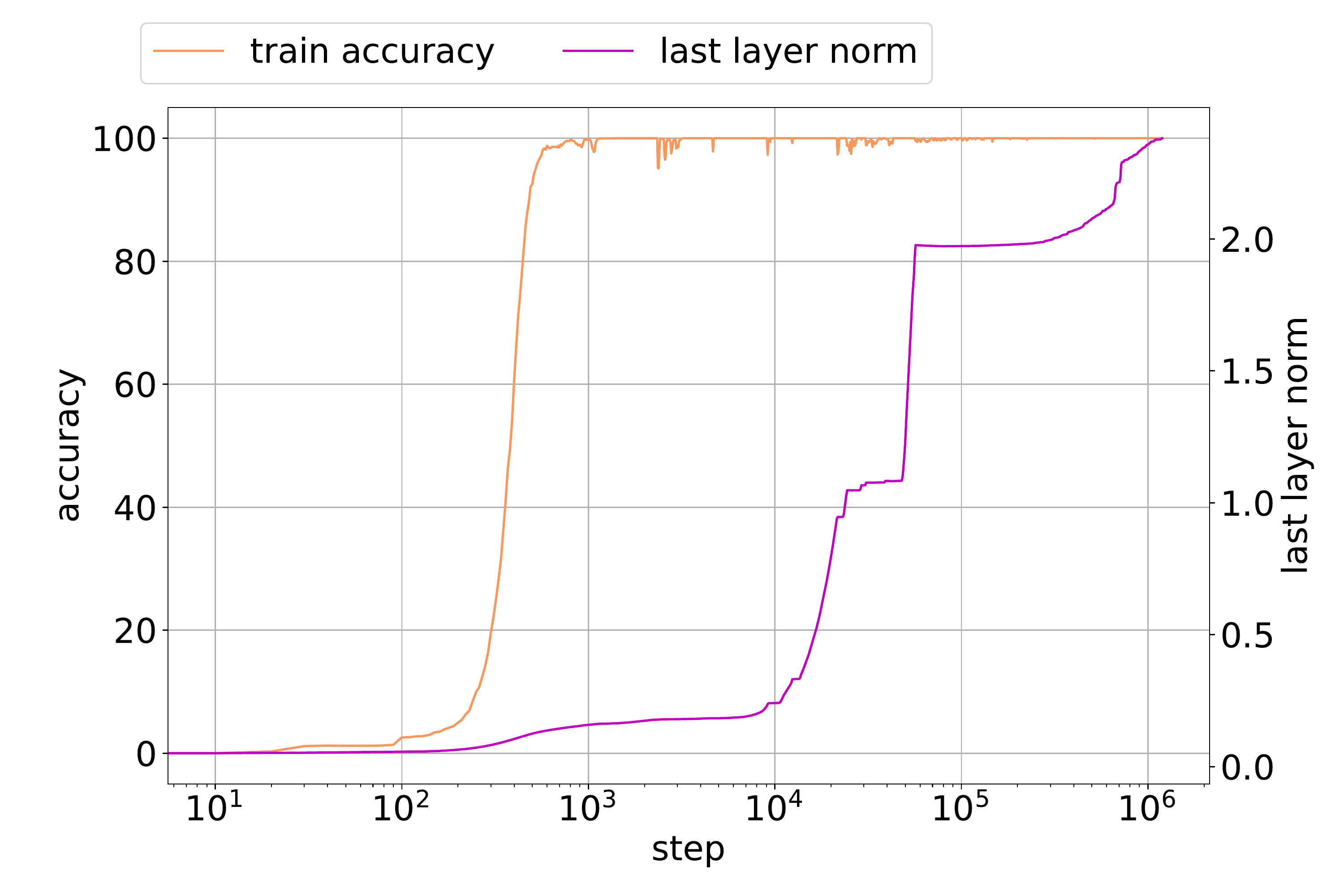} &
     \includegraphics[width=0.32\linewidth]{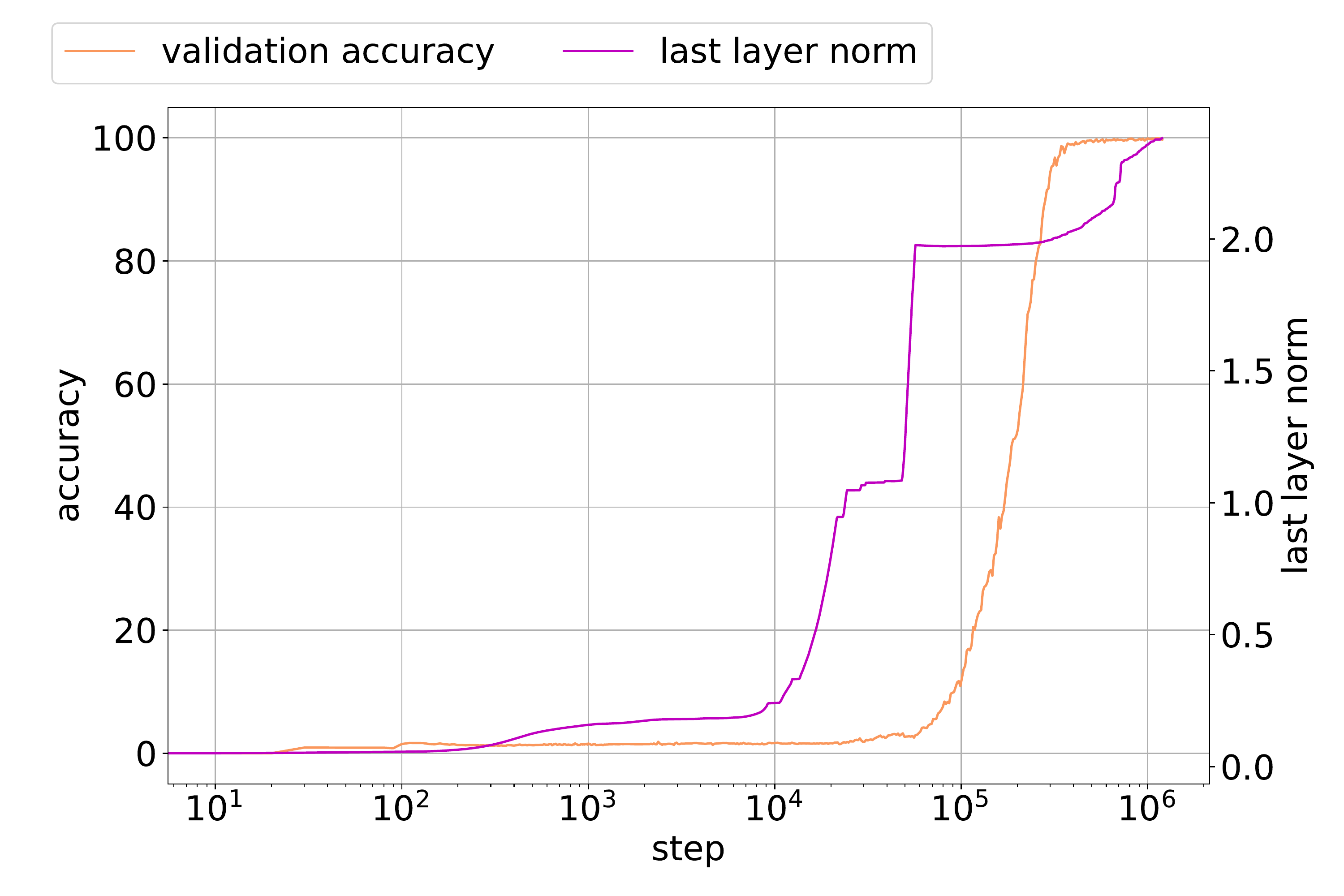} & 
      \includegraphics[width=0.32\linewidth]{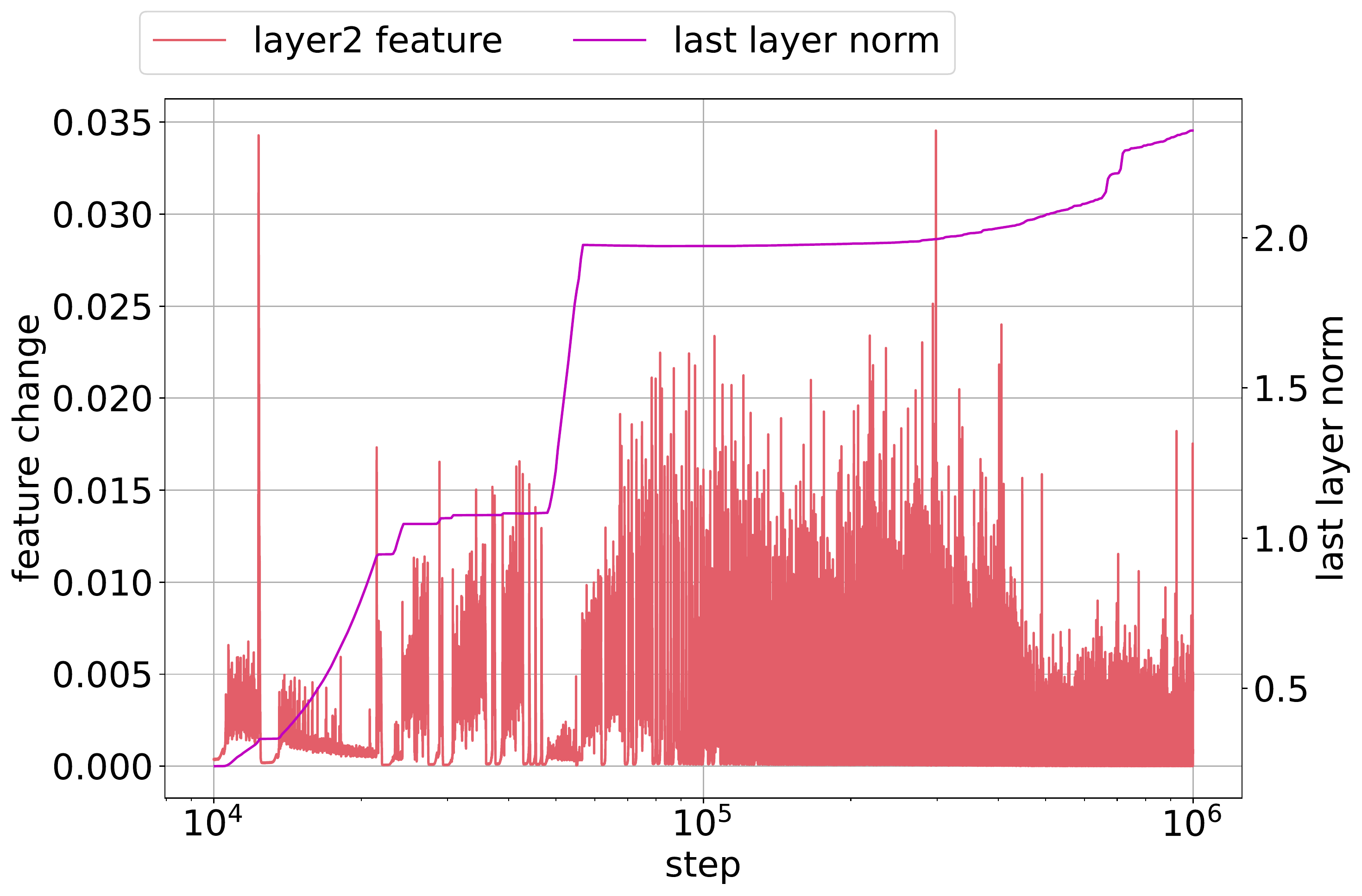} \\
      (b)  & (d) & (f) 

  \end{tabular}
 \caption{Division dataset: Last layer weight norm growth versus a) loss on training data b) accuracy on training data (c) loss on validation data d) accuracy on validation data e) normalized relative change in features of first Transformer layer (f) normalized relative change in features of second Transformer layer. Note that the feature change plots are shown starting at 10K step to emphasize the feature change behavior during norm growth and plateau phases, revealing that the features stop changing during the norm growth phase and resume changing during the plateaus.} 
 \label{fig:grok_div_lln_50p}
\end{figure*}

Figure~\ref{fig:grok_div_lln_50p} shows the metrics of interest that we record on training and validation samples for modular division dataset. Specifically, we measure 1) \textit{train loss}; 2) \textit{train accuracy}; 3) \textit{validation loss}; 4) \textit{validation accuracy}; 5) \textit{last layer norm}: denoting the norm of the classification layer's weights and 6) \textit{feature  change}: the relative change of features of the l-th layer ($h^l$) after the t-th gradient update step $\frac{\|h^l_{t+1} - h^l_{t}\| }{\|h^l_{t}\|}$.  We observe from Figure~\ref{fig:grok_div_lln_50p}b that the model is able to reach high training accuracy around step 300 while validation accuracy starts improving after $10^{5}$ steps as seen in Figure~\ref{fig:grok_div_lln_50p}d. Power et al.~\cite{power2021grokking} originally showed this phenomenon and refer to it as grokking. We observe that while the validation accuracy does not exhibit any change until much later in training, the validation loss shown in Figure~\ref{fig:grok_div_lln_50p}c exhibits a double descent behavior with an initial decrease, then a growth before rapidly decreasing to zero. \\
Seemingly, some of these observations can be explained by the arguments in ~\cite{soudry2018implicit} and their extensions to adaptive optimizers~\cite{wang2021implicit}. Namely, at the point of reaching perfect classification of the training set, the cross-entropy (CE) loss by design pressures the classification layer to grow in norm at relatively fast rate. Simultaneously, the implicit bias of the optimizer coupled with the CE loss, pushes the direction of the classification layer to coincide with that of the maximum margin classifier, albeit at a much slower rate. 

These insights motivate us to measure the classifier's last layer norm during training. We observe in Figure~\ref{fig:grok_div_lln_50p}a that once classification reaches perfect accuracy on the training set, the classification layers norm exhibits a distinct cyclic behavior, alternating between rapid growth and plateau, with a sharp phase transition between phases. Simultaneously, the training loss retains a low value in periods of rapid norm growth, and then wildly fluctuating in periods of norm plateau. 
Figure~\ref{fig:grok_div_lln_50p}e and Figure~\ref{fig:grok_div_lln_50p}f shows the evolution of the relative change in features output by each layer in the Transformer. We observe that the feature maps are not updated much during the norm growth phase. However, at the phase transition, we observe that the feature maps receive a rapid update, which suggests that the internal representation of the model is updating.\\

\paragraph{Is Slingshot a general phenomenon?}
In an attempt to ascertain the generality of Slingshot Effects as an optimization artifact, we run similar experiments with additional architectures, datasets, optimizers, and hyperparameters. We use all algorithmic datasets as proposed in~\cite{power2021grokking}, as well as frequently used vision benchmarks such as CIFAR-10~\cite{krizhevsky09learningmultiple}, and even synthetic Gaussian dataset.
For architectures, we use Transformers, MLPs and deep linear models (see figure \ref{fig:slingshot_intro}). We find abundant evidence of Slingshot Effects in all of our experiments with Adam, AdamW and RMSProp. We are unable to observe Slingshot Effects with Adagrad~\cite{JMLR:v12:duchi11a} and also with stochastic gradient descent (SGD) or SGD with momentum, pointing to the generality of the mechanism across architectures and datasets. We refer the reader to Appendix~\ref{appendix:slingshot_optim_validation} for the full, detailed description of the experiments.   

\paragraph{Why does Slingshot happen?}\label{para:why_slingshot} We hypothesize that the norm growth continues until the curvature of the loss surface becomes large, effectively ``flinging" the weights to a different region in parameter space as small gradient directions get amplified, reminiscent of the mechanics of a slingshot flinging a projectile. We attempt to quantify how far a model is flung by measuring the cosine distance between a checkpoint during optimization and initial parameters. Specifically, we divide the model parameters into representation (pre-classifier) parameters and classifier (last layer) parameters and calculate how far these parameters have moved from initialization. We show that checkpoints collected after a model experiences Slingshot has a larger representation cosine distance. We defer the reader to the appendix for further details.

\begin{figure*}[h]
\centering
  \begin{tabular}{cc}
    \includegraphics[width=0.4\linewidth]{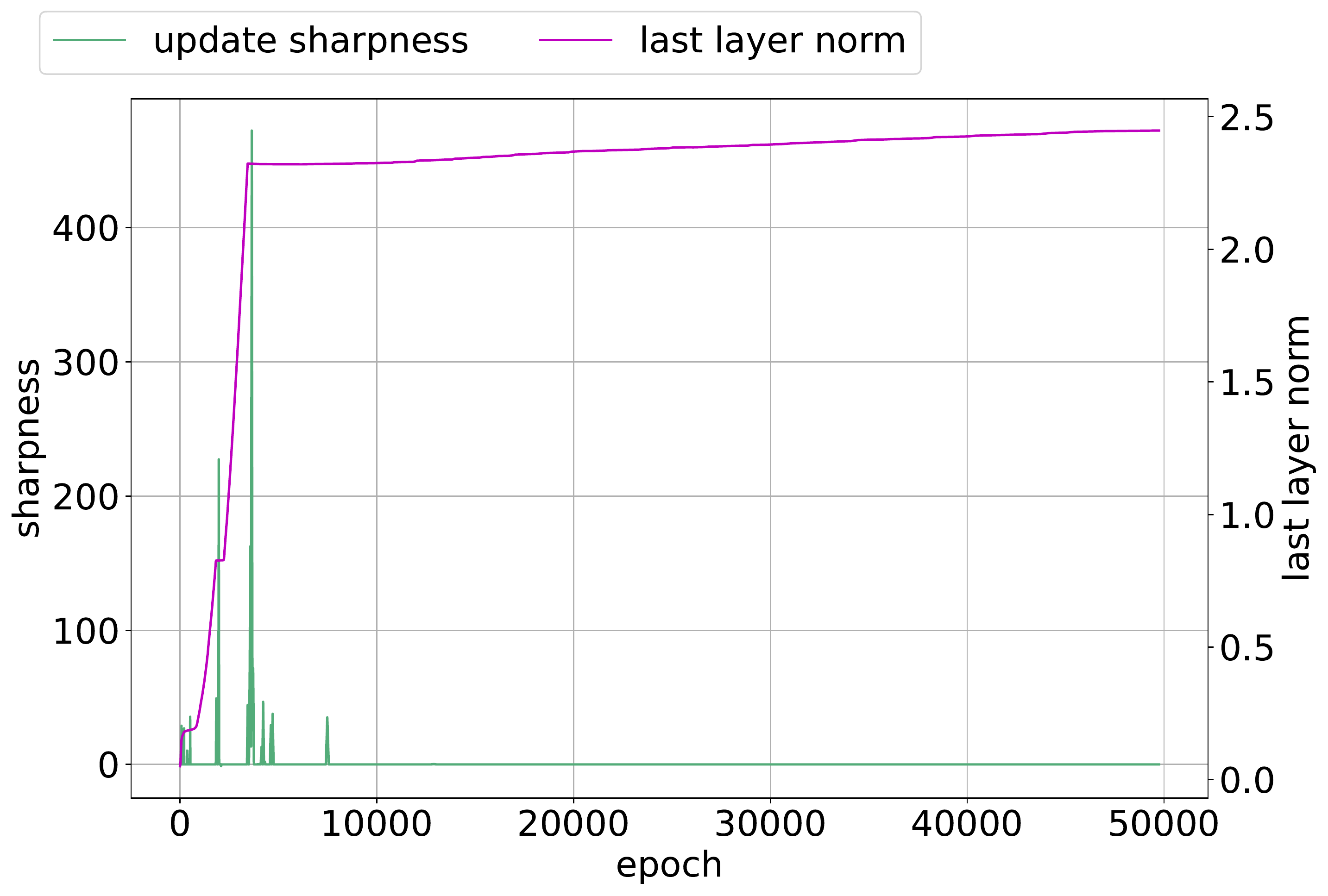} & 
    \includegraphics[width=0.4\linewidth]{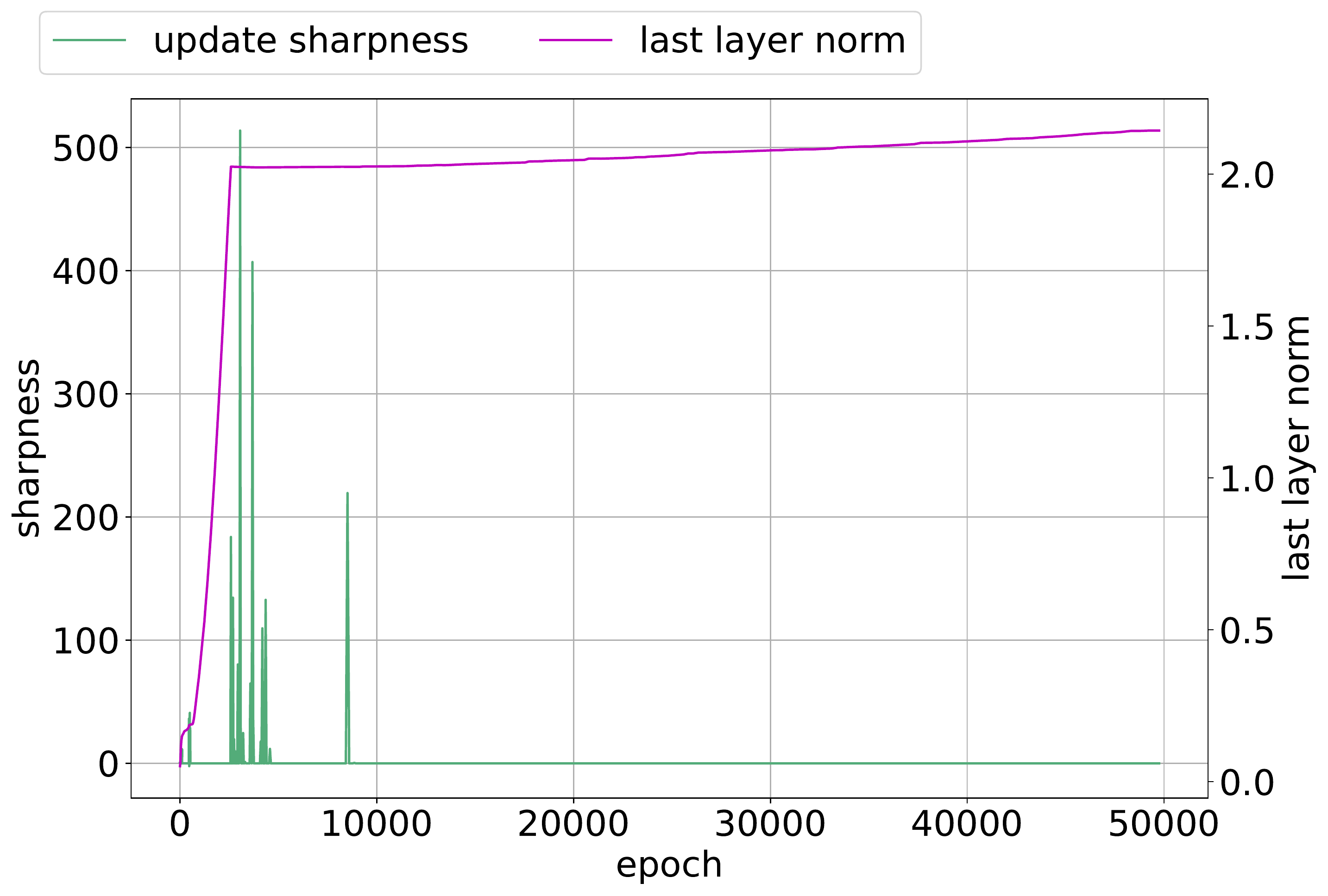} \\
    (a)  & (c) \\
    \includegraphics[width=0.4\linewidth]{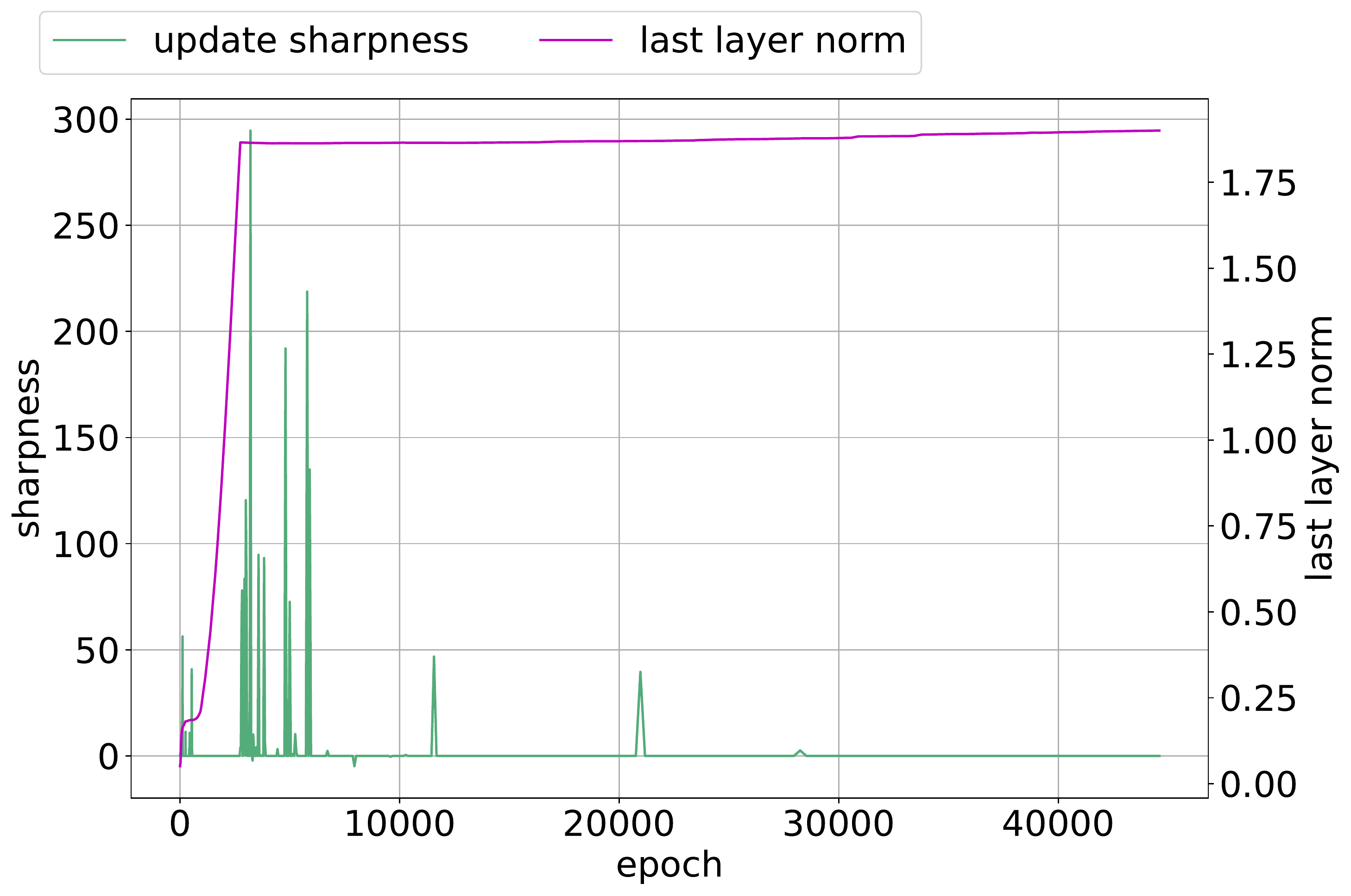} &
    \includegraphics[width=0.4\linewidth]{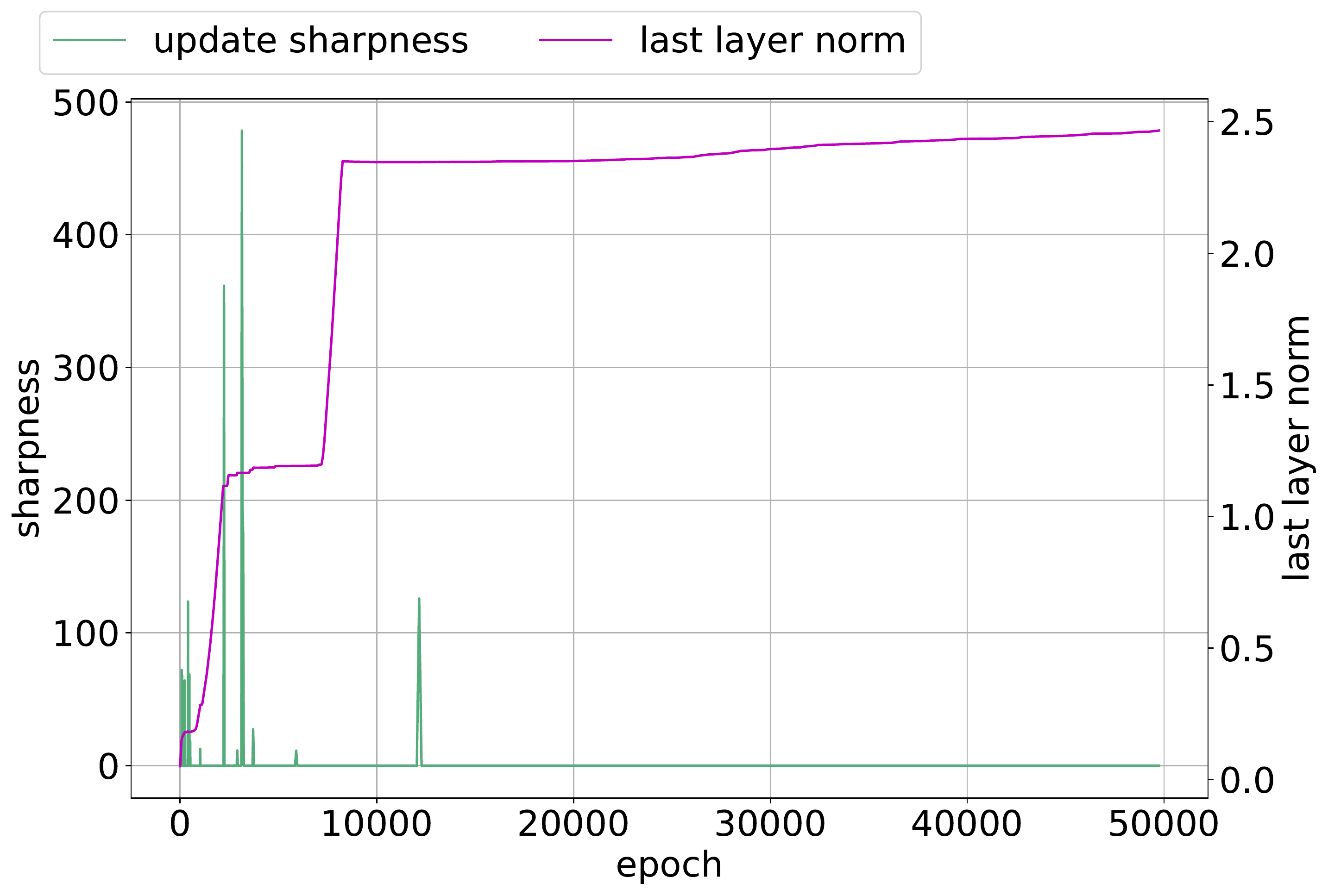} \\
    (b)  & (d)

  \end{tabular}
  \caption{Curvature metric (denoted as "update sharpness") evolution vs norm growth on (a) addition, (b) subtraction, (c) multiplication, and (d) division dataset. Note the spike in the sharpness metric near the phase transitions between norm growth and plateau.} 
  \label{fig:grok_update_sharpness}
\end{figure*}

By design, adaptive optimizers adapt the learning rate on a per parameter basis. In toy, convex scenarios, the $\epsilon$ parameter provably determine whether the algorithm will converge stably.  To illustrate this, we take inspiration from \cite{cohen2021gradient}, and consider a quadratic cost function $\mathcal{L}(A,B,C) = \frac{1}{2}x^\top A x + B^\top x + C, A \in \mathcal{R}^{d\times d},x,B \in \mathcal{R}^d, C \in \mathcal{R}$, where we assume $A$ is symmetric and positive definite. Note that the global minimum of this cost is given by $x^\star = -A^{-1}B$. The gradient of this cost with respect to $x$ is given by $g = Ax + B$. Consider optimizing the cost with adaptive optimization steps of the simple form $x_{t+1} = x_t - \mu\frac{g}{|g| + \epsilon} = x_t - \mu\frac{Ax_t + B}{|Ax_t + B| + \epsilon}$ where $\mu$ is a learning rate, and the division and absolute operations are taken element wise. Starting from some $x_0$, the error $e_{t} = x_{t} - x^\star$ evolves according to:
\begin{equation}
e_{t+1} = \big(I - \mu \text{diag}(\frac{1}{|Ae_t| + \epsilon})A\big)e_t \overset{\text{def}}{=}\mathcal{M}_t e_t
\end{equation}

Note that the condition $\|A\|_s < \frac{2\epsilon}{\mu}$ where $\|\cdot\|_s$ denotes the spectral norm,  implies that the mapping $\mathcal{M}_t$ is a contraction for all values of $t$, and hence convergence to the global optimum is guaranteed (This is in contrast to gradient descent, where the requirement is $\|A\|_s <\frac{2}{\mu}$). Note that the choice of $\epsilon$ crucially controls the requirement on the curvature of the cost, represented by the the spectrum of $A$ in this case. In other words, the smaller $\epsilon$, the more restrictive the requirements on the top eigenvalue of $A$. In~\cite{cohen2021gradient}, it was observed that full batch gradient descent increases the spectral norm of the hessian to its maximum allowed value. We therefore hypothesize that for deep networks, a small value for $\epsilon$ requires convergence to a low curvature local minimum, causing a Slingshot Effect when this does not occur. Moreover, we may reasonably predict that increasing the value of $\epsilon$ would lift the restriction on the curvature, and with it evidence of Slingshot Effects. 

Figure~\ref{fig:grok_update_sharpness} shows evidence consistent with the hypothesis that Slingshot Effects occur in the vicinity of high loss curvature, by measuring the local loss surface curvature along the optimization trajectory. Let $\mathcal{H}_t$ denote the local hessian matrix of the loss, and $u_t$ the parameter update at time $t$ given the optimization algorithm of choice. We use the local curvature along the trajectory of the optimizer, given by $\frac{1}{\|u_t\|^2}u_t^\top \mathcal{H}_t u_t$, as a curvature measure. Across the arithmetic datasets from~\cite{power2021grokking}, whenever the last layer weight norm plateaus, the curvature measure momentarily peaks and settles back down. 

\paragraph{Varying $\epsilon$}
\label{para:grok_vary_eps}

We next observe from Figure~\ref{fig:grok_div_lln_50p}a that the training loss value also spikes up around the time step when the weight norm transitions from growth to plateau. A low training loss value suggests that the gradients (and their moments) used as inputs to the optimizer are small, which in turn can cause the $\epsilon$ hyperparameter value to play a role in calculating updates. Our hypothesis here is that the Slingshot Effect should eventually disappear with a sufficiently large $\epsilon$. To confirm this hypothesis, we run an experiment where we vary $\epsilon$ while retaining the rest of the setup described in the previous section.

\begin{figure*}[t]
\centering
  \begin{tabular}{ccc}
      $\epsilon=10^{-08}$  & $\epsilon=10^{-07}$ & $\epsilon=10^{-05}$ \\
      \includegraphics[width=0.33\linewidth]{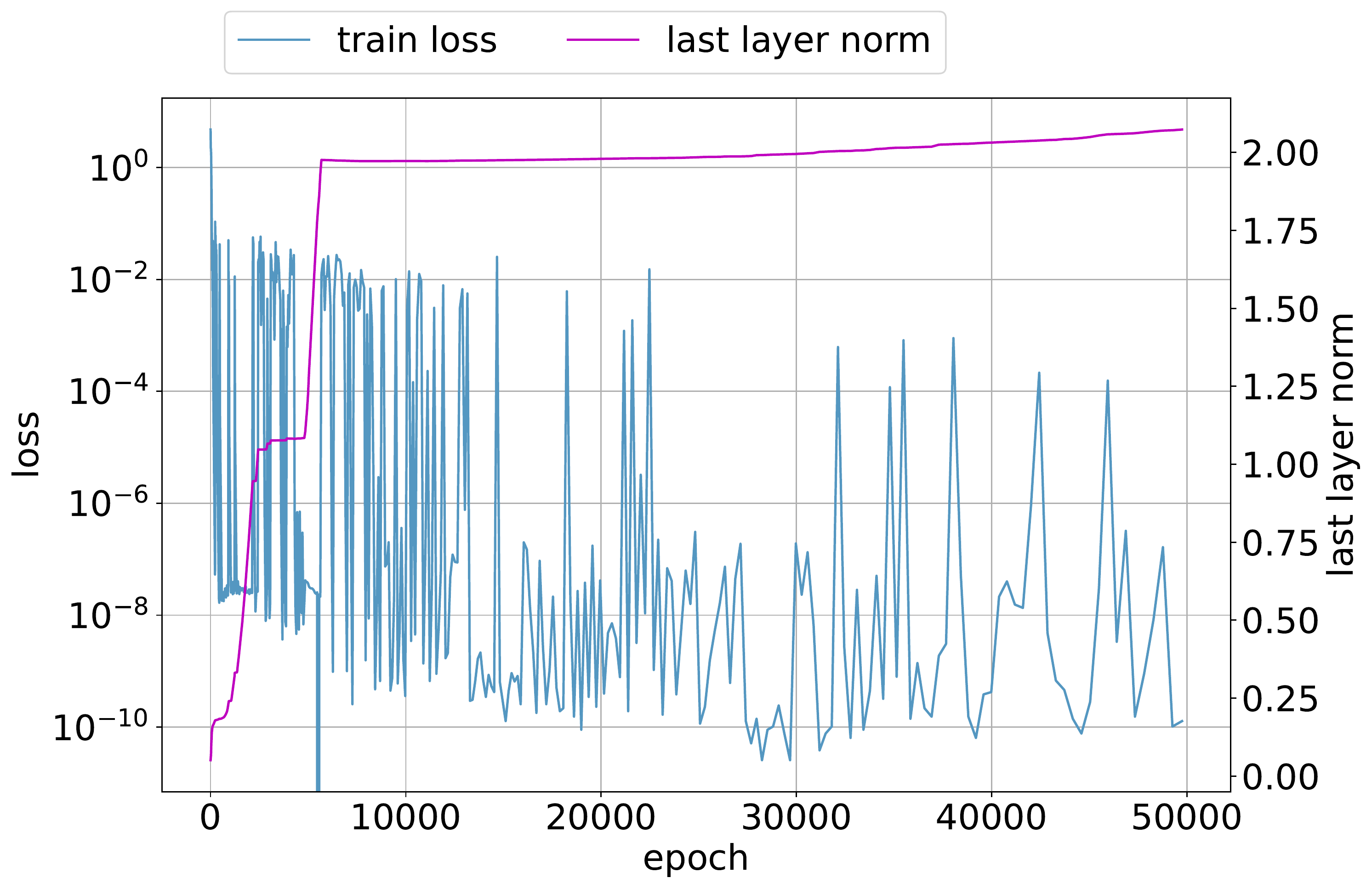} & 
      \includegraphics[width=0.33\linewidth]{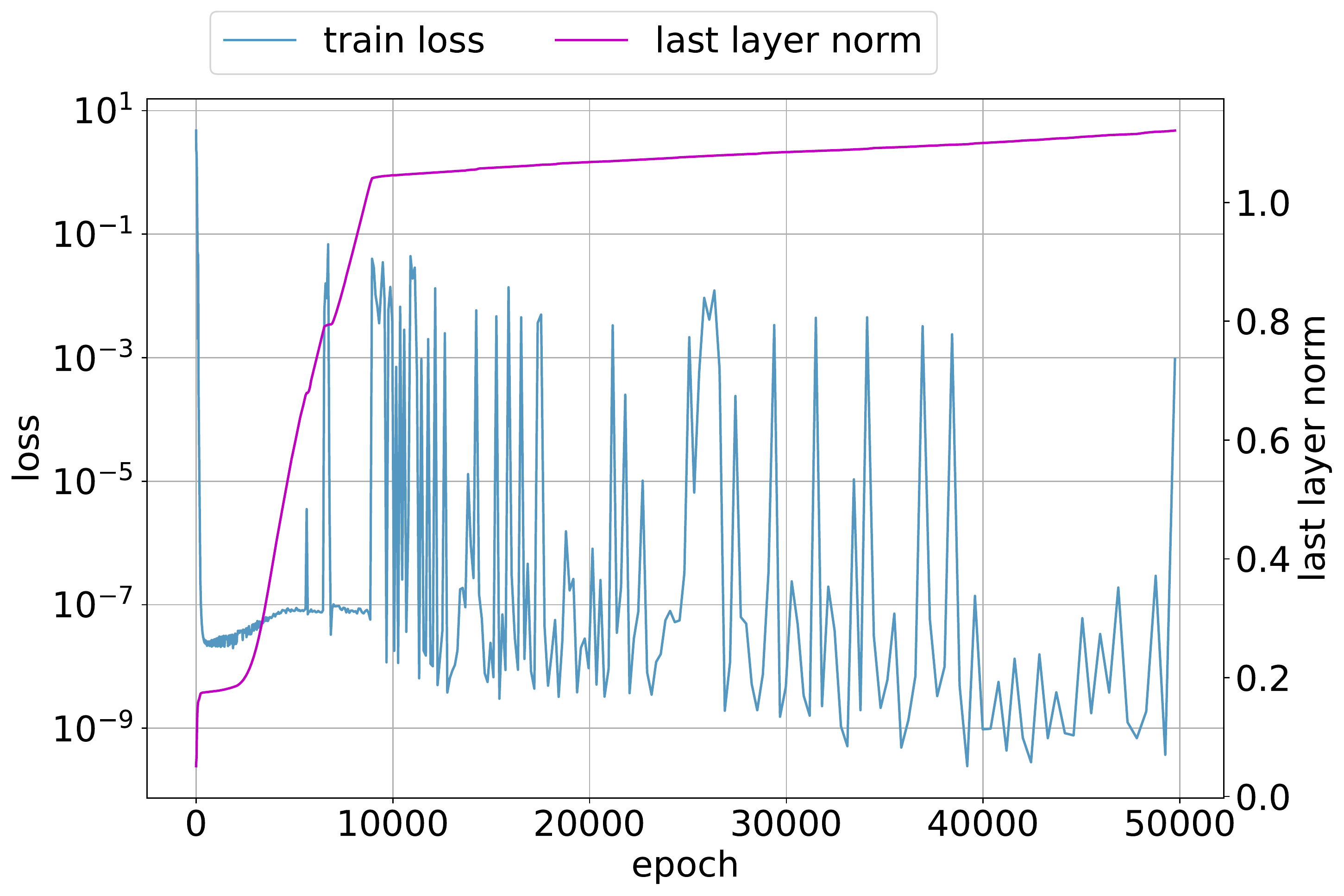} &
      \includegraphics[width=0.33\linewidth]{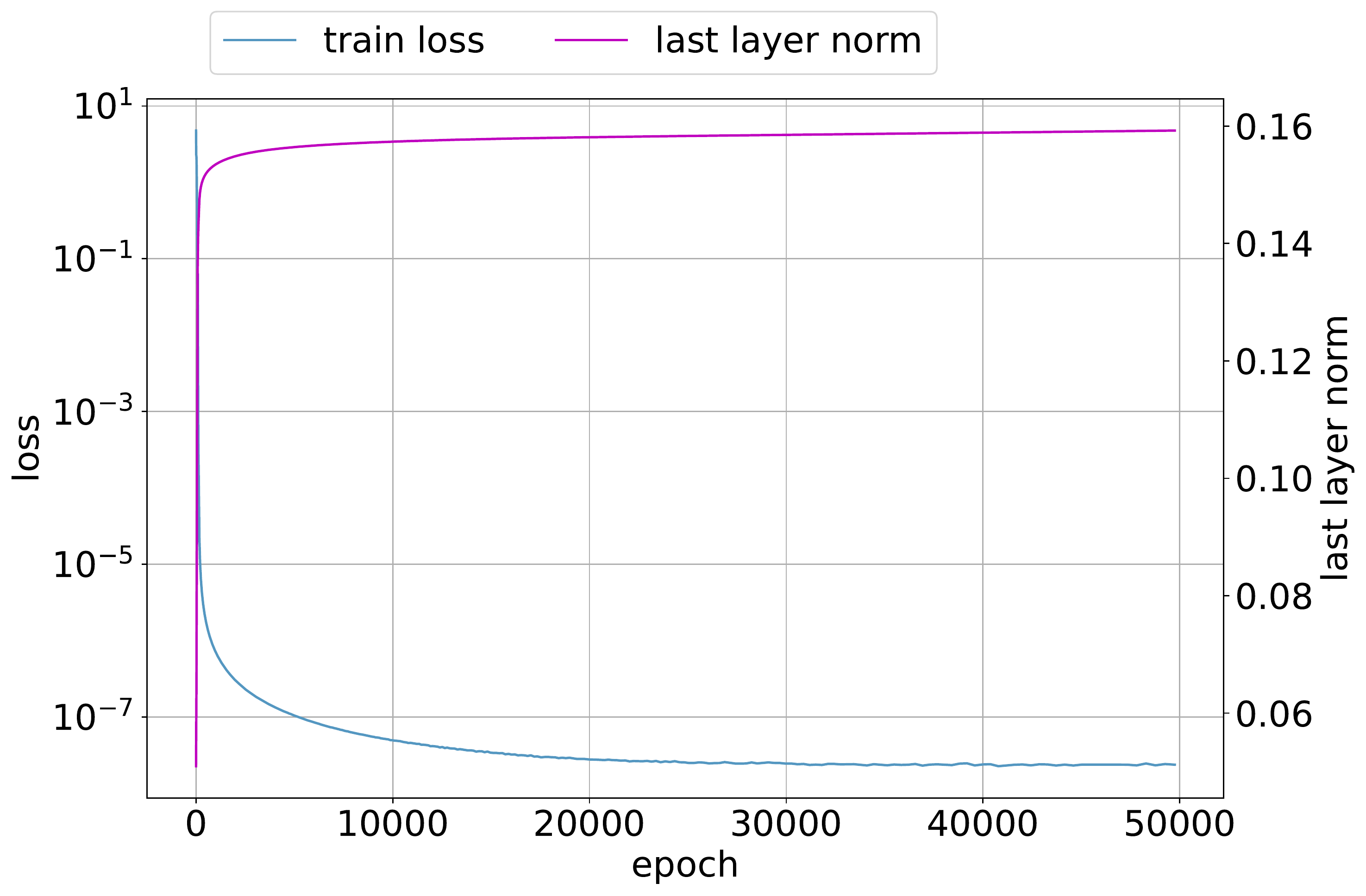} \\
      (a)  & (b) & (c) \\
      & training loss vs epochs \\
      \includegraphics[width=0.33\linewidth]{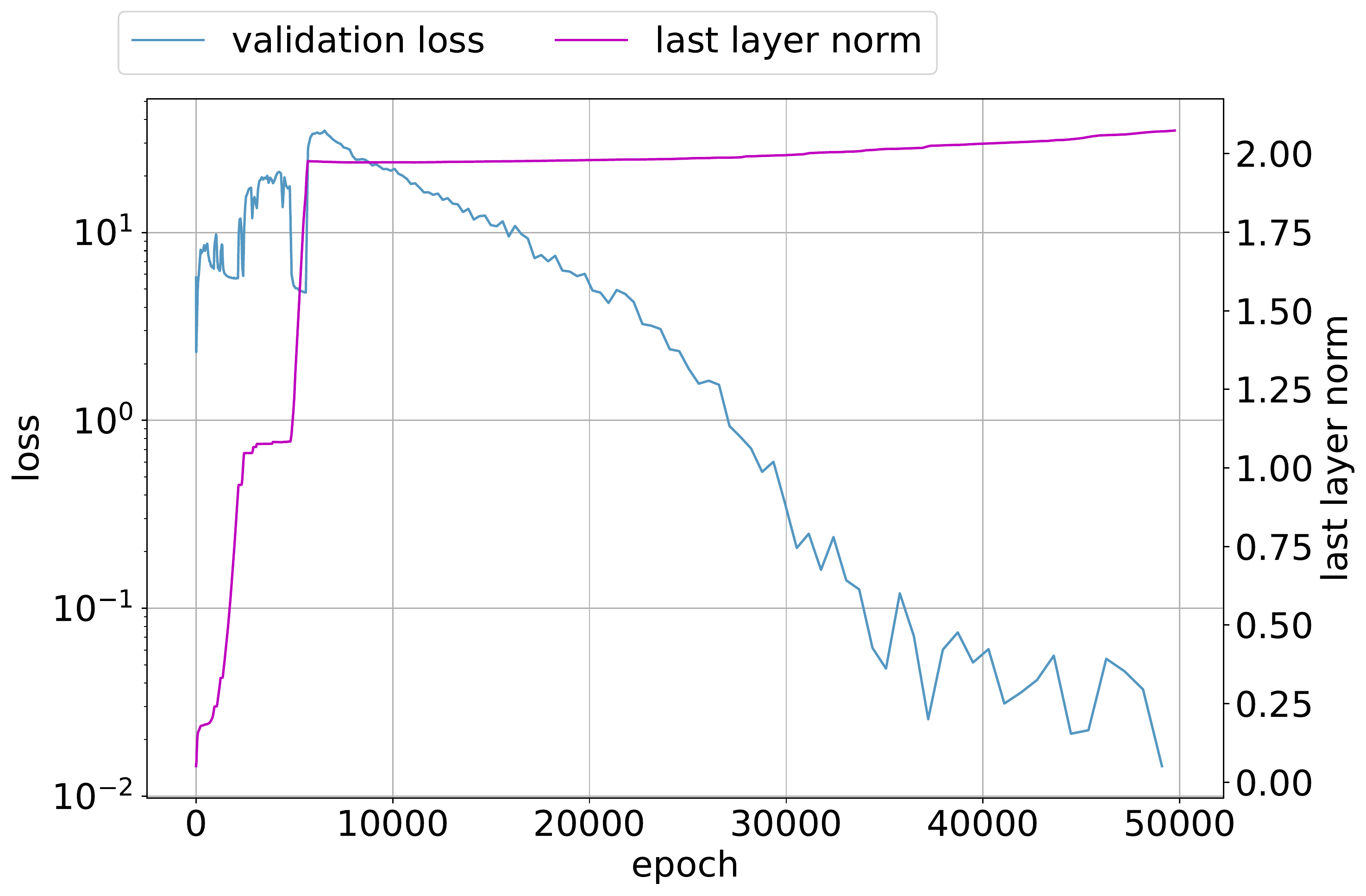} & 
      \includegraphics[width=0.33\linewidth]{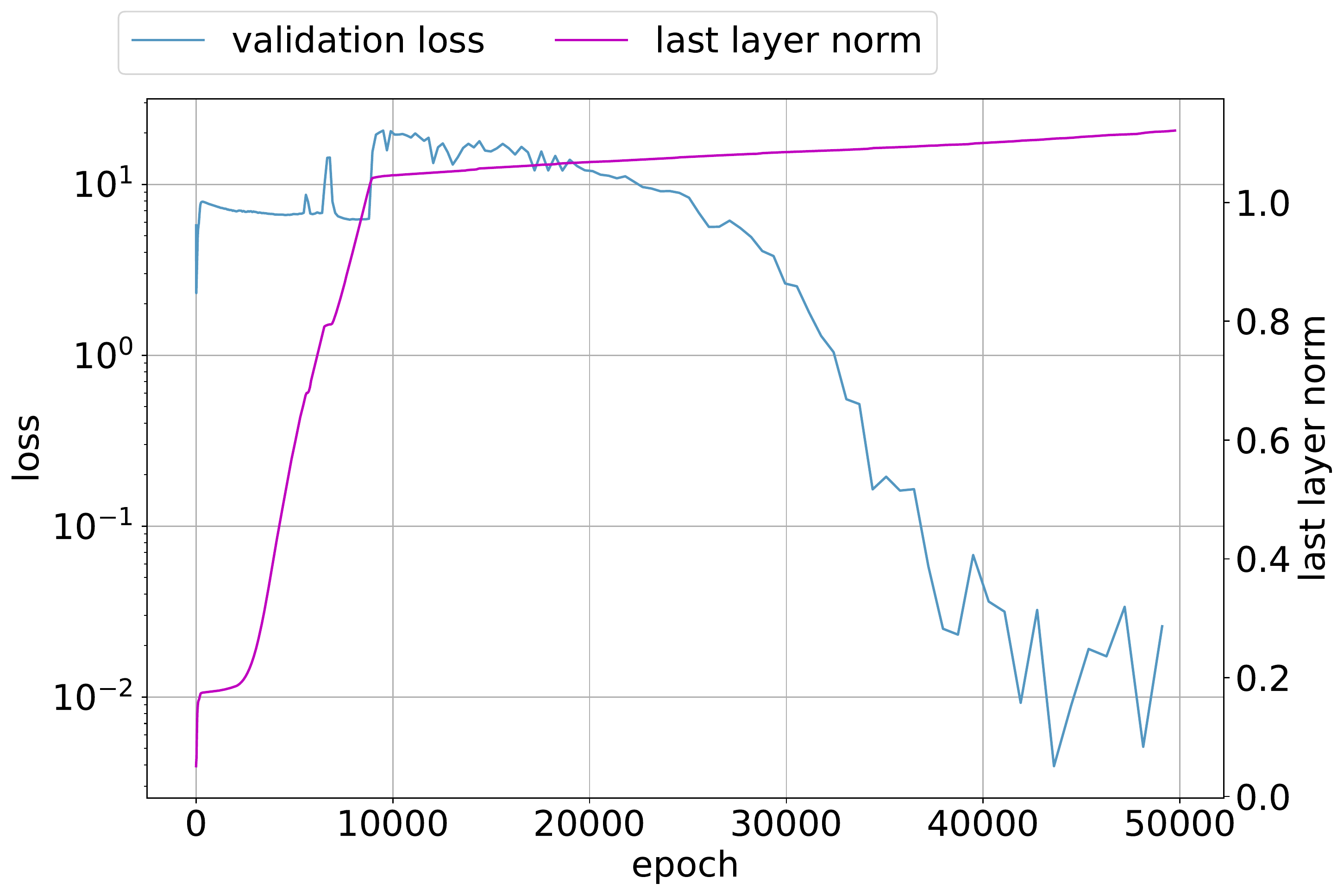} &
      \includegraphics[width=0.33\linewidth]{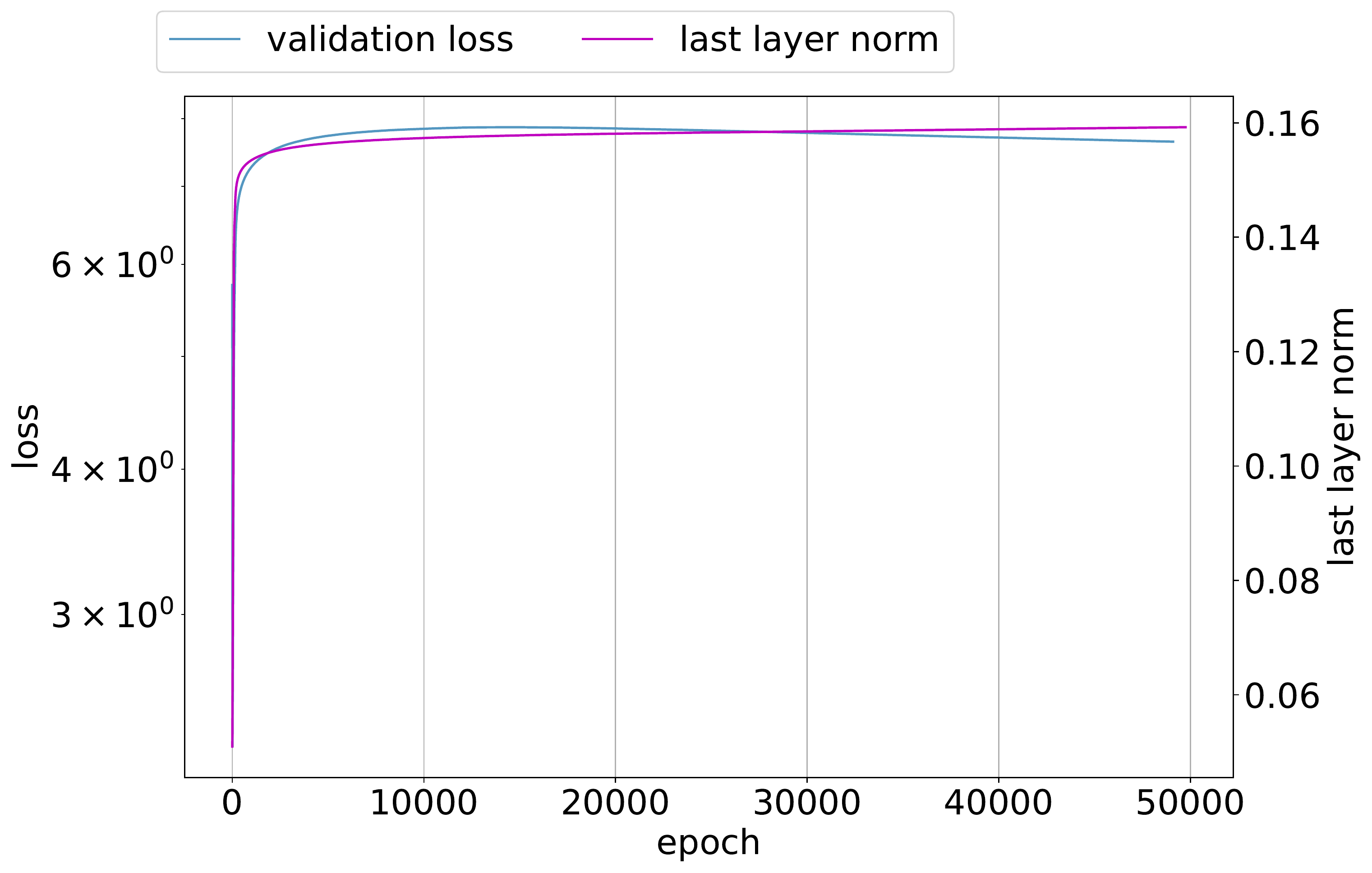} \\
      (d)  & (e) & (f) \\
      & validation loss vs epochs \\
      \includegraphics[width=0.33\linewidth]{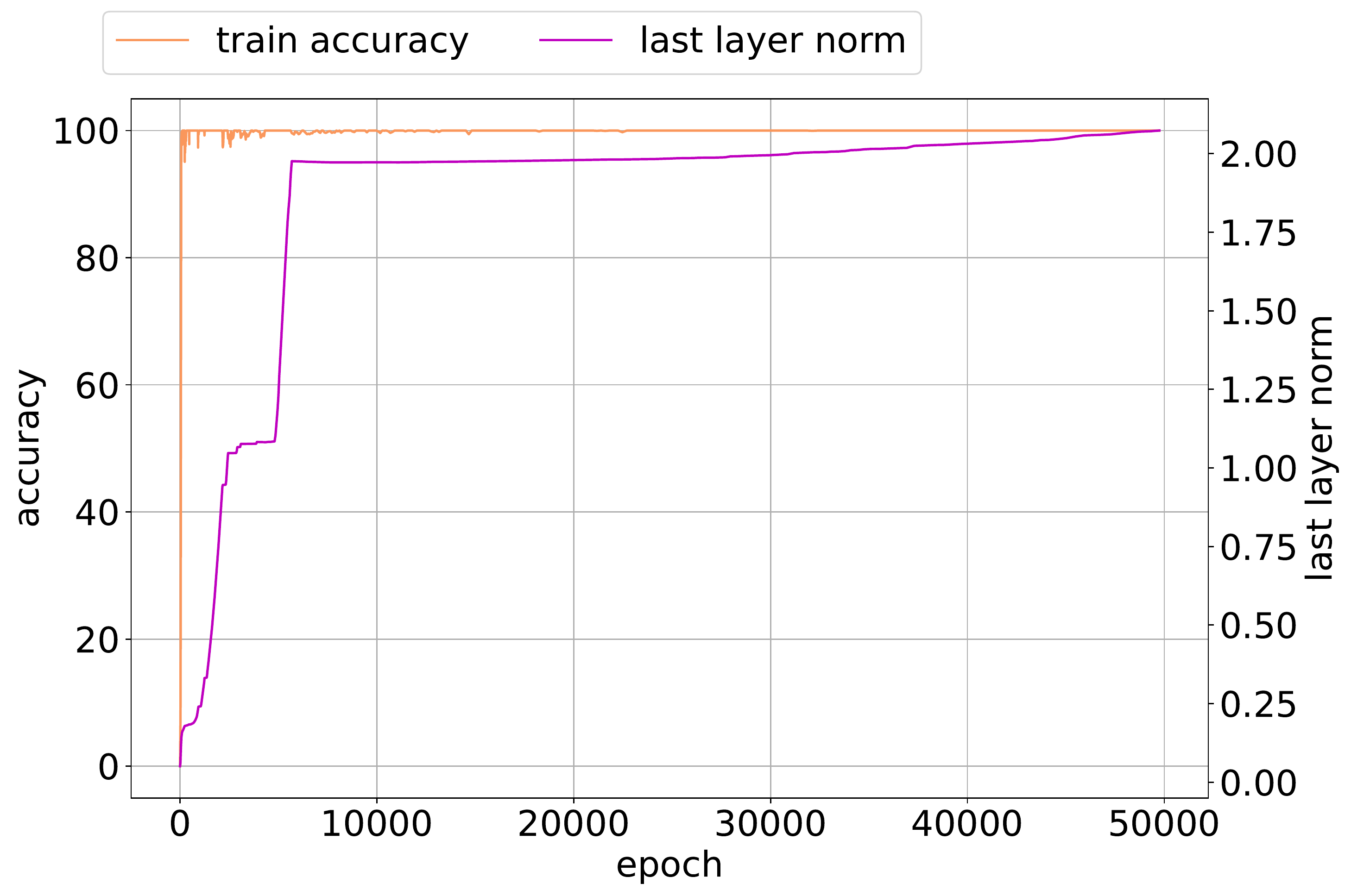} &    \includegraphics[width=0.33\linewidth]{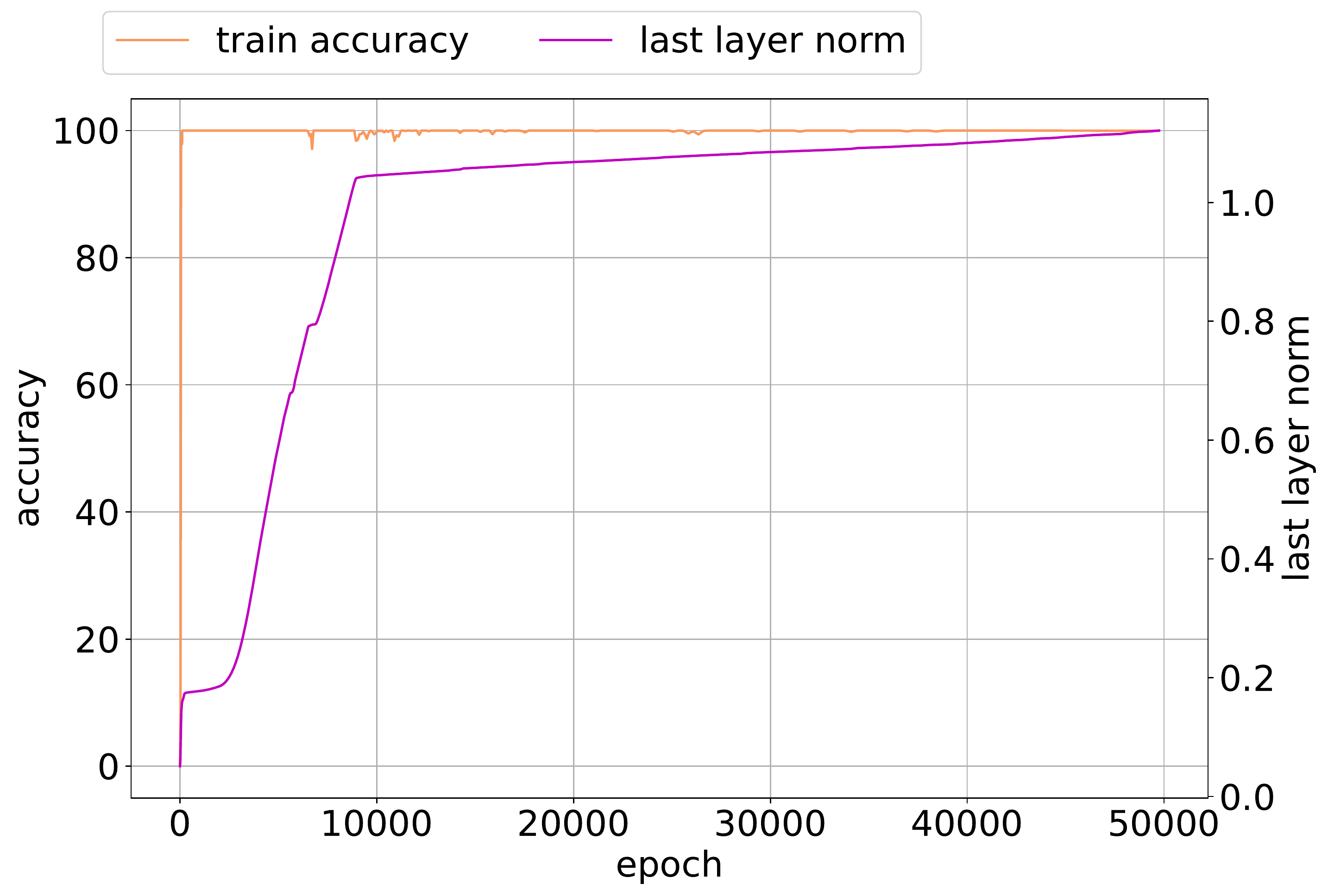} &    \includegraphics[width=0.33\linewidth]{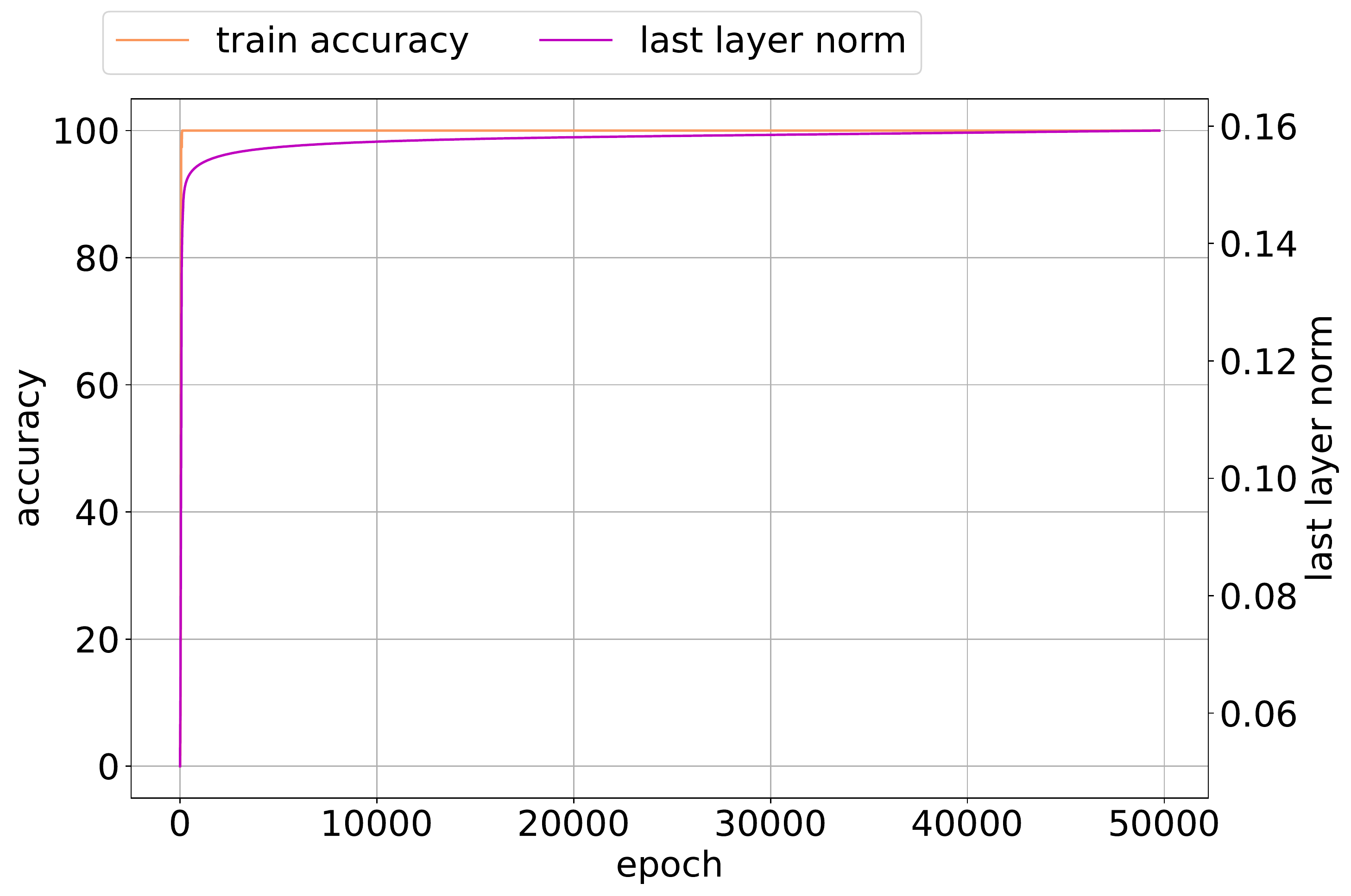} \\
      (g)  & (h) & (i) \\
      & training accuracy vs epochs \\
      \includegraphics[width=0.33\linewidth]{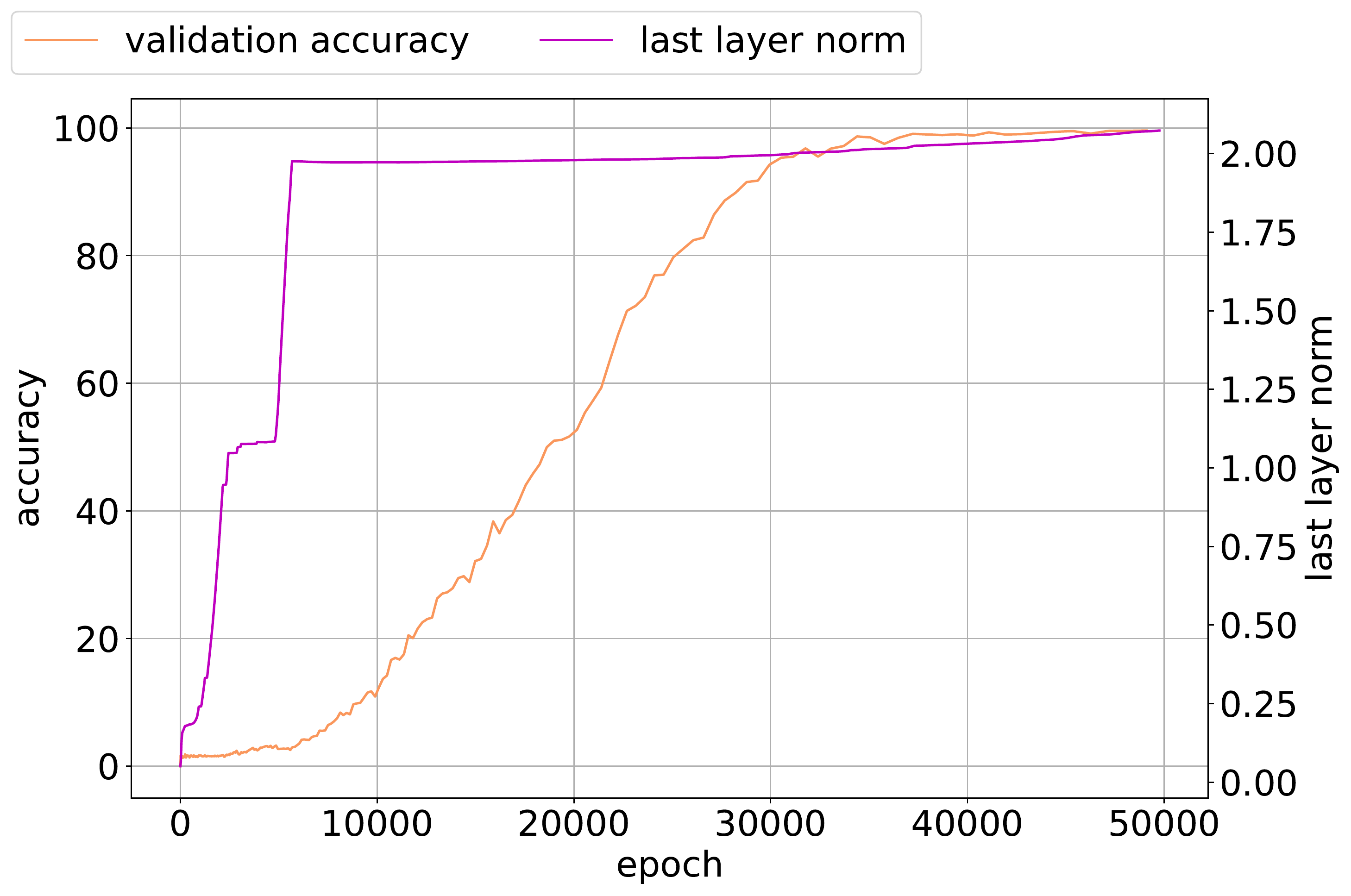} &
      \includegraphics[width=0.33\linewidth]{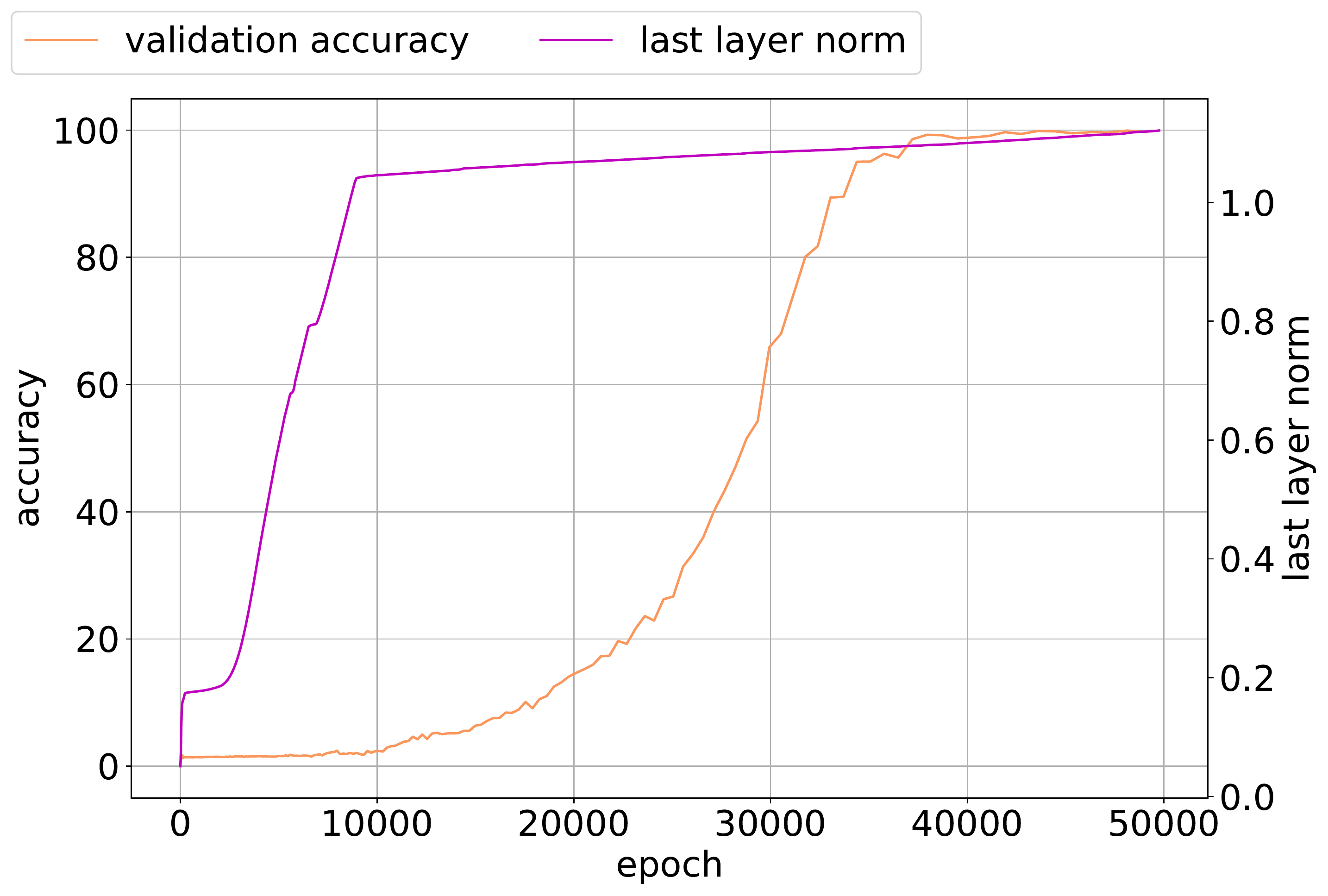} &
      \includegraphics[width=0.33\linewidth]{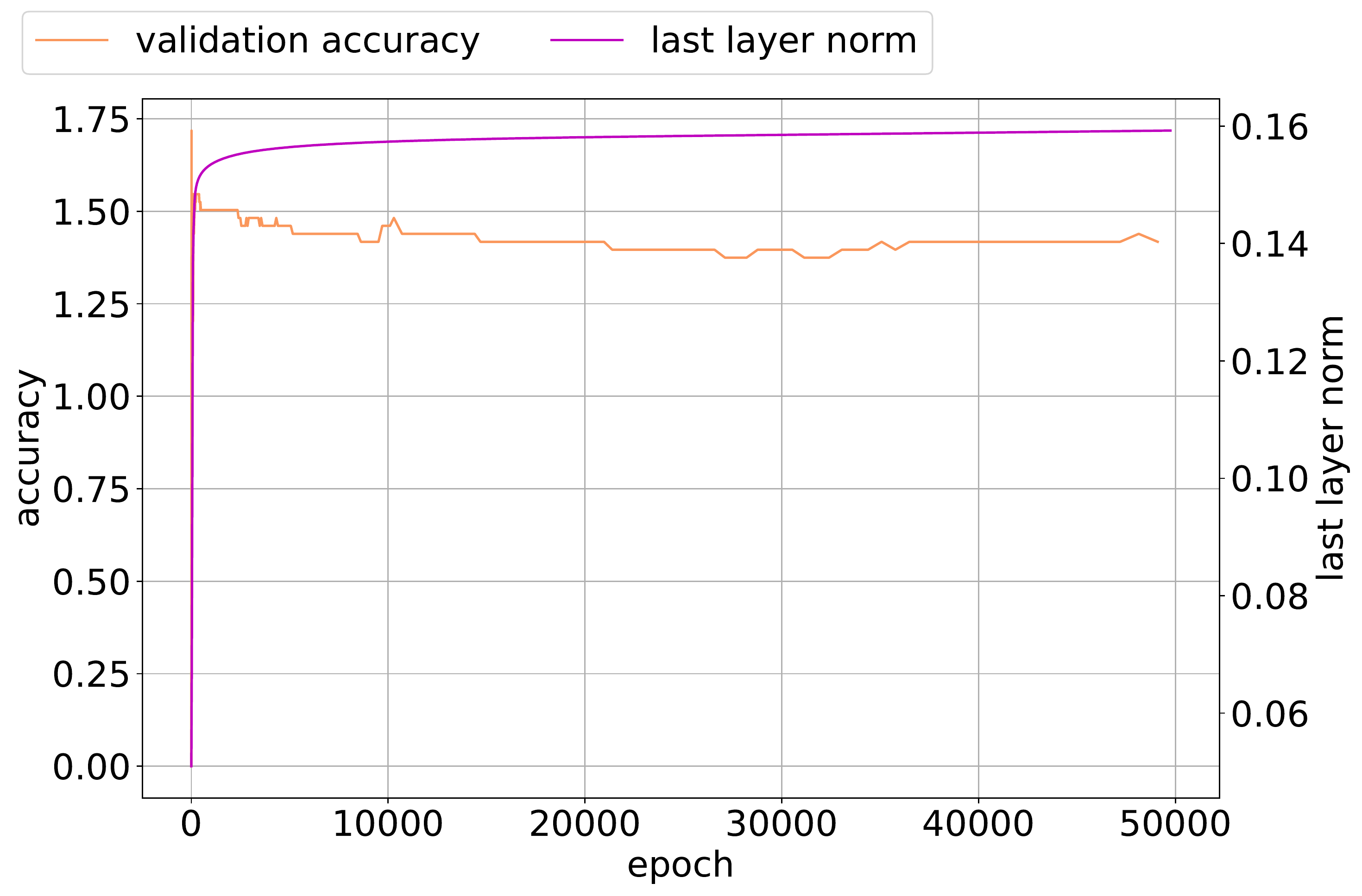} \\
      (j)  & (k) & (l) \\
      & validation accuracy vs epochs \\
  \end{tabular}
 \caption {Varying $\epsilon$ in Adam on the Division dataset. Observe that as $\epsilon$ increases, there is no Slingshot Effect or grokking behavior. Figure (a) corresponds to default $\epsilon$ suggested in~\cite{kingma2014adam}  where the model trained with smallest value undergoes multiple Slingshot cycles. } 
 \label{fig:grok_vary_eps}
\end{figure*}

Figure~\ref{fig:grok_vary_eps} shows the results for various values of $\epsilon$ considered in this experiment. We first observe that the number of Slingshot Effect cycles is higher for smaller values of $\epsilon$. Secondly, smaller values of $\epsilon$ cause grokking to appear at an earlier time step when compared to larger values. More intriguingly, models that show signs of grokking also experience Slingshot Effects while models that do not experience Slingshot Effects do not show any signs of grokking. Lastly, the model trained with the largest $\epsilon = 10^{-5}$ shows no sign of generalization even after receiving 500K updates.

\subsection{Effects on Generalization}
\label{section:slinghsot:slingshot_gen}

\label{sec:slingshot_gen}
\begin{figure*}[t]
\centering
  \includegraphics[width=0.9\linewidth]{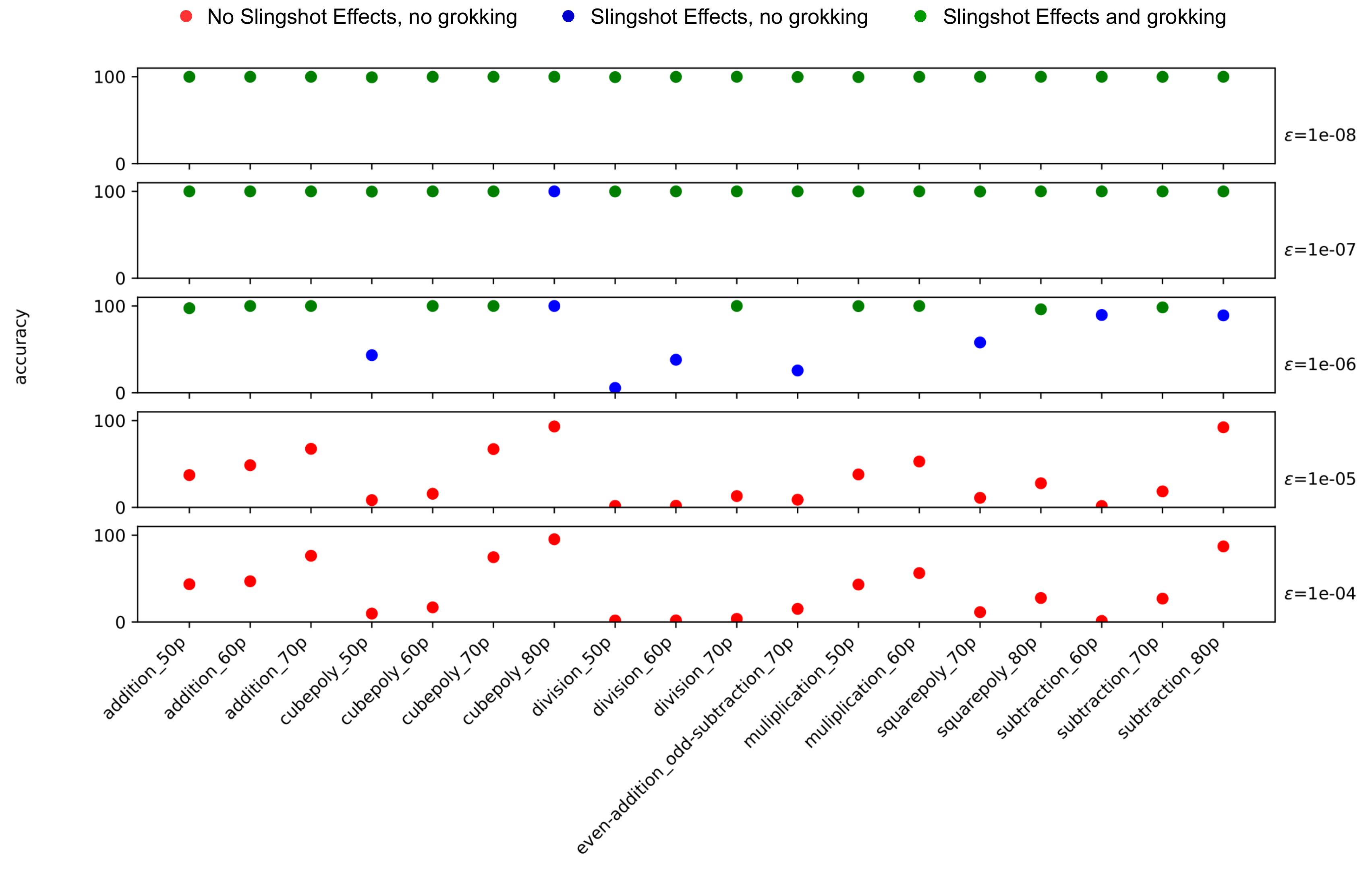}
   \caption{Extended analysis on multiple grokking datasets. Points shown in green represent both Slingshot Effects and grokking, points shown blue indicate Slingshot Effects but not grokking while points in red indicate no Slingshot Effects and no grokking. $\epsilon$ in Adam is varied as shown in text. Observe that as $\epsilon$ increases, there are no Slingshot Effects or grokking behavior.} 
  \label{fig:grok_macro_analysis}
\end{figure*}

In order to understand the relationship between Slingshot Effects and neural networks generalization, we experiment with various models and datasets.
We observe that models that exhibit Slingshot tend to generalize better, which suggests the benefit of training models for a long time with Adam~\cite{kingma2014adam} and AdamW~\cite{loshchilov2017decoupled}. More surprisingly, we observe that Slingshots and grokking tend to come in tandem.

\paragraph{Transformers with algorithmic datasets} 
We follow the setting in Power et al.~\cite{power2021grokking} and
generate several datasets that represent algorithmic operations and consider several training and validation splits. This dataset creation approach is consistent with the methodology used to demonstrate grokking~\cite{power2021grokking}. The Transformer is trained with AdamW~\cite{loshchilov2017decoupled} with a learning rate of 0.001, weight decay set to 0, and with learning rate warmup for 500K steps. We consider $\epsilon$ of AdamW as a hyperparameter in this experiment. Figure~\ref{fig:grok_macro_analysis} summarizes the results for this experiment where the x-axis indicates the algorithmic operation followed by the training data split size.
As can be seen in Figure~\ref{fig:grok_macro_analysis}, Slingshot Effects are seen with lower values of $\epsilon$ and disappear with higher values of $\epsilon$ which confirms the observations made in Section~\ref{sec:slingshot_mechanism} with modular division dataset. 
In addition, models that exhibit Slingshot Effects and grokking (shown in green) tend to generalize better than models that do not experience Slingshot Effects and grokking (shown in red). 

\begin{figure*}[h]
\centering
    \includegraphics[width=0.62\linewidth]{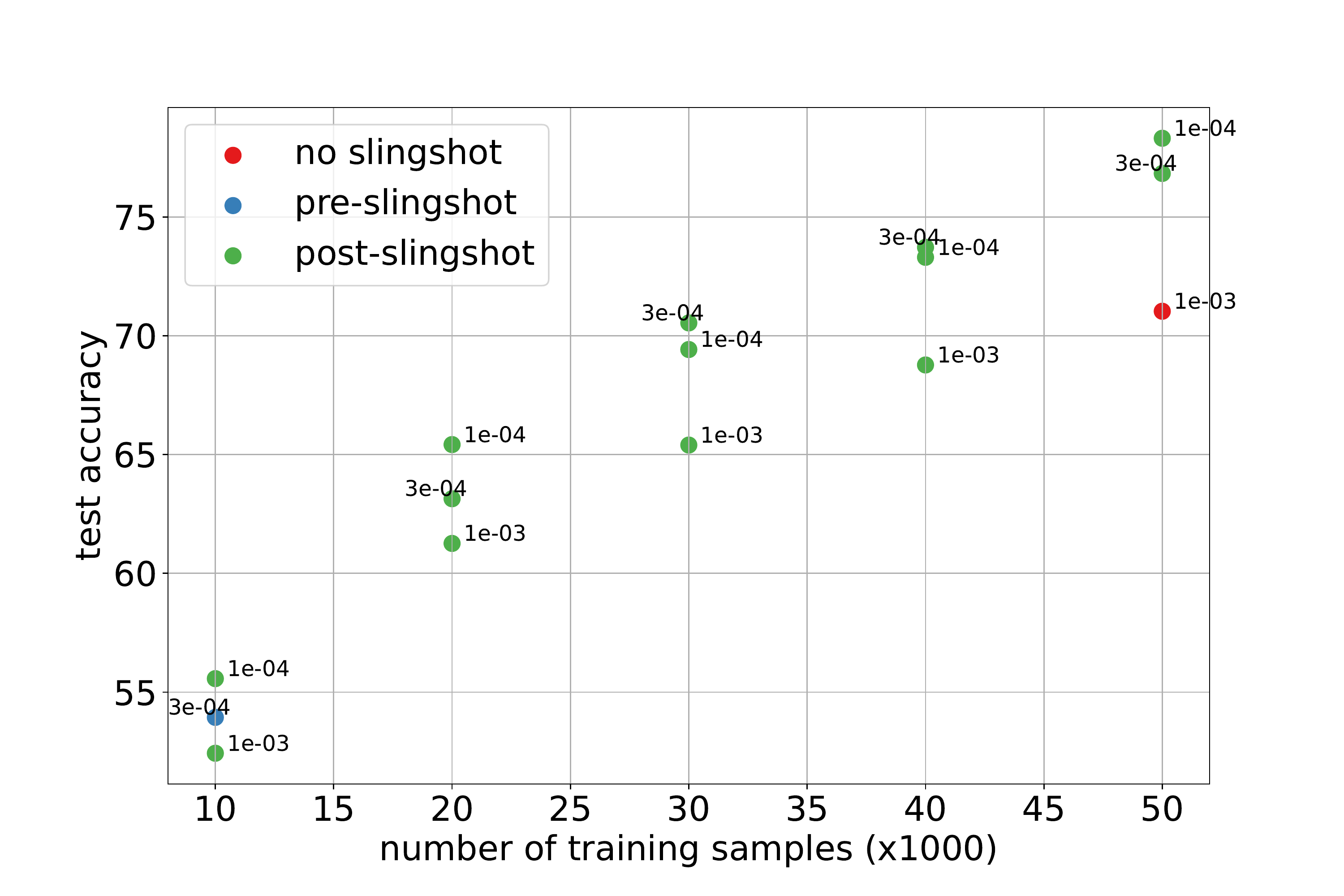}
\caption{Slingshot Effects on subsets of CIFAR-10 dataset. We train ViTs with multiple learning rates to verify the impact this parameter has on Slingshot. Power et al~\cite{power2021grokking} note that grokking occurs over a narrow range of learning rates. Note that the points marked in: (i) green correspond to test accuracy for an experiment after the Slingshot Effect begins, (ii) blue are for trials where best checkpoint is observed prior to start of a Slingshot Effect and (iii) red are for trials with no Slingshot Effect.}
\label{fig:gen_summary}
\end{figure*}

\paragraph{ViT with CIFAR-10} For further validation of Slingshot Effects and generalization, we train a Vision Transformer (ViT)~\cite{dosovitskiy2020an} on CIFAR-10~\cite{krizhevsky09learningmultiple}. The ViT consists of 12 layers, width 384 and 12 attention heads trained on fixed subsets of CIFAR-10 dataset~\cite{krizhevsky09learningmultiple}. The ViT model described above is trained with 10K, 20K, 30K, 40K and 50K (full dataset) training samples. We train the models with the following learning rates: $10^{-04}$, $3.10e^{-04}4$ and $10^{-03}$ and with a linear learning rate warmup for the 1 epoch of optimization. We consider multiple learning rates to study the impact of this hyperparameter on Slingshot taking inspiration from~\cite{power2021grokking} where the authors report observing grokking over a narrow range of learning rates . Figure~\ref{fig:gen_summary} shows a plot of the highest test accuracy for a set of hyperparameters (learning rate, number of training samples) as a function of the number of training samples from which we make the following observations. The best test accuracy for a given set of hyperparameters is typically achieved after Slingshot phase begins during optimization. The checkpoints that achieve the highest test accuracy are labeled as "post-slingshot" and shown in green in Figure~\ref{fig:gen_summary}. While post-Slingshot checkpoints seem to enjoy higher test accuracy, there are certain combinations of hyperparameters that lead to models that show better test accuracy prior to the start of the first Slingshot phase. We label these points as "pre-slingshot" (shown in blue) in Figure~\ref{fig:gen_summary}. The above observations appear to be consistent with our finding that training long periods of time may lead to better generalization seen  with grokking datasets~\cite{power2021grokking}.

\paragraph{Non-Transformer Models} We conduct experiments with MLPs on synthetic data where the synthetic data is a low dimensional embedding projected to higher dimensions via random projections. This design choices is critical with showing the existence of the Slingshot Effect with synthetically generated data. We find that using  low dimensional data does not lead to any Slingshots. With this dataset, we show that generalization occurs late in training with Adam. Specifically, we tune $\epsilon$ in Adam and show that the optimizer is highly sensitive to this hyperparameter. These observations are consistent with the behavior reported above with Transformers and on algorithmic datasets as well as standard vision benchmark such as CIFAR-10. We refer the reader to Appendix~\ref{appendix:synth_data} for complete description and details of these experiments.

\subsection{Drawbacks and Limitations}

While the Slingshot Mechanism exposes an interesting implicit bias of Adam that often promotes generalization, due to it's arresting of the norm growth and ensuing feature learning, it also leads to some training instability and prolonged training time.
In the Appendix we show that it is possible to achieve similar levels of generalization with Adam on the modular division dataset~\cite{power2021grokking} using the same Transformer setup as above, while maintaining stable learning, in regimes that do not show a clear Slingshot Effect.
First we employ weight decay, which causes the training loss values to converge to a higher value than the unregularized model. In this regime the model does not become unstable, but instead regularization leads to comparable generalization, and much more quickly. However, it is important to tune the regularization strength appropriately.
Similarly, we find that it is possible to normalize the features and weights using the following scheme to explicitly control norm growth:
$
    w = \frac{w}{\lVert  w  \rVert},
    f(x) = \frac{f(x)}{\lVert  f(x)  \rVert},
$
where $w$ and $f(x)$ are the weights and inputs to the classification layer respectively, the norm used above is the $L_{2}$ norm, and $x$ is the input to the neural network. This scheme also results in stable training and similar levels of generalization. In all cases the effects rely on keeping the weight norms from growing uncontrollably, which may be the most important factor for improving generalization.
These results suggest that while the Slingshot Mechanism may be an interesting self-correcting scheme for controlling norm growth, there are likely more efficient ways to leverage adaptive optimizers to similar levels of generalization without requiring the instability that is a hallmark of the Slingshot effect.\\
Finally, we lack a satisfactory theoretical explanation for the Slingshot Mechanism, and hence removed all attempts at a more rigorous mathematical definition, which we feel would only serve as a distraction.   

\section{Conclusion}

We have empirically shown that optimizing deep networks with cross entropy loss and adaptive optimizers produces the Slingshot Mechanism, a curious optimization anomaly unlike anything described in the literature. We have provided ample evidence that Slingshot Effects can be observed with different neural architectures and datasets. Furthermore, we find that Grokking \cite{power2021grokking} almost always occurs in the presence of Slingshot Effects and associated regions of instability in the Terminal Phase of Training (TPT). These results in their pure form absent explicit regularization, reveal an intriguing inductive bias of adaptive gradient optimizers that becomes salient in the TPT, characterized by cyclic stepwise effects on the optimization trajectory. These effects often promote generalization in ways that differ from non-adaptive optimizers like SGD, and warrant further study to be able to harness efficiently. There are open question remaining to be answered, for instance \textbf{1)} What's the causal factor of the plateau of weight norm growth? \textbf{2)} Are there better ways of promoting generalization without relying on this accidental training instability? Answering these questions w ill allow us to decouple optimization and regularization, and ultimately to control and improve them independently.


\bibliographystyle{plain}
\bibliography{main}

\begin{thebibliography}{10}

\bibitem{AllenZhu2019ACT}
Zeyuan Allen-Zhu, Yuanzhi Li, and Zhao Song.
\newblock A convergence theory for deep learning via over-parameterization.
\newblock {\em ArXiv}, abs/1811.03962, 2019.

\bibitem{Barakat2021ConvergenceAD}
Anas Barakat and Pascal Bianchi.
\newblock Convergence and dynamical behavior of the adam algorithm for
  nonconvex stochastic optimization.
\newblock {\em SIAM J. Optim.}, 31:244--274, 2021.

\bibitem{jax2018github}
James Bradbury, Roy Frostig, Peter Hawkins, Matthew~James Johnson, Chris Leary,
  Dougal Maclaurin, George Necula, Adam Paszke, Jake Vander{P}las, Skye
  Wanderman-{M}ilne, and Qiao Zhang.
\newblock {JAX}: composable transformations of {P}ython+{N}um{P}y programs.
\newblock 2018.

\bibitem{cohen2021gradient}
Jeremy~M. Cohen, Simran Kaur, Yuanzhi Li, J.~Zico Kolter, and Ameet Talwalkar.
\newblock Gradient descent on neural networks typically occurs at the edge of
  stability.
\newblock {\em arXiv preprint arXiv: Arxiv-2103.00065}, 2021.

\bibitem{dosovitskiy2020an}
Alexey Dosovitskiy, Lucas Beyer, Alexander Kolesnikov, Dirk Weissenborn,
  Xiaohua Zhai, Thomas Unterthiner, Mostafa Dehghani, Matthias Minderer, Georg
  Heigold, Sylvain Gelly, Jakob Uszkoreit, and Neil Houlsby.
\newblock An image is worth 16x16 words: Transformers for image recognition at
  scale.
\newblock {\em arXiv preprint arXiv: Arxiv-2010.11929}, 2020.

\bibitem{JMLR:v12:duchi11a}
John Duchi, Elad Hazan, and Yoram Singer.
\newblock Adaptive subgradient methods for online learning and stochastic
  optimization.
\newblock {\em Journal of Machine Learning Research}, 12(61):2121--2159, 2011.

\bibitem{Hoffer2017TrainLG}
Elad Hoffer, Itay Hubara, and Daniel Soudry.
\newblock Train longer, generalize better: closing the generalization gap in
  large batch training of neural networks.
\newblock {\em ArXiv}, abs/1705.08741, 2017.

\bibitem{ioffe2015batch}
Sergey Ioffe and Christian Szegedy.
\newblock Batch normalization: Accelerating deep network training by reducing
  internal covariate shift.
\newblock In {\em International conference on machine learning}, pages
  448--456. PMLR, 2015.

\bibitem{kingma2014adam}
Diederik~P. Kingma and Jimmy Ba.
\newblock Adam: A method for stochastic optimization.
\newblock {\em arXiv preprint arXiv: Arxiv-1412.6980}, 2014.

\bibitem{krizhevsky09learningmultiple}
Alex Krizhevsky.
\newblock Learning multiple layers of features from tiny images.
\newblock 2009.

\bibitem{lewkowycz2020large}
Aitor Lewkowycz, Yasaman Bahri, Ethan Dyer, Jascha Sohl-Dickstein, and Guy
  Gur-Ari.
\newblock The large learning rate phase of deep learning: the catapult
  mechanism.
\newblock {\em arXiv preprint arXiv:2003.02218}, 2020.

\bibitem{loshchilov2017decoupled}
Ilya Loshchilov and Frank Hutter.
\newblock Decoupled weight decay regularization.
\newblock {\em arXiv preprint arXiv:1711.05101}, 2017.

\bibitem{lyu2019gradient}
Kaifeng Lyu and Jian Li.
\newblock Gradient descent maximizes the margin of homogeneous neural networks.
\newblock {\em arXiv preprint arXiv:1906.05890}, 2019.

\bibitem{Papyan2020PrevalenceON}
Vardan Papyan, X.~Y. Han, and David~L. Donoho.
\newblock Prevalence of neural collapse during the terminal phase of deep
  learning training.
\newblock {\em Proceedings of the National Academy of Sciences of the United
  States of America}, 117:24652 -- 24663, 2020.

\bibitem{paszke2019}
Adam Paszke, Sam Gross, Francisco Massa, Adam Lerer, James Bradbury, Gregory
  Chanan, Trevor Killeen, Zeming Lin, Natalia Gimelshein, Luca Antiga, Alban
  Desmaison, Andreas Kopf, Edward Yang, Zachary DeVito, Martin Raison, Alykhan
  Tejani, Sasank Chilamkurthy, Benoit Steiner, Lu~Fang, Junjie Bai, and Soumith
  Chintala.
\newblock Pytorch: An imperative style, high-performance deep learning library.
\newblock In H.~Wallach, H.~Larochelle, A.~Beygelzimer, F.~d\textquotesingle
  Alch\'{e}-Buc, E.~Fox, and R.~Garnett, editors, {\em Advances in Neural
  Information Processing Systems 32}, pages 8024--8035. Curran Associates,
  Inc., 2019.

\bibitem{scikit-learn}
F.~Pedregosa, G.~Varoquaux, A.~Gramfort, V.~Michel, B.~Thirion, O.~Grisel,
  M.~Blondel, P.~Prettenhofer, R.~Weiss, V.~Dubourg, J.~Vanderplas, A.~Passos,
  D.~Cournapeau, M.~Brucher, M.~Perrot, and E.~Duchesnay.
\newblock Scikit-learn: Machine learning in {P}ython.
\newblock {\em Journal of Machine Learning Research}, 12:2825--2830, 2011.

\bibitem{power2021grokking}
Alethea Power, Yuri Burda, Harri Edwards, Igor Babuschkin, and Vedant Misra.
\newblock Grokking: Generalization beyond overfitting on small algorithmic
  datasets.
\newblock In {\em ICLR MATH-AI Workshop}, 2021.

\bibitem{simonyan2014deep}
Karen Simonyan and Andrew Zisserman.
\newblock Very deep convolutional networks for large-scale image recognition.
\newblock {\em arXiv preprint arXiv: Arxiv-1409.1556}, 2014.

\bibitem{soudry2018implicit}
Daniel Soudry, Elad Hoffer, Mor~Shpigel Nacson, Suriya Gunasekar, and Nathan
  Srebro.
\newblock The implicit bias of gradient descent on separable data.
\newblock {\em The Journal of Machine Learning Research}, 19(1):2822--2878,
  2018.

\bibitem{tieleman2012lecture}
Tijmen Tieleman and Geoffrey Hinton.
\newblock Lecture 6.5-rmsprop, coursera: Neural networks for machine learning.
\newblock {\em University of Toronto, Technical Report}, 6, 2012.

\bibitem{vaswani2017attention}
Ashish Vaswani, Noam Shazeer, Niki Parmar, Jakob Uszkoreit, Llion Jones,
  Aidan~N. Gomez, Lukasz Kaiser, and Illia Polosukhin.
\newblock Attention is all you need.
\newblock {\em arXiv preprint arXiv: Arxiv-1706.03762}, 2017.

\bibitem{wang2021implicit}
Bohan Wang, Qi~Meng, Wei Chen, and Tie-Yan Liu.
\newblock The implicit bias for adaptive optimization algorithms on homogeneous
  neural networks.
\newblock In {\em International Conference on Machine Learning}, pages
  10849--10858. PMLR, 2021.

\bibitem{Zhang2020WhyGC}
J.~Zhang, Tianxing He, Suvrit Sra, and Ali Jadbabaie.
\newblock Why gradient clipping accelerates training: A theoretical
  justification for adaptivity.
\newblock {\em arXiv: Optimization and Control}, 2020.

\end{thebibliography}

\clearpage

\begin{center}
\textbf{\LARGE The Slingshot Mechanism: An Empirical Study of Adaptive Optimizers and the \emph{Grokking Phenomenon} - Appendix}
\end{center}

\begin{appendices}
\tableofcontents

\newpage
\section{Slingshot Effects across Architectures, Optimizers and Datasets}
\label{appendix:slingshot_optim_validation}

This section provides further evidence of the prevalence of Slingshot across architectures and optimizers on subsets of CIFAR-10, testing setups beyond the specific setup consider by Power et al.~\cite{power2021grokking}. In these experiments, we focus solely on characterizing the optimization properties of various setups described below. The small sample sizes are used in order to more easily find regimes where different architectures can converge to fit the training data fairly quickly.


We use cross-entropy loss to optimize the models with AdamW~\cite{loshchilov2017decoupled} in the following experiments. The following experiments are implemented in PyTorch~\cite{paszke2019}.

\begin{figure*}[h]
\centering
  \begin{tabular}{cc}
      \includegraphics[width=0.45\linewidth]{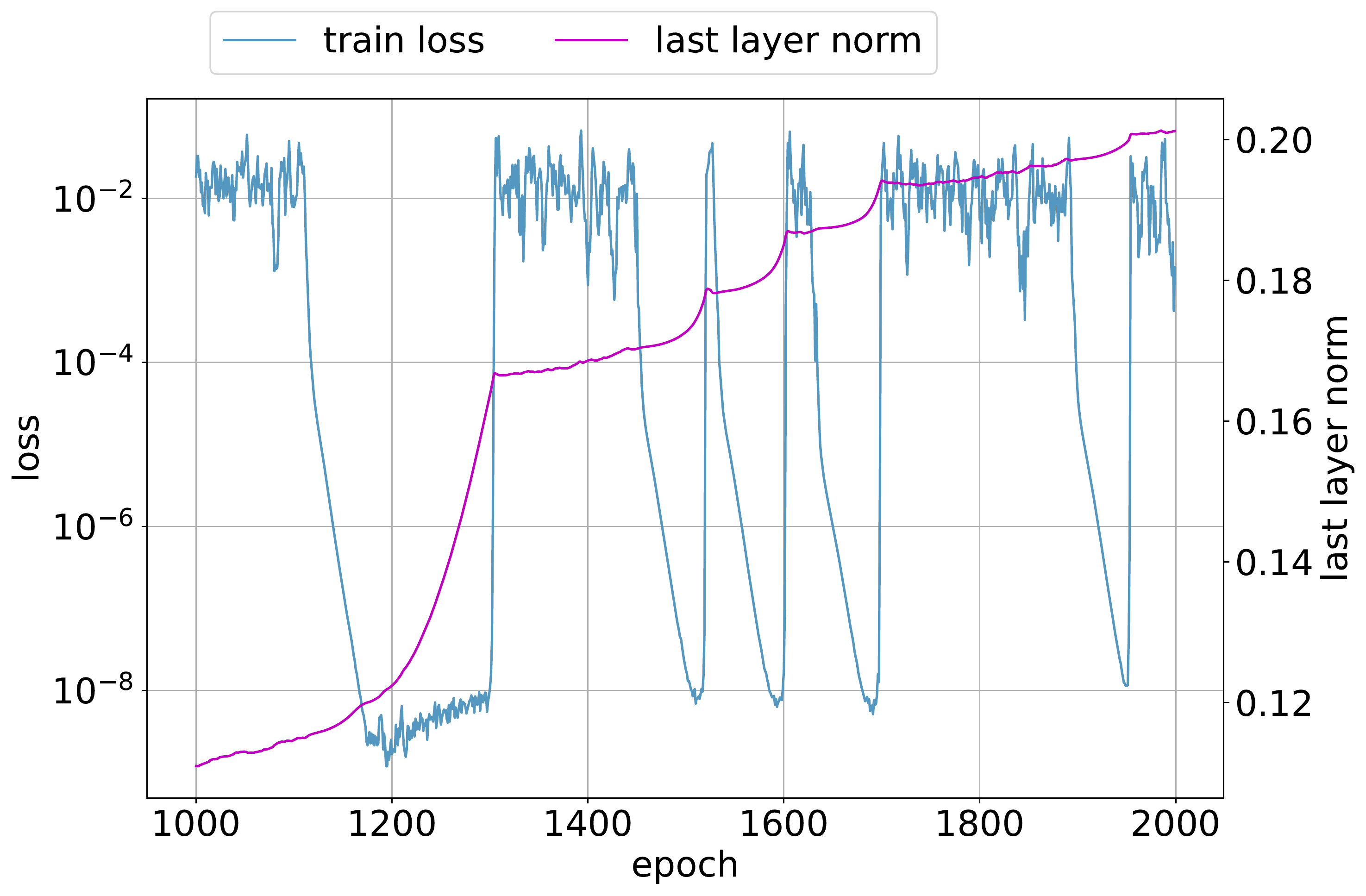} & 
      \includegraphics[width=0.45\linewidth]{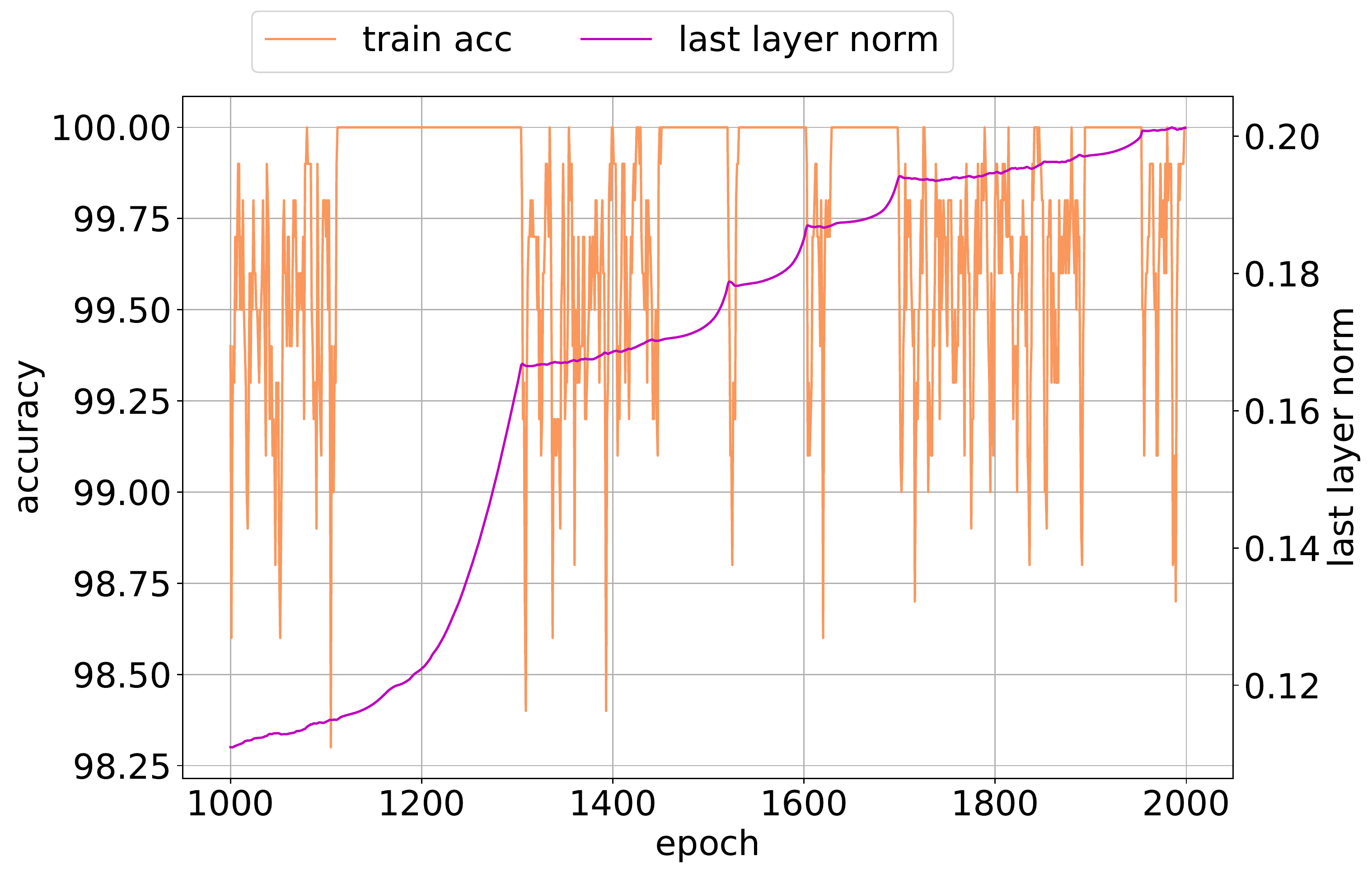} \\
      (a)  & (b) 

  \end{tabular}
 \caption{Vision Transformer on 1000 samples from CIFAR-10: Norm growth versus a) loss on training data b) accuracy on training data for a ViT trained on 1000 samples.}
 \label{fig:grok_cifar10}
\end{figure*} 

\subsubsection{Vision Transformers on 1000 samples from CIFAR-10} For further validation, we train a Vision Transformer (ViT)~\cite{dosovitskiy2020an} with 12 layers that has 10 million parameters on a small sample of the CIFAR-10 dataset~\cite{krizhevsky09learningmultiple}. In this setup, we use a learning rate to 0.001, no weight decay, $\beta_{1} = 0.9$, $\beta_{2} = 0.95$, $\epsilon=1e-08$ and minibatch size of 128. We choose a sample size of 1000 training samples for computational reasons, as we wish to observe multiple cycles of the Slingshot Mechanism extremely late in training. The input images are standardized to be in the range $[0, 1]$. No data augmentation is used in our training pipeline. Due to the extremely small sample size, we focus our attention on the training metrics since no generalization is expected. Figure~\ref{fig:grok_cifar10}a (respectively Figure~\ref{fig:grok_cifar10}b) shows a plot of training loss (respectively training accuracy) and last layer norm evolution during the latter stages of training. Multiple Slingshot stages are observed in these plots (5 clear cycles), which can be seen by the sharp transition of the weight norm from high growth to plateau.

\begin{figure*}[!h]
\centering
  \begin{tabular}{cc}
      \includegraphics[width=0.45\linewidth]{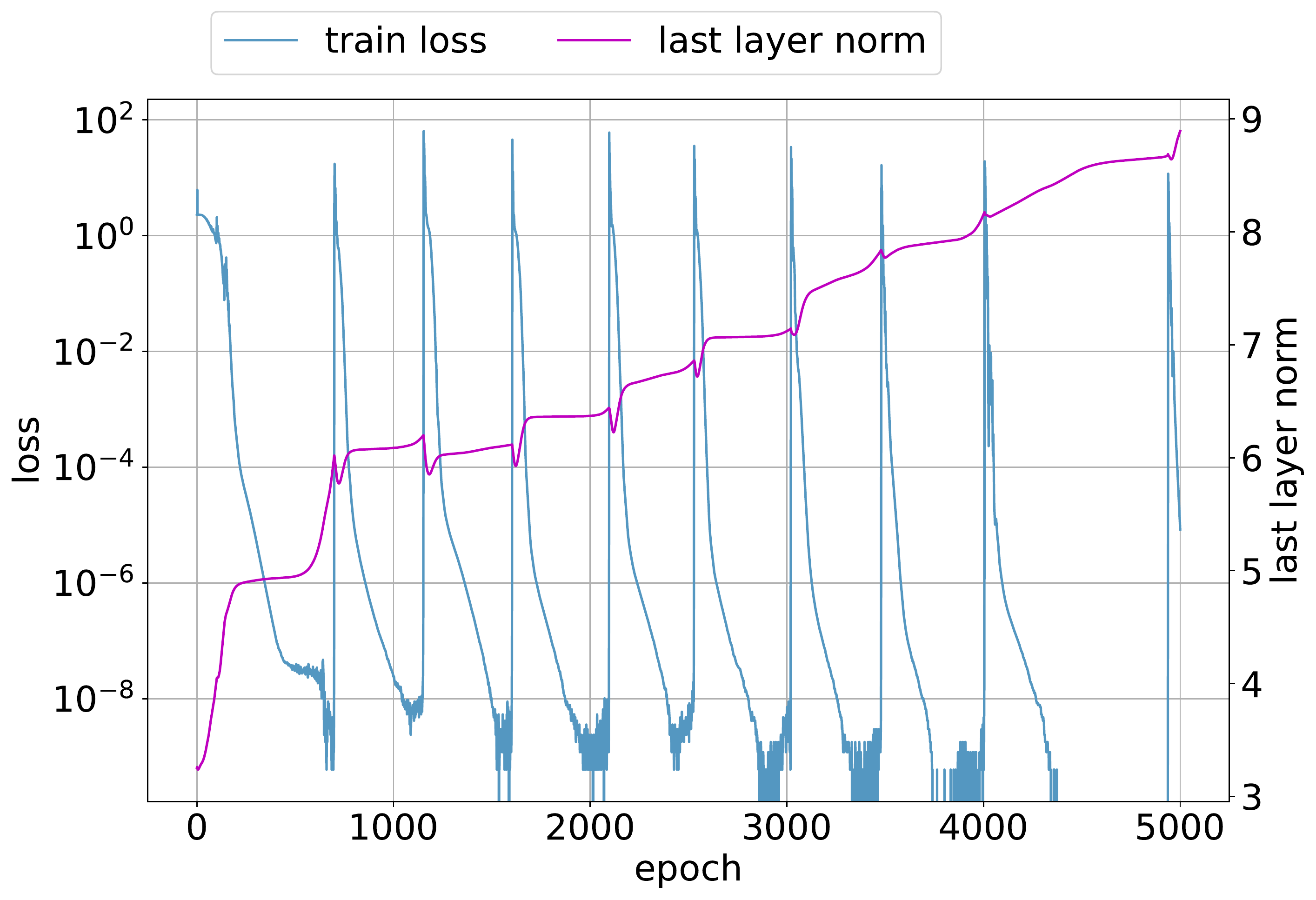} & 
      \includegraphics[width=0.45\linewidth]{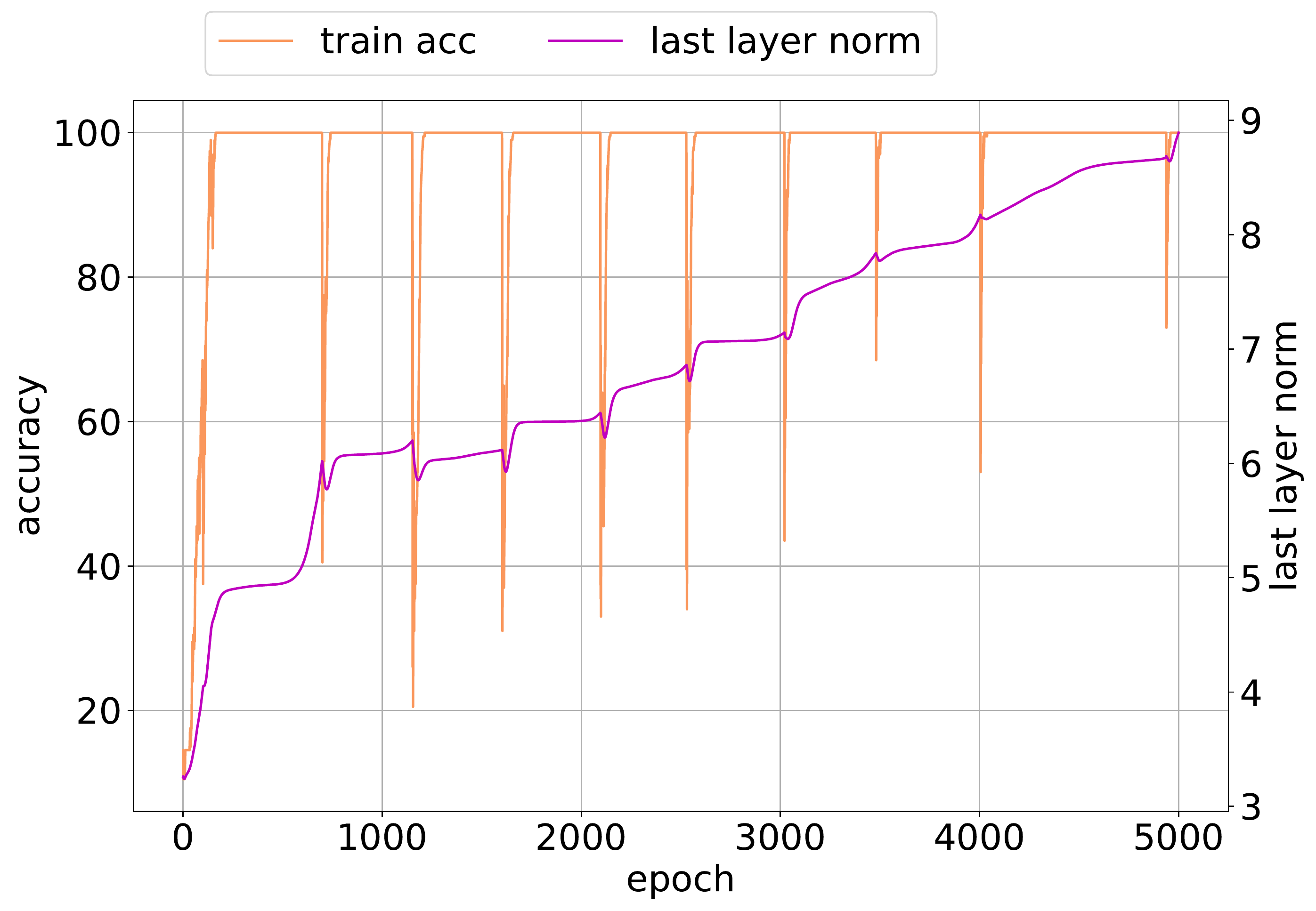} \\
  (a)  & (b) 

  \end{tabular}
 \caption{CNN on CIFAR-10 dataset: Norm growth versus a) loss on training data b) accuracy on training data for a VGG11-like model without batch normalization trained on 200 samples.}
 \label{fig:grok_vgg_cifar10}
\end{figure*}

\subsubsection{CNN on 200 samples from CIFAR-10} 
We consider a VGG-like architecture~\cite{simonyan2014deep} that has been adapted for CIFAR-10 dataset.\footnote{We use the VGG11 architecture without batch normalization~\cite{ioffe2015batch} from \url{https://github.com/kuangliu/pytorch-cifar} in this experiment.} The model is trained with 200 randomly chosen samples from CIFAR-10 training split and with full-batch AdamW~\cite{loshchilov2017decoupled}. The hyperparameters used for the optimizer include a learning rate of $0.001$, weight decay$=0$, $\beta_{1} = 0.9$, $\beta_{2} = 0.95$, and $\epsilon=1e-08$. As with ViT, no data augmentation is used in these experiments other than standardizing the input to be in the range $[0, 1]$. We observe the prescenece of multiple Slingshot stages with CNN from Figure~\ref{fig:grok_vgg_cifar10}a and Figure~\ref{fig:grok_vgg_cifar10}b. These experiments suggest that Slingshot Effects are not restricted to Transformers architecture alone.

\begin{figure*}[!h]
\centering
  \begin{tabular}{cc}
      \includegraphics[width=0.45\linewidth]{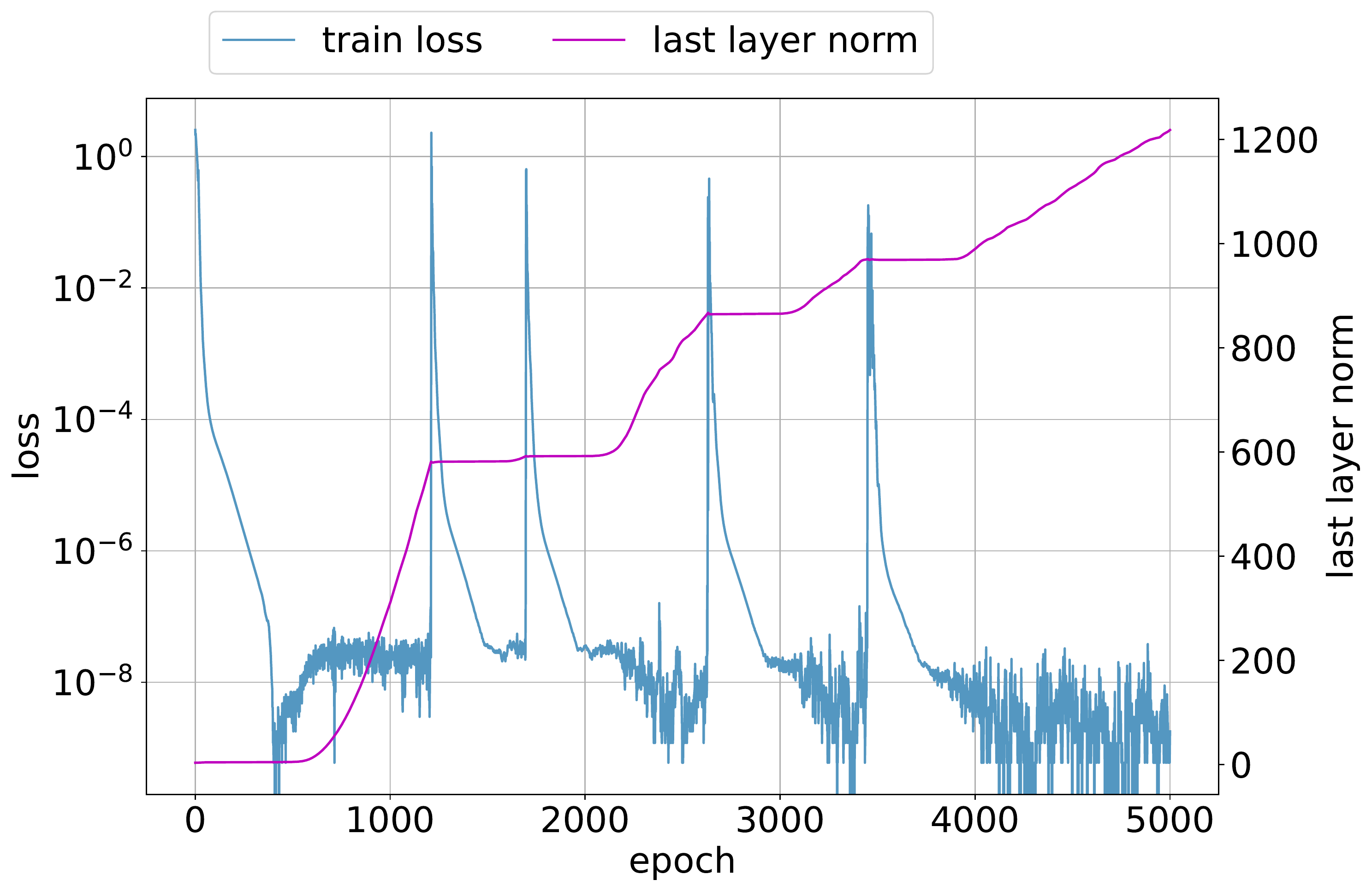} & 
      \includegraphics[width=0.45\linewidth]{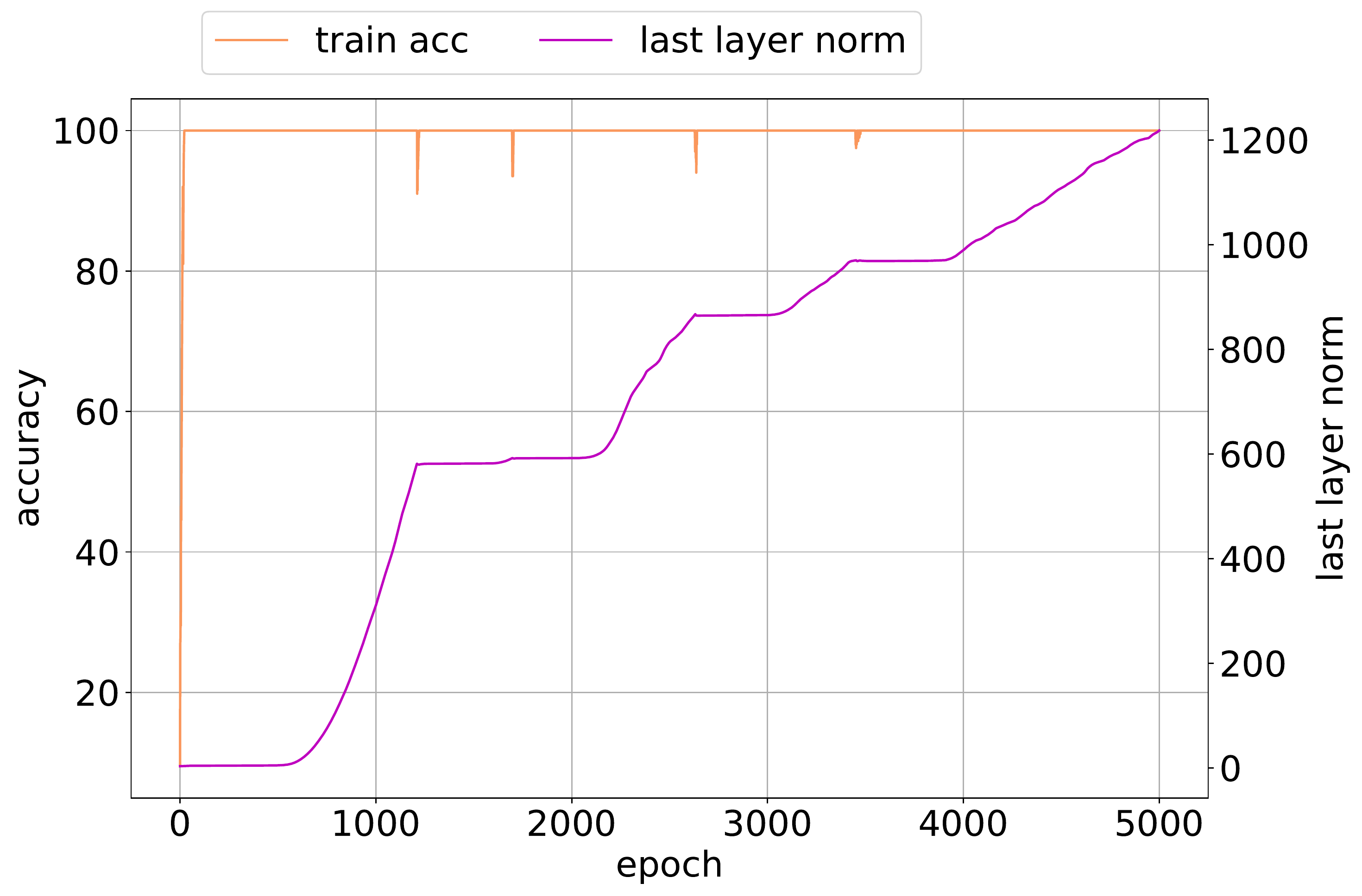} \\
  (a)  & (b) 

  \end{tabular}
 \caption{CNN on 200 samples from CIFAR-10: Norm growth versus a) loss on training data b) accuracy on training data for a VGG11-like model without batch normalization trained on 200 samples.}
 \label{fig:grok_vgg_bn_cifar10}
\end{figure*}

\paragraph{With BatchNorm} We repeat the CNN-based described above but with a VGG-like model that includes batch normalization~\cite{ioffe2015batch}.\footnote{We use the VGG11 architecture with batch normalization~\cite{ioffe2015batch} from \url{https://github.com/kuangliu/pytorch-cifar} in this experiment.} The training setup is identical to the one described for CNN wihtout batch normalization. We observe the prescenece of multiple Slingshot stages with CNN from Figure~\ref{fig:grok_vgg_bn_cifar10}a and Figure~\ref{fig:grok_vgg_bn_cifar10}b. The weight norm does not decrease during training as opposed to the weight norm dynamics for CNN wihtout batch normalization seen in Figure~\ref{fig:grok_vgg_cifar10}. These experiments suggest that Slingshot Effects can be seen with standard neural network training components including batch normalization.

\begin{figure*}[!h]
\centering
  \begin{tabular}{cc}
      \includegraphics[width=0.45\linewidth]{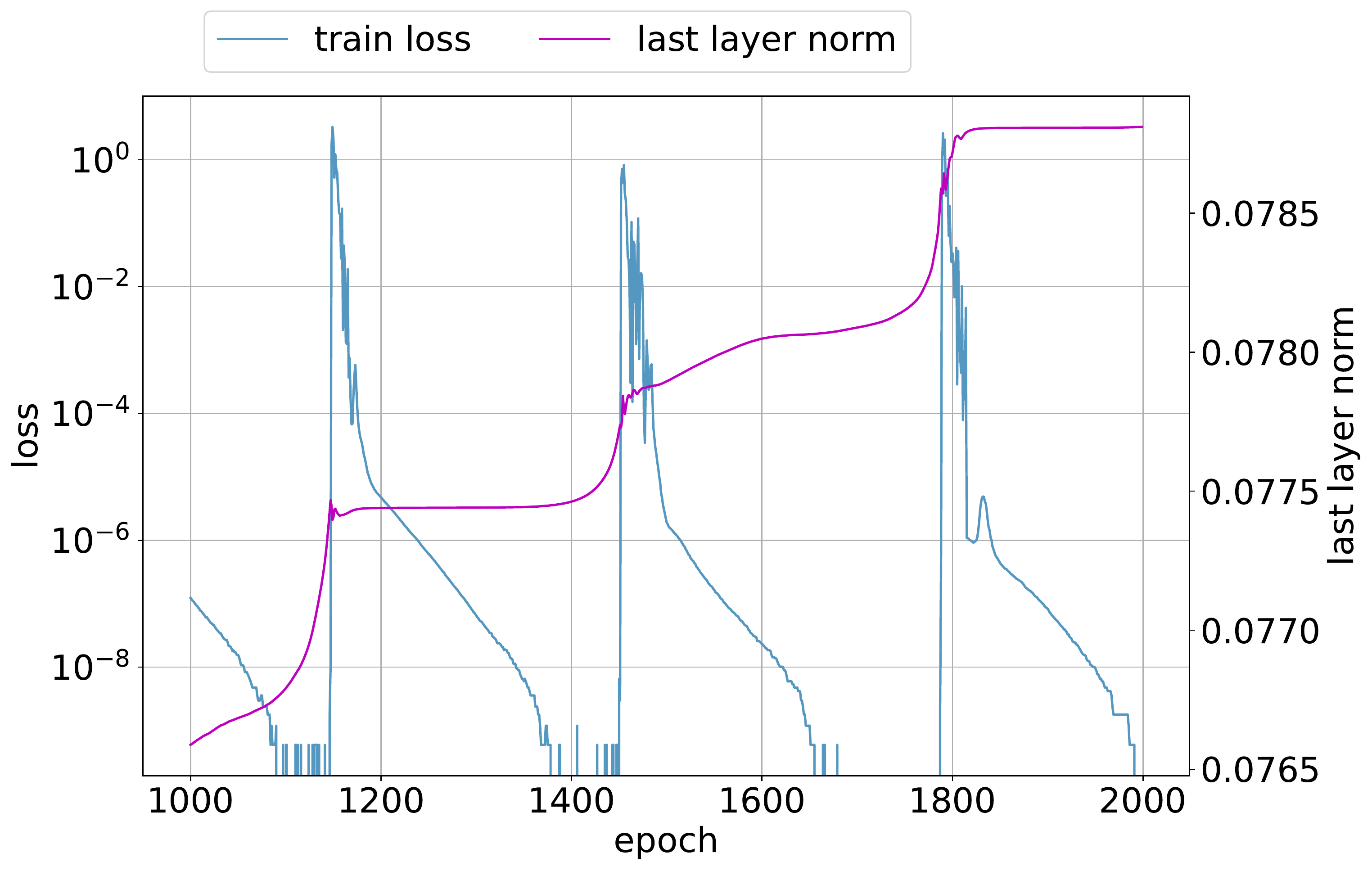} & 
      \includegraphics[width=0.45\linewidth]{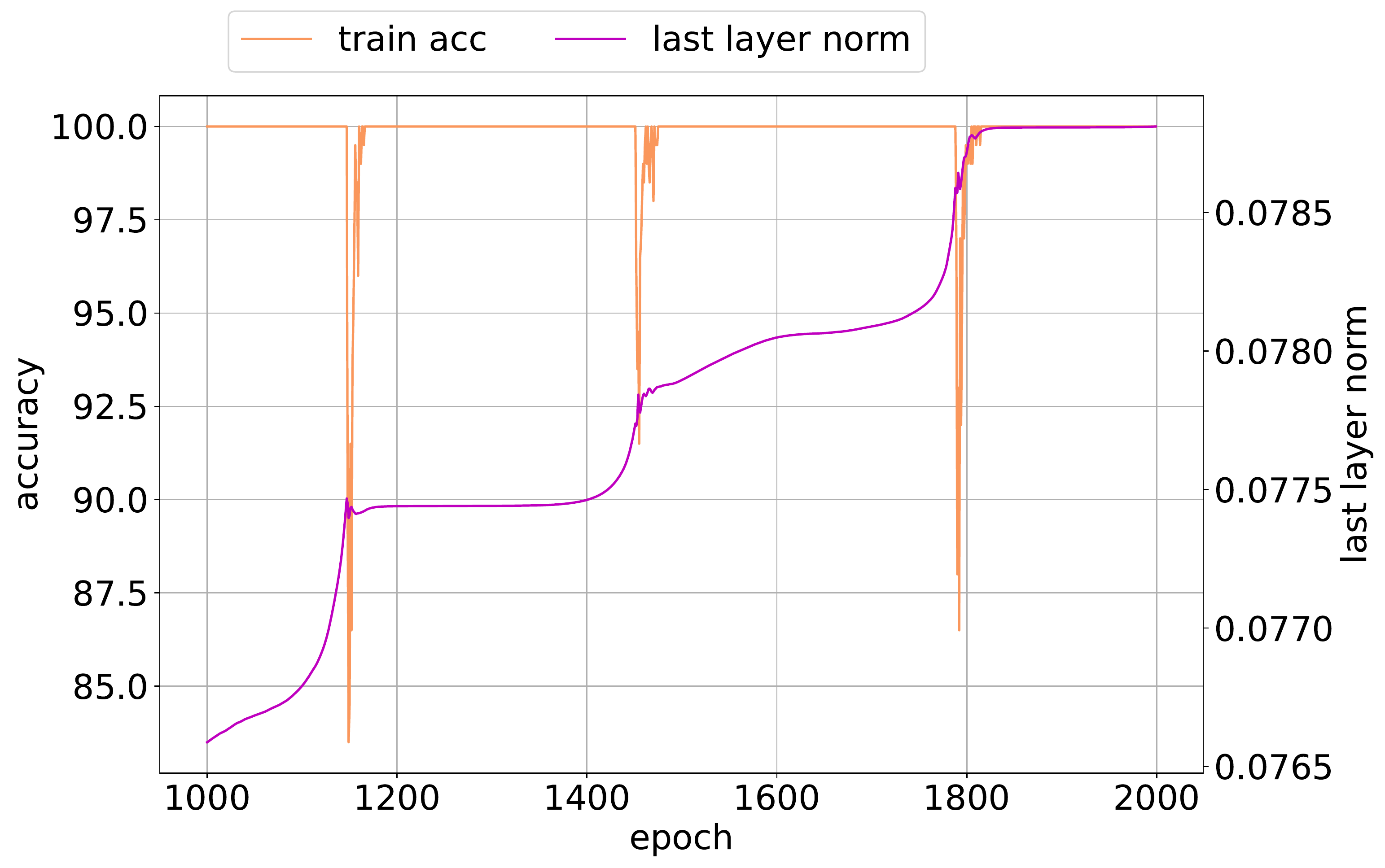} \\
  (a)  & (b) 

  \end{tabular}
 \caption{MLP on 200 samples from CIFAR-10: Norm growth versus a) loss on training data b) accuracy on training data for a model trained on 200 samples.}
 \label{fig:grok_mlp_cifar10}
\end{figure*}

\subsubsection{MLPs on 200 samples from CIFAR-10} The next architecture we consider is a deep (6 layers) fully connected network trained on a small sample of 200 samples belonging to the CIFAR-10 dataset~\cite{krizhevsky09learningmultiple} with full-batch AdamW~\cite{loshchilov2017decoupled} optimizer. The optimizer's hyperparameters are set as following: learning rate $=0.001$, weight decay $=0$, $\beta_{1} = 0.9$, $\beta_{2} = 0.95$, and $\epsilon=1e-08$. As with the ViT setup above we do no use data augmentation for training this model. Figure~\ref{fig:grok_mlp_cifar10}a (respectively Figure~\ref{fig:grok_mlp_cifar10}ab) shows a plot of training loss (respectively training accuracy) and last layer norm evolution during the latter stages of training. Multiple Slingshot stages are observed in this setup as well. These experiments further suggest that the Slingshot mechanism is prevalent in simple models as well.

\begin{figure*}
\centering
  \begin{tabular}{ccc}
      \includegraphics[width=0.32\linewidth]{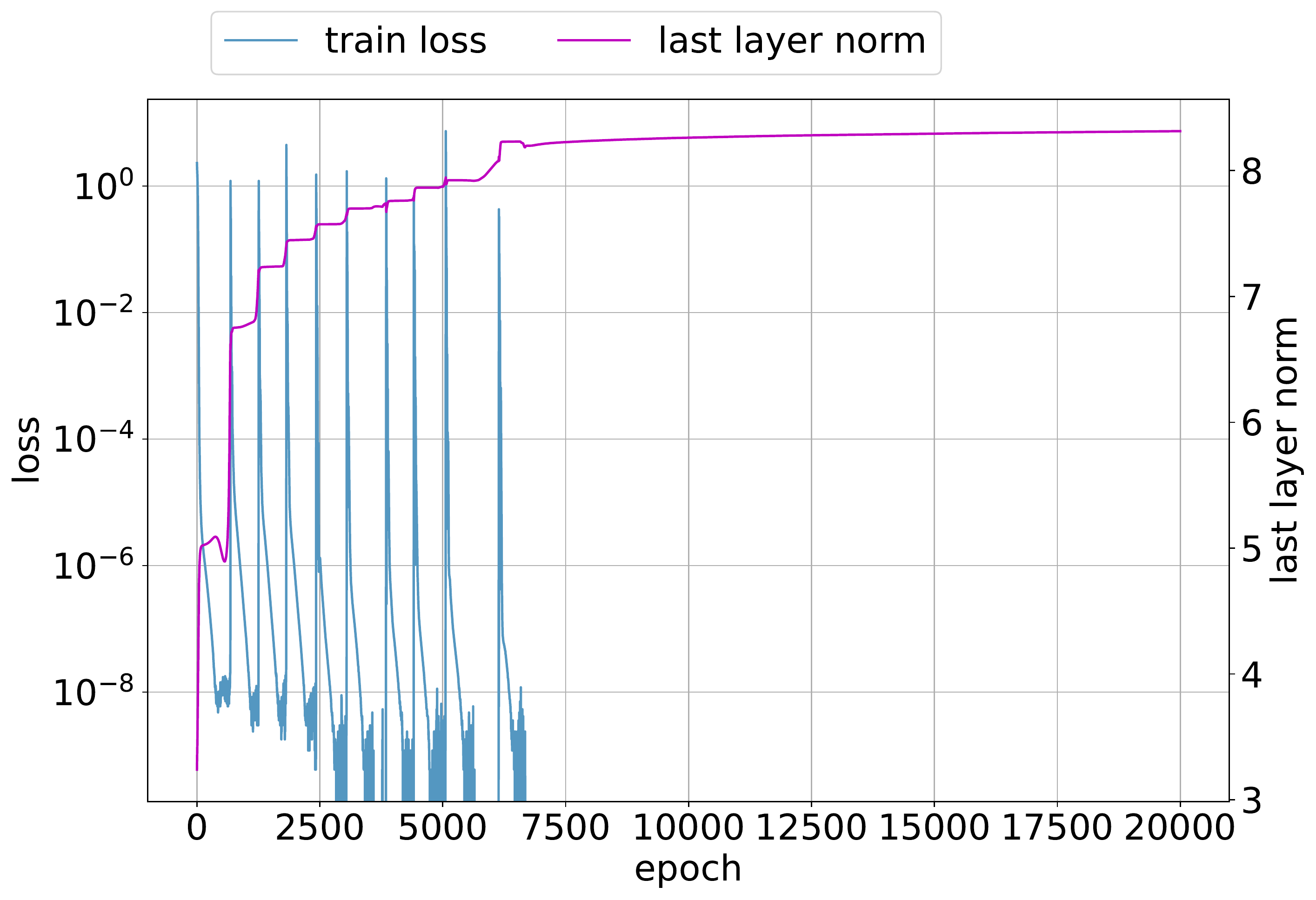} & 
      \includegraphics[width=0.32\linewidth]{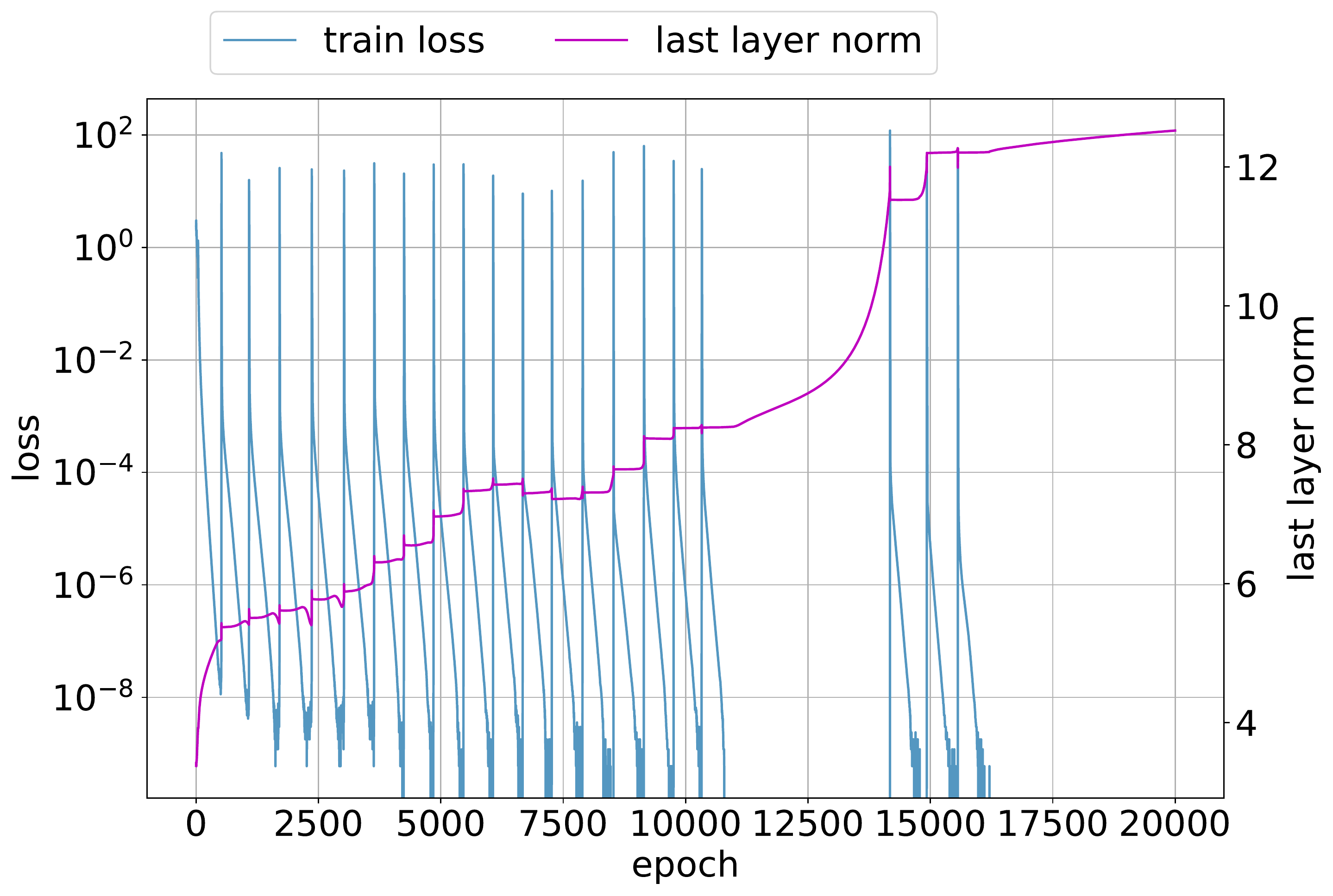} &
      \includegraphics[width=0.32\linewidth]{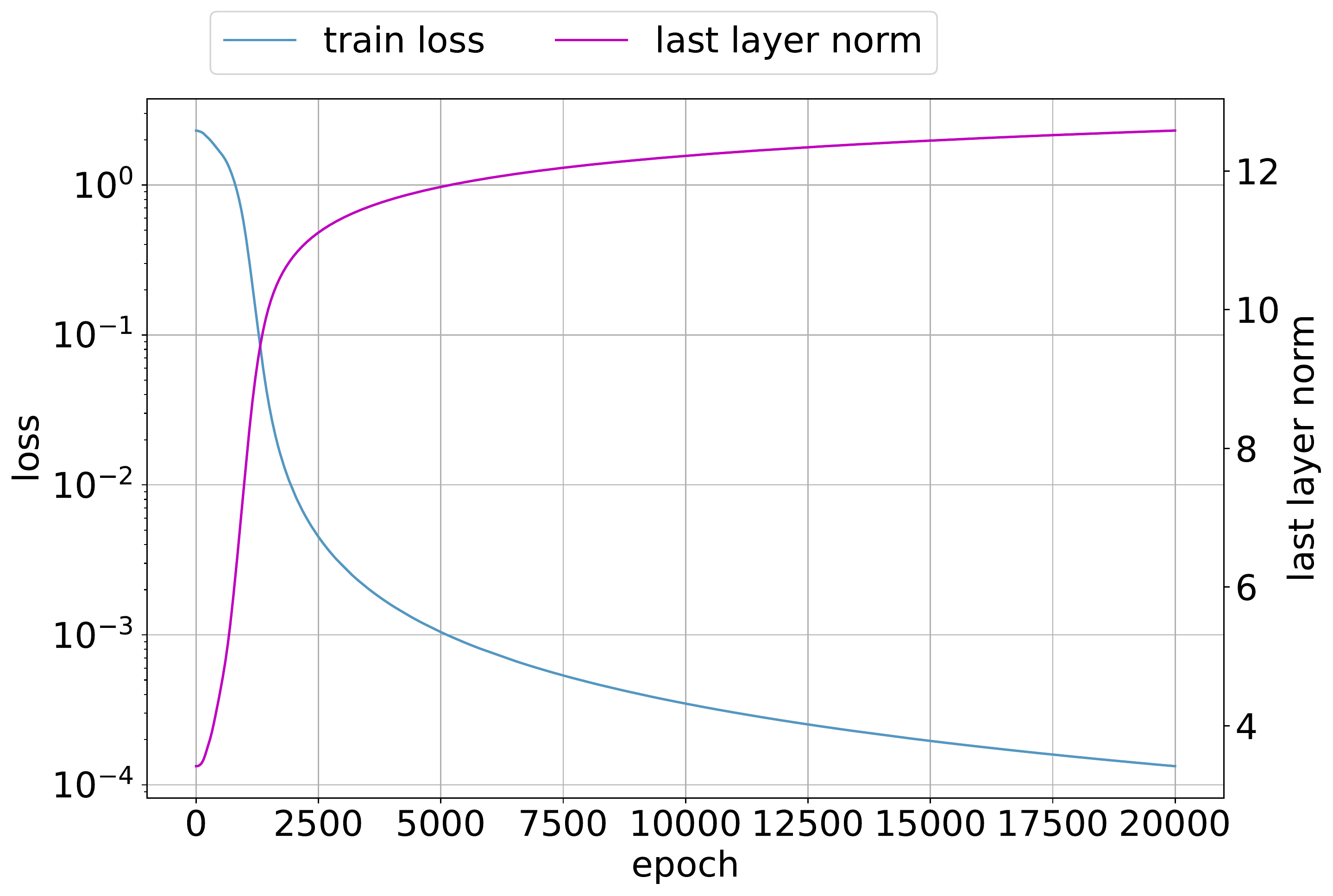} \\
      (a)  & (c) & (e) \\
     \includegraphics[width=0.32\linewidth]{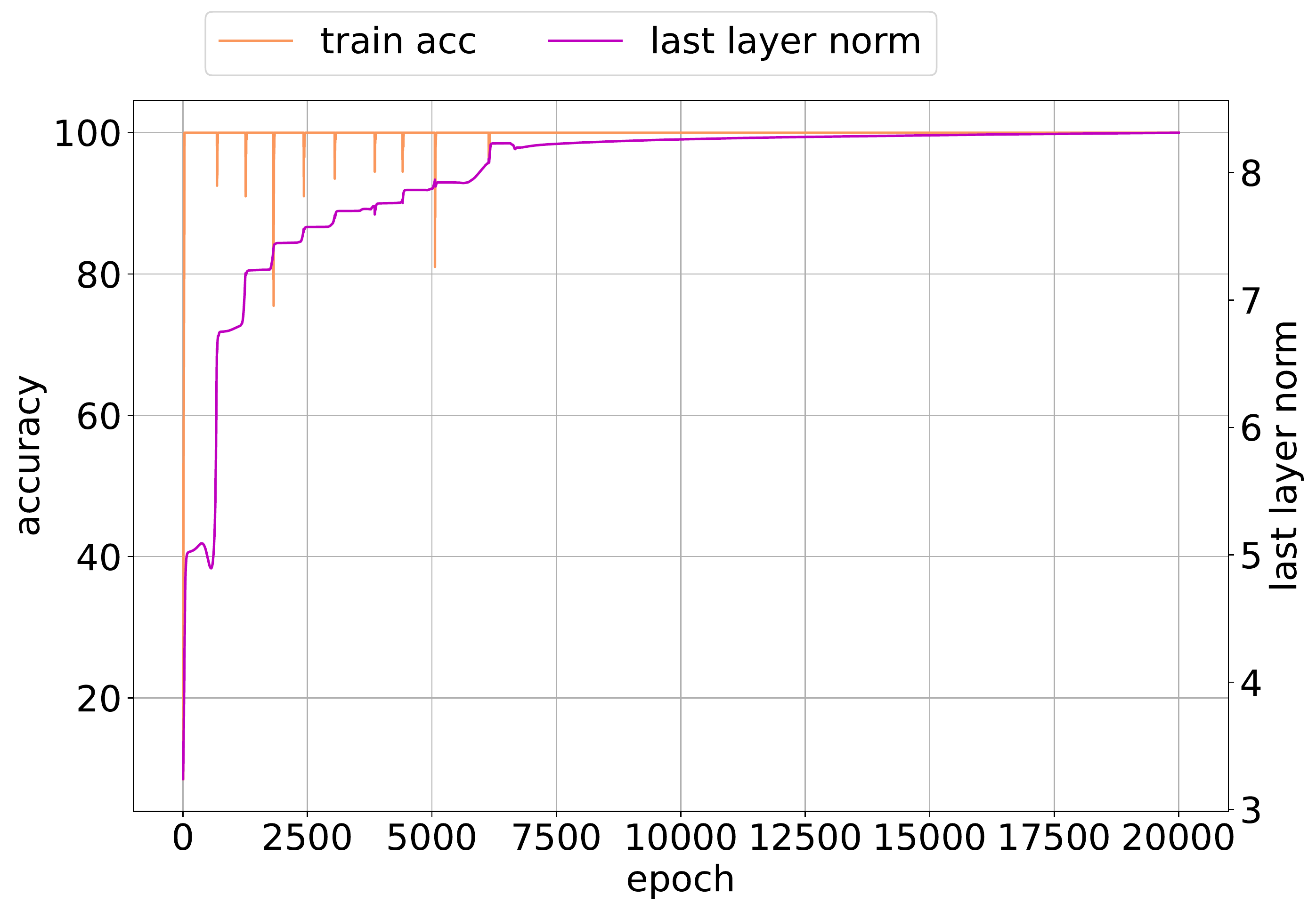} & 
      \includegraphics[width=0.32\linewidth]{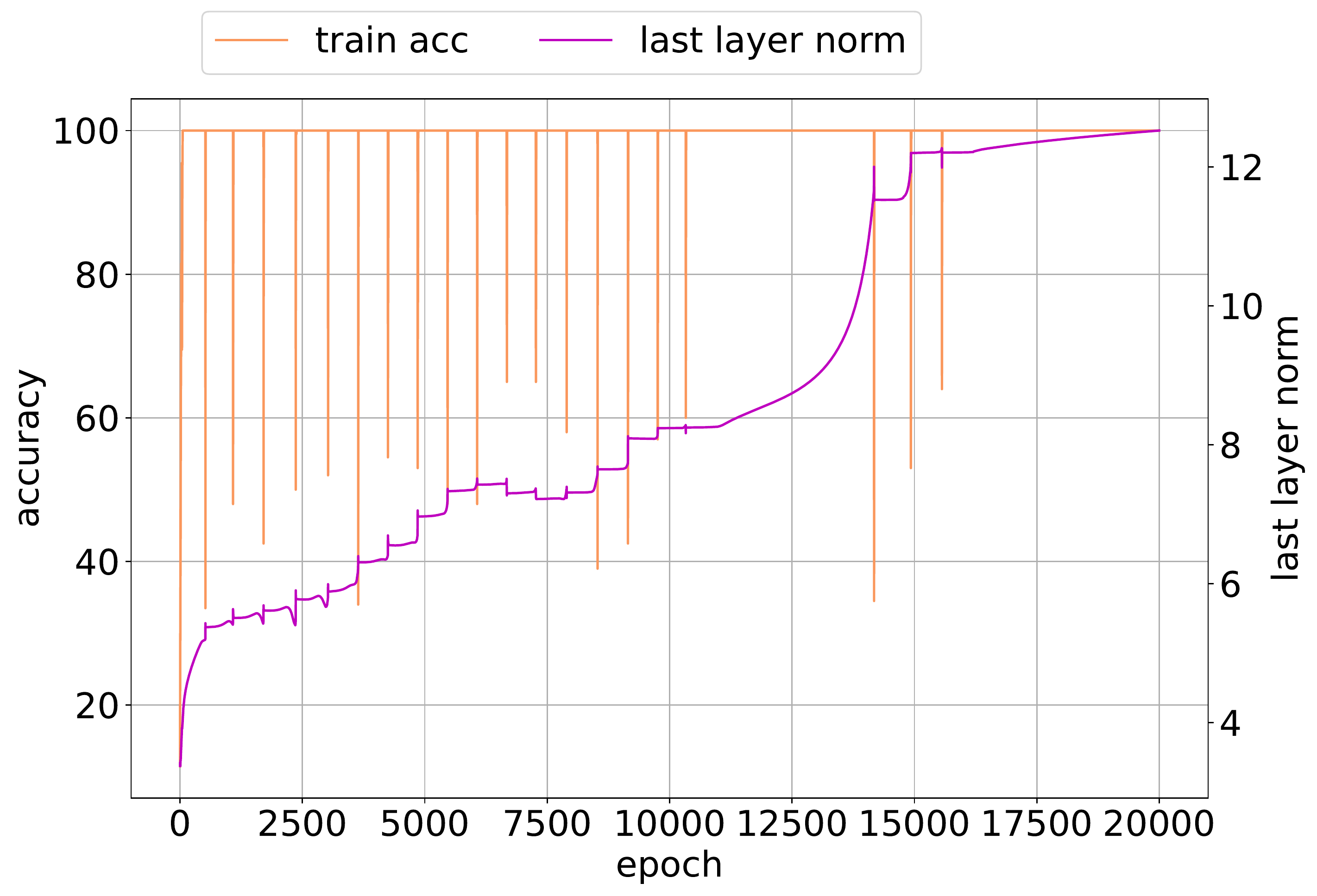} &
      \includegraphics[width=0.32\linewidth]{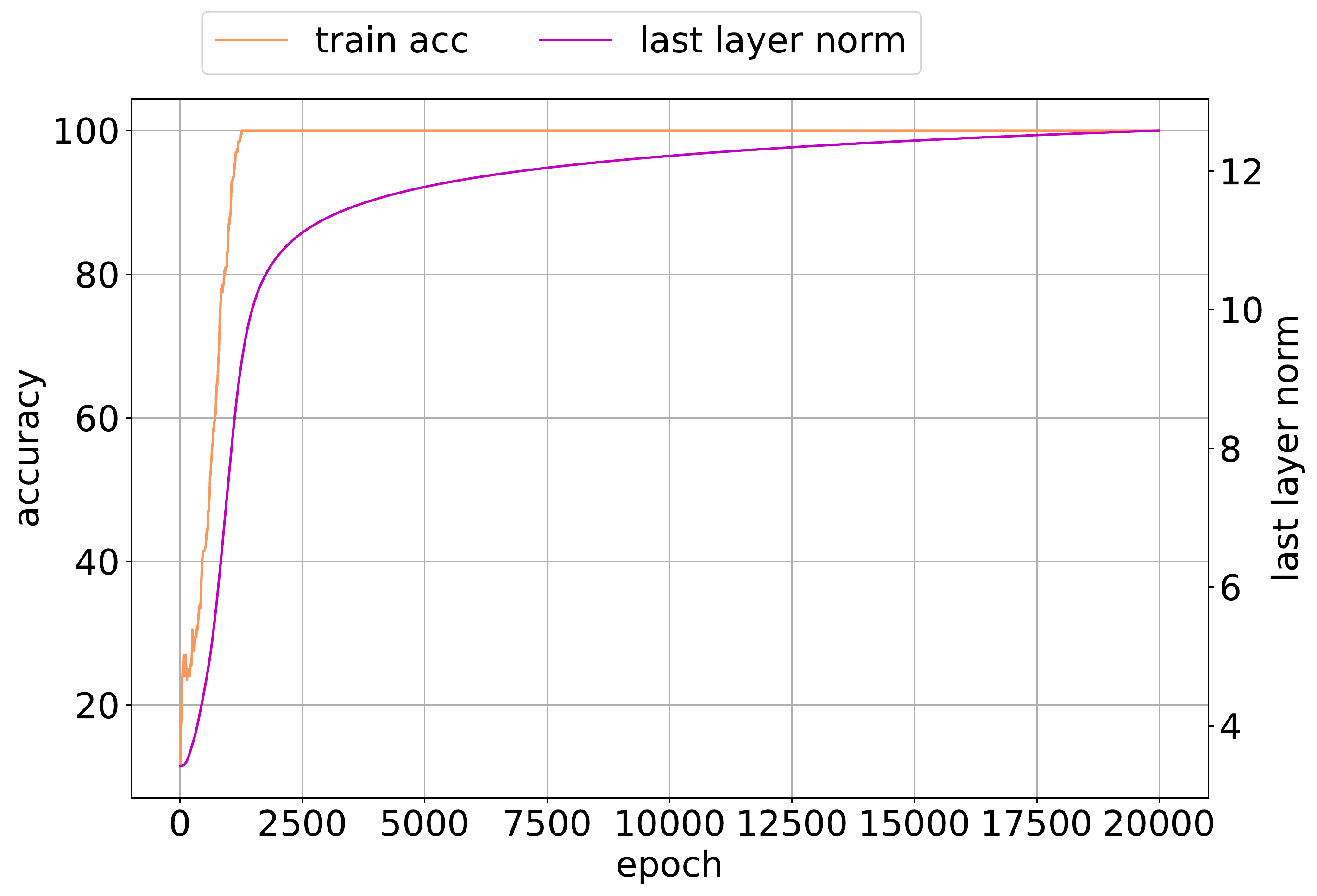} \\
      (b)  & (d) & (f) 

  \end{tabular}
 \caption{Optimizer choice on deep linear models on 200 samples from CIFAR-10: (a) Training loss for AdamW (b) Training accuracy for AdamW (c) Training loss for RMSProp (d) Training accuracy for RMSProp (e) Training loss for Gradient Descent and (f) Training accuracy for Gradient Descent. All optimizers train a 6-layer linear model full-batch on 200 CIFAR-10 samples.} 
 \label{fig:grok_optim_cifar10}
\end{figure*}

\subsubsection{Deep linear models} We train a 6 layer linear model with 200 samples belonging to CIFAR-10~\cite{krizhevsky09learningmultiple} with full-batch AdamW~\cite{loshchilov2017decoupled}.  The optimizer's hyperparameters are set as following: learning rate $=0.001$, weight decay $=0$, $\beta_{1} = 0.9$, $\beta_{2} = 0.95$, and $\epsilon=1e-08$. Figure~\ref{fig:grok_optim_cifar10}a and Figure~\ref{fig:grok_optim_cifar10}b show the training loss and accuracy behavior observed during optimization. Multiple Slingshot stages are observed with this architecture as well.

\subsubsection{Different Optimizers} In this set of experiments, we study the training loss behavior of deep linear models optimized full-batch with AdamW~\cite{loshchilov2017decoupled}, RMSProp~\cite{tieleman2012lecture} and full-batch gradient descent (GD). The six layer model is trained with 200 samples. The hyperparameters used for optimizing the model with various optimizers are described in Table~\ref{table:optim_hparams}. Figure~\ref{fig:grok_optim_cifar10} shows the training loss and accuracy behavior of the three optimizers considered in this experiment. We observe Slingshot behavior with AdamW and RMSProp from Figure~\ref{fig:grok_optim_cifar10} while Slingshot behavior is absent with standard gradient descent. This observation suggests that the normalization used in adaptive optimizers to calculate the update from gradients may lead to Slingshot behavior.


\begin{table}
  \caption{Optimizers hyperparameters. Learning rate is set to $0.001$ and weight decay to $0$ for all optimizers}
  \label{table:optim_hparams}
  \centering
  \begin{tabular}{cc}
    \toprule
    Optimizer & Other hyperparameters \\
    \midrule
    Adam      & $\beta_1=0.9, \beta_2=0.95$ \\
    RMSProp   & $\alpha=0.95$, momentum=$0.0$  \\
    GD       & momentum=$0.9$  \\
    \bottomrule
  \end{tabular}
\end{table}

\subsection{Slingshot with MLP and Synthetic Dataset}\label{appendix:synth_data}

\begin{figure*}[h]
\centering
  \begin{tabular}{ccc}
    \includegraphics[width=0.33\linewidth]{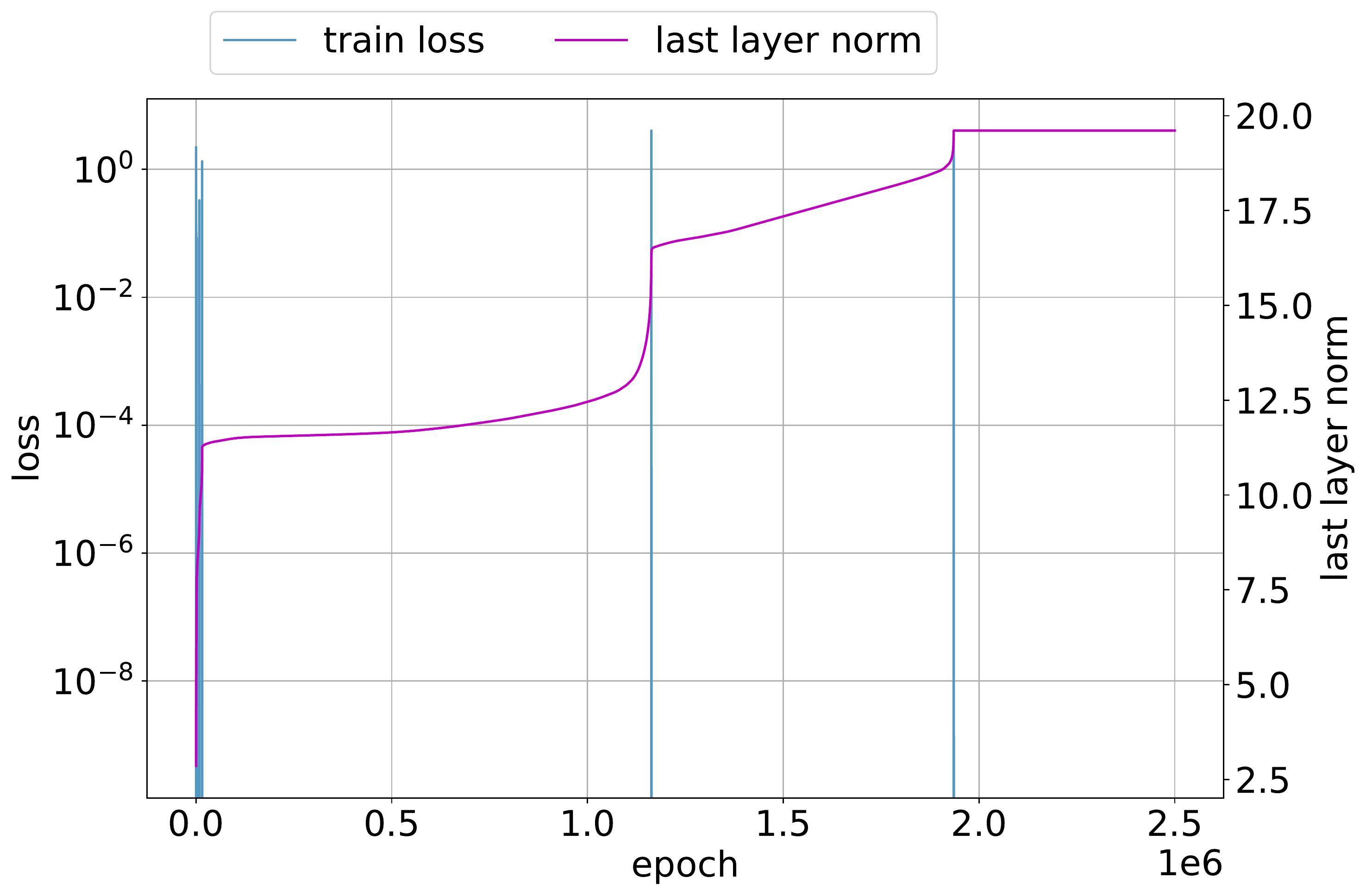} & 
    \includegraphics[width=0.33\linewidth]{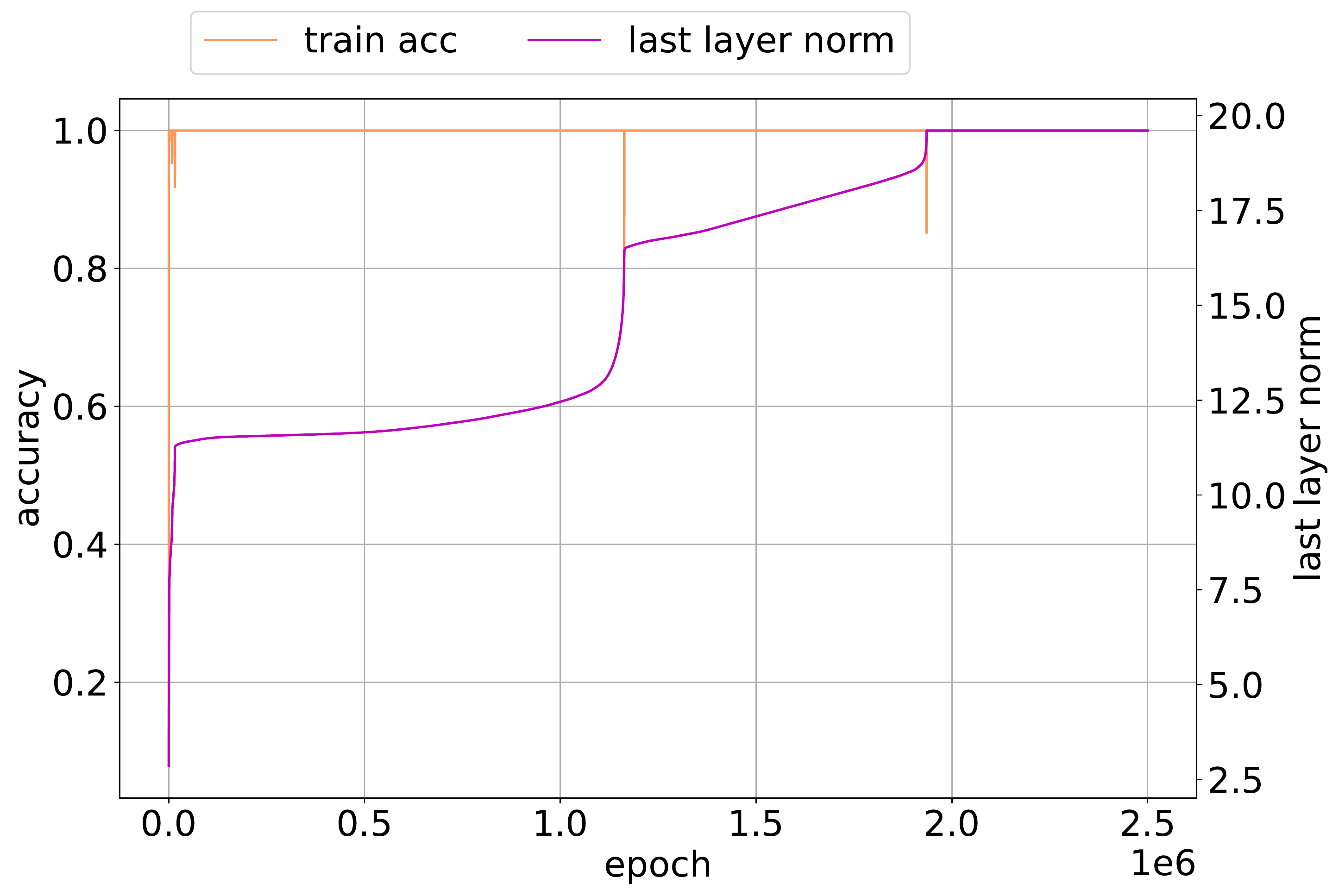} &
    \includegraphics[width=0.33\linewidth]{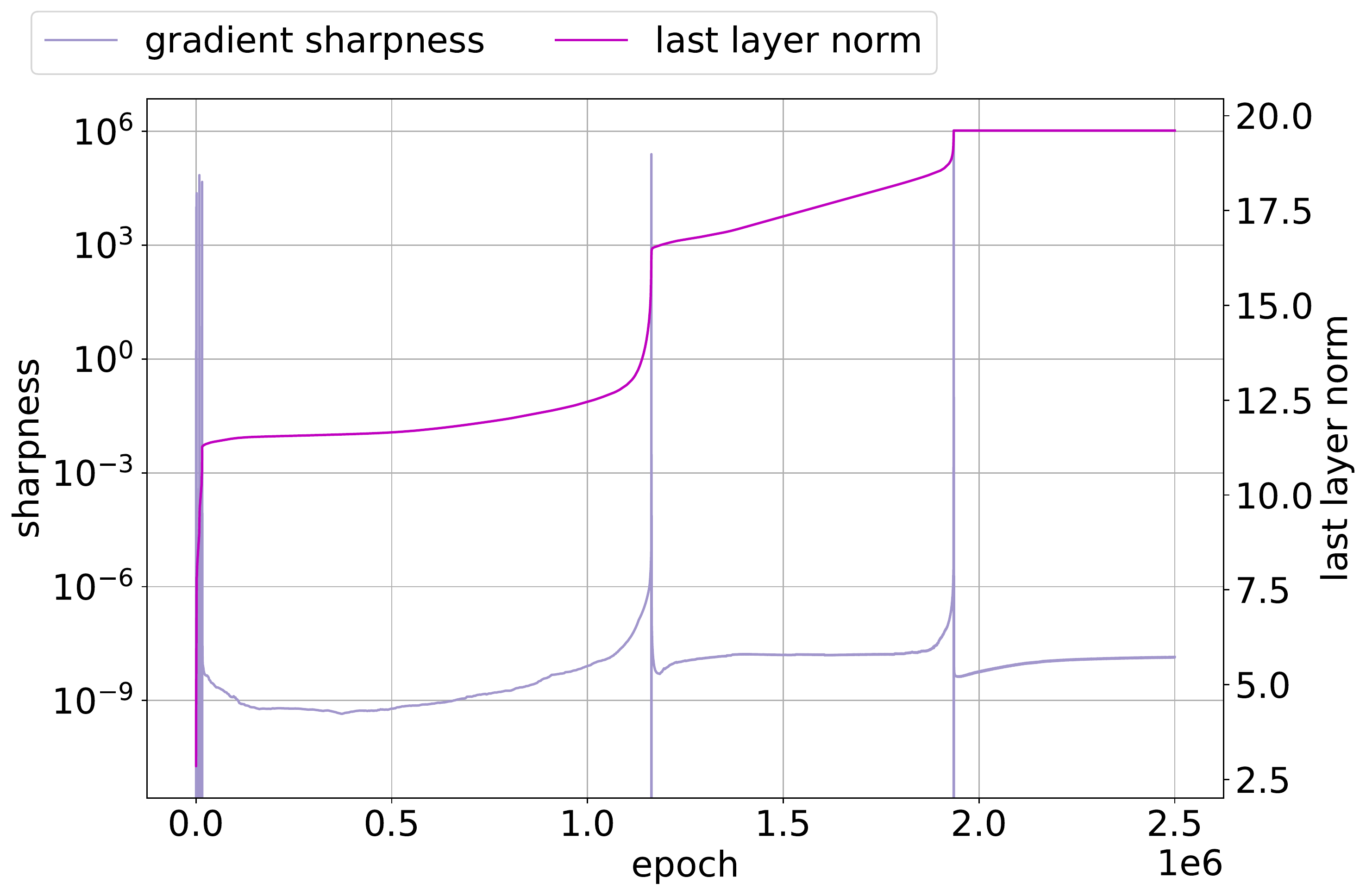} \\
      (a)  & (c) & (e) \\
     \includegraphics[width=0.33\linewidth]{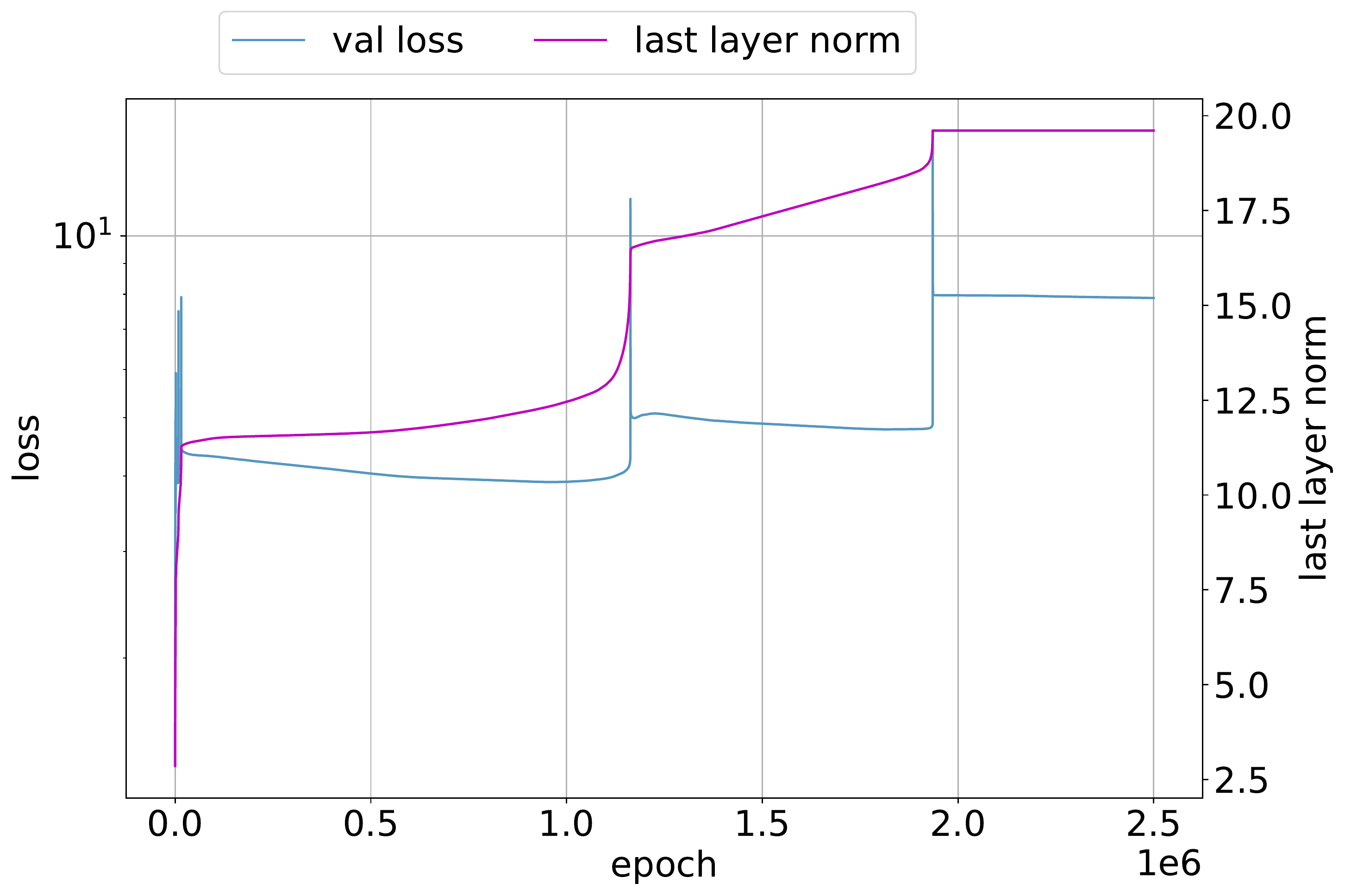} &
      \includegraphics[width=0.33\linewidth]{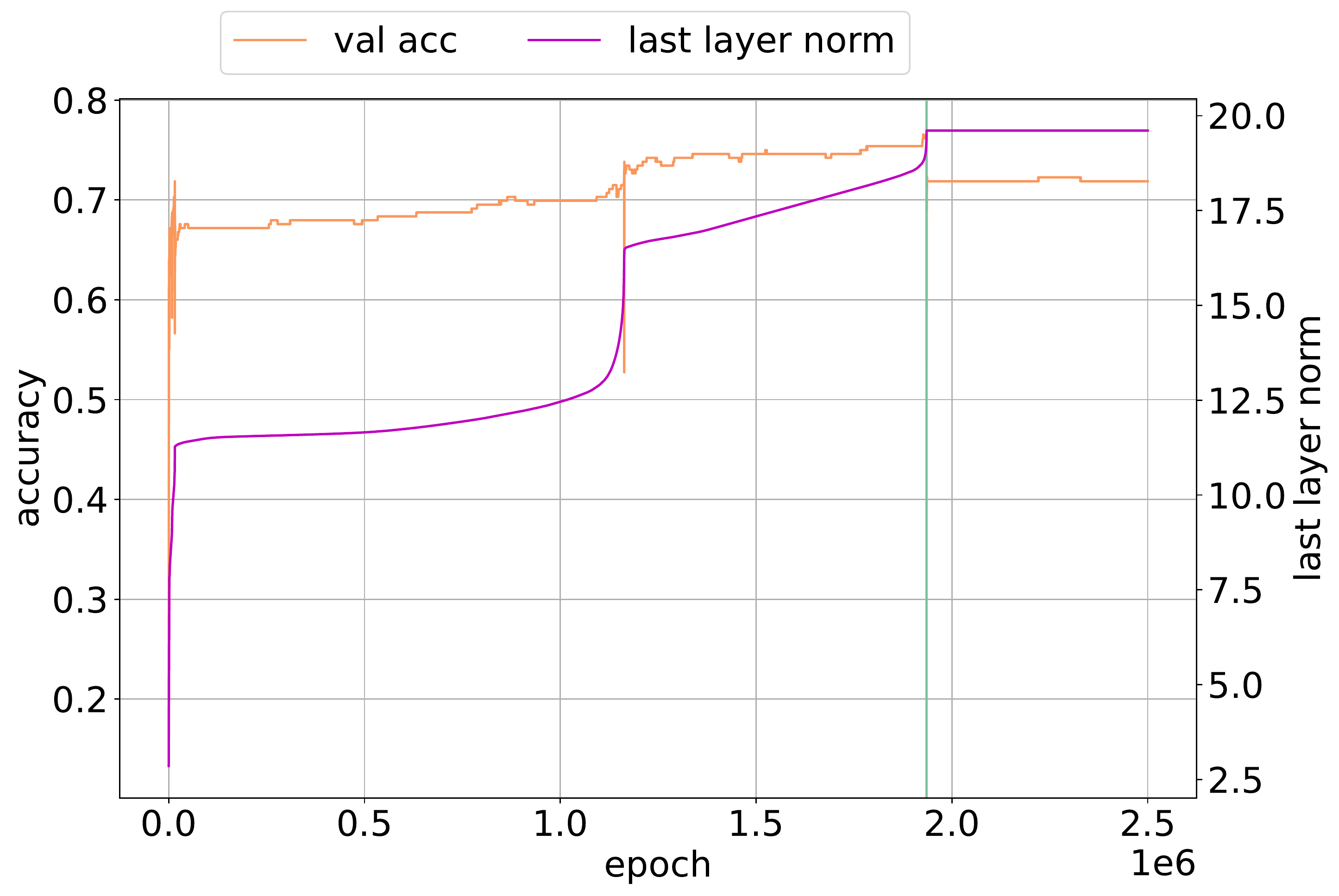} &
      \includegraphics[width=0.33\linewidth]{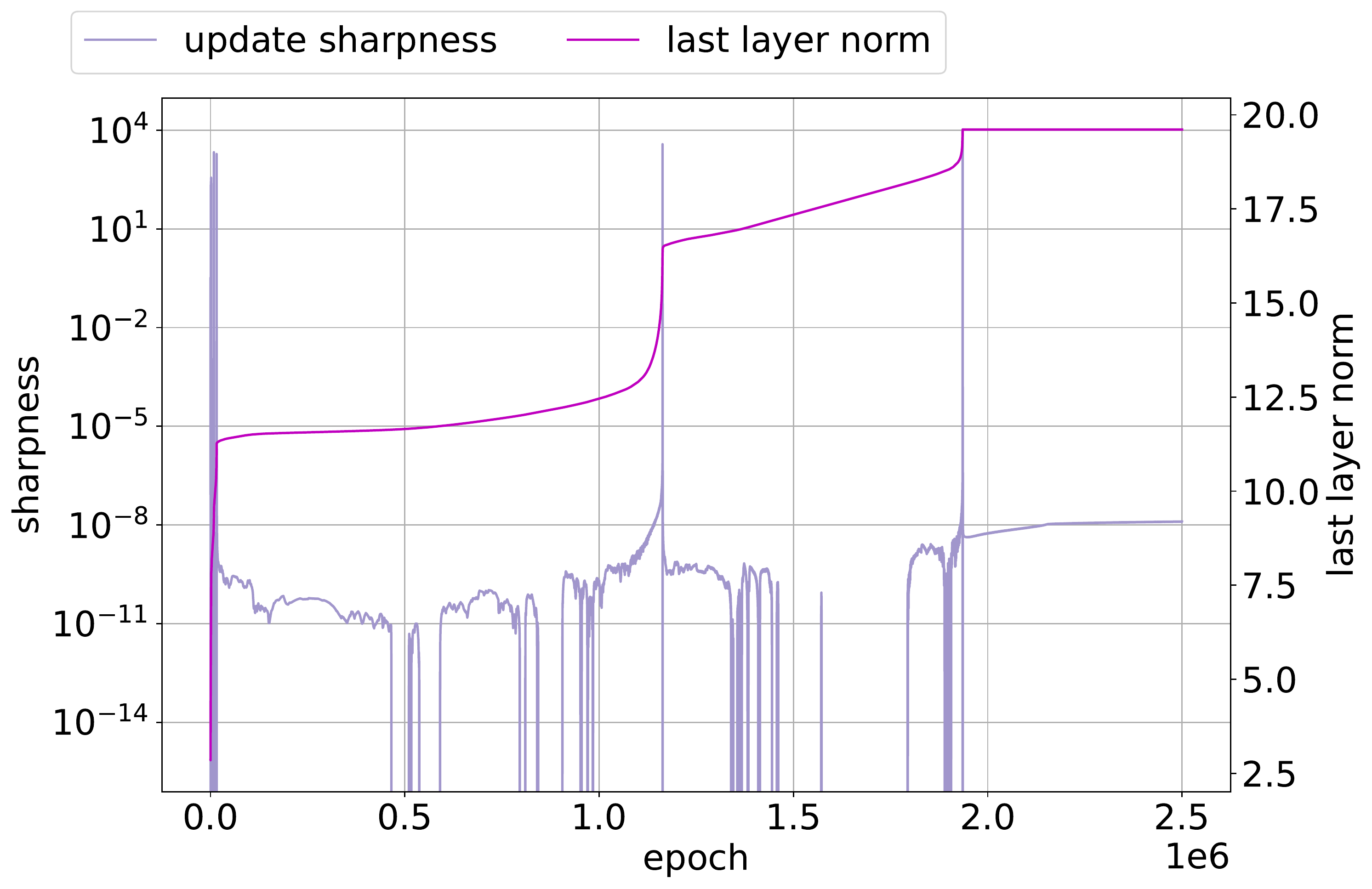} \\
      (b)  & (d) & (f)

  \end{tabular}
 \caption{Slingshot generalization on synthetic dataset: Norm growth versus a) loss on training data b) accuracy on training data (c) loss on validation data d) accuracy on validation data. Note that the vertical line in green shows location of maximum test accuracy. Adam hyperparameters are $\beta_{1}=0.9$, $\beta_{1}=0.95$, $\epsilon=1e^{-08}$} 
 \label{fig:gen_synth_eps8}
\end{figure*}


In this section, we provide empirical evidence that Slingshot Effects are observed with a synthetic dataset in a fully-connected architecture. The small dimensional dataset, like the Grokking dataset of Power et al.~\cite{power2021grokking}, allows us to easily measure of sharpness, given by $\frac{1}{\|u_t\|^2}u_t^\top \mathcal{H}_t u_t$ where $u_{t}$ is the optimizer's update vector and $\mathcal{H}_t$ is the Hessian at step $t$, to examine the interplay between Slingshot Effects and generalization.

\begin{figure*}[h!]
\centering
  \begin{tabular}{cc}
    \includegraphics[width=0.4\linewidth]{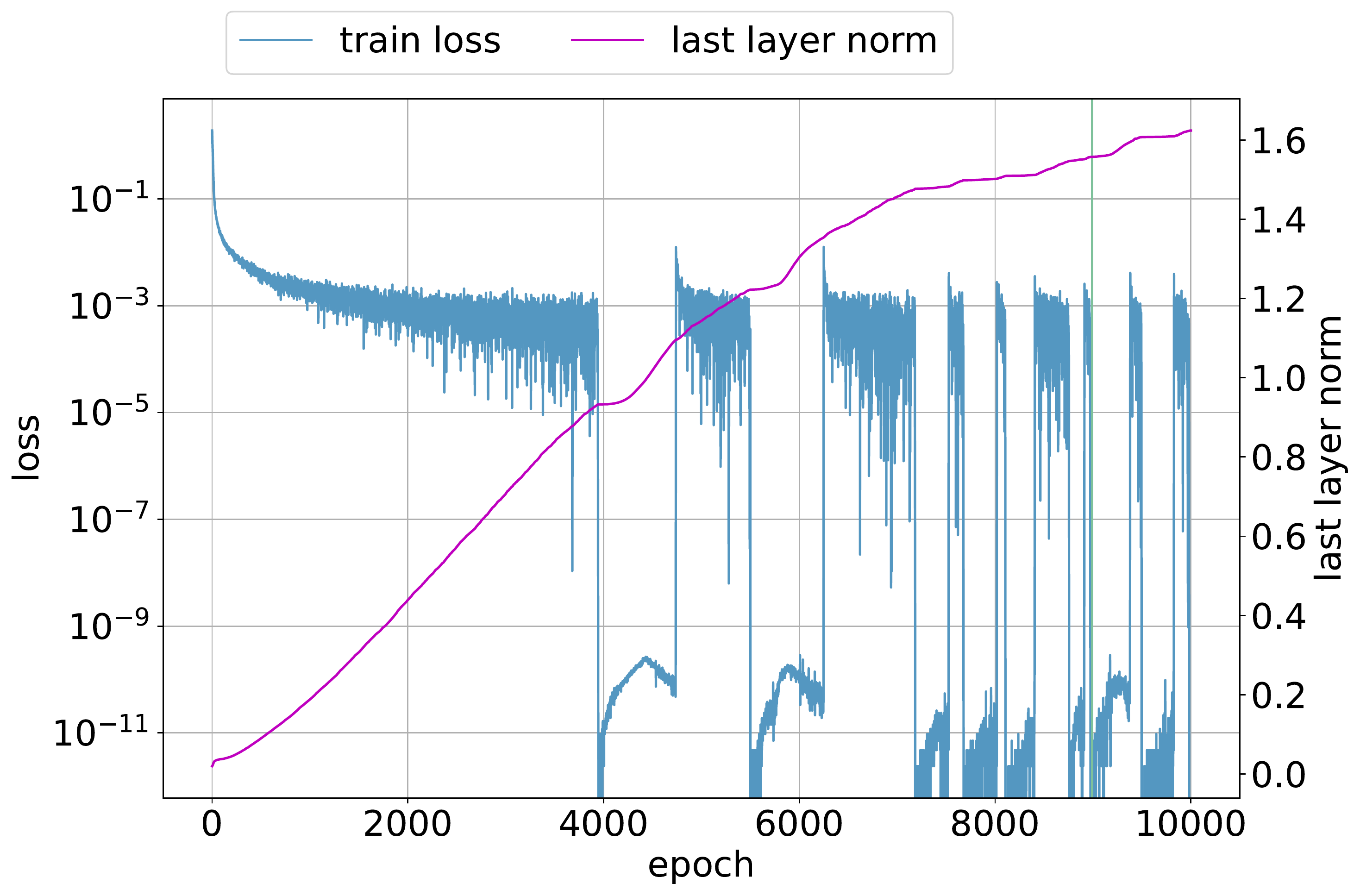} & 
    \includegraphics[width=0.4\linewidth]{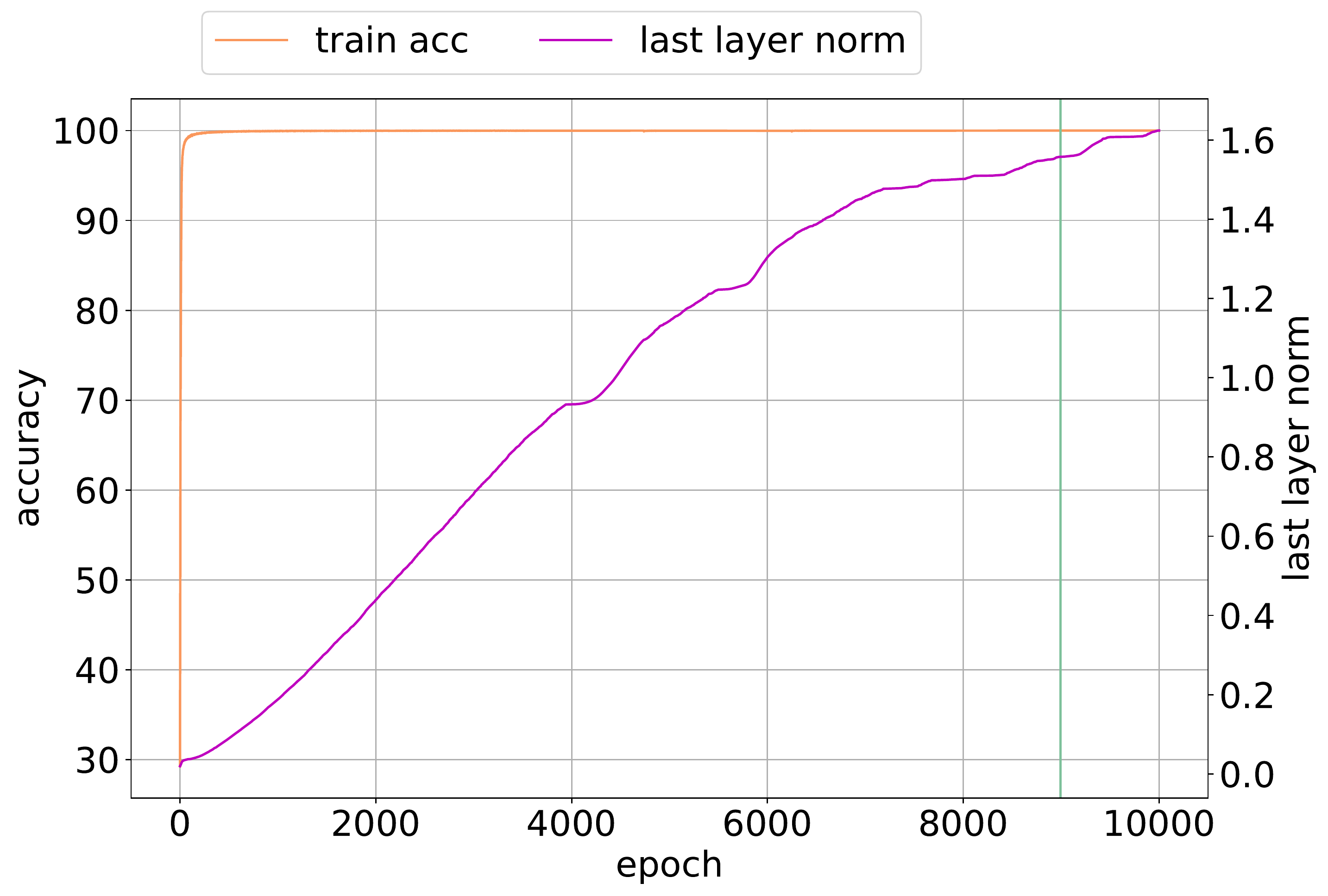} \\
      (a)  & (c) \\
     \includegraphics[width=0.4\linewidth]{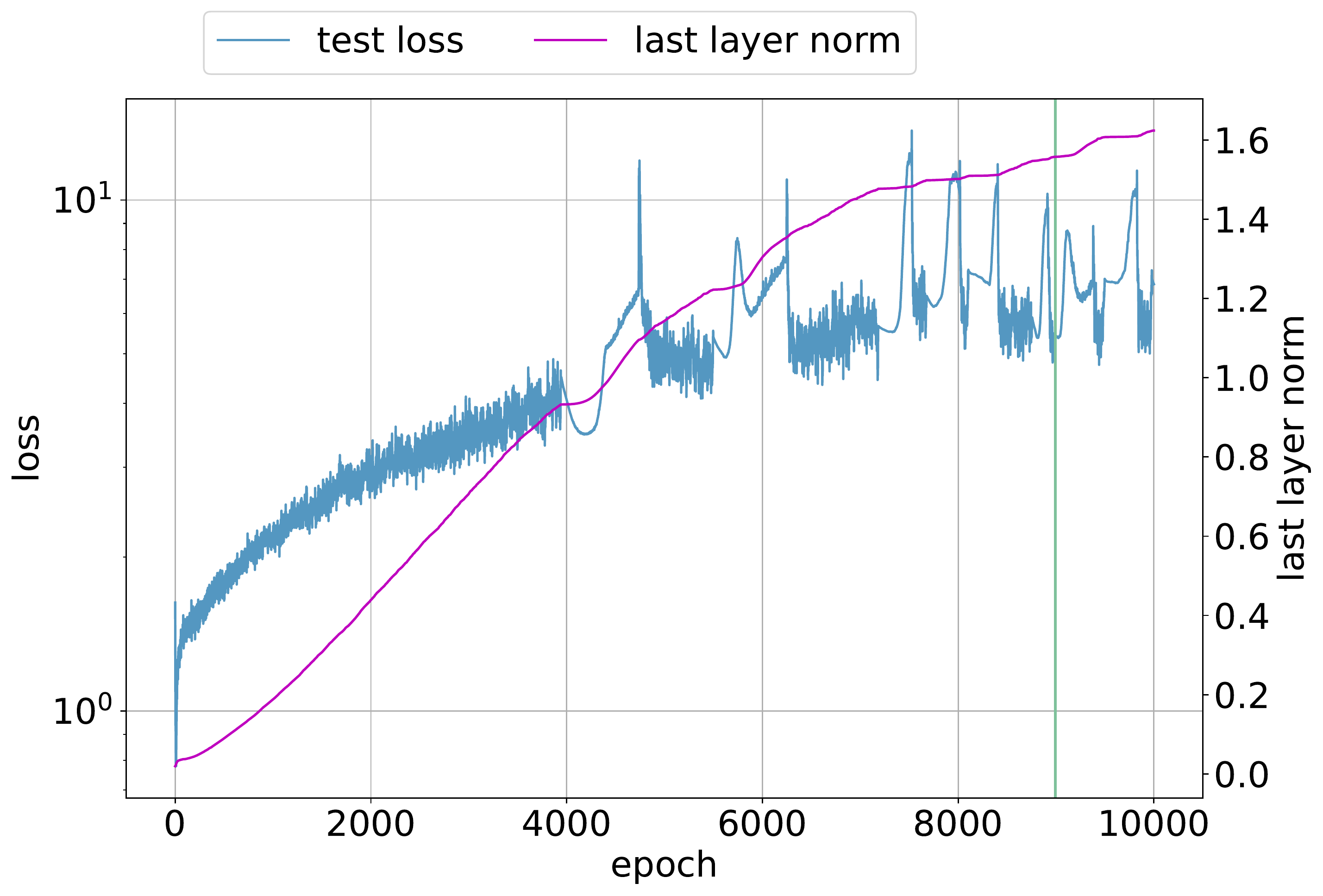} &
      \includegraphics[width=0.4\linewidth]{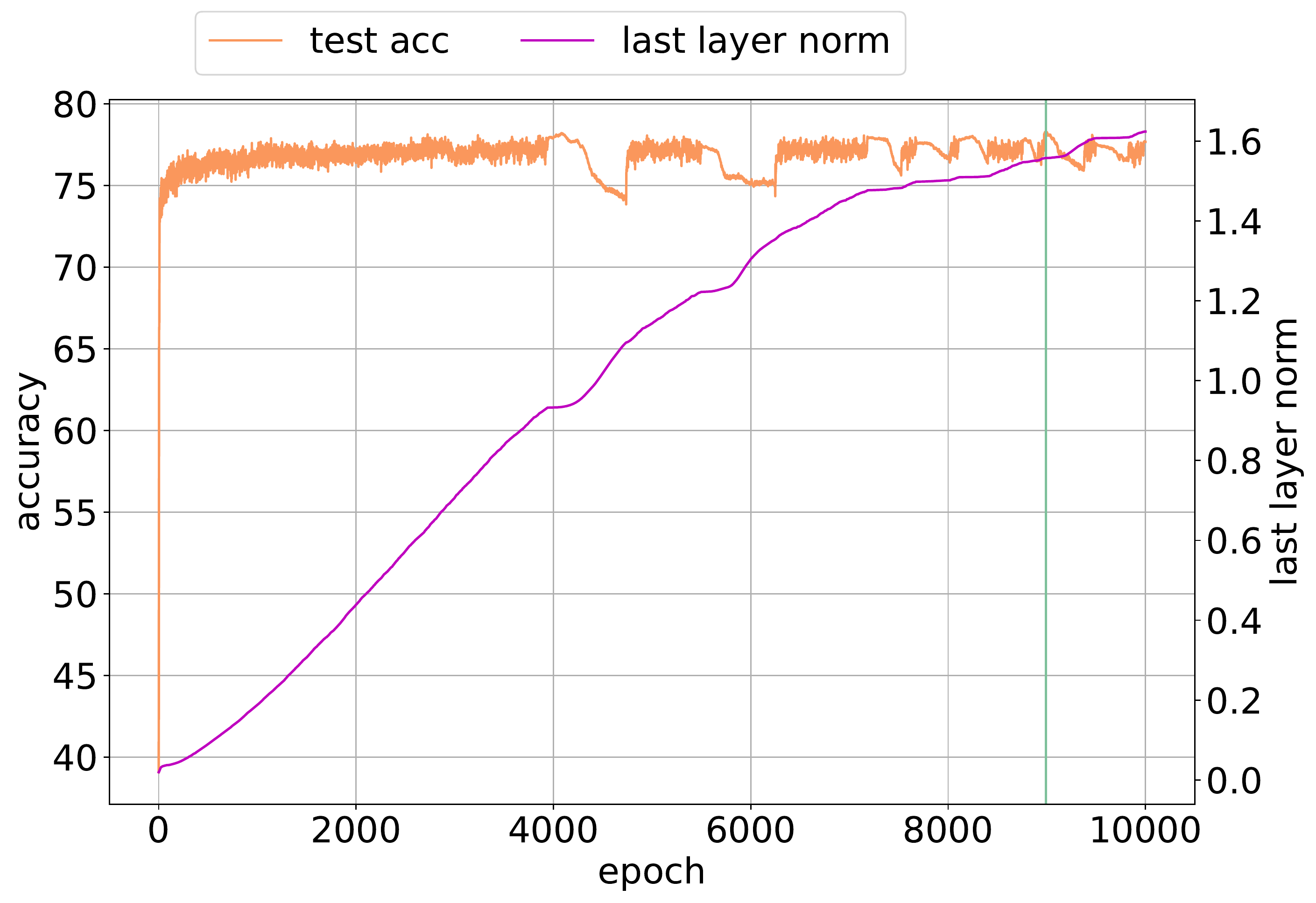} \\
      (b)  & (d)

  \end{tabular}
 \caption{Slingshot generalization on full CIFAR-10 dataset: Norm growth versus a) loss on training data b) accuracy on training data (c) loss on test data d) accuracy on test data.} 
 \label{fig:gen_cifar10_50k}
\end{figure*}

\paragraph{Vision Transformers and Full CIFAR-10}
In Appendix~\ref{appendix:slingshot_optim_validation}, we have empirically shown that the existence of the Slingshot phenomenon on a small subset of CIFAR-10 dataset~\cite{krizhevsky09learningmultiple} with Vision Transformers (ViTs). We now study the impact that Slingshot has on the generalization ability of ViTs by training a model on all 50000 samples in CIFAR-10 training dataset. The ViT used here is a larger model than the one considered in~\ref{appendix:slingshot_optim_validation} to account for larger dataset size. The ViT model consists of 12 layers, width 384 and 12 attention heads and is optimized by AdamW~\cite{loshchilov2017decoupled}. For this experiment, we set the learning rate to 0.0001, weight decay to 0, $\beta_{1} = 0.9$, $\beta_{1} = 0.95$ and $\epsilon = 10^{-08}$, minibatch size of 512 and linear learning rate warmup for $1$ epoch of optimization. Figure~\ref{fig:gen_cifar10_50k} shows the results of experiment with full CIFAR-10 dataset. Multiple Slingshots can be observed in these plots similar to the plots described in Appendix~~\ref{appendix:slingshot_optim_validation}. We observe from Figure~\ref{fig:gen_cifar10_50k}d that the test accuracy peaks in epochs following a Slingshot with the maximum recorded test accuracy occurring very late in optimization. This observation suggests that the Slingshot can have a favorable effect on generalization consistent with the behavior observed in the main paper  with division dataset.

\label{appendix:slingshot_gen_cifar10}

\subsubsection{Abalation Study} 
In this section, we train a toy model on a synthetically generated dataset with the aim of analysing the effect of different hyper parameters on the Slingshot Mechanism. 
We construct a 128-dimensional dataset with Scikit-learn~\cite{scikit-learn} that has 3 informative dimensions that represents a 8-class classification problem. The class centers are the edges of a 3-dimensional hypercube around which clusters are data are sampled from a standard normal distribution. The other 125-dimensions are also filled at random to create a high-dimensional dataset used in our experiments. We generate 256 training and validation samples for this dataset and use a minibatch size of 128 in all the experiments described in the following.

\paragraph{Architecture and Optimizer} Figure~\ref{fig:gen_synth_eps8} shows the training and validation metrics when we optimize a 4-layer fully-connected network (FCN) with Adam using a learning rate of 0.001, $\beta_{1}=0.9$, $\beta_{1}=0.95$, no weight decay and $\epsilon=1e^{-08}$. Note that we use this value of $\epsilon$ in our first experiment as this is the default value proposed in Kingma and Ba~\cite{kingma2014adam}. These experiments are implemented in JAX~\cite{jax2018github}.

\begin{figure*}[h]
\centering
    \includegraphics[width=0.75\linewidth]{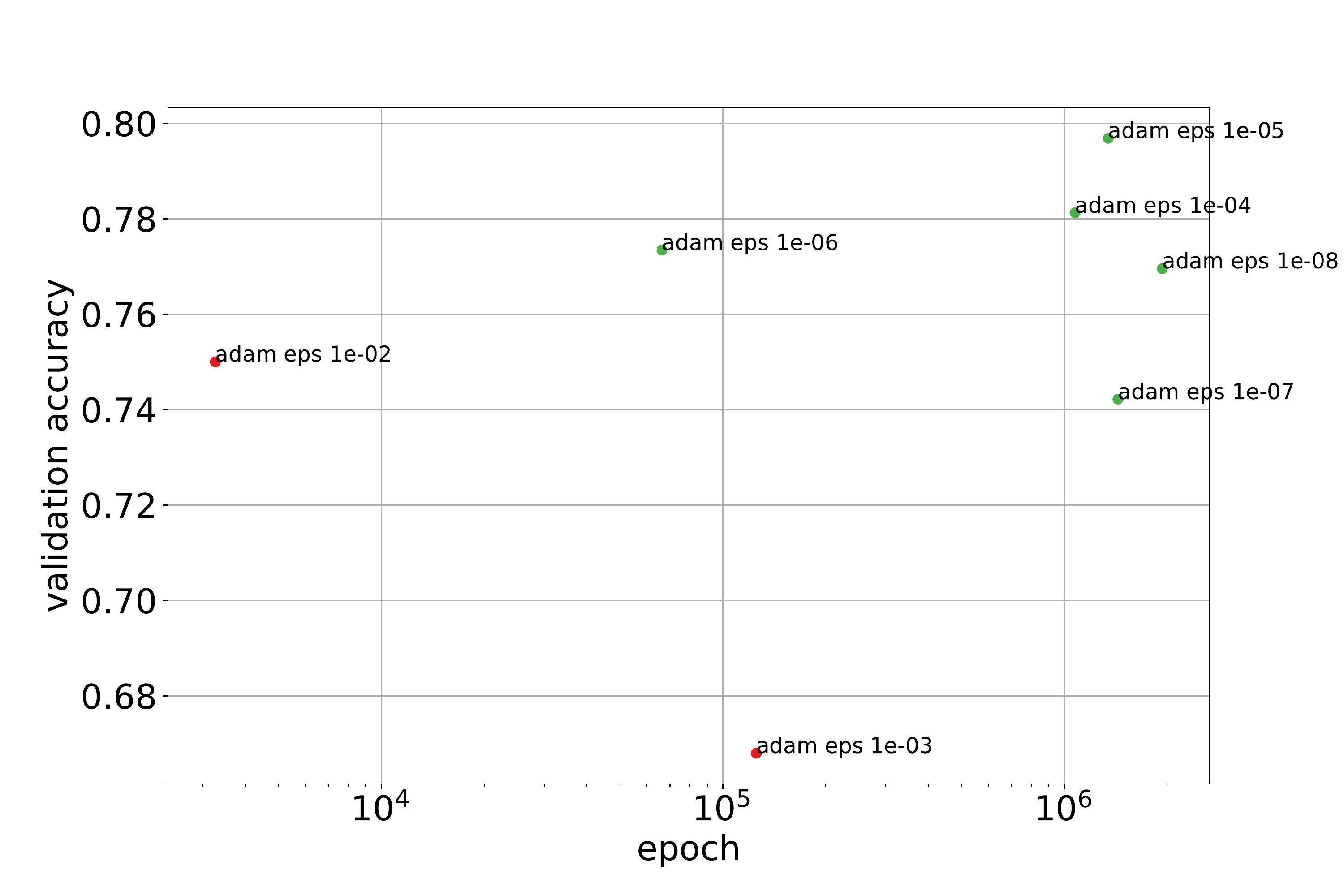}
\caption{Slingshot on syntehtic dataset. Note that the points marked in: (i) green correspond to Adam-trained models that undergo Slingshot, (ii) red correspond to Adam-trained models that do not experience Slingshot; Adam's hyperparameters are given by $\beta_{1}=0.9$, $\beta_{2}=0.95$, no weight decay and $\epsilon$ shown in parentheses.}
\label{fig:gen_synth_adam}
\end{figure*}


\paragraph{Tuning $\epsilon$} In the next set of experiments with synthetic data, we tune $\epsilon$ value for Adam to understand its impact on test accuracy. Figure~\ref{fig:gen_synth_adam} shows a plot of the maximum validation accuracy achieved by models trained with Adam as a function of time (epoch). We observe that Adam reaches its best test accuracy late in optimization with $\epsilon=10^{-5}$ yielding the highest validation accuracy. Furthermore, the best accuracy is achieved with a model that experiences Slingshot during optimization. This observation is consistent with our findings for ViT training with CIFAR-10 dataset described in the main paper and Appendix~\ref{appendix:slingshot_gen_cifar10}.

\begin{figure*}[h!]
\centering
  \begin{tabular}{ccc}
      loss & accuracy & sharpness \\
      \includegraphics[width=0.33\linewidth]{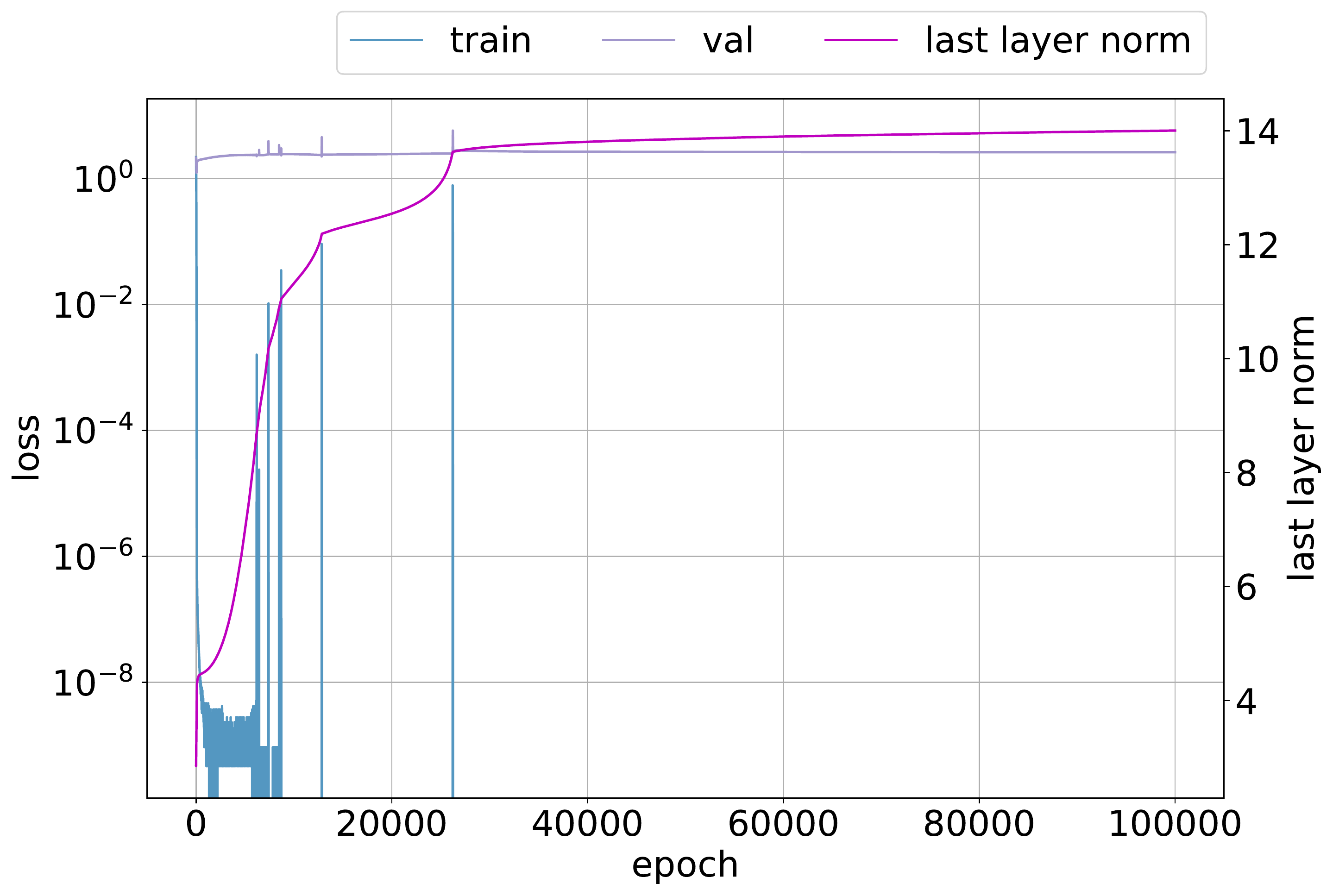} & 
      \includegraphics[width=0.33\linewidth]{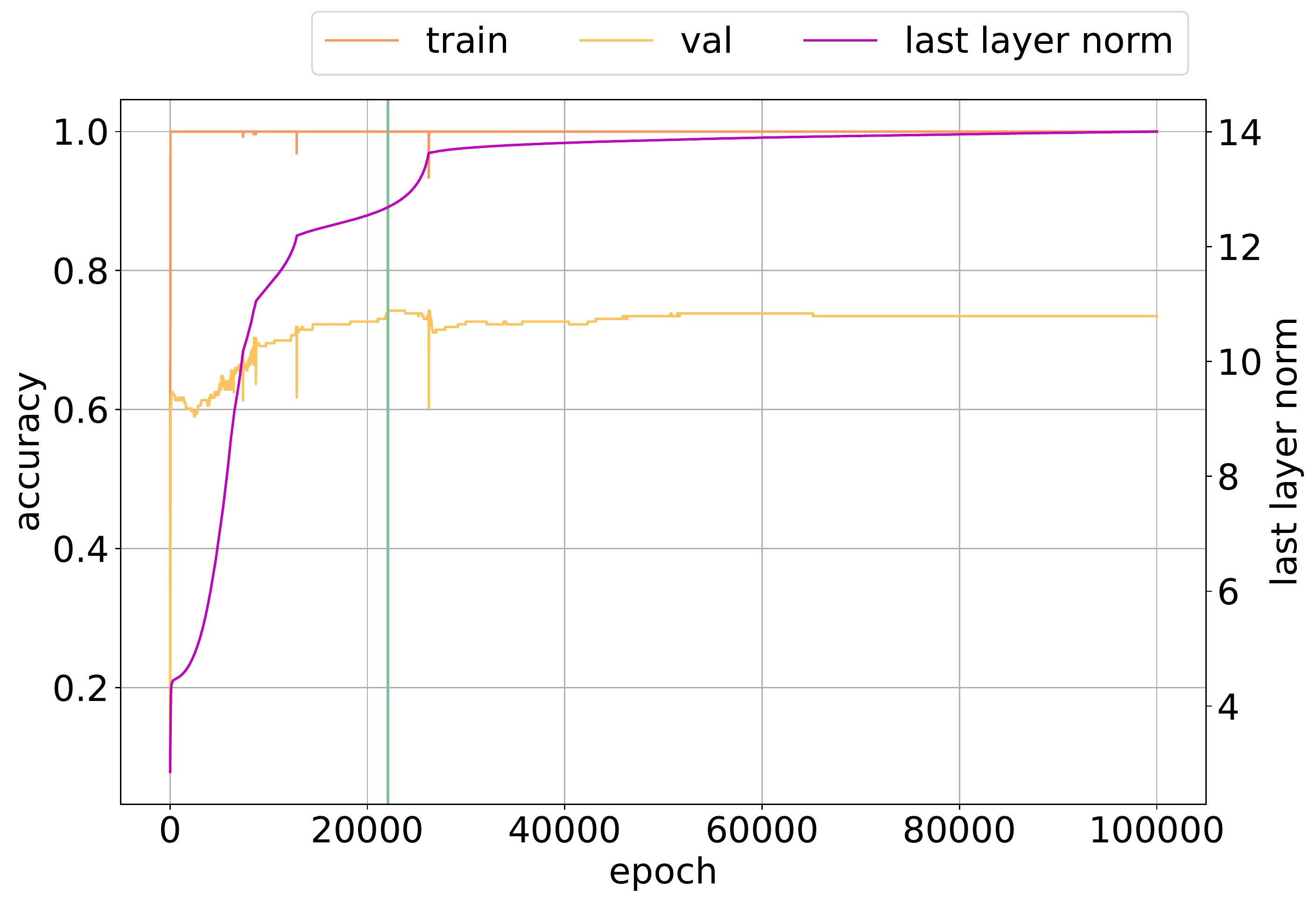} &
      \includegraphics[width=0.33\linewidth]{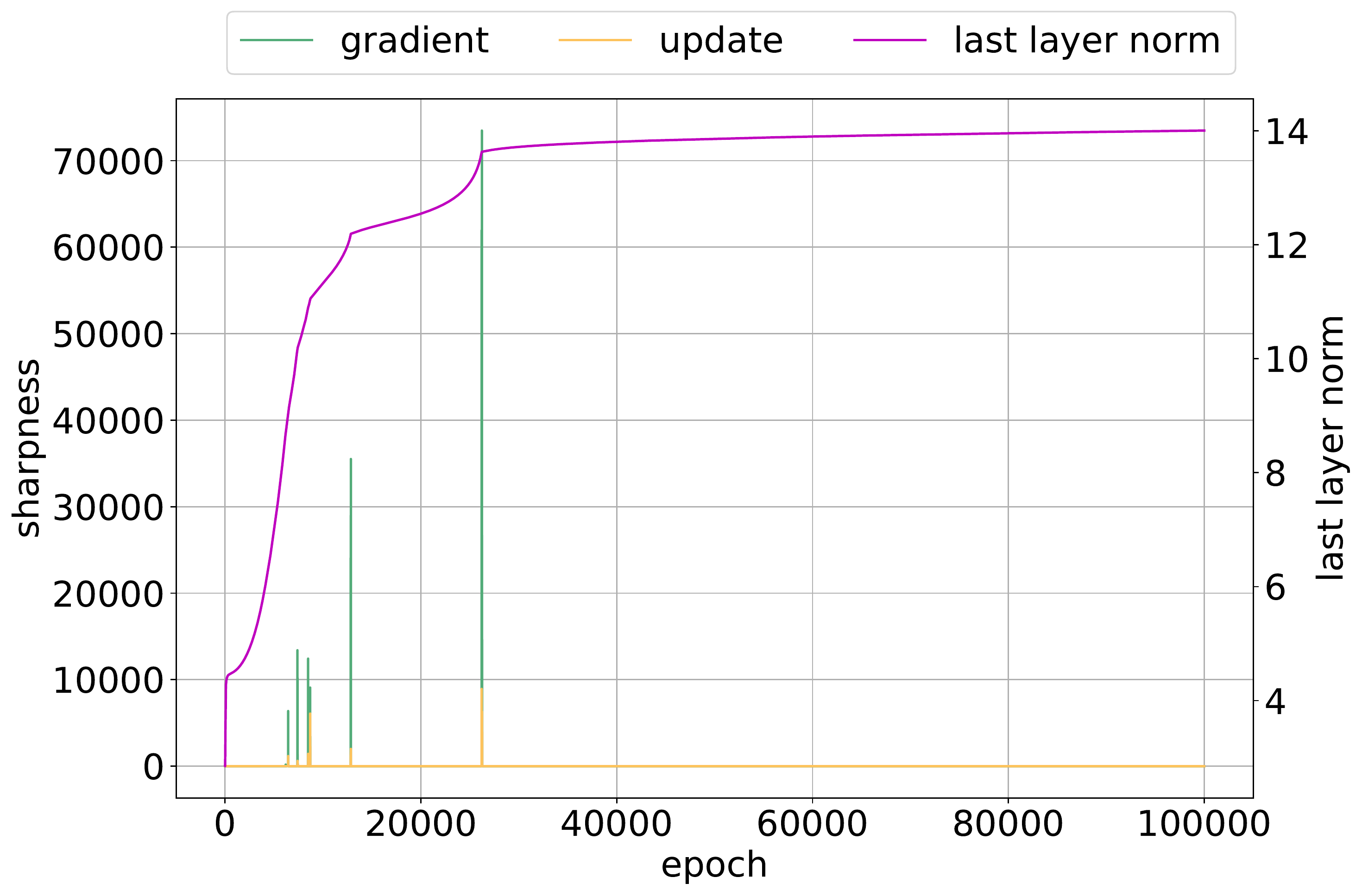} \\
      (a)  & (b) & (c) \\
      & $\beta_{1}=0, \beta_{2}=0$. Observe multiple Slingshots\\
      & \\
      \includegraphics[width=0.33\linewidth]{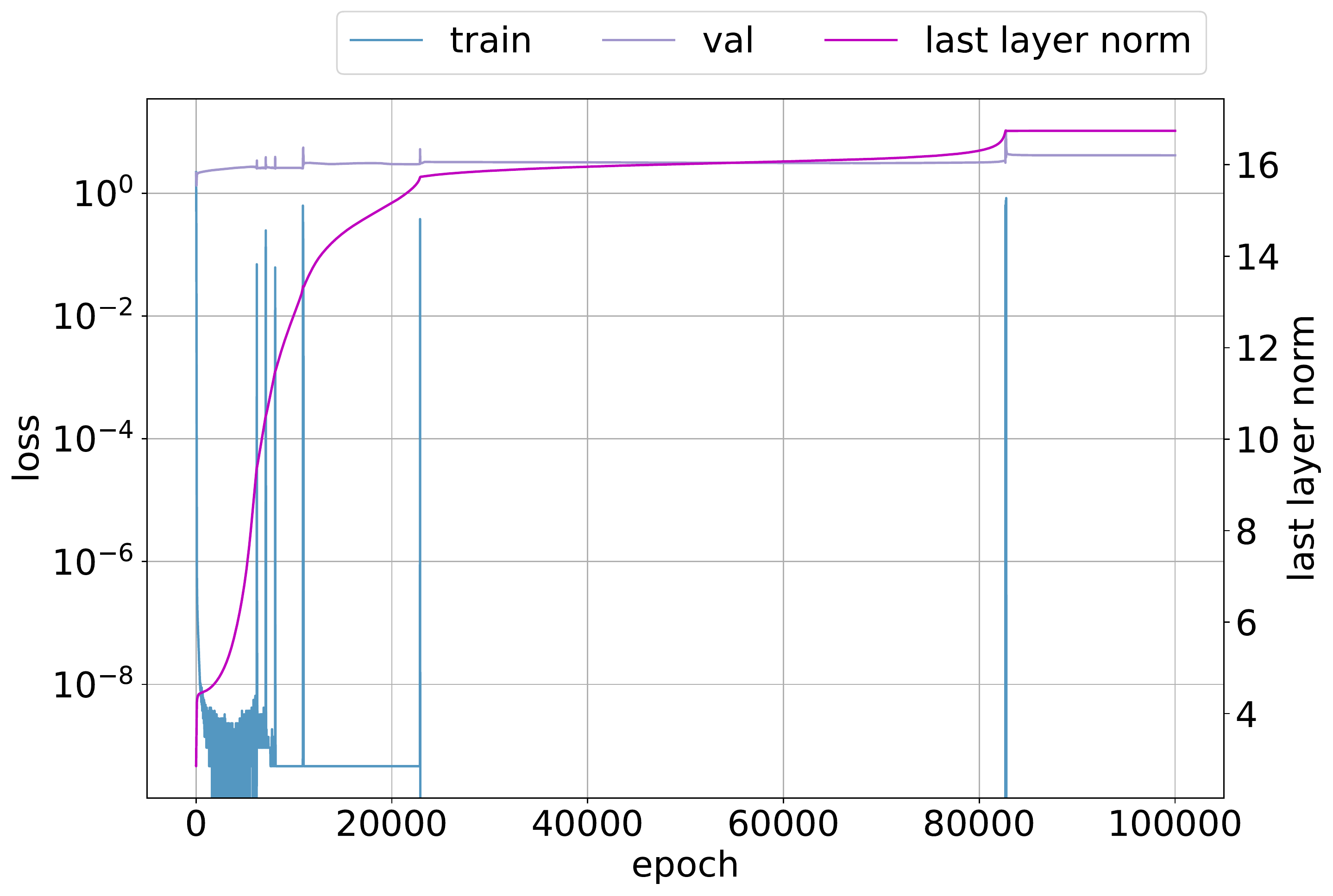} & 
      \includegraphics[width=0.33\linewidth]{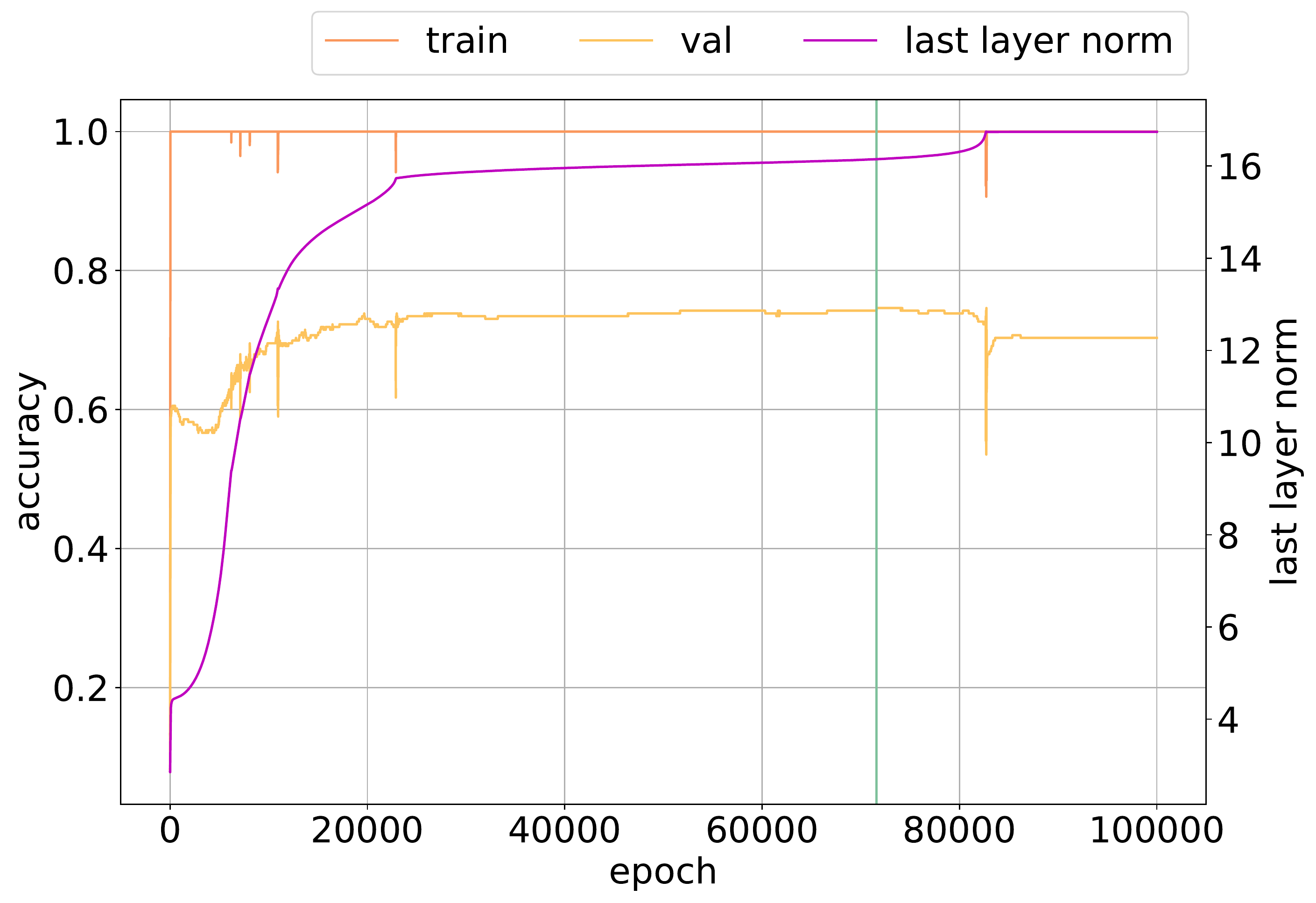} &
      \includegraphics[width=0.33\linewidth]{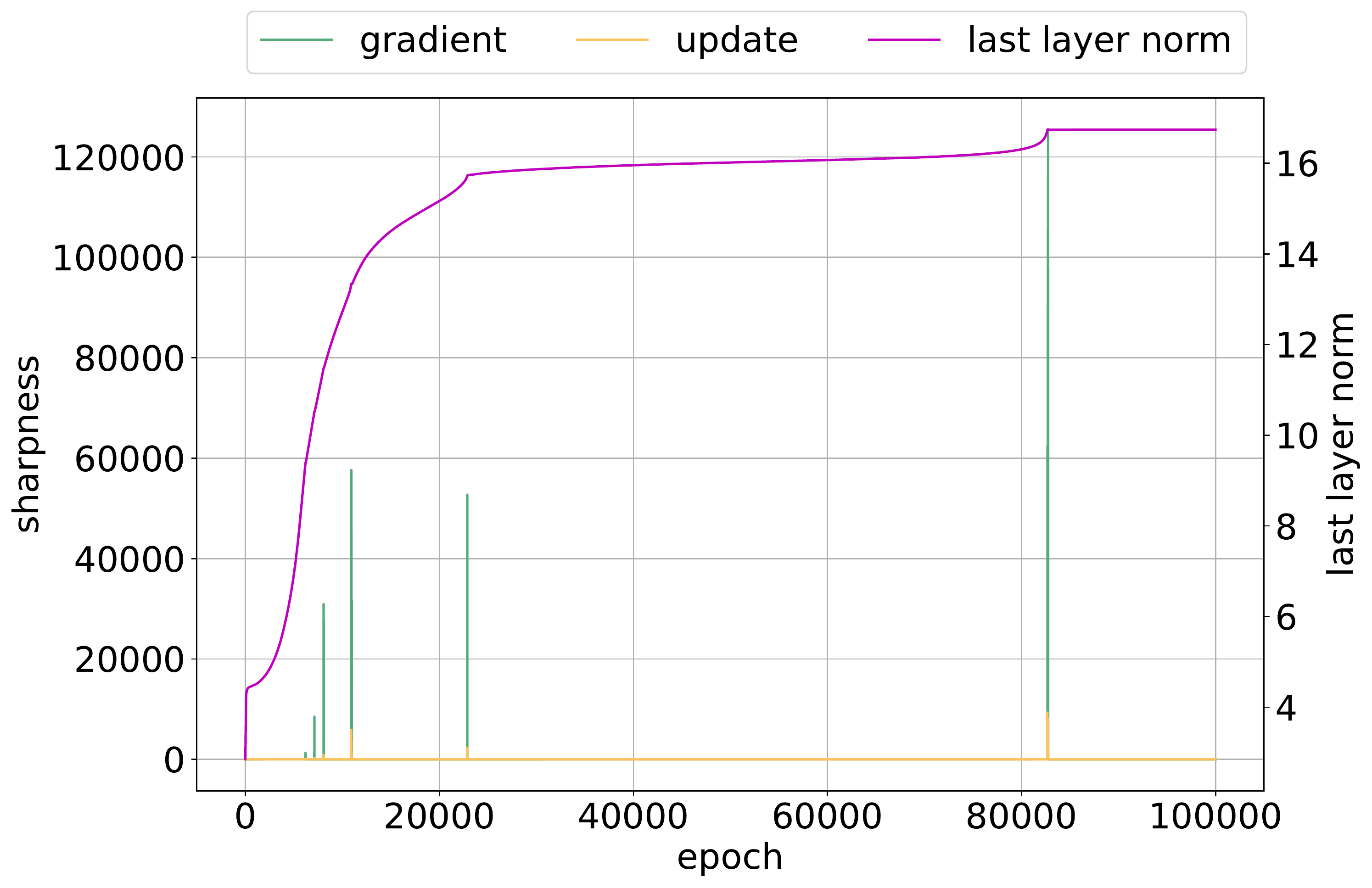} \\
      (d)  & (e) & (f) \\
      & $\beta_{1}=0.5, \beta_{2}=0.5$. Observe multiple Slingshots \\
      & \\
      \includegraphics[width=0.33\linewidth]{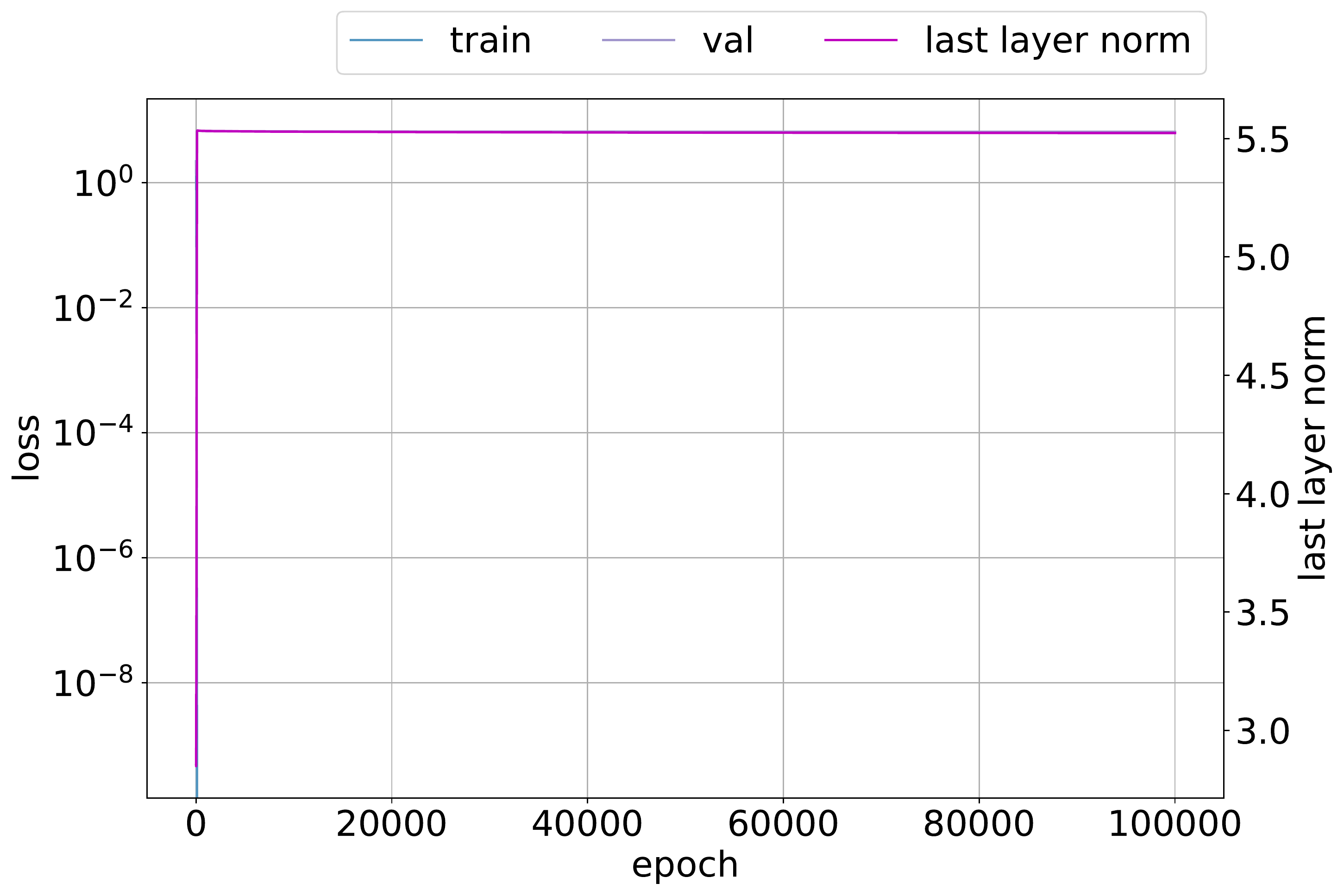} & 
      \includegraphics[width=0.33\linewidth]{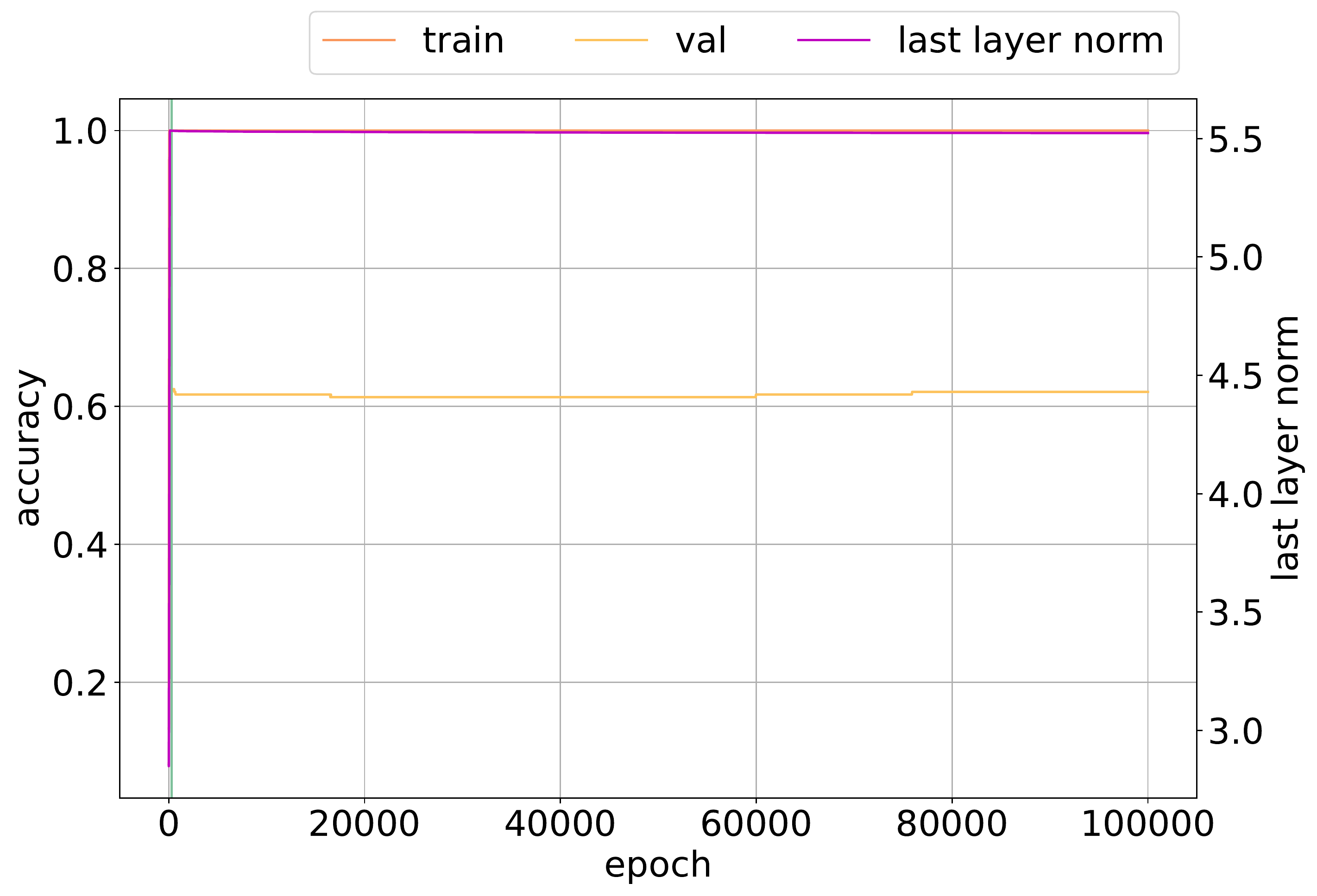} &
      \includegraphics[width=0.33\linewidth]{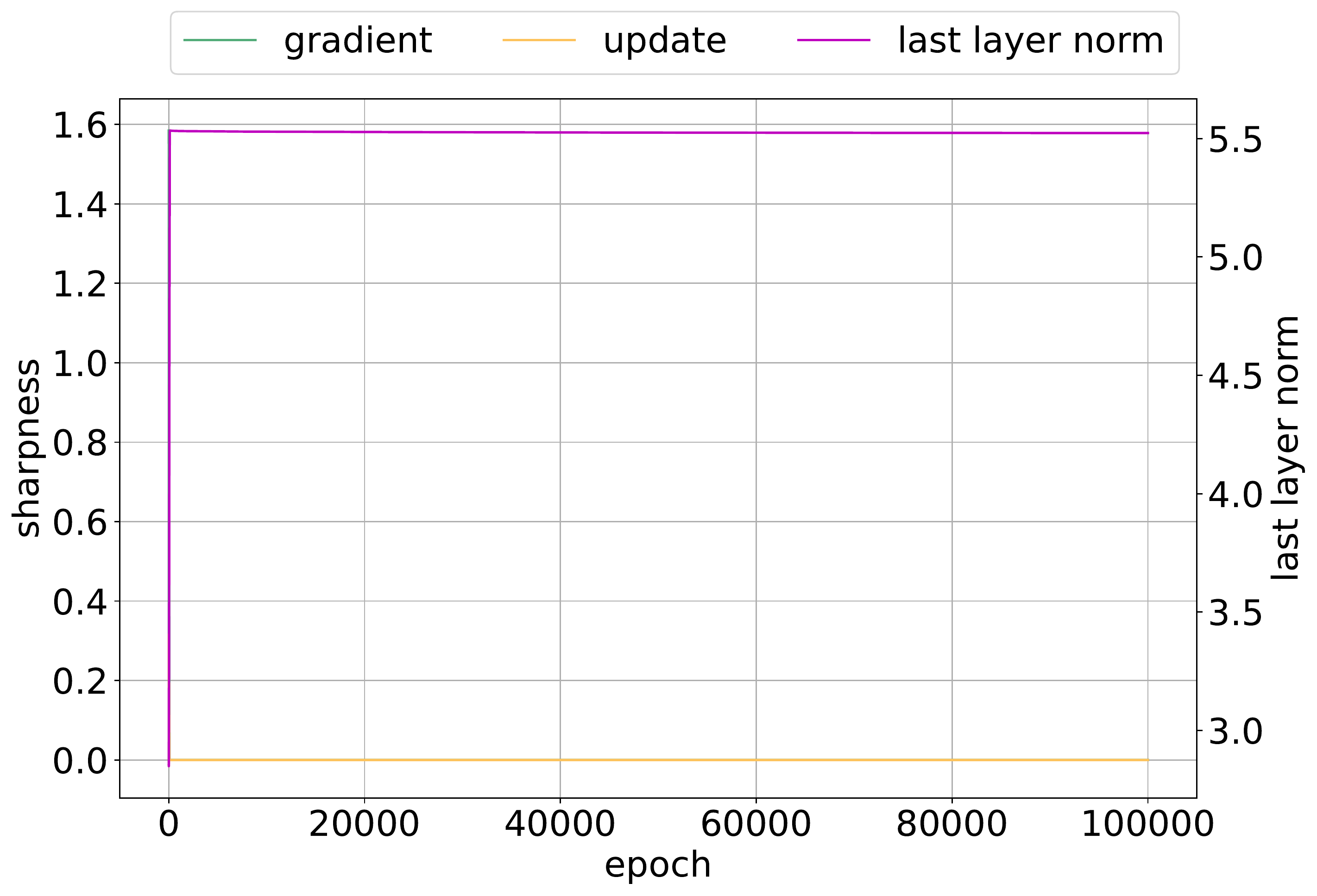} \\
      (g)  & (h) & (i) \\
      & $\beta_{1}=0.9, \beta_{2}=0.8$. Observe no Slingshot \\
      & \\
      \includegraphics[width=0.33\linewidth]{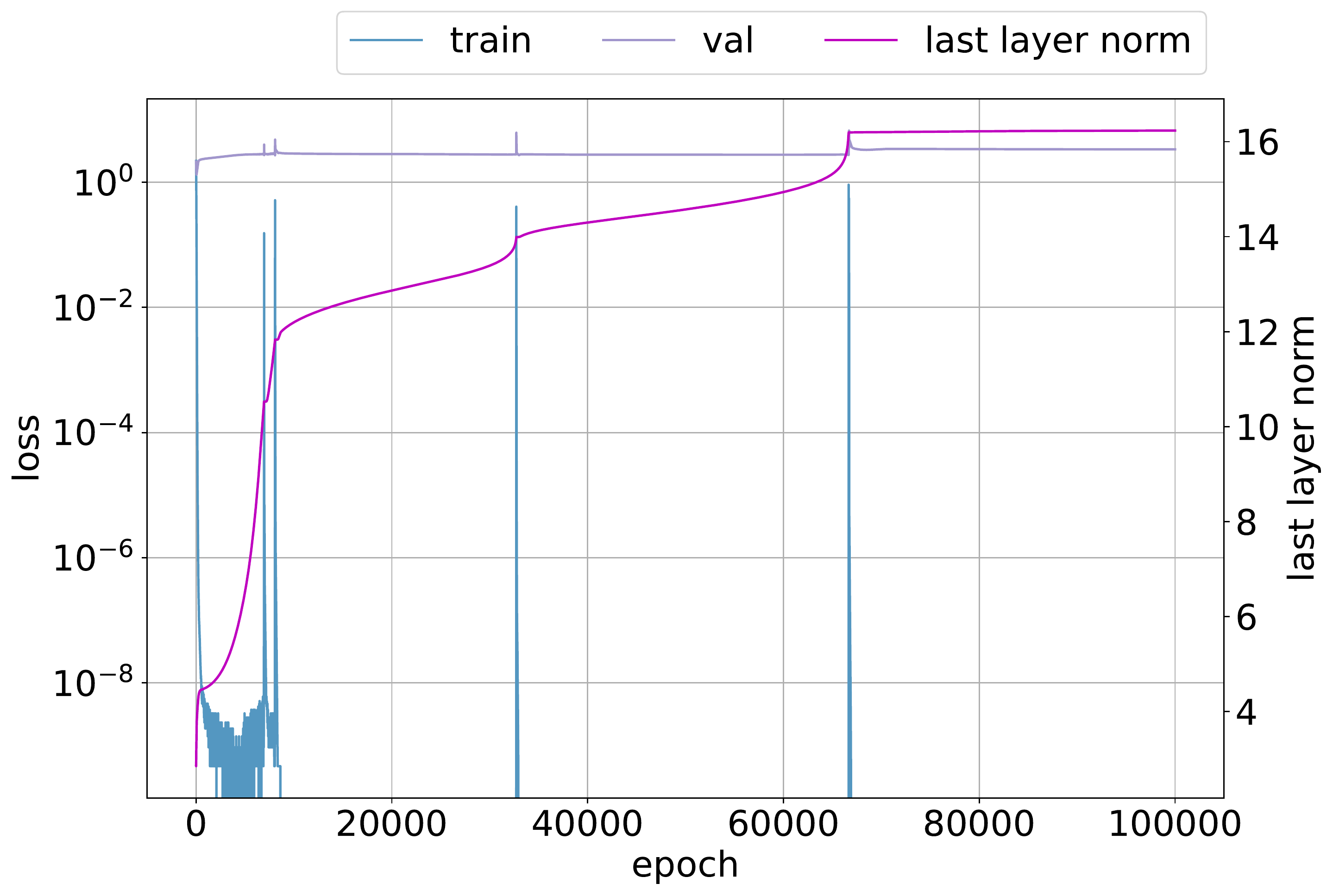} & 
      \includegraphics[width=0.33\linewidth]{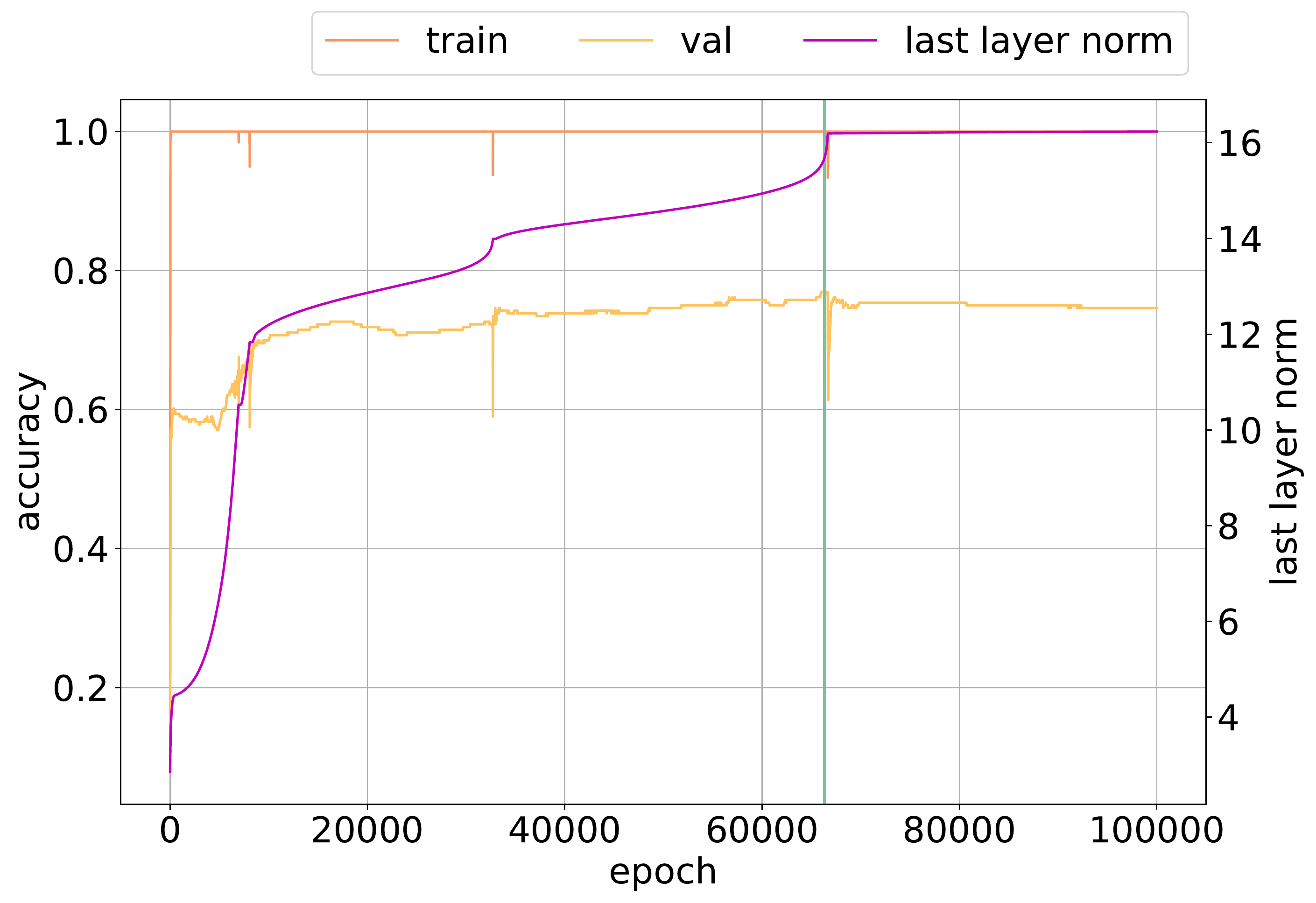} &
      \includegraphics[width=0.33\linewidth]{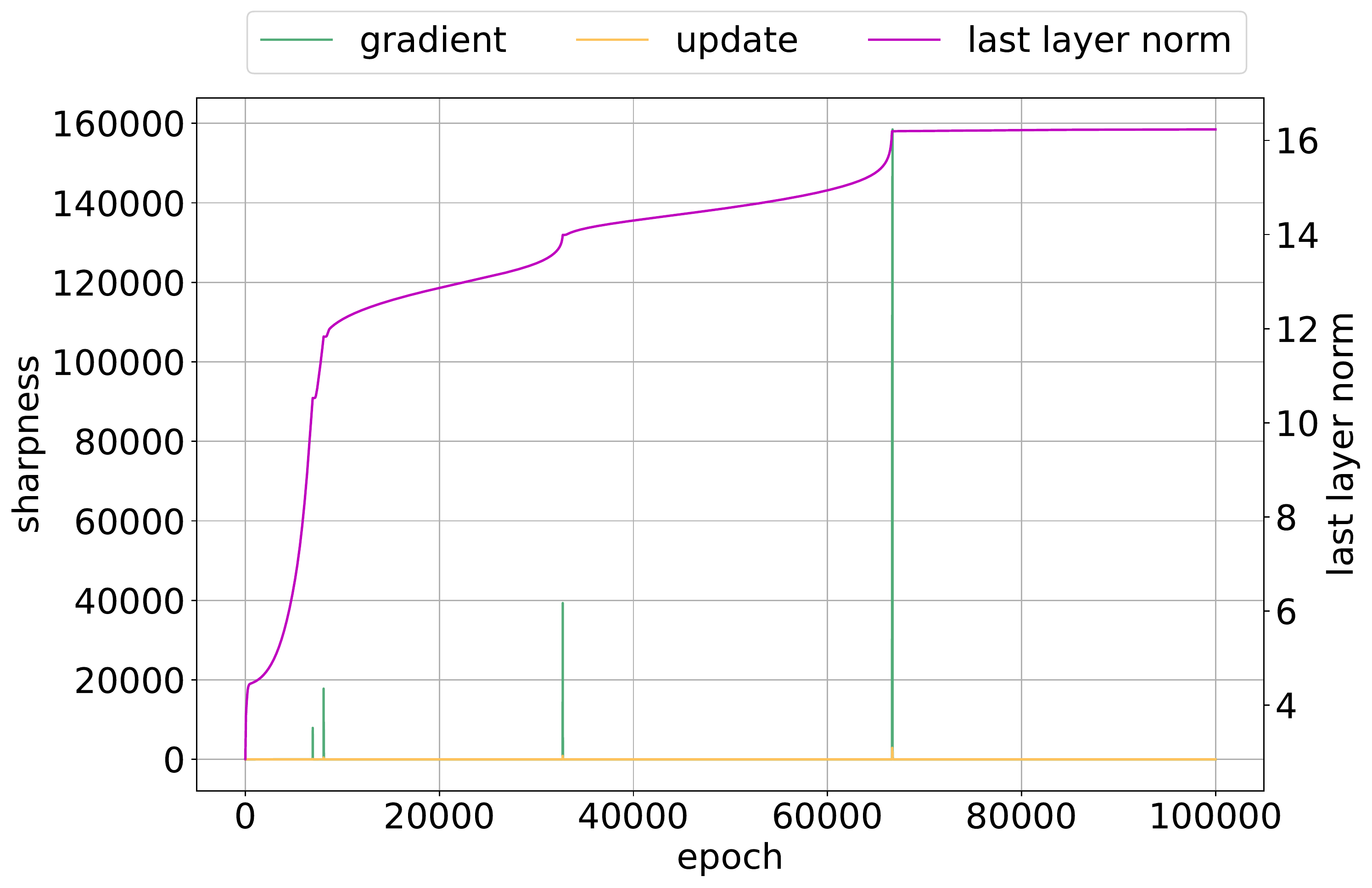} \\
      (j)  & (k) & (l) \\
      & $\beta_{1}=0.9, \beta_{2}=0.95$. Observe multiple Slingshots \\
      & \\
  \end{tabular}
 \caption {Varying $\beta_{1}, \beta_{2}$ in Adam on synthetic dataset. FCN is trained with Adam using learning rate $0.001$ and $\epsilon=10^{-06}$. The validation accuracy of models that experience Slingshot reach their highest accuracy later in training.} 
 \label{fig:slingshot_vary_b1b2}
\end{figure*}

\begin{figure*}[h!]
\centering
  \includegraphics[width=0.9\linewidth]{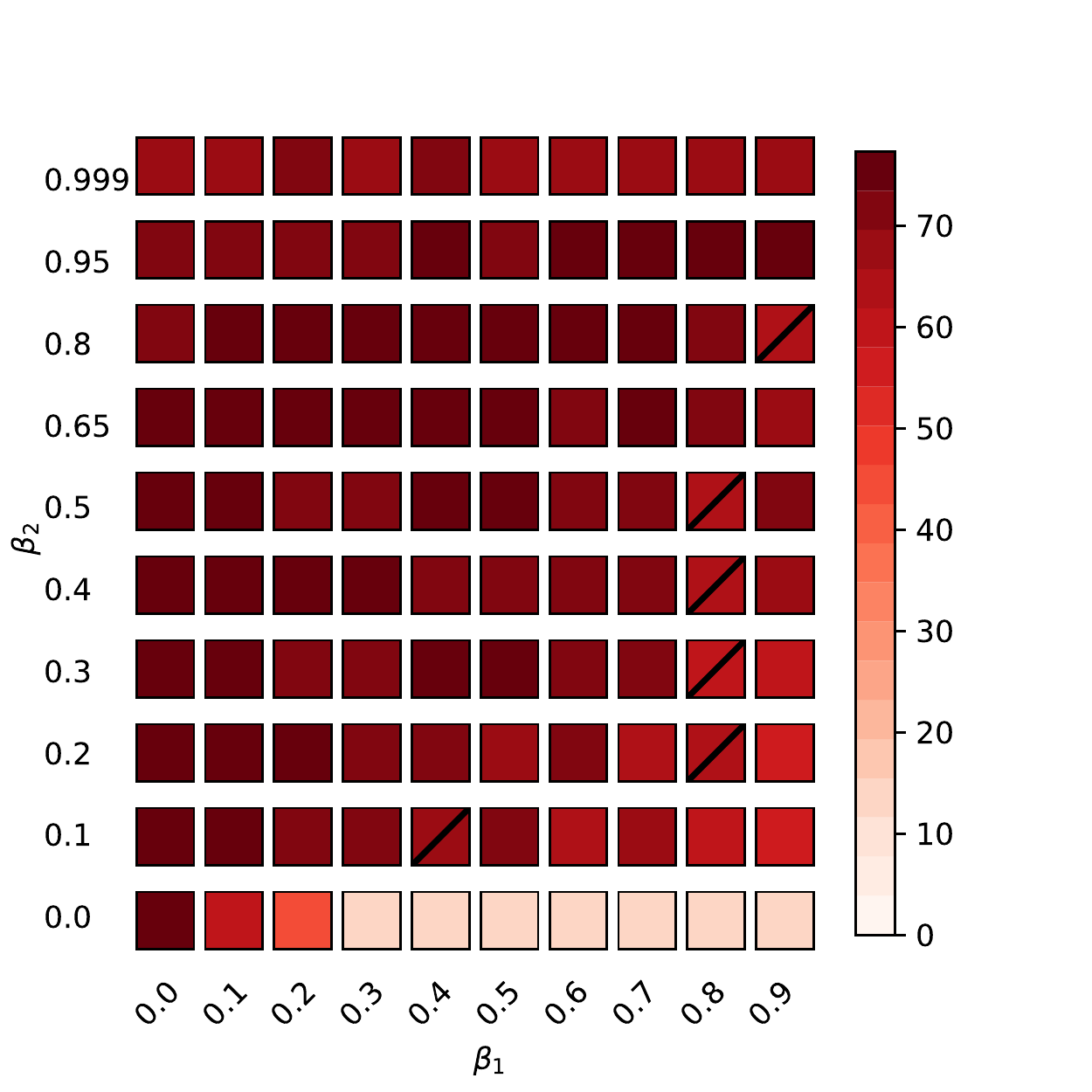}
   \caption{Extended analysis of $\beta_{1}, \beta_{2}$ in Adam on synthetic dataset. Plot shows the highest validation accuracy achieved with various values of $\beta_{1}, \beta_{2}$ with learning rate set to $0.001$ and $\epsilon=10^{-06}$. Hyperparameters that do not induce Slingshot Effects are marked with a diagonal line in black. Models trained with $\beta_{1} > 0.2$ and $\beta_{2} = 0$ diverged during training due to instability. These trials have their validation accuracy set to chance level.}
  \label{fig:synth_slingshot_summary}
\end{figure*}

\paragraph{Influence of $\beta_{1}$ and $\beta_{2}$}\label{para:b1b2} In these experiments, we aim to study the impact of Adam/AdamW optimizer's $\beta_{1}$ and $\beta_{2}$ hyperparameters on Slingshot. We use the synthetic data described above and set the learning rate of $0.001$ and $\epsilon=10^{-08}$ for this analysis. Figure~\ref{fig:slingshot_vary_b1b2} and Figure~\ref{fig:synth_slingshot_summary} shows the results of this study. We observe from Figure~\ref{fig:slingshot_vary_b1b2} that the Slingshot Mechanism is fairly robust to the values of $\beta_{1}$ and $\beta_{2}$. Figure~\ref{fig:slingshot_vary_b1b2}a-Figure~\ref{fig:slingshot_vary_b1b2}c show that Slingshot is even observed with $\beta_{1}$ and $\beta_{2}$ set to $0$ which effectively disables exponential moving averaging of gradient moments in Adam~\cite{kingma2014adam}. Figure~\ref{fig:slingshot_vary_b1b2}g-Figure~\ref{fig:slingshot_vary_b1b2}i provide an example of hyperparameters that fail to induce Slingshot. We observe from Figure~\ref{fig:slingshot_vary_b1b2} that models that experience Slingshot tend to reach their best test accuracy during the later stages of training. Specifically, we observe from Figure~\ref{fig:slingshot_vary_b1b2}b, Figure~\ref{fig:slingshot_vary_b1b2}e and Figure~\ref{fig:slingshot_vary_b1b2}k that the best validation accuracy occurs after $60000$ epochs. These examples provide further evidence about an interesting implicit bias of Adam. Figure~\ref{fig:synth_slingshot_summary} shows more examples of hyperparameters that do not induce Slingshot Effects. Finally, we observe from Figure~\ref{fig:synth_slingshot_summary} that hyperparameters that provide higher validation accuracy are from models that experience Slingshot Effects.

\newpage

\section{Slingshot and Grokking}
\label{appendix:xformers_setup}

We use the empirical setup described by Power et al.~\cite{power2021grokking} to describe the Slingshot Mechanism. The following section describes relevant details including datasets, architecture and optimizer used in our experiments.

\paragraph{Architecture} The model used a decoder-only Transformer~\cite{vaswani2017attention} with causal attention masking. The architecture used in all our experiments consists of 2 decoder layers with each layer of width 128 and 4 attention heads.

\paragraph{Optimization} We train the architecture described above with Adam optimizer~\cite{kingma2014adam, loshchilov2017decoupled} in most of our experiments unless noted otherwise. The learning rate is set to $0.001$ and with linear learning rate warmup for the first $10$ steps. We use $\beta_{1} = 0.9$, $\beta_{2} = 0.98$ for Adam's hyperparameters.  The Transformers are optimized with cross-entropy (CE) loss that is calculated on the output tokens for a given binary operation.

\paragraph{Algorithmic Datasets} The Transformer is trained on small algorithmic datasets that consists of sequences that represent a mathematical operation. The following operations are used in our experiments:
\begin{itemize}
    \item[] $c = a + b \pmod {p}$ for $0 \leq a, b < p$
    \item[] $c = a - b \pmod {p}$ for $0 \leq a, b < p$
    \item[] $c = a * b \pmod {p}$ for $0 \leq a, b < p$
    \item[] $c = a \div b \pmod {p}$ for $0 \leq a, b < p$
    \item[] $c = a^{2} + b \pmod {p}$ for $0 \leq a, b < p$
    \item[] $c = a^{3} + b \pmod {p}$ for $0 \leq a, b < p$
    \item[] $c = a^{2} + b^{2} \pmod {p}$ for $0\leq a, b < p$
    \item[] $c = a^{2} + b^{2} + ab \pmod {p}$ for $0\leq a, b < p$
    \item[] $c = a^{2} + b^{2} + ab + b \pmod {p}$ for $0\leq a, b < p$
    \item[] $c = a^{3} + ab \pmod {p}$ for $0 \leq a, b < p$
    \item[] $c = a^{3} + ab^{2} + b \pmod {p}$ for $0 \leq a, b < p$
    \item[] $c = [a \div b \pmod {p}$ if $b$ is odd, otherwise $a-b \pmod {p}$] for $0\leq a,b < p$
    \item[] $c = a\cdot b$ for $a, b\in S_{5}$
    \item[] $c = a\cdot b\cdot a^{-1}$ for $a, b\in S_{5}$
    \item[] $c = x\cdot b\cdot a$ for $a, b\in S_{5}$
    \item[] $c = [a + b \pmod {p}$ if $a$ is even, otherwise $a * b \pmod {p}$] for $0\leq a,b < p$
    \item[] $c = [a + b \pmod {p}$ if $a$ is even, otherwise $a - b \pmod {p}$] for $0\leq a,b < p$
\end{itemize}

where $p=97$ and with the dataset split in training and validation data. Each equation in the dataset is of the form $ (a) (op) (b) (=) c $ where (x) represents the token used to represent x. We refer to Power et al.~\cite{power2021grokking} for a detailed description of the datasets

\subsection{Analysis of Parameter Dynamics}

\begin{figure*}[h!]
\centering
  \begin{tabular}{ccc}
      loss & accuracy & cosine distance \\
      \includegraphics[width=0.33\linewidth]{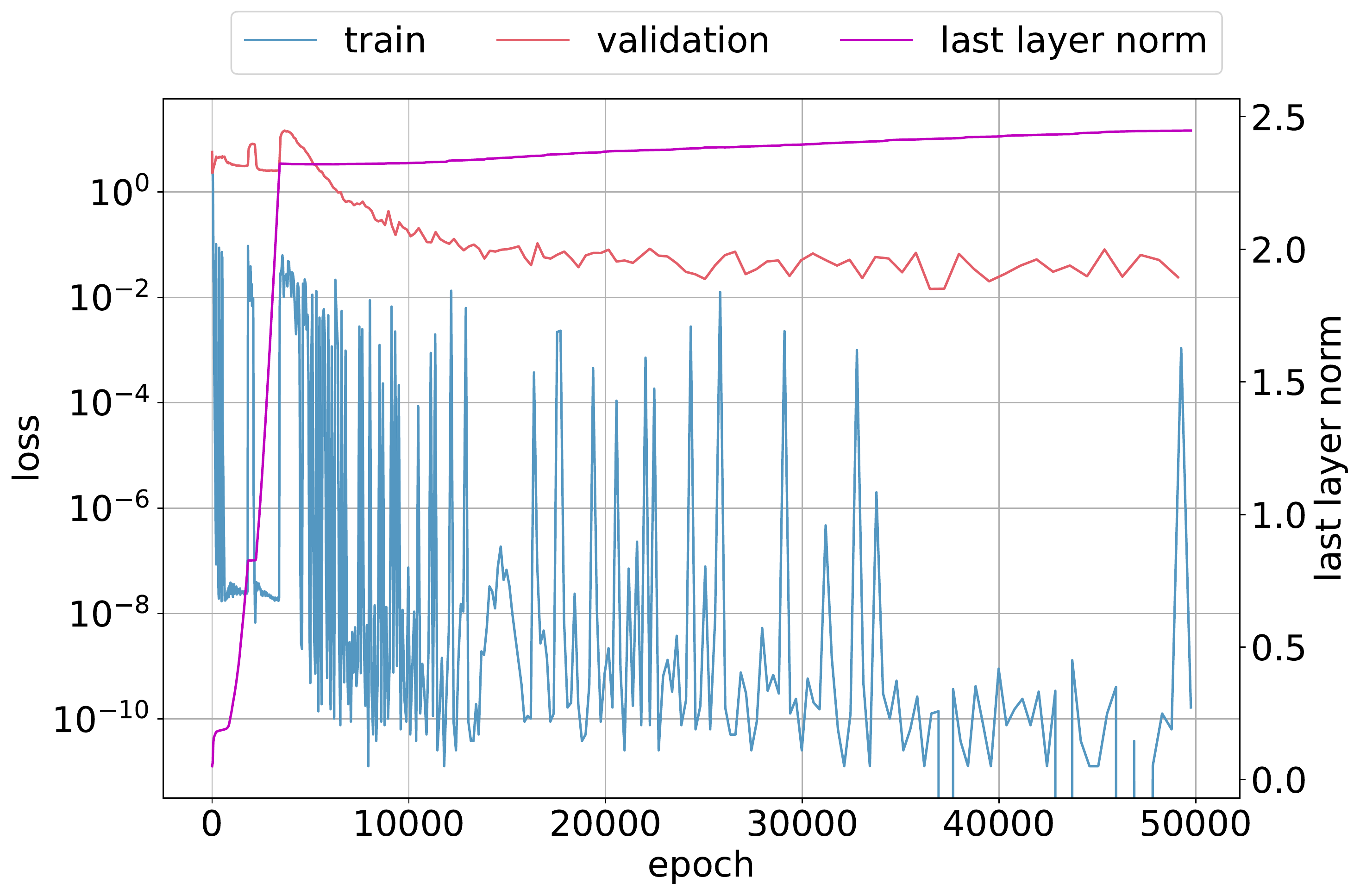} & 
      \includegraphics[width=0.33\linewidth]{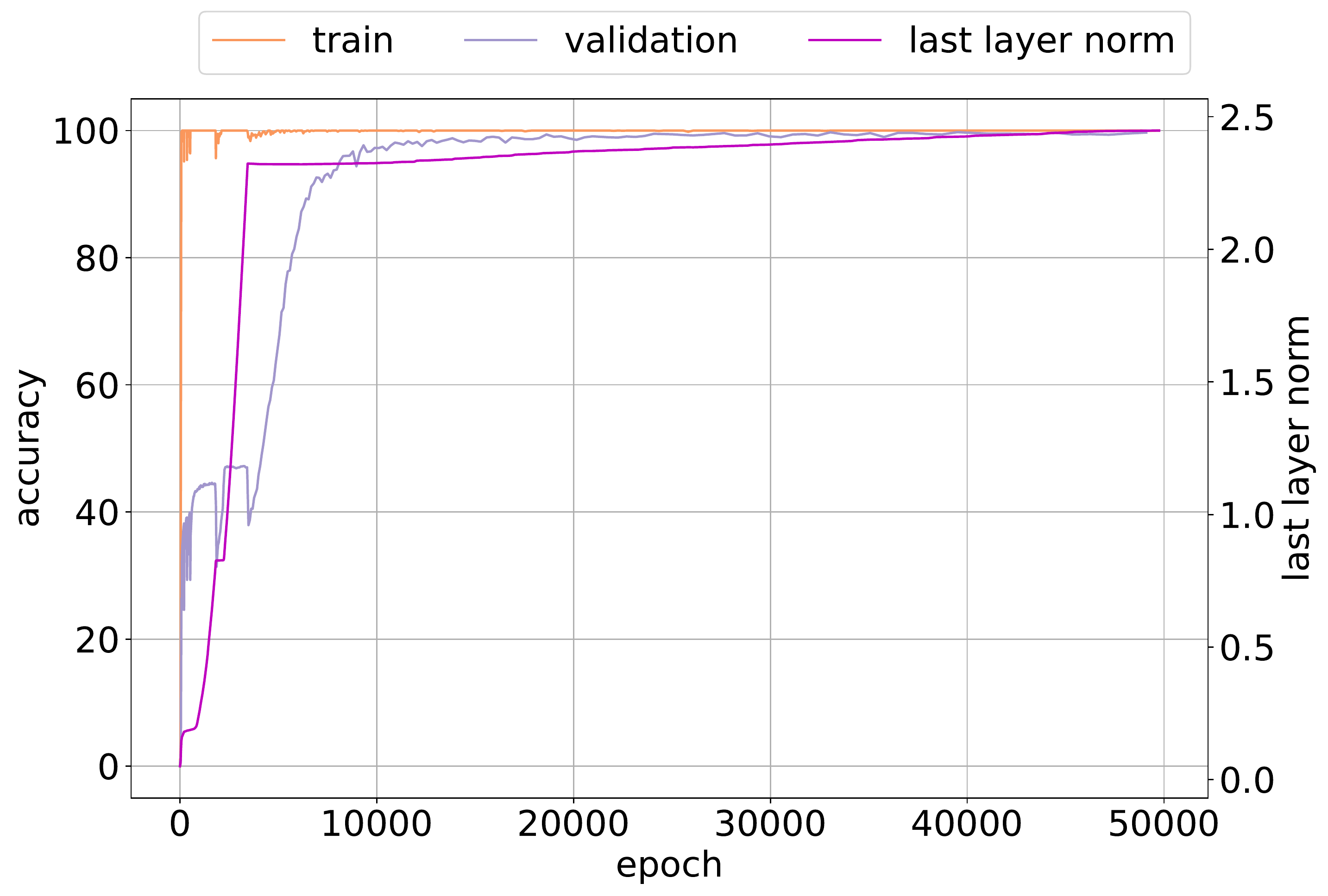} & 
      \includegraphics[width=0.33\linewidth]{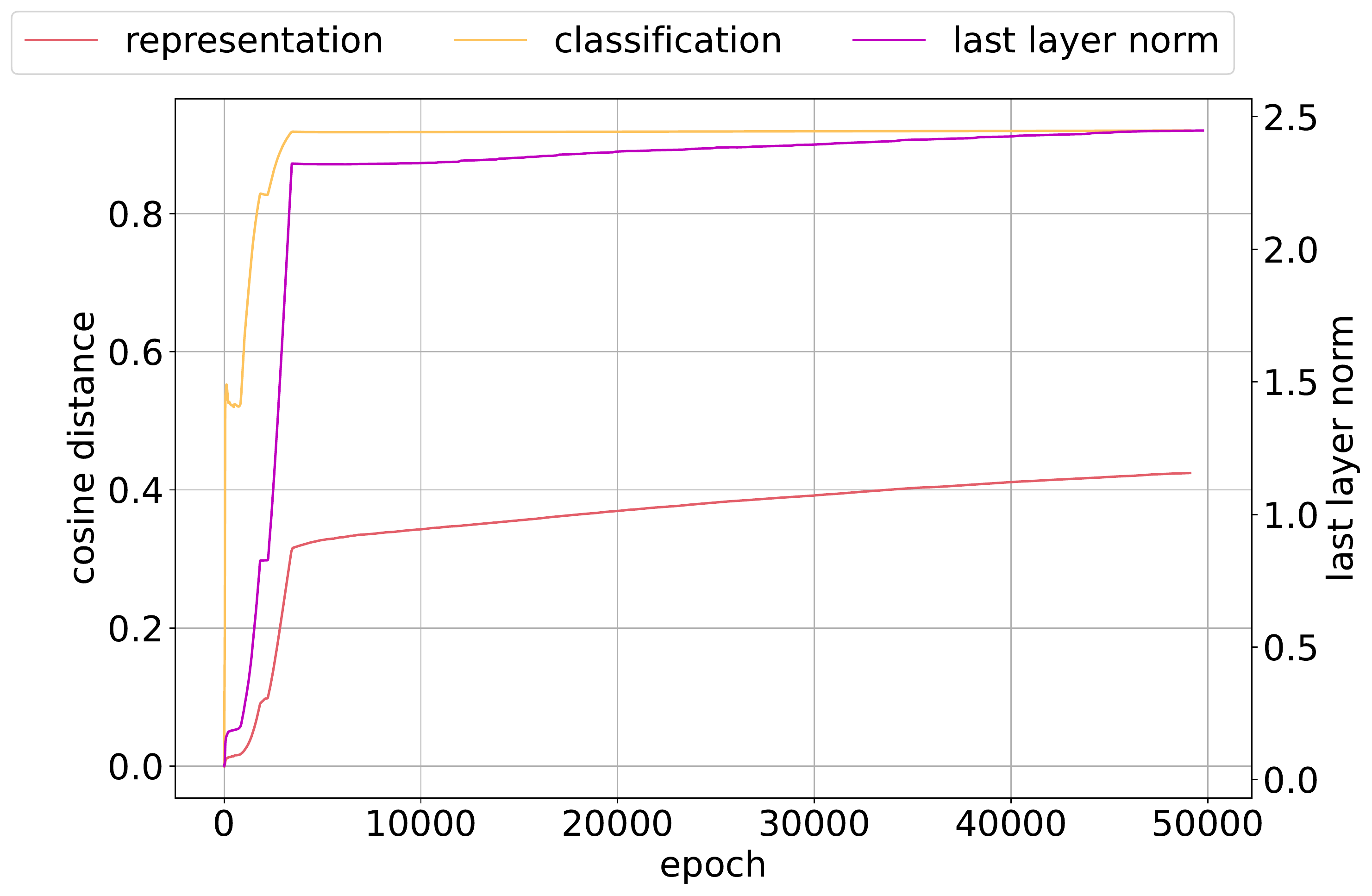} \\
      (a)  & (b) & (c) \\
      & $epsilon=10^{-08}$\\
      & \\
      \includegraphics[width=0.33\linewidth]{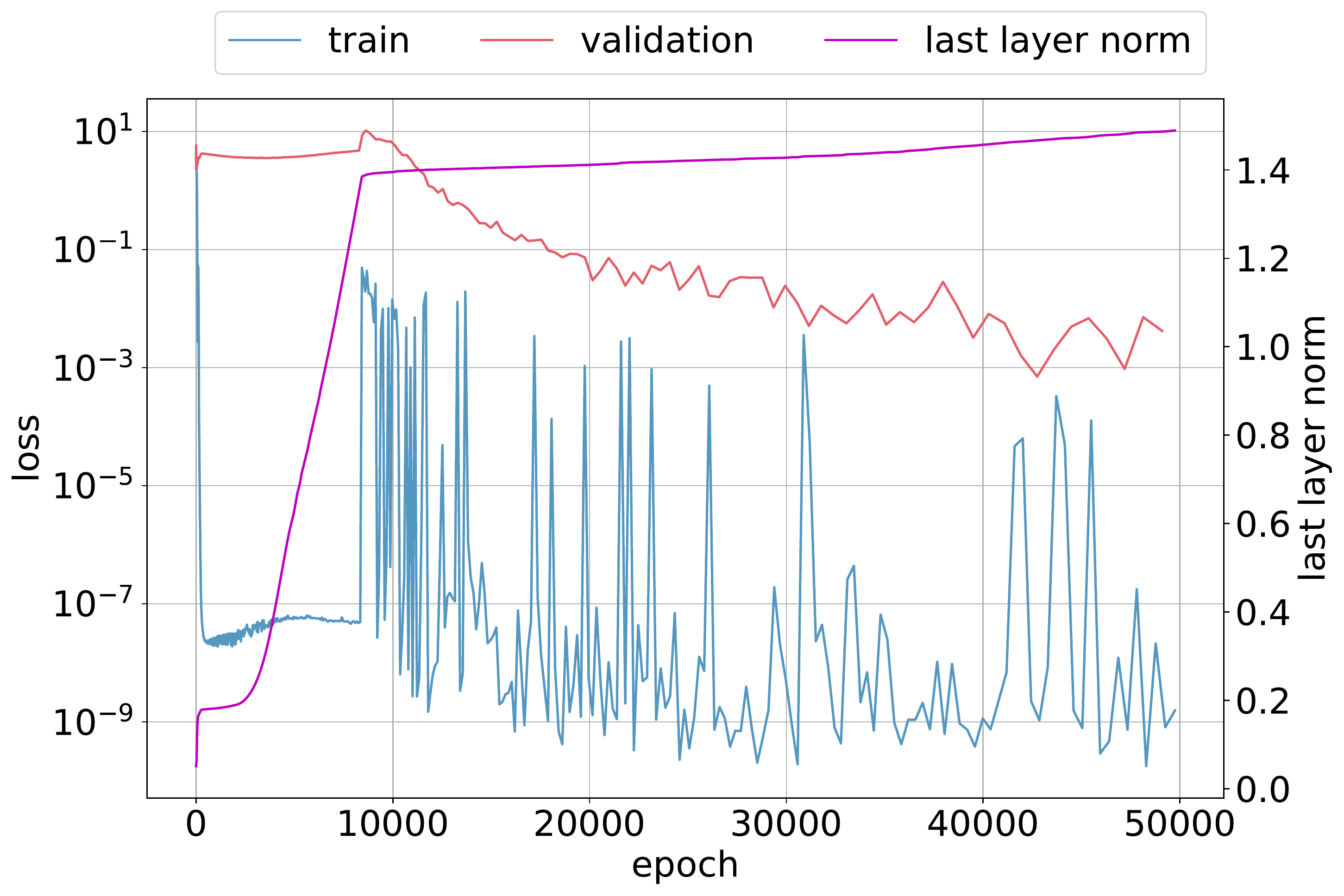} & 
      \includegraphics[width=0.33\linewidth]{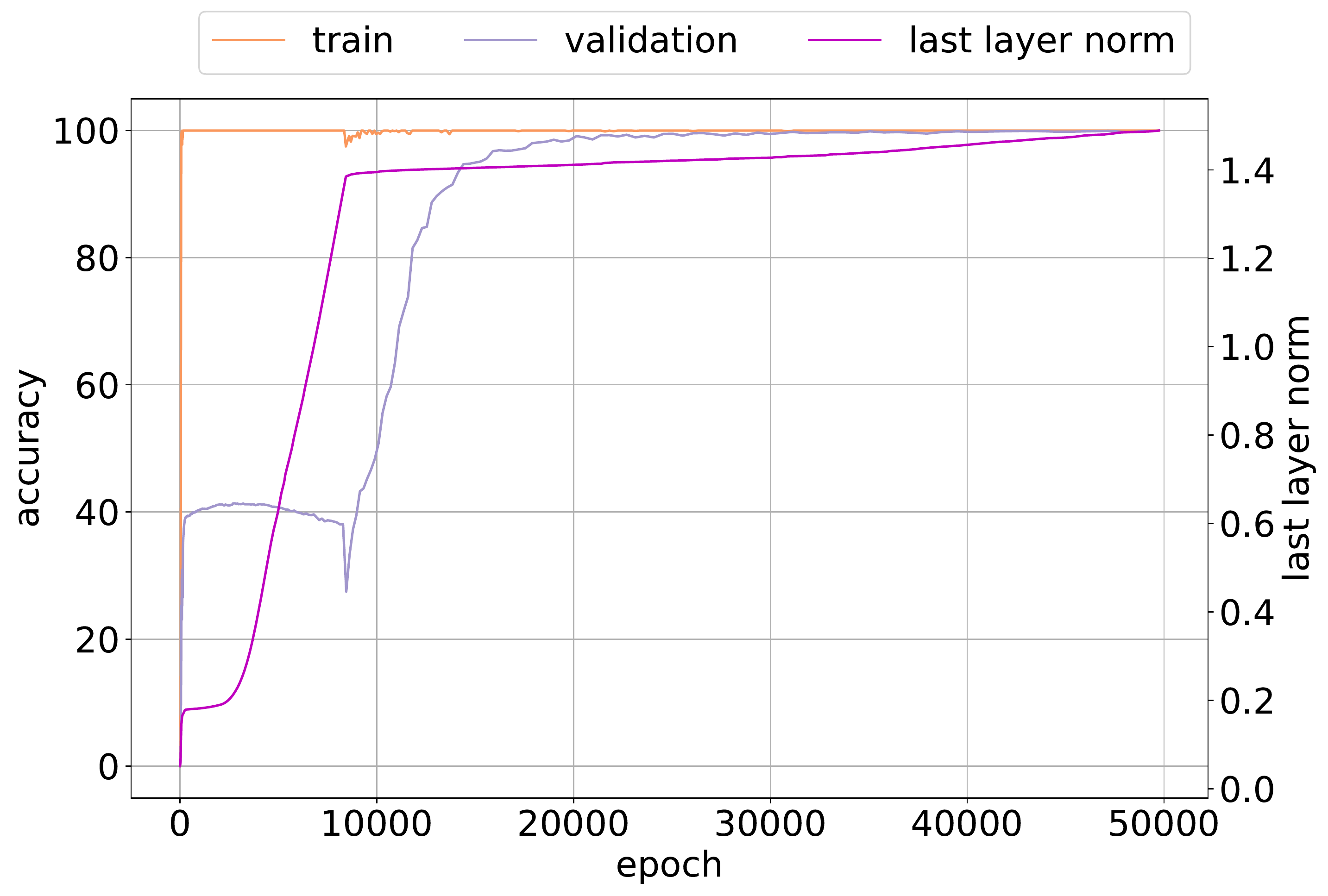} & 
      \includegraphics[width=0.33\linewidth]{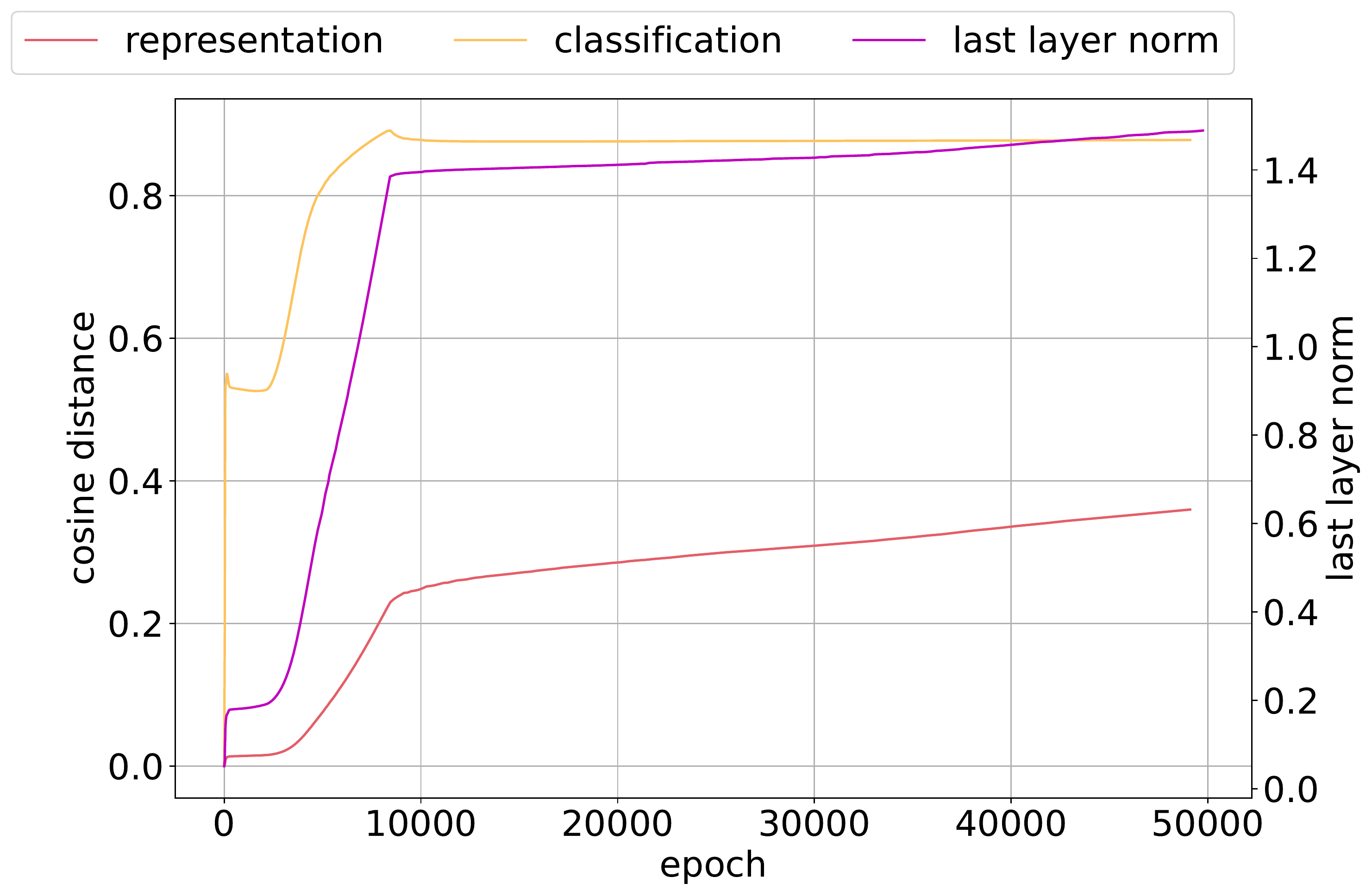} \\
      (d)  & (e) & (f) \\
      & $epsilon=10^{-07}$\\
      & \\
      \includegraphics[width=0.33\linewidth]{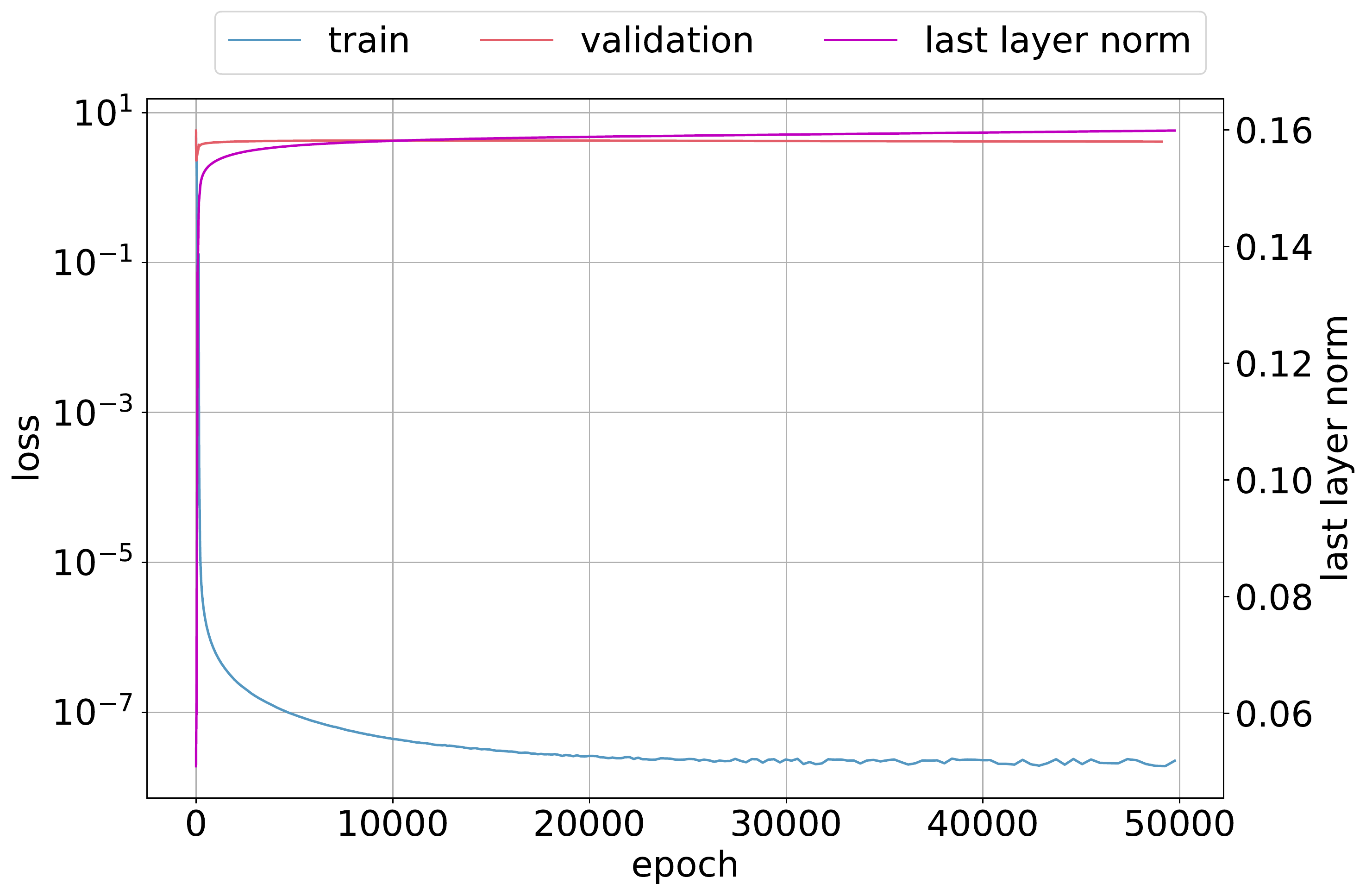} & 
      \includegraphics[width=0.33\linewidth]{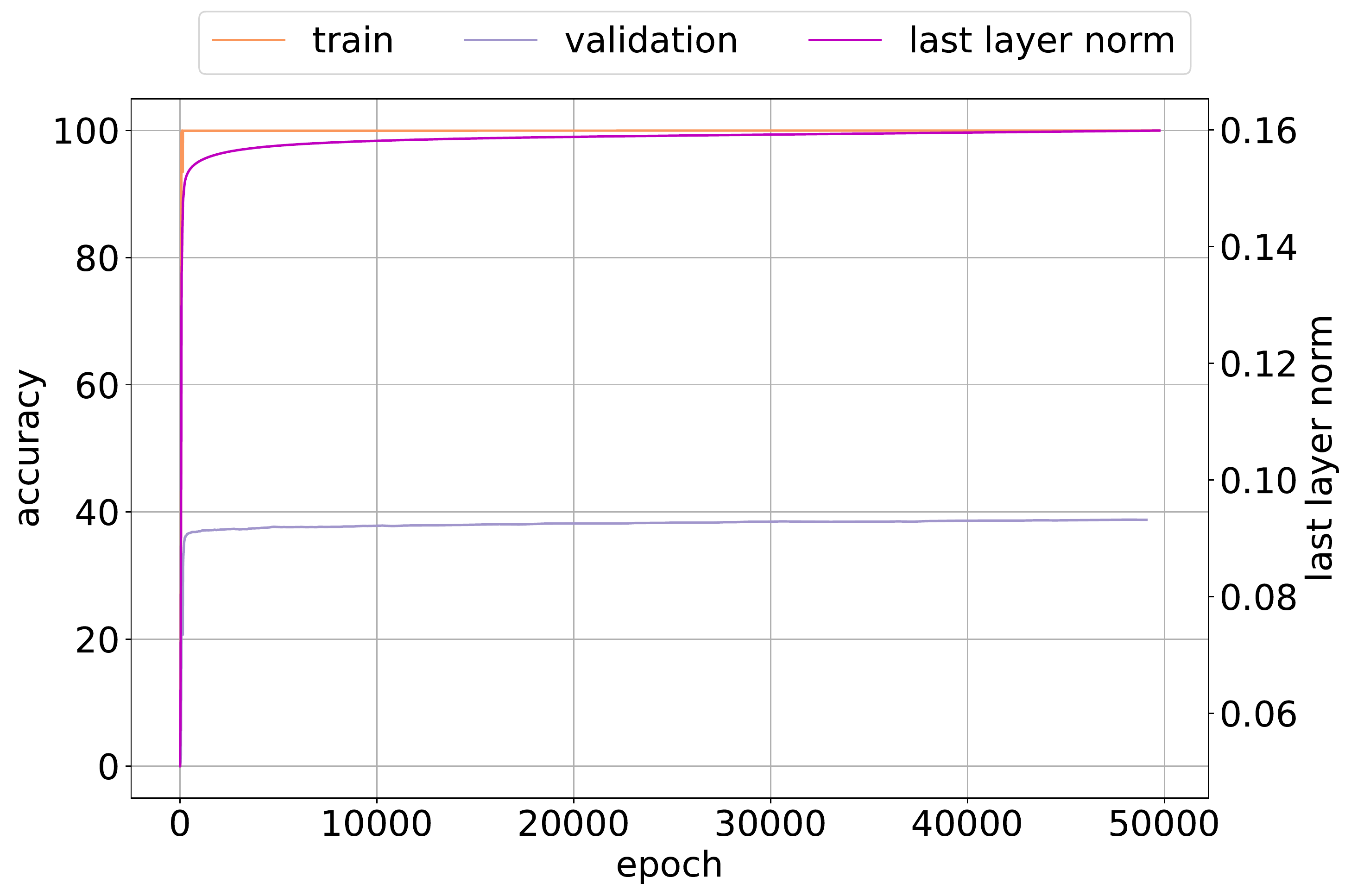} & 
      \includegraphics[width=0.33\linewidth]{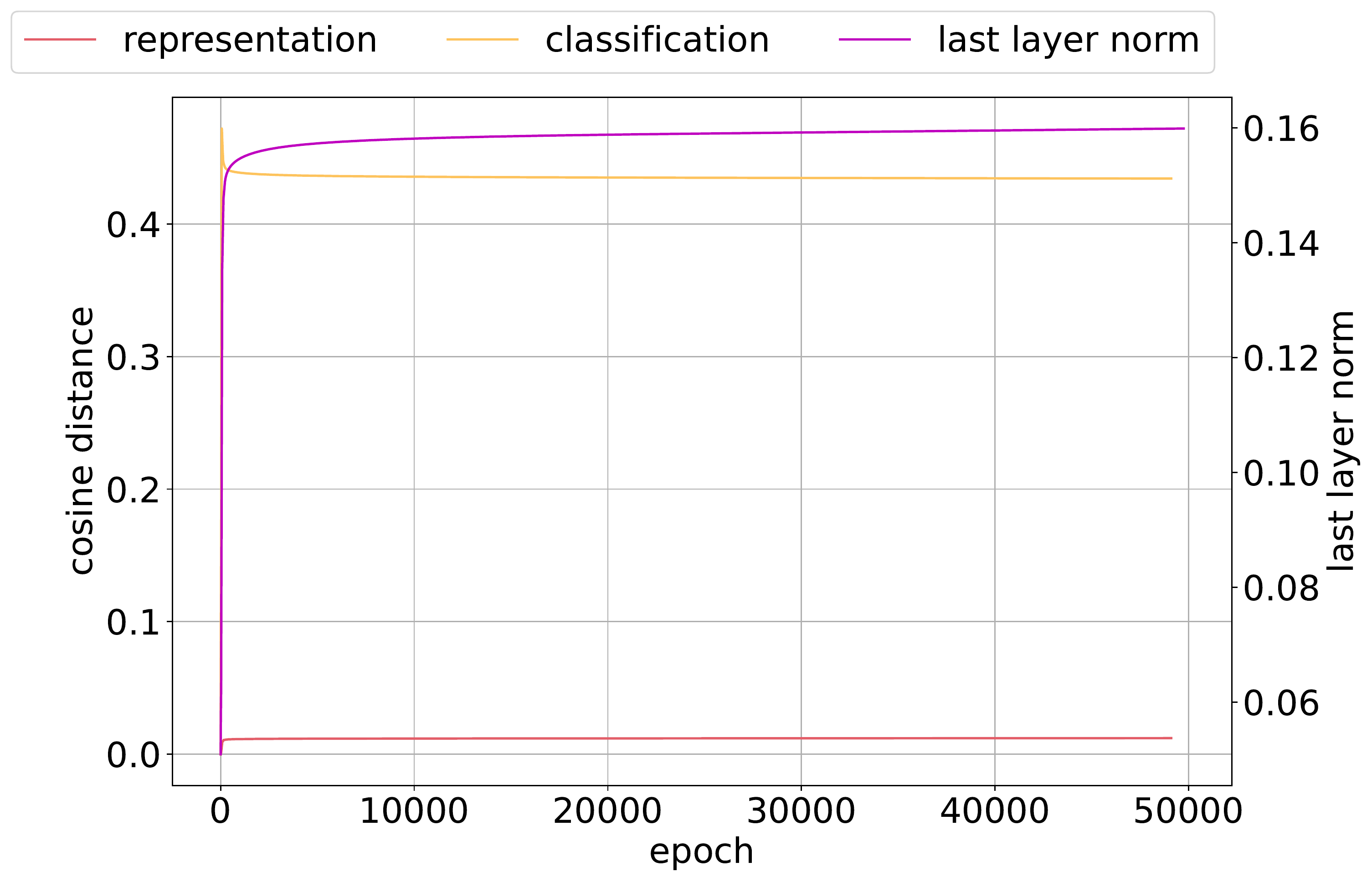} \\
      (g)  & (h) & (i) \\
      & $epsilon=10^{-05}$\\
      & \\
      \includegraphics[width=0.33\linewidth]{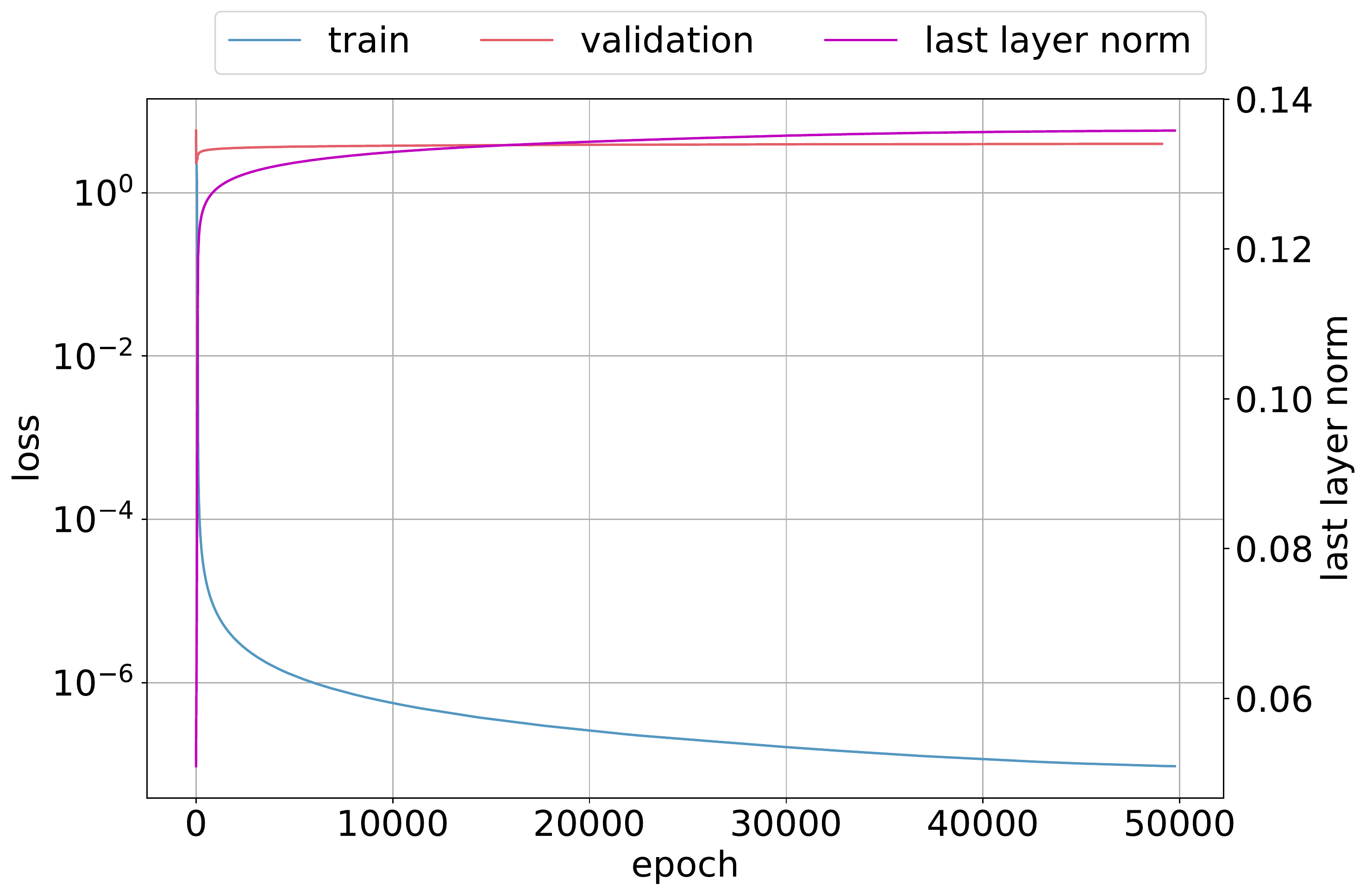} & 
      \includegraphics[width=0.33\linewidth]{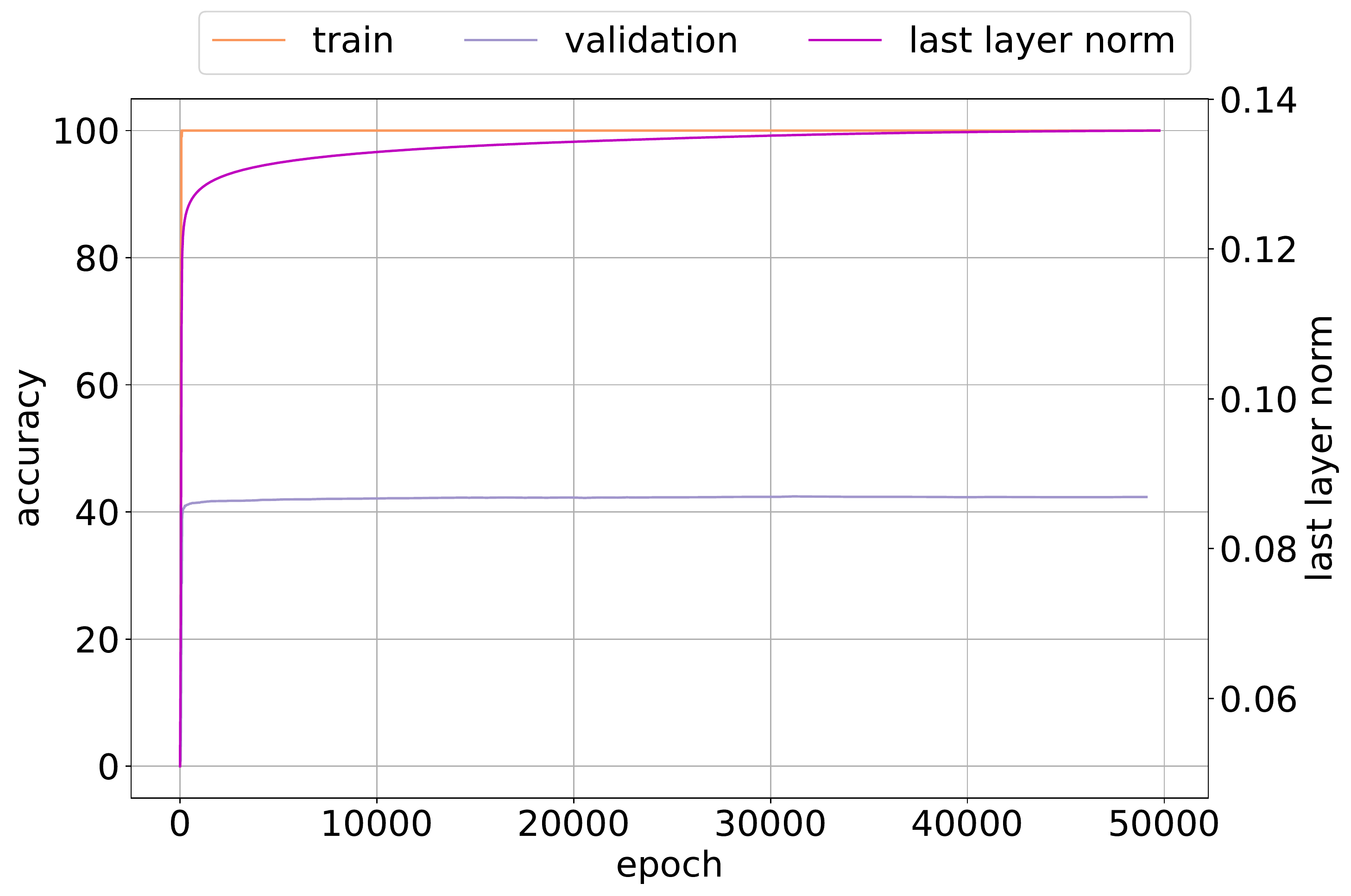} & 
      \includegraphics[width=0.33\linewidth]{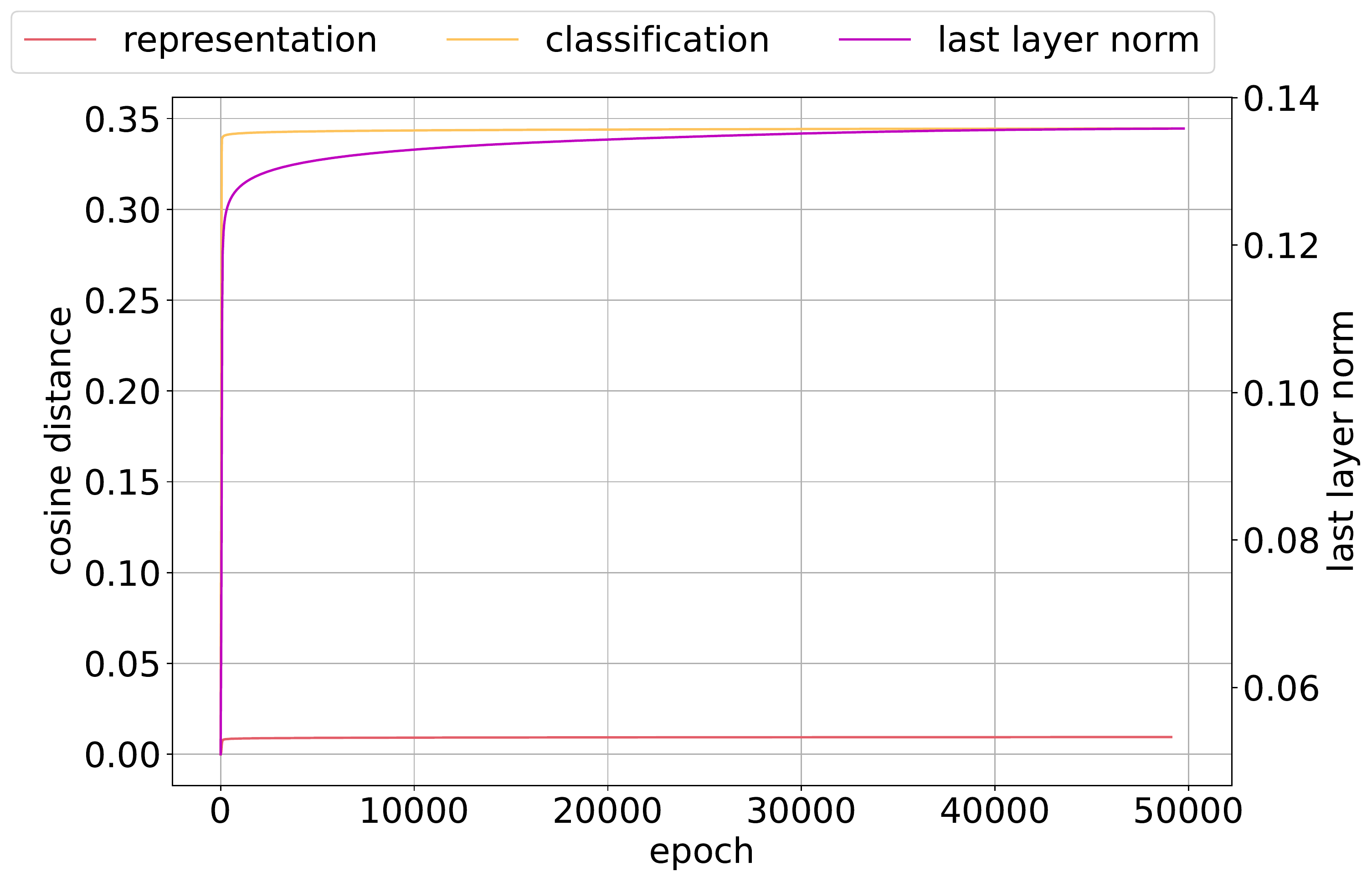} \\
      (j)  & (k) & (l) \\
      & $epsilon=10^{-04}$\\
      & \\
  \end{tabular}
 \caption {Cosine distance evolution for Transformer described in Appendix~\ref{appendix:xformers_setup} trained on modular addition. Observe that the cosine distance from initialization increases with models that experience Slingshot Effects.}
 \label{fig:slingshot_dist2init_add}
\end{figure*}

\begin{figure*}[h!]
\centering
  \begin{tabular}{ccc}
      loss & accuracy & cosine distance \\
      \includegraphics[width=0.33\linewidth]{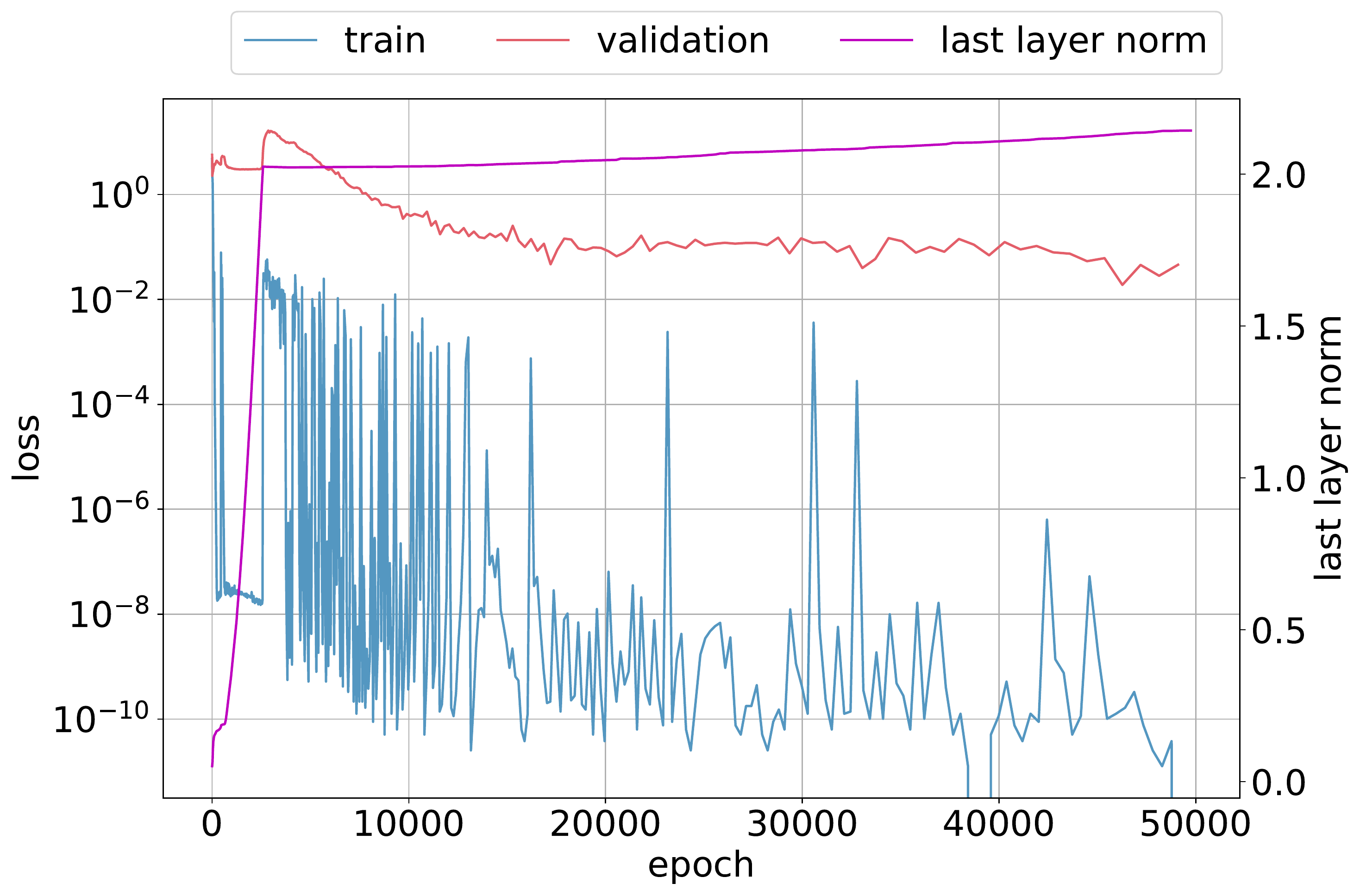} & 
      \includegraphics[width=0.33\linewidth]{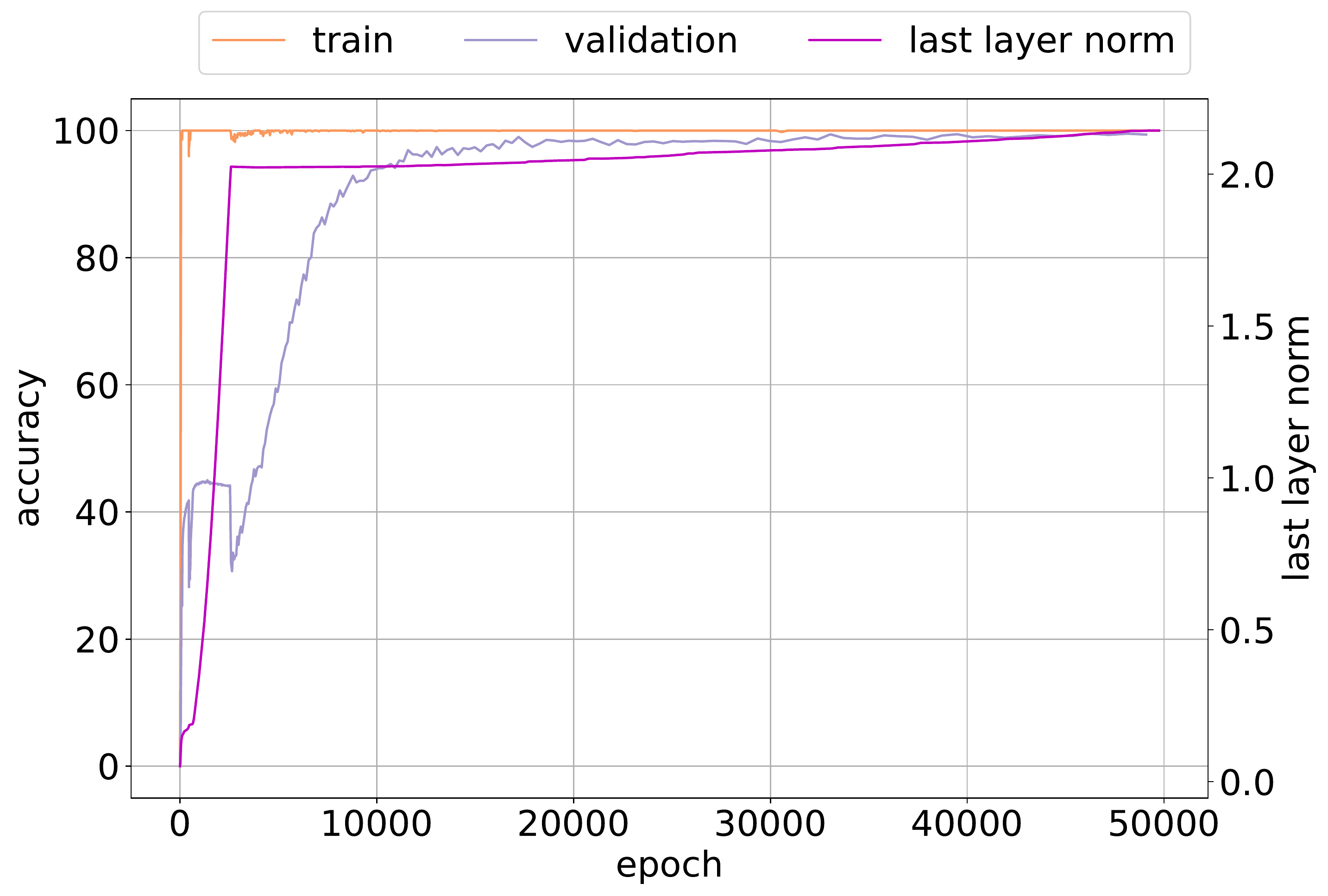} & 
      \includegraphics[width=0.33\linewidth]{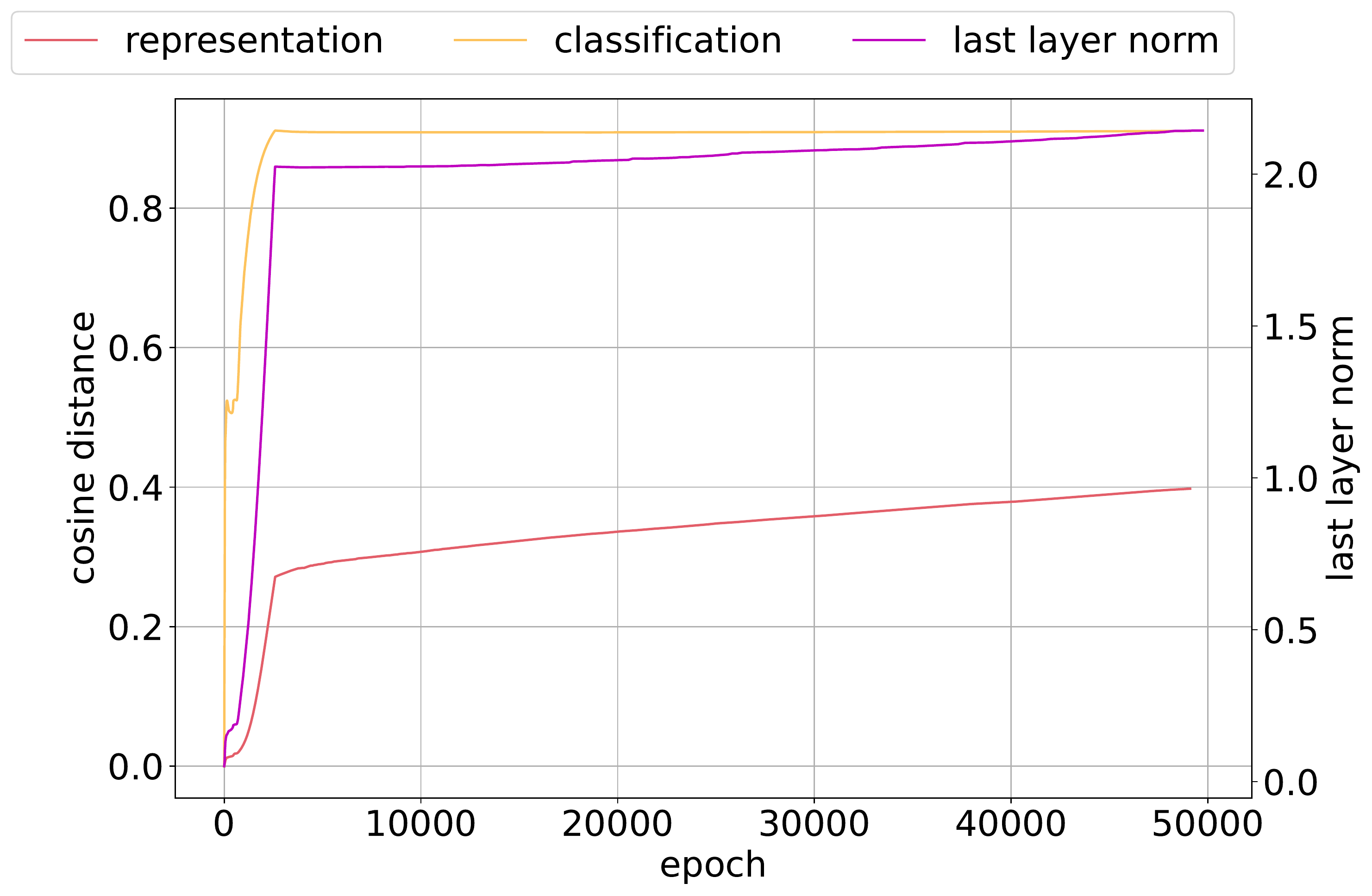} \\
      (a)  & (b) & (c) \\
      & $epsilon=10^{-08}$\\
      & \\
      \includegraphics[width=0.33\linewidth]{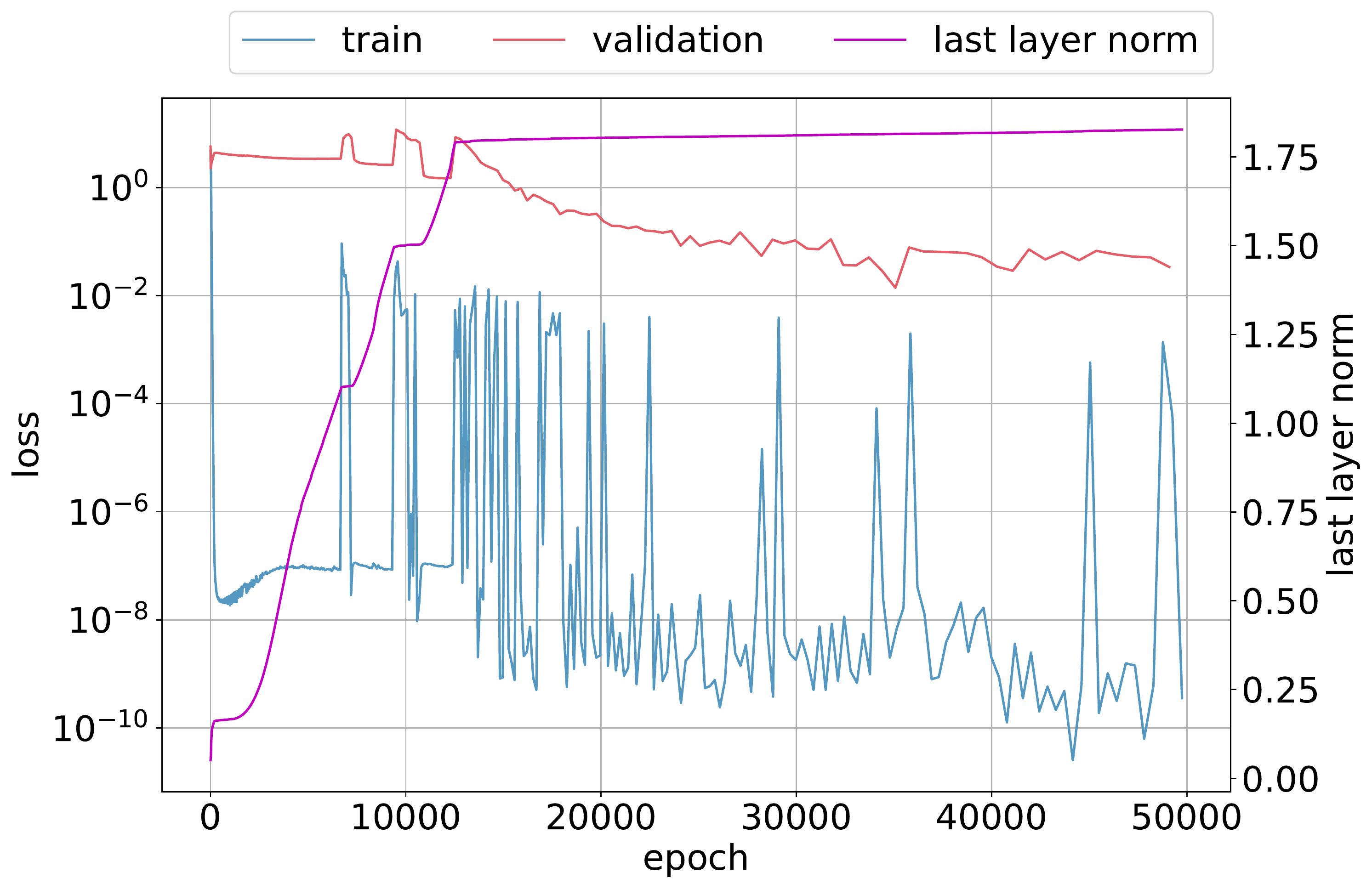} & 
      \includegraphics[width=0.33\linewidth]{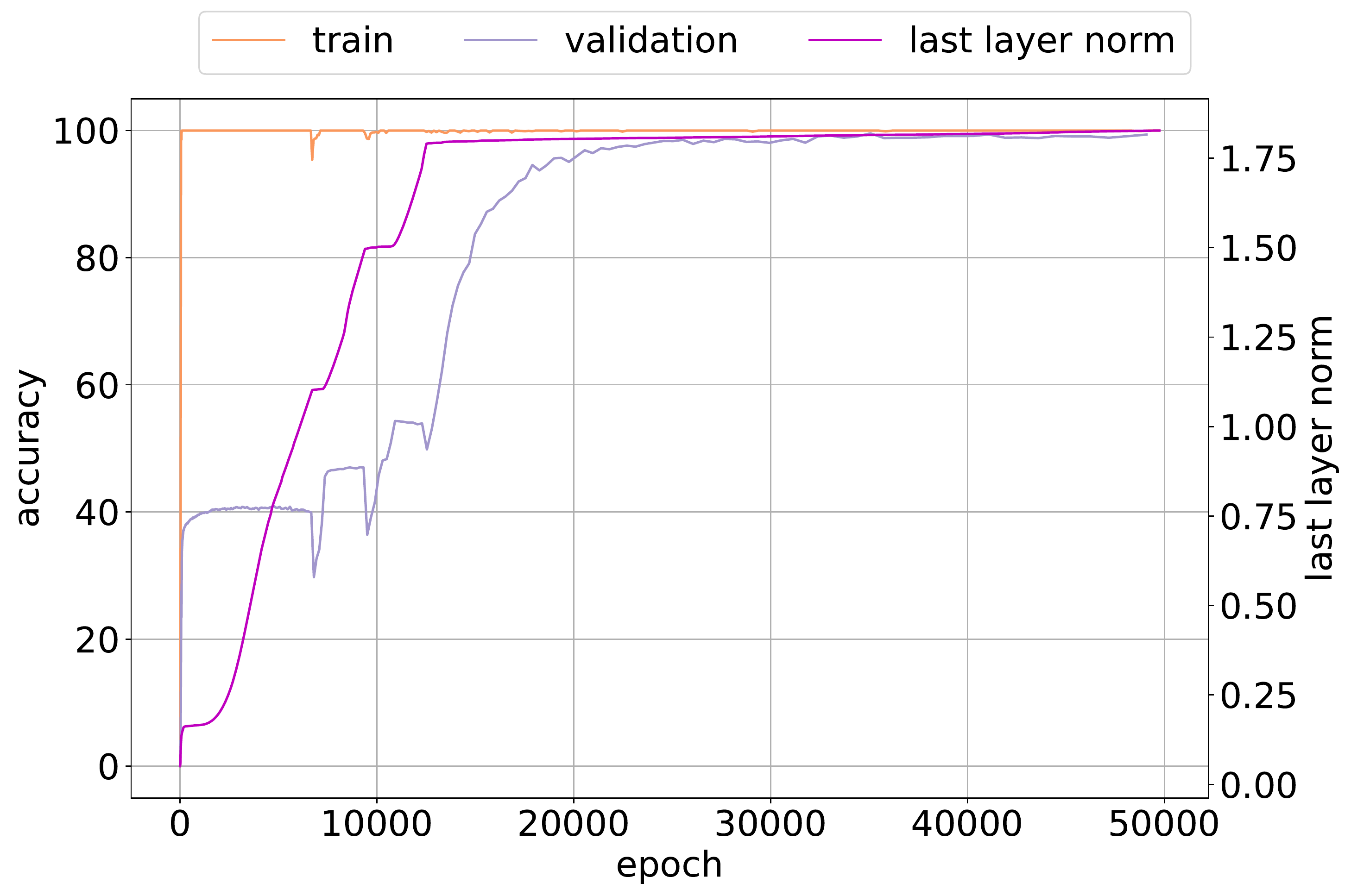} & 
      \includegraphics[width=0.33\linewidth]{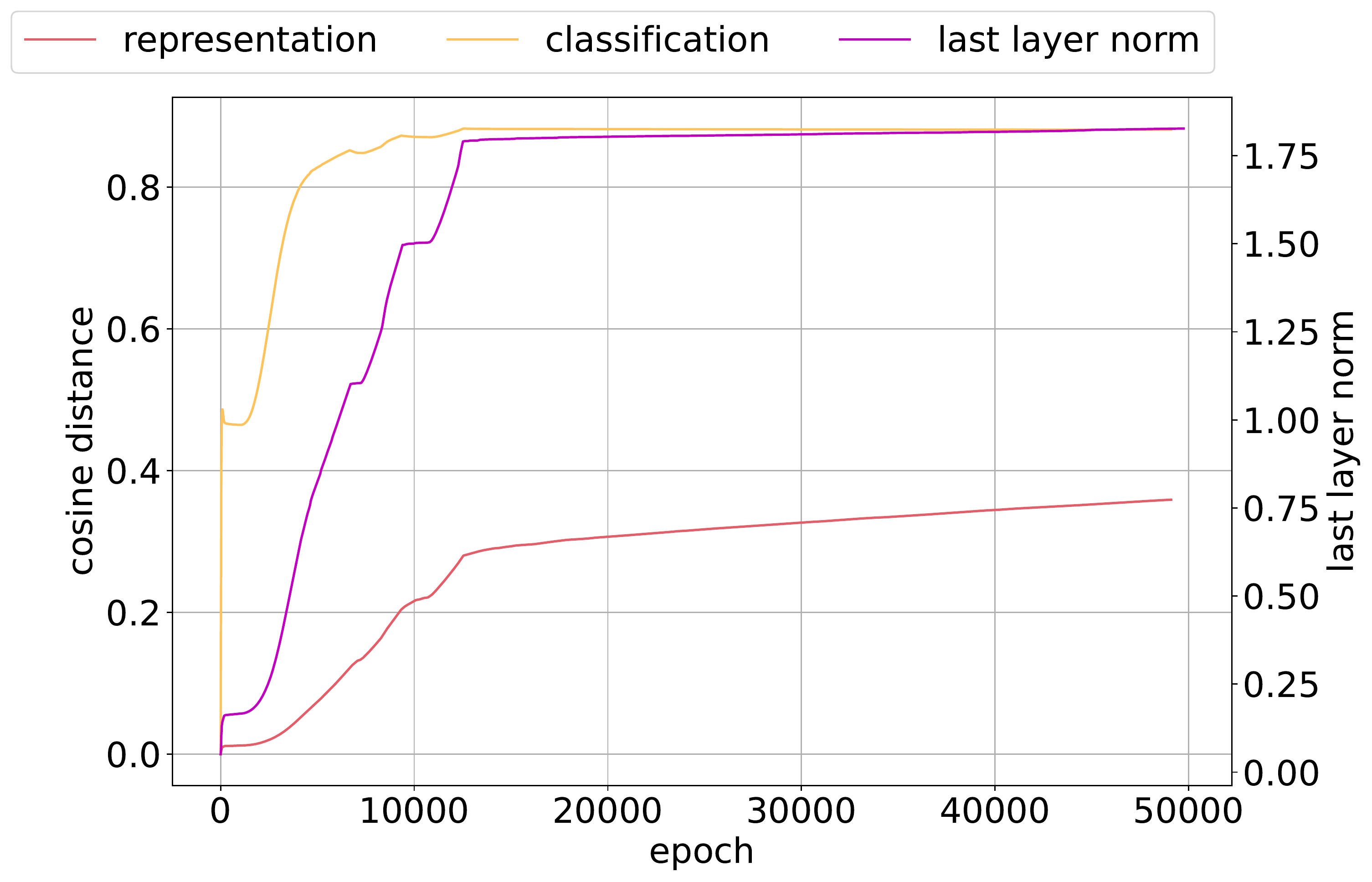} \\
      (d)  & (e) & (f) \\
      & $epsilon=10^{-07}$\\
      & \\
      \includegraphics[width=0.33\linewidth]{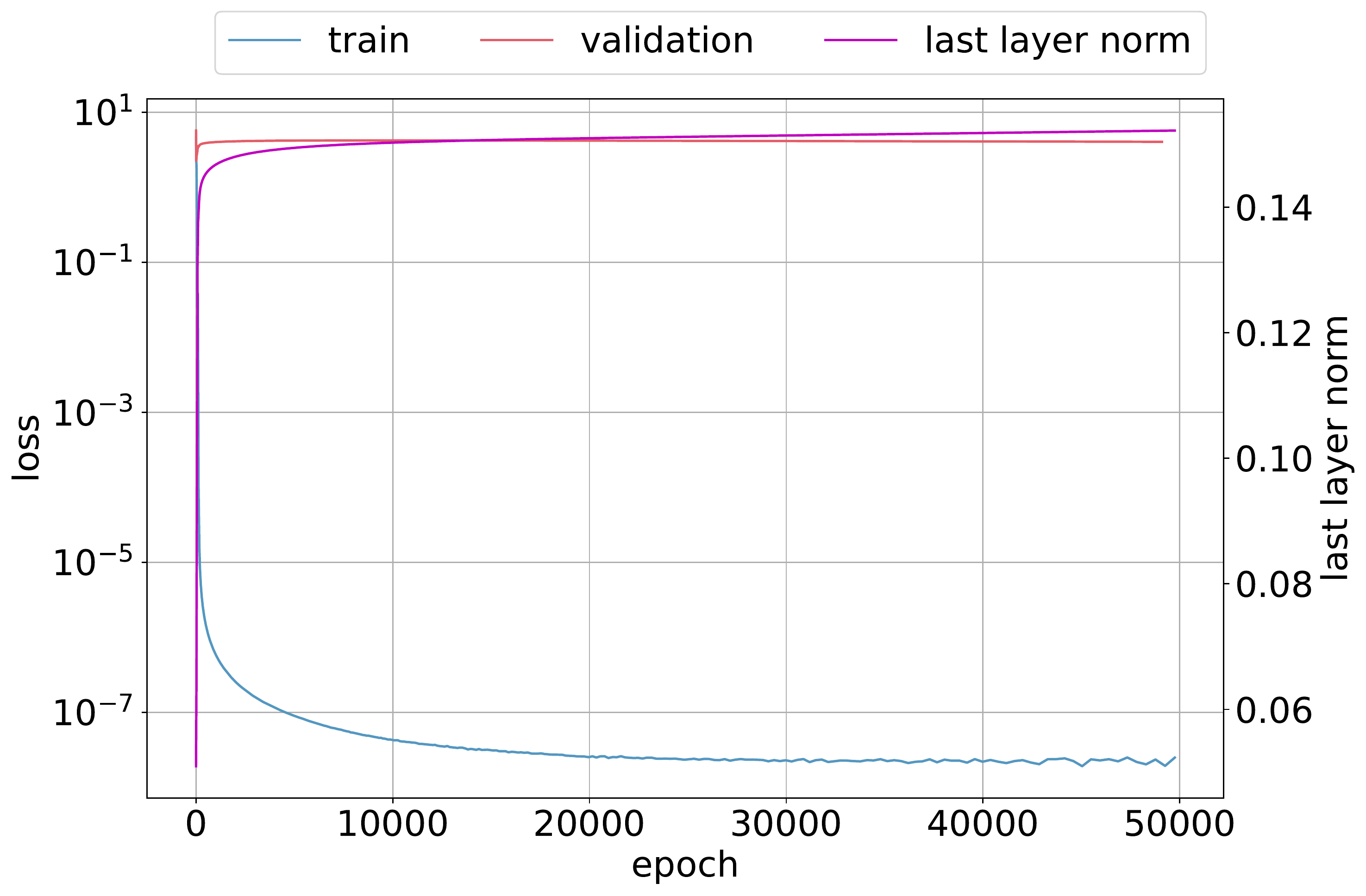} & 
      \includegraphics[width=0.33\linewidth]{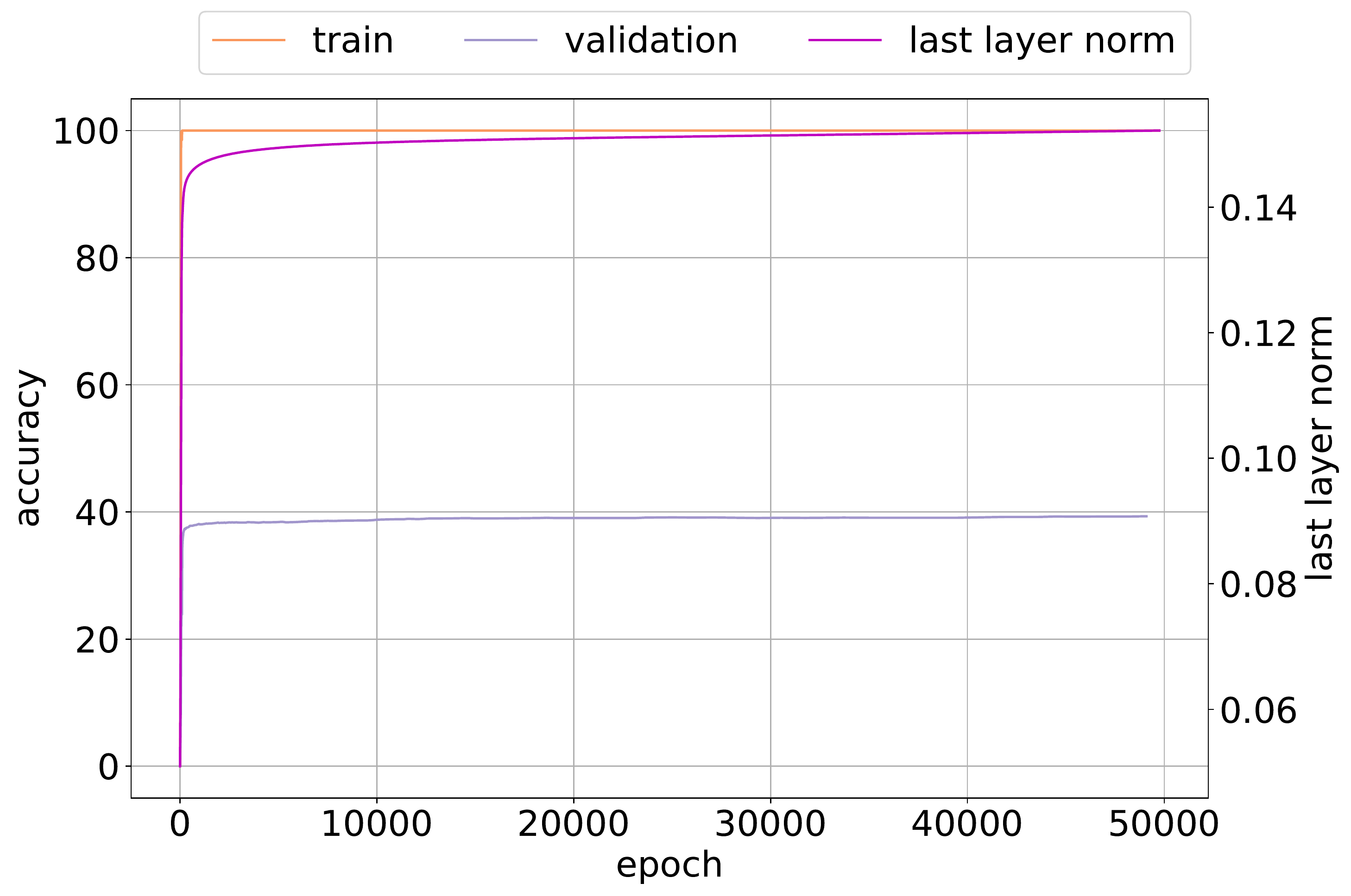} & 
      \includegraphics[width=0.33\linewidth]{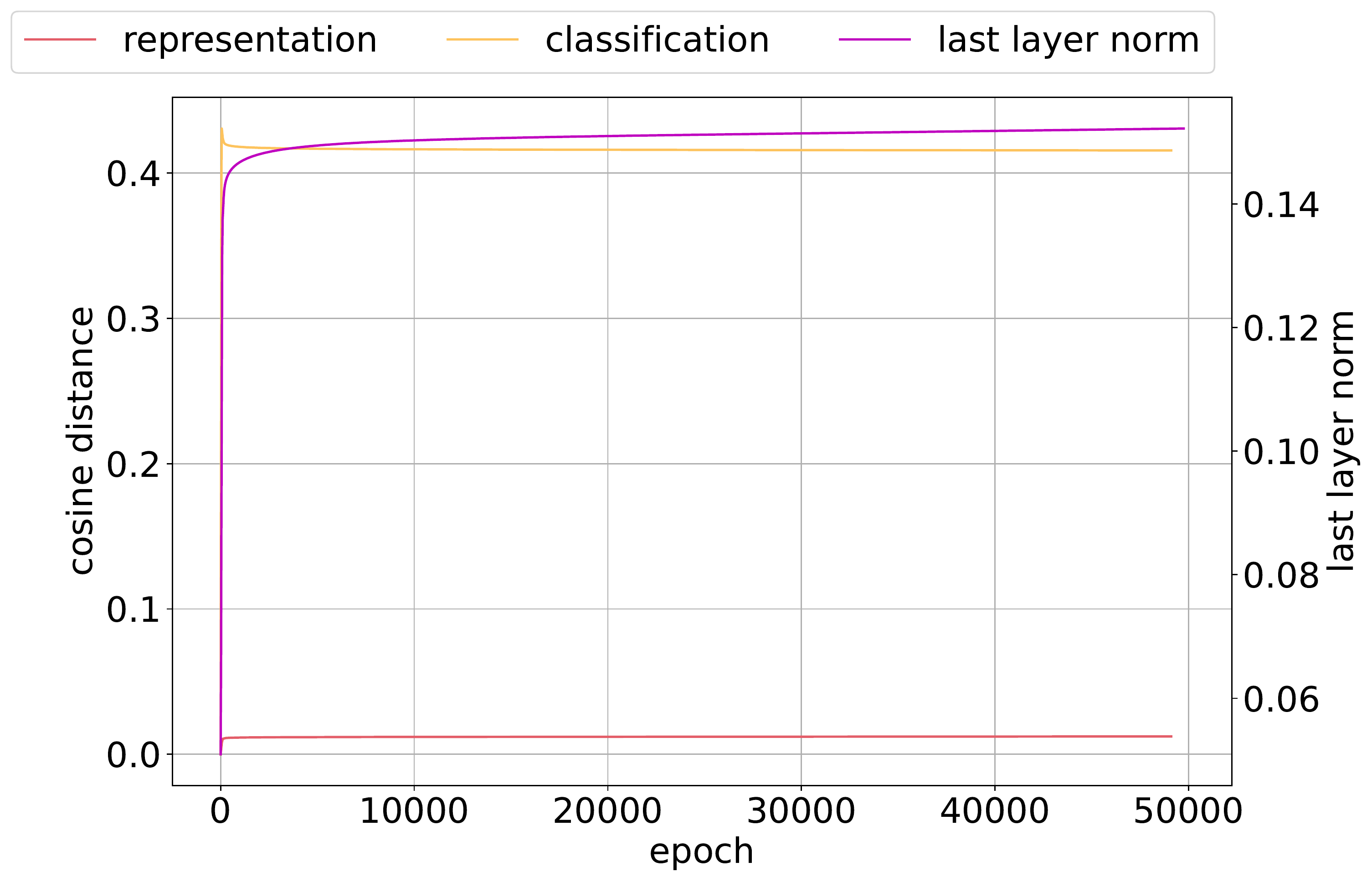} \\
       (g)  & (h) & (i) \\
      & $epsilon=10^{-05}$\\
      & \\
      \includegraphics[width=0.33\linewidth]{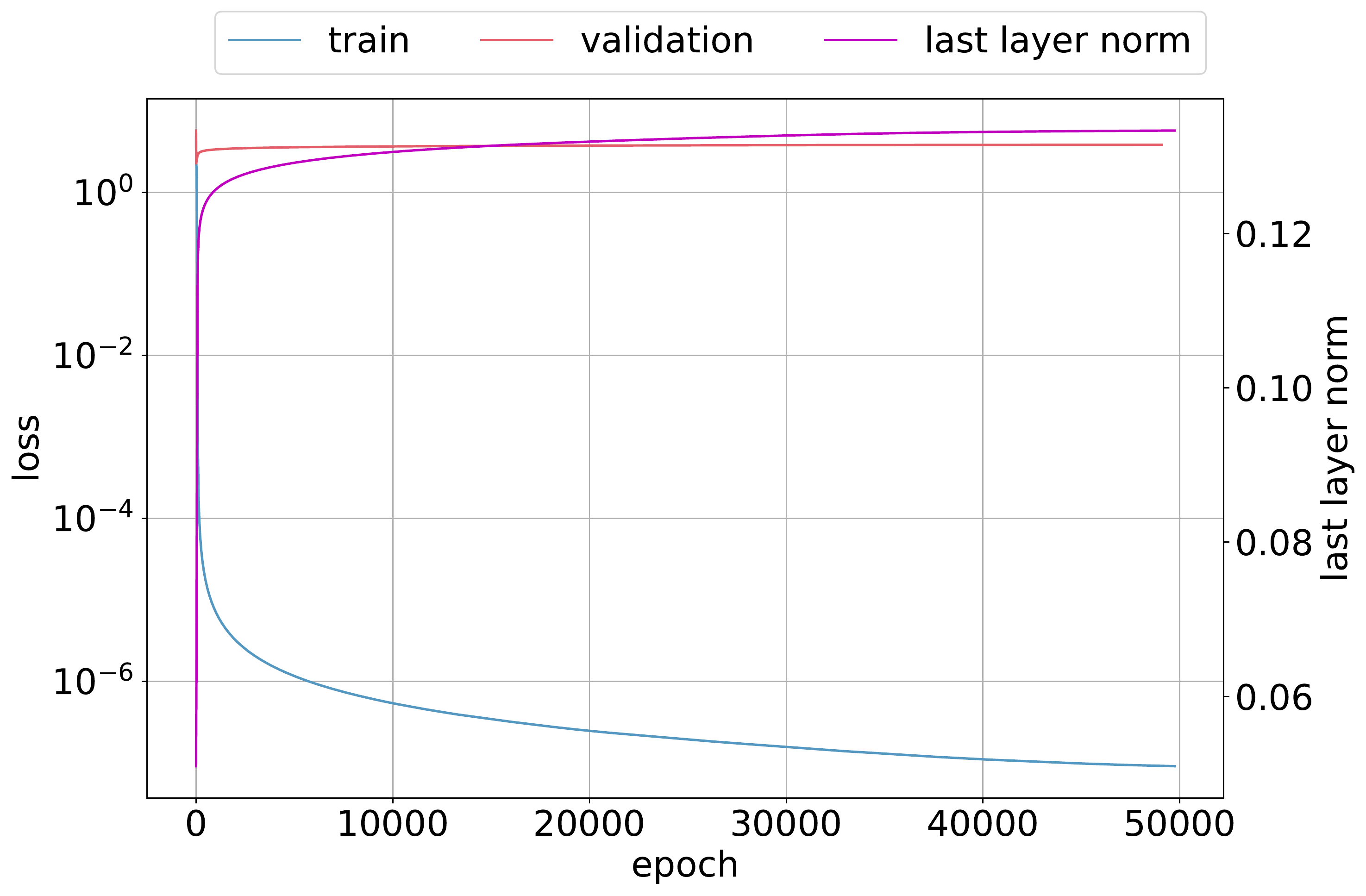} & 
      \includegraphics[width=0.33\linewidth]{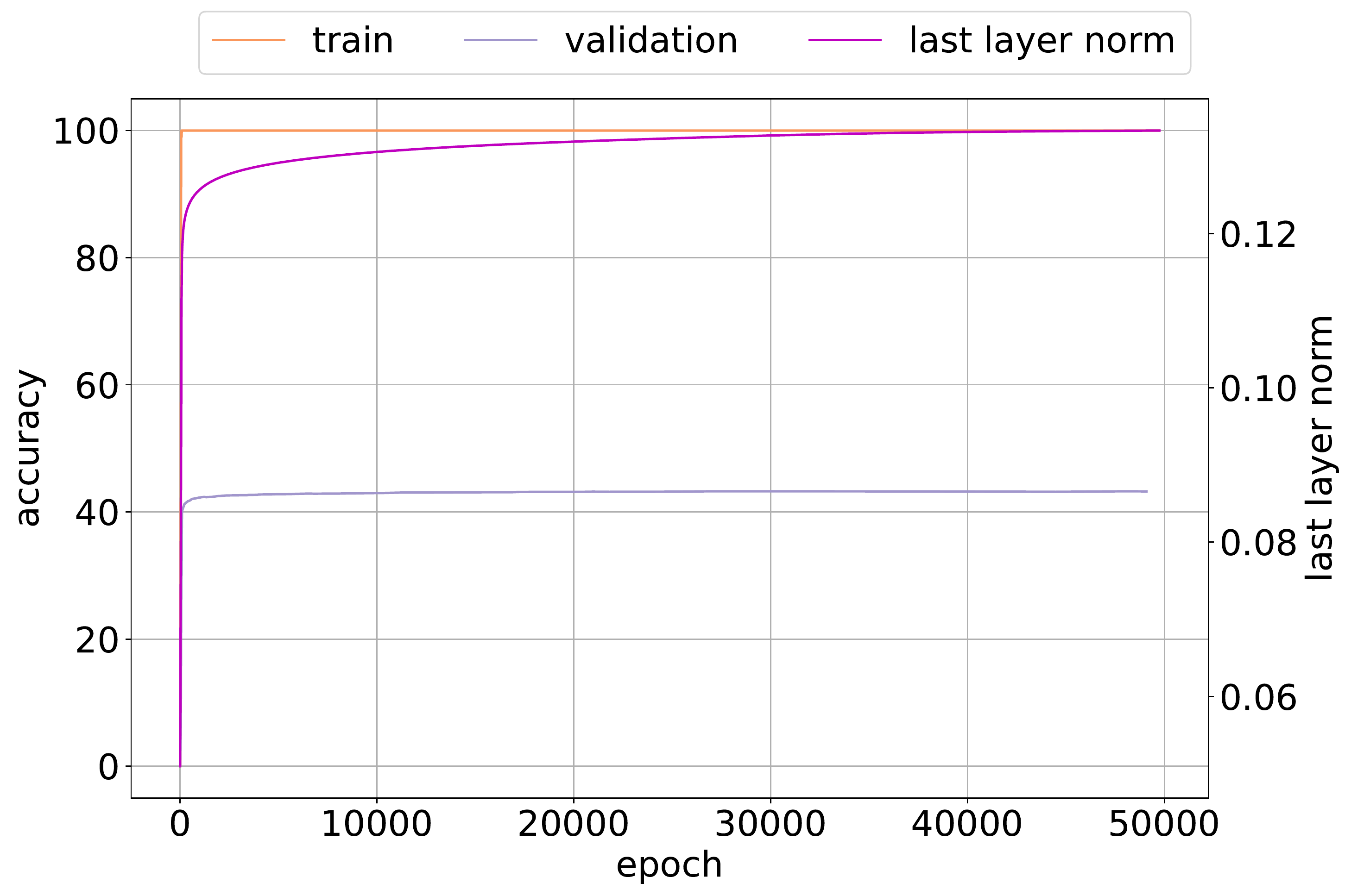} & 
      \includegraphics[width=0.33\linewidth]{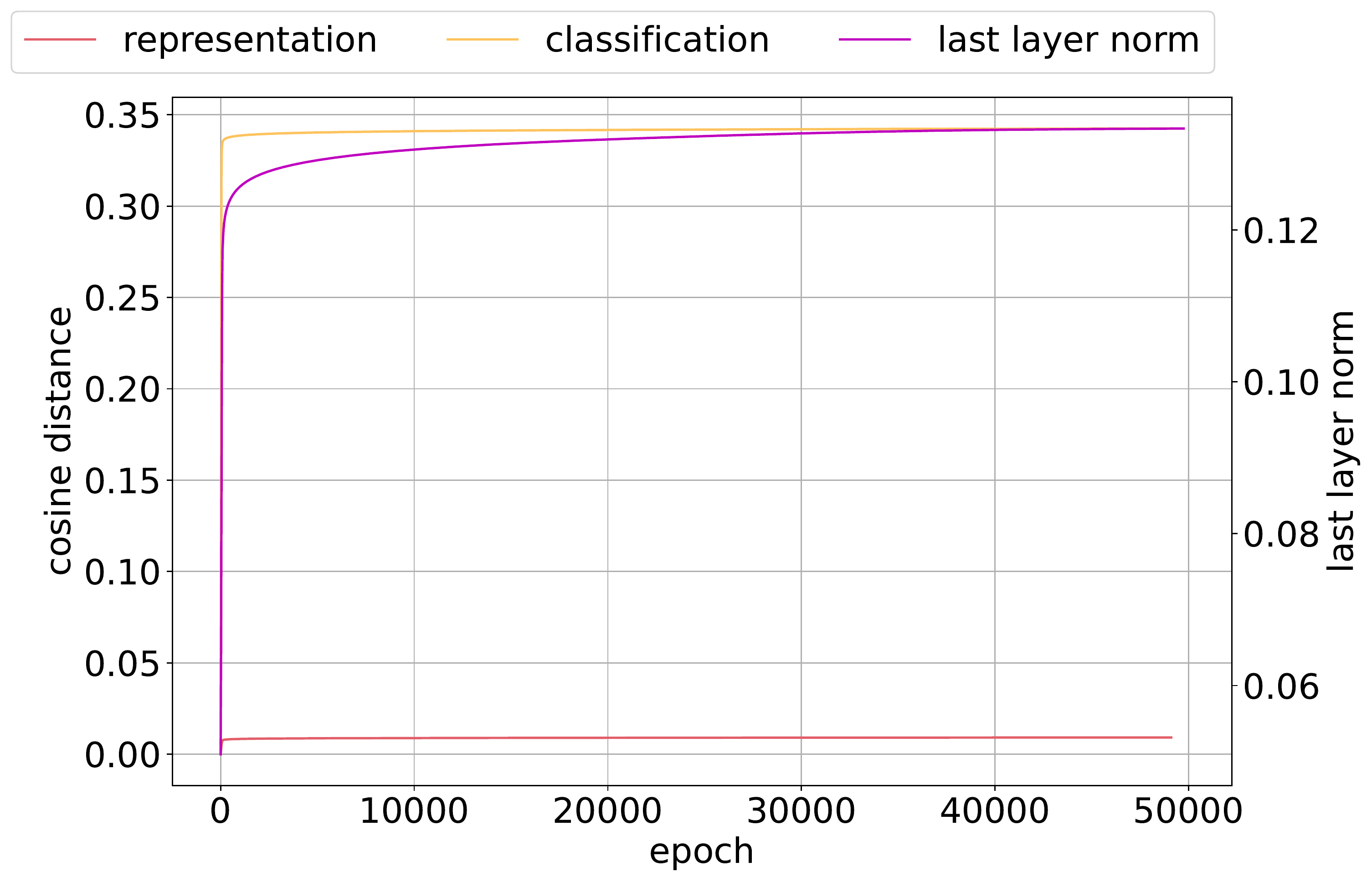} \\
      (j)  & (k) & (l)\\
      & $epsilon=10^{-04}$\\
      & \\
  \end{tabular}
 \caption {Cosine distance evolution for Transformer described in Appendix~\ref{appendix:xformers_setup} trained on modular multiplication. Observe that the cosine distance from initialization increases with models that experience Slingshot Effects.}
 \label{fig:slingshot_dist2init_mul}
\end{figure*}

\begin{figure*}[h!]
\centering
  \begin{tabular}{ccc}
      loss & accuracy & cosine distance \\
      \includegraphics[width=0.33\linewidth]{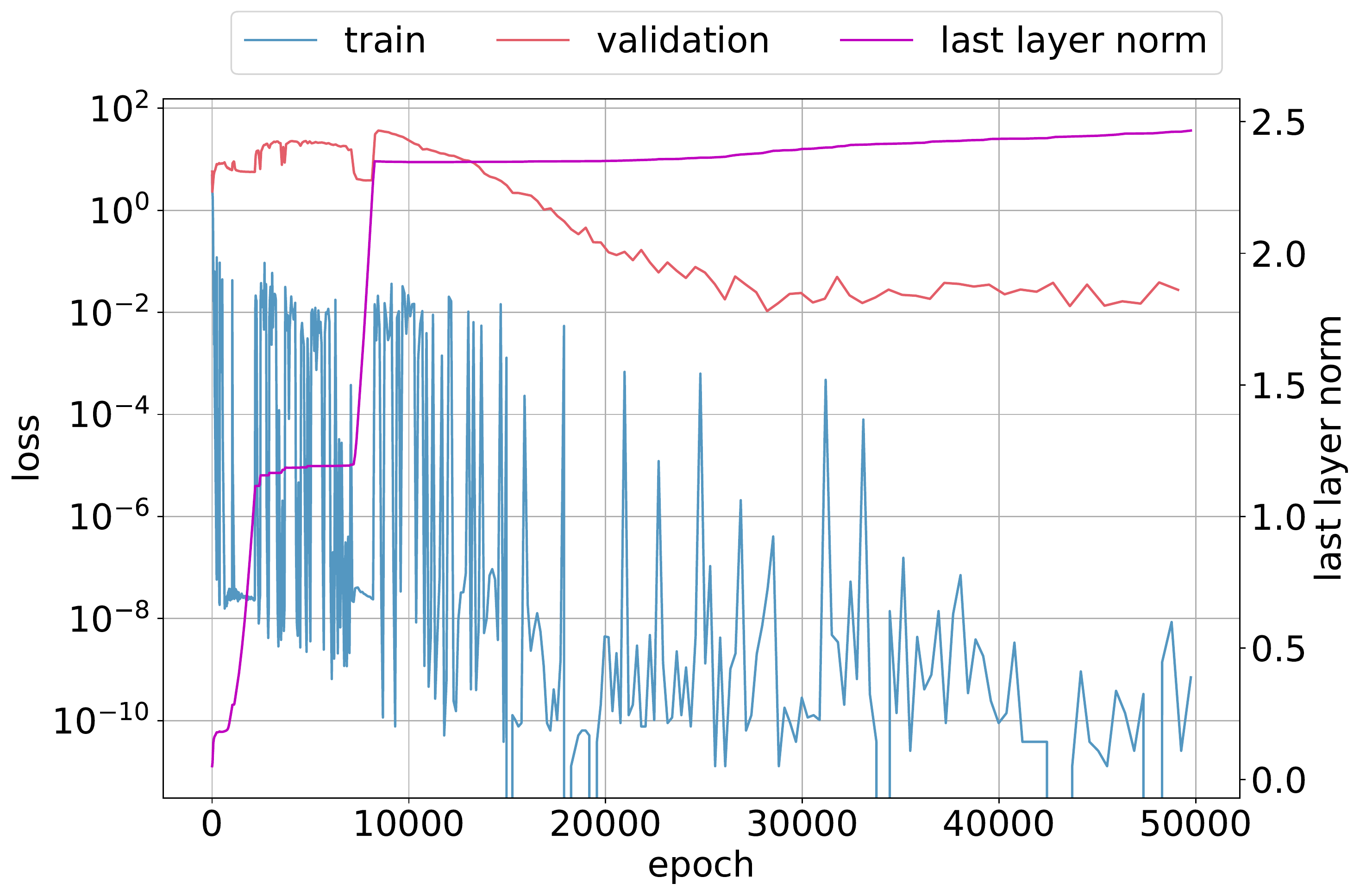} & 
      \includegraphics[width=0.33\linewidth]{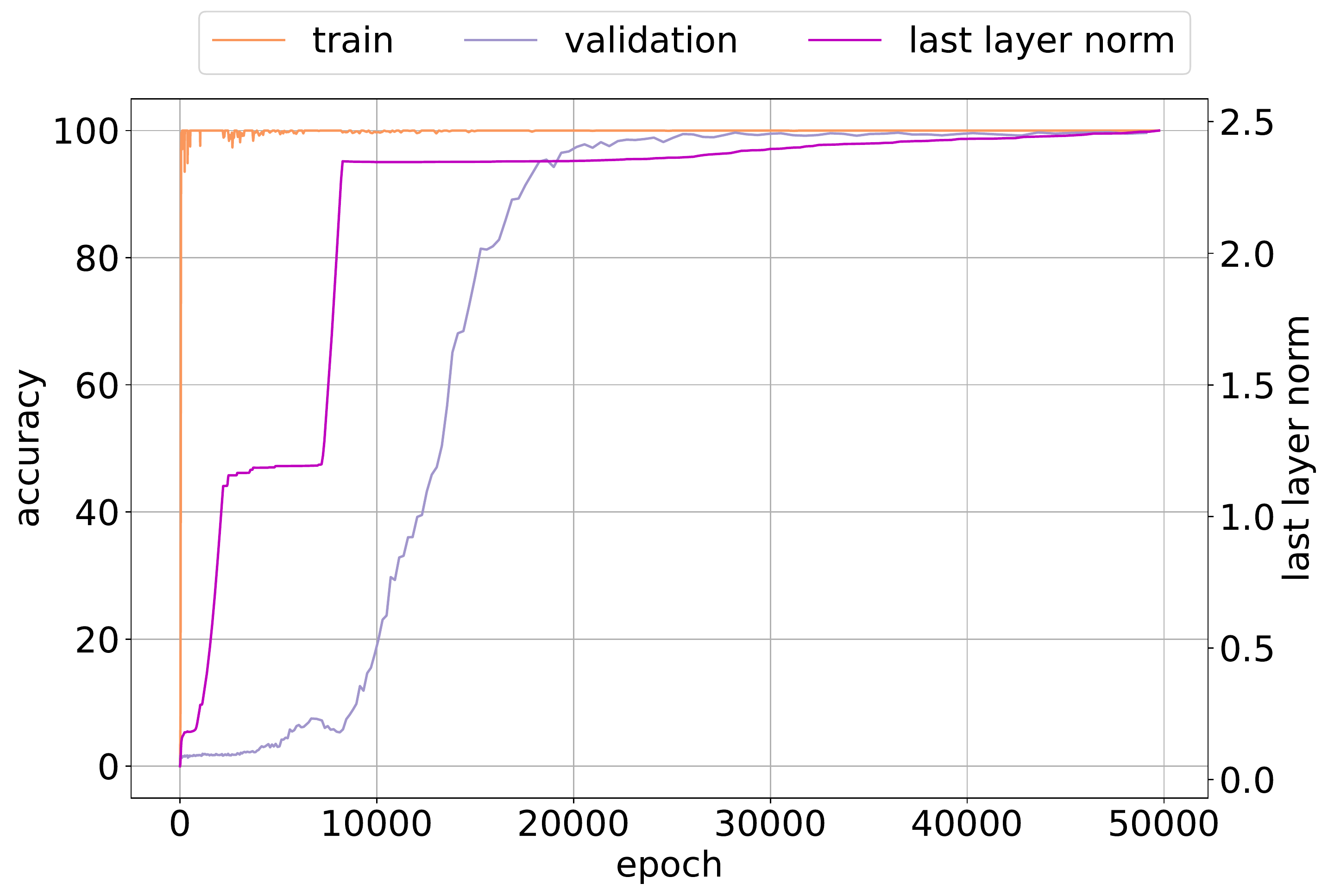} & 
      \includegraphics[width=0.33\linewidth]{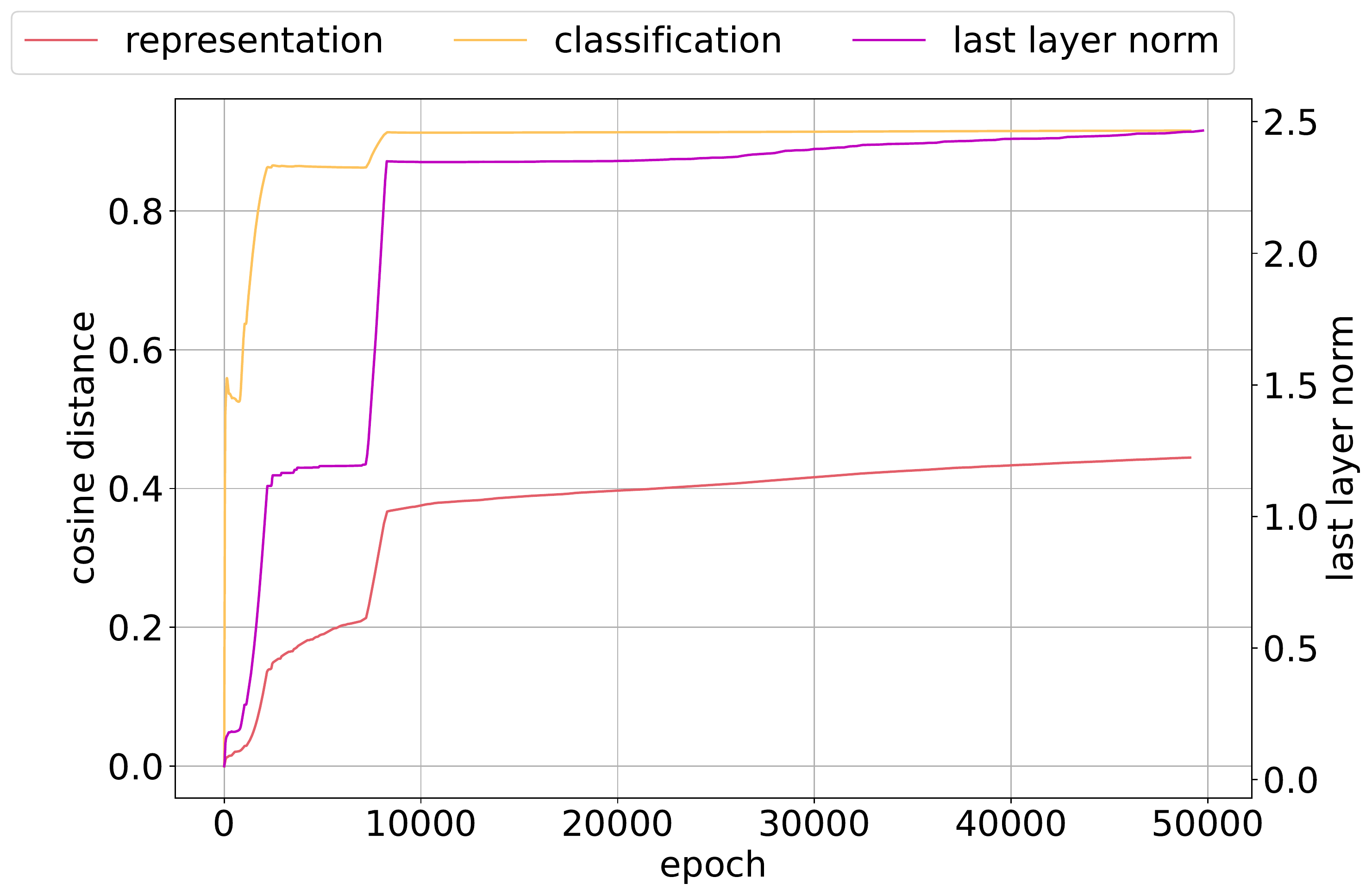} \\
      (a)  & (b) & (c) \\
      & $epsilon=10^{-08}$\\
      & \\
      \includegraphics[width=0.33\linewidth]{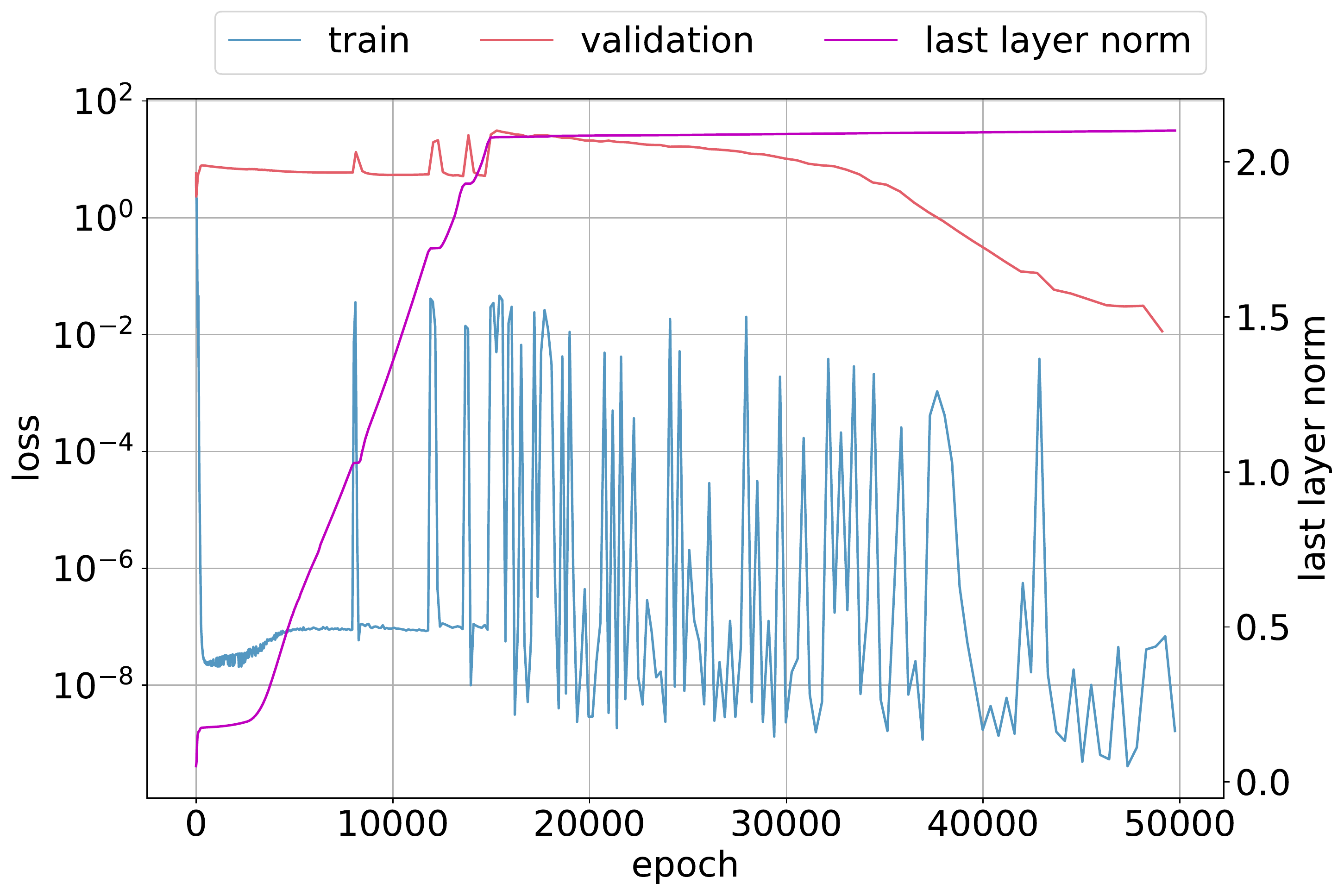} & 
      \includegraphics[width=0.33\linewidth]{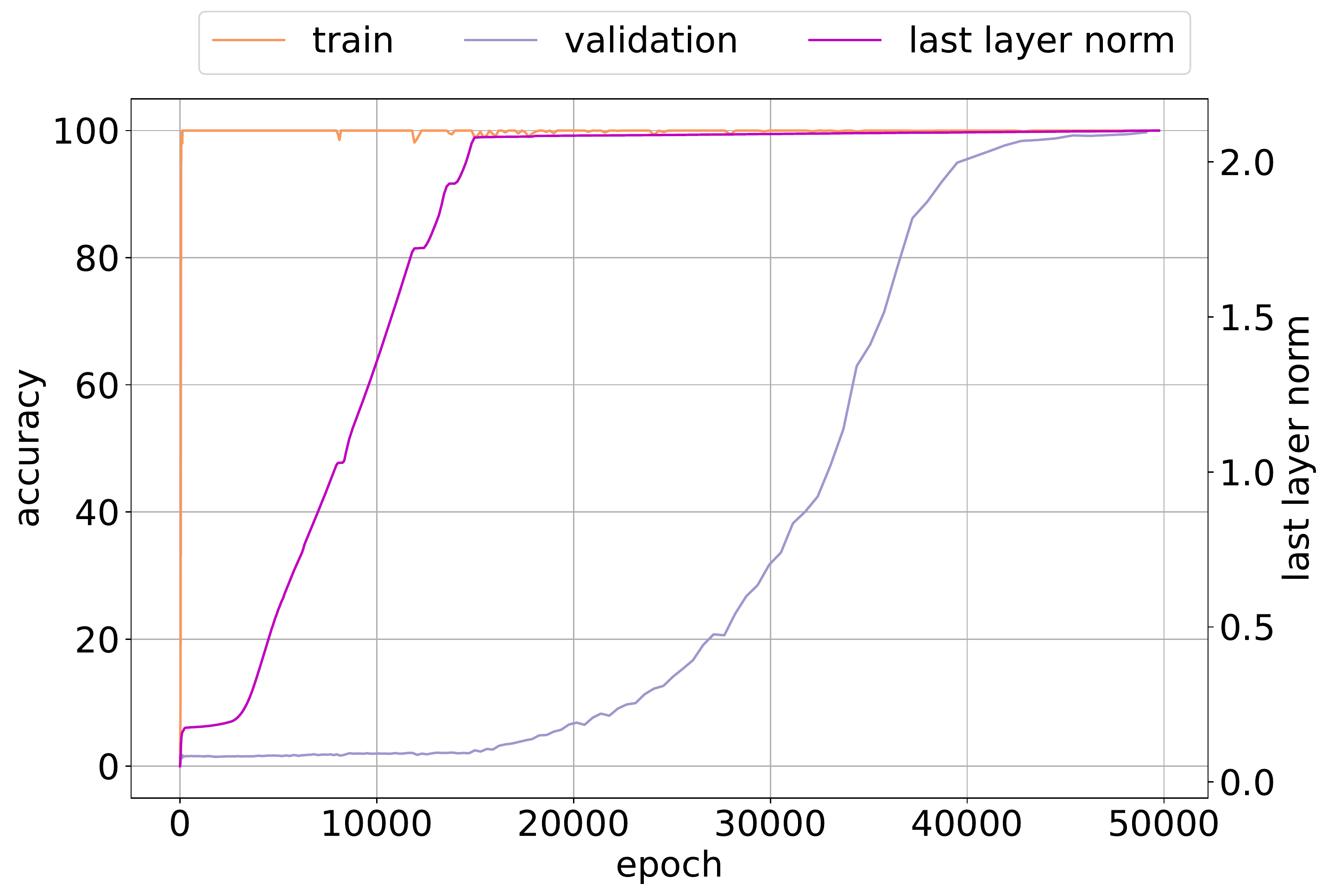} & 
      \includegraphics[width=0.33\linewidth]{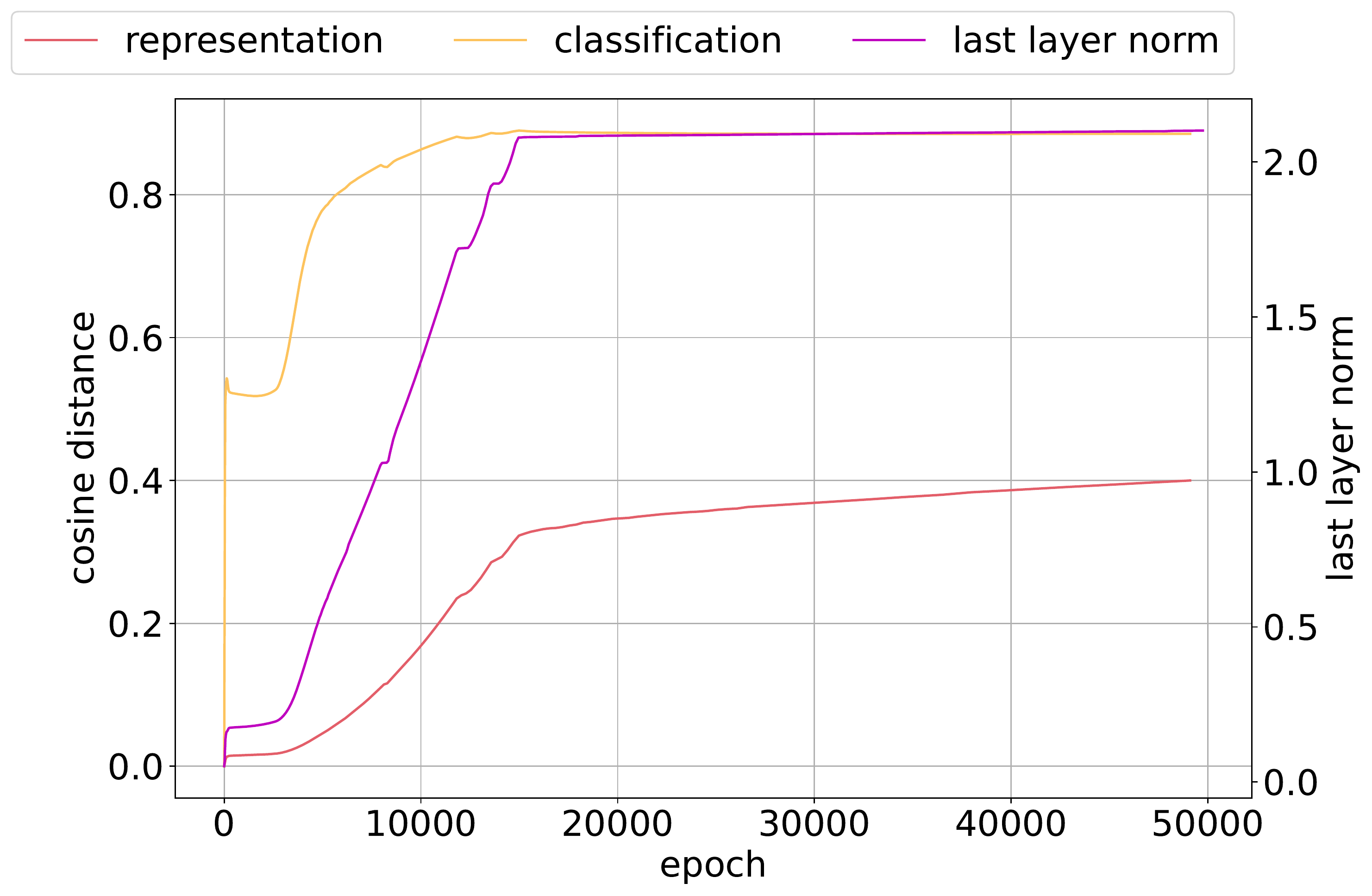} \\
      (d)  & (e) & (f) \\
      & $epsilon=10^{-07}$\\
      & \\
      \includegraphics[width=0.33\linewidth]{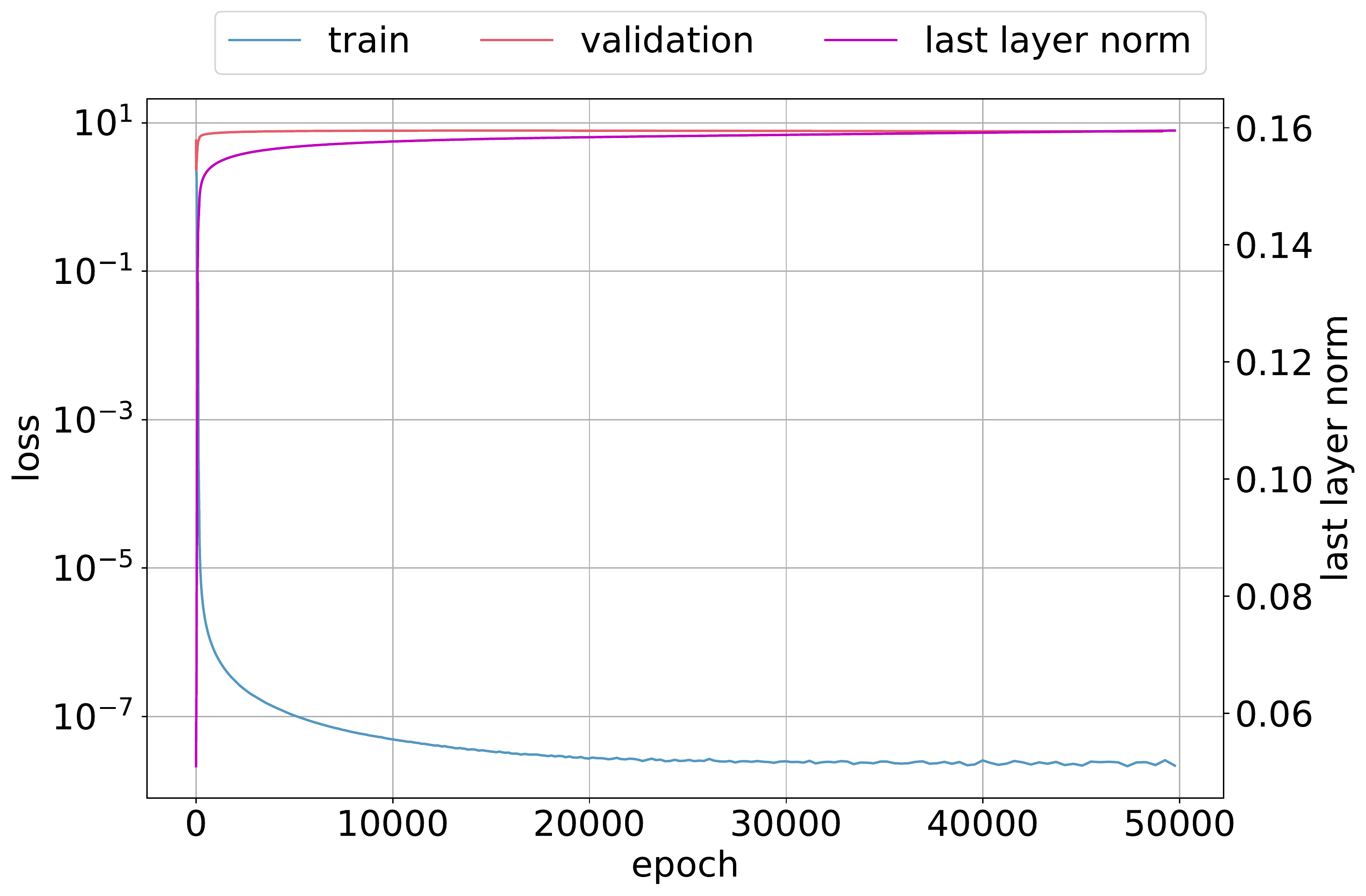} & 
      \includegraphics[width=0.33\linewidth]{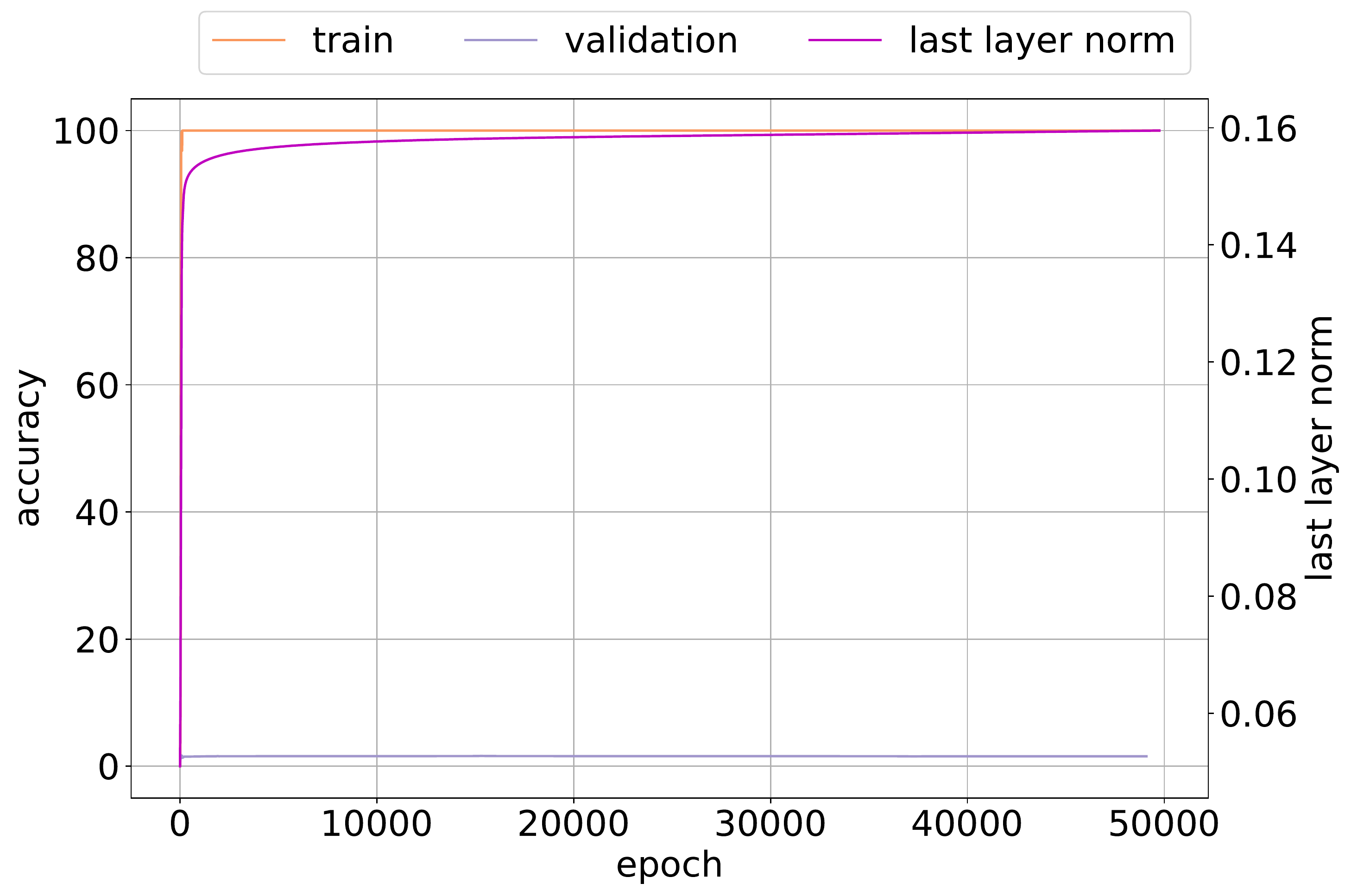} & 
      \includegraphics[width=0.33\linewidth]{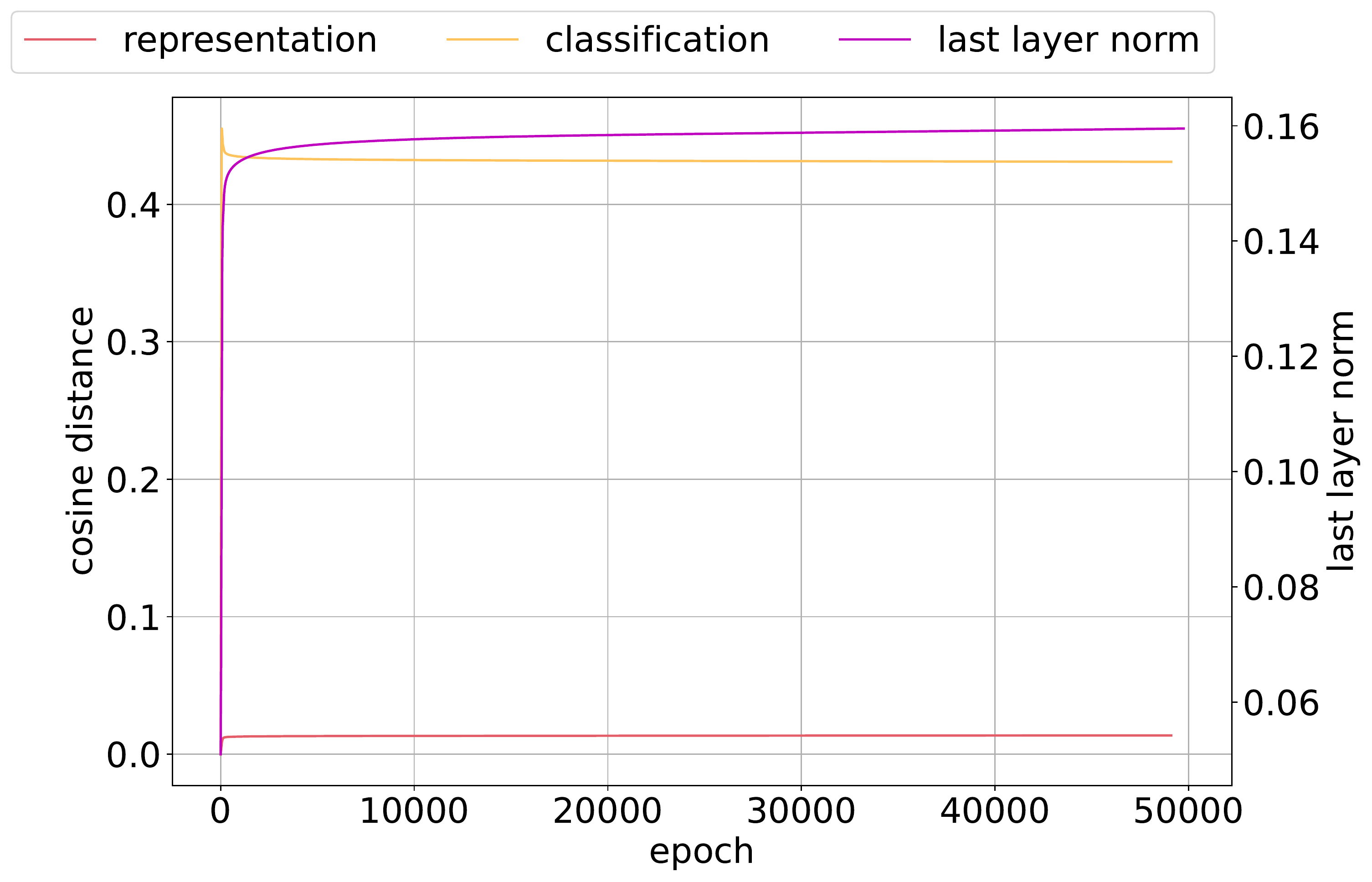} \\
      (g)  & (h) & (i) \\
      & $epsilon=10^{-05}$\\
      & \\
      \includegraphics[width=0.33\linewidth]{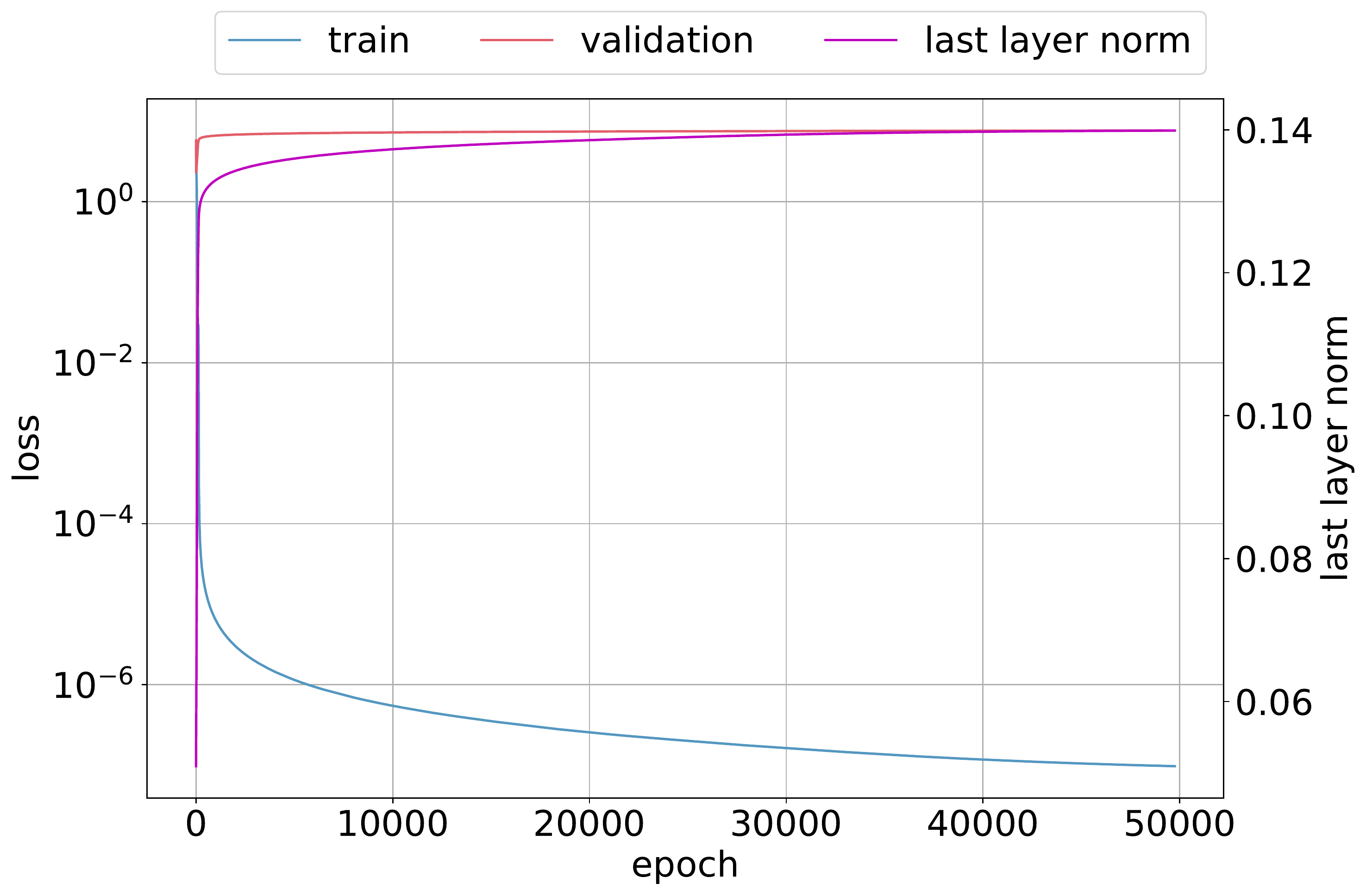} & 
      \includegraphics[width=0.33\linewidth]{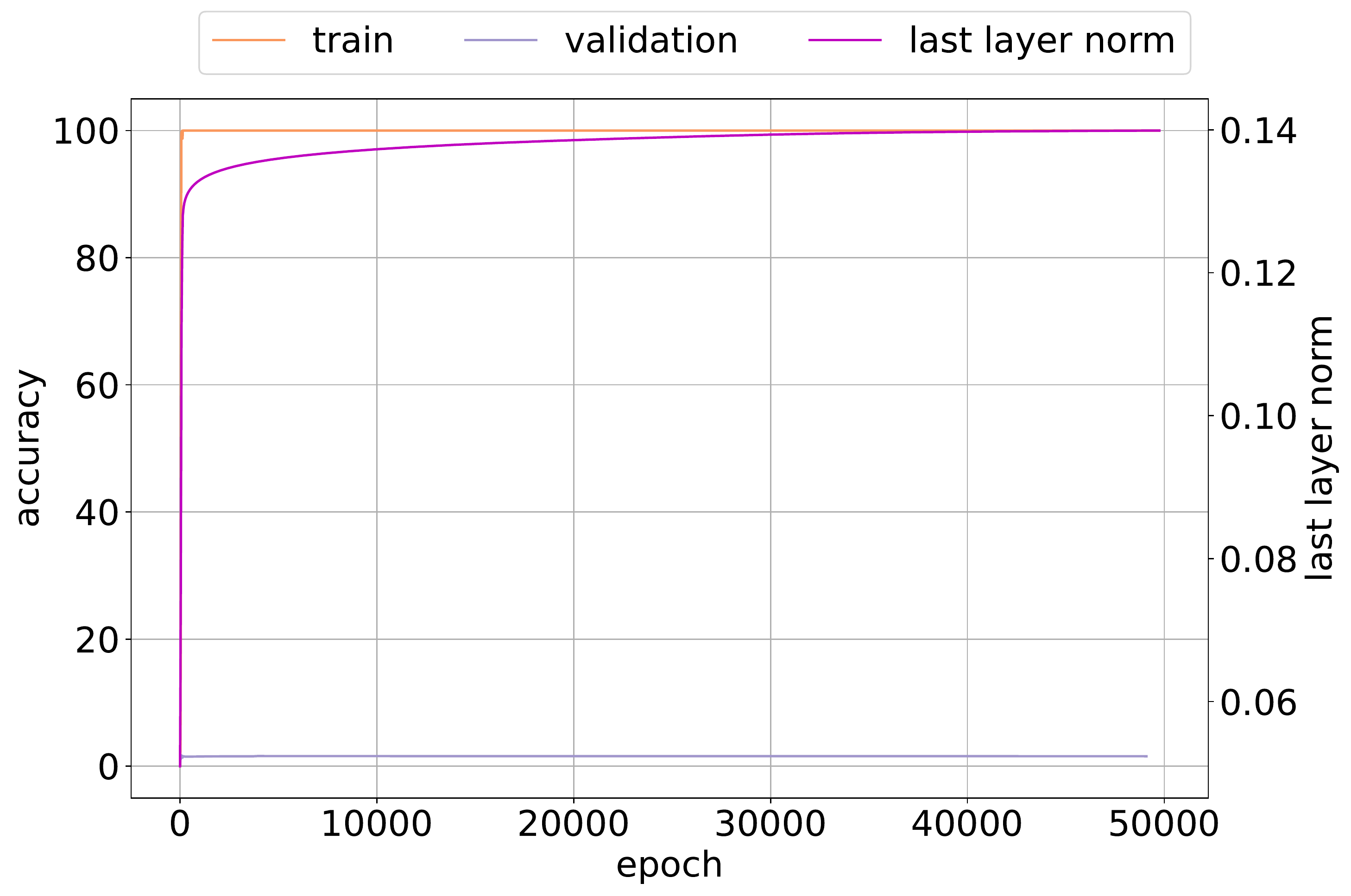} & 
      \includegraphics[width=0.33\linewidth]{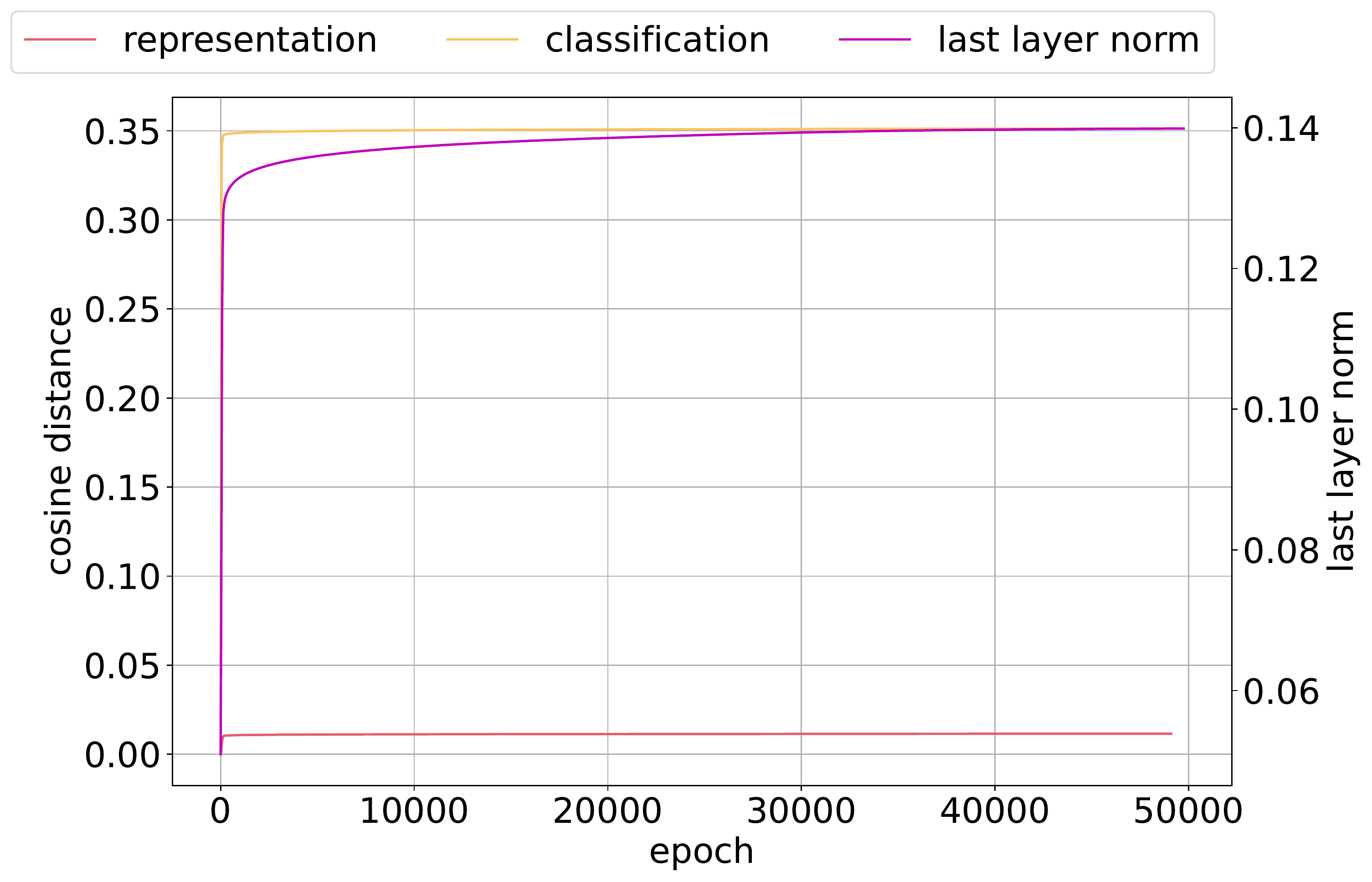} \\
      (j)  & (k) & (l) \\
      & $epsilon=10^{-04}$\\
      & \\
  \end{tabular}
 \caption {Cosine distance evolution for Transformer described in Appendix~\ref{appendix:xformers_setup} trained on modular division. Observe that the cosine distance from initialization increases with models that experience Slingshot Effects.}
 \label{fig:slingshot_dist2init_div}
\end{figure*}

A common observation is that intermediate representations tend to evolve beyond simple scale increase during phase transitions from norm growth to plateau.
In order to empirically quantify this effect, we train the Transformer described in Appendix~\ref{appendix:xformers_setup} with modular addition, multiplication and division datasets using Adam with learning rate set to $0.001$ and $\beta_{1}=0.9$ and $\beta_{2}=0.98$.
We calculate the cosine distance between the representation and classification parameters from their initial values where the cosine distance is given by
\begin{equation*}
d^{repr}= 1.0 - \frac{w^{repr}_{t}}{\lVert  w^{repr}_{t}  \rVert} \cdot \frac{w^{repr}_{0}}{\lVert  w^{repr}_{0}  \rVert}
\end{equation*}
\begin{equation*}
    d^{clf}= 1.0 - \frac{w^{clf}_{t}}{\lVert  w^{clf}_{t}  \rVert} \cdot \frac{w^{clf}_{0}}{\lVert  w^{clf}_{0}  \rVert}
\end{equation*}

where $d^{repr}$ ($d^{clf}$) denotes cosine distance for representation (respectively classification) parameters, $w^{repr}_{t}$ (resp. $w^{clf}_{t}$) denotes representation (resp. classification) parameters at time $t$ with $w^{repr}_{0}$ ($w^{clf}_{t}$) indicating the initial representation (resp. classification) parameters where the norm used above is the Euclidean norm.

Figure~\ref{fig:slingshot_dist2init_add} shows the dynamics of the loss, accuracy and cosine distance recorded during training. We observe that the classification parameters move farther away from initialization faster than the representation parameters. More interestingly, we observe from Figure~\ref{fig:slingshot_dist2init_add}c and Figure~\ref{fig:slingshot_dist2init_add}f that the representation parameters travel farther from initialization for training runs that experience Slingshot. These trials use $\epsilon=10^{-08}$ and $\epsilon=10^{-07}$ and experience Slingshot Effects. In contrast, we see from Figure~\ref{fig:slingshot_dist2init_add}i and Figure~\ref{fig:slingshot_dist2init_add}l that the representation distance remains low for models trained with $\epsilon=10^{-05}$ and $\epsilon=10^{-04}$. The models trained with higher $\epsilon$ values do not experience Slingshot Effects. These results suggest that Slingshot may have a beneficial effect in moving the representation parameters away from initialization which eventually helps with model generalization. Figure~\ref{fig:slingshot_dist2init_mul} and Figure~\ref{fig:slingshot_dist2init_div} show a similar trend for multiplication and division datasets respectively.

\subsection{SGD Optimization}
\label{appendix:grok_sgd}

\begin{figure*}[h!]
\centering
  \begin{tabular}{cc}
      \includegraphics[width=0.45\linewidth]{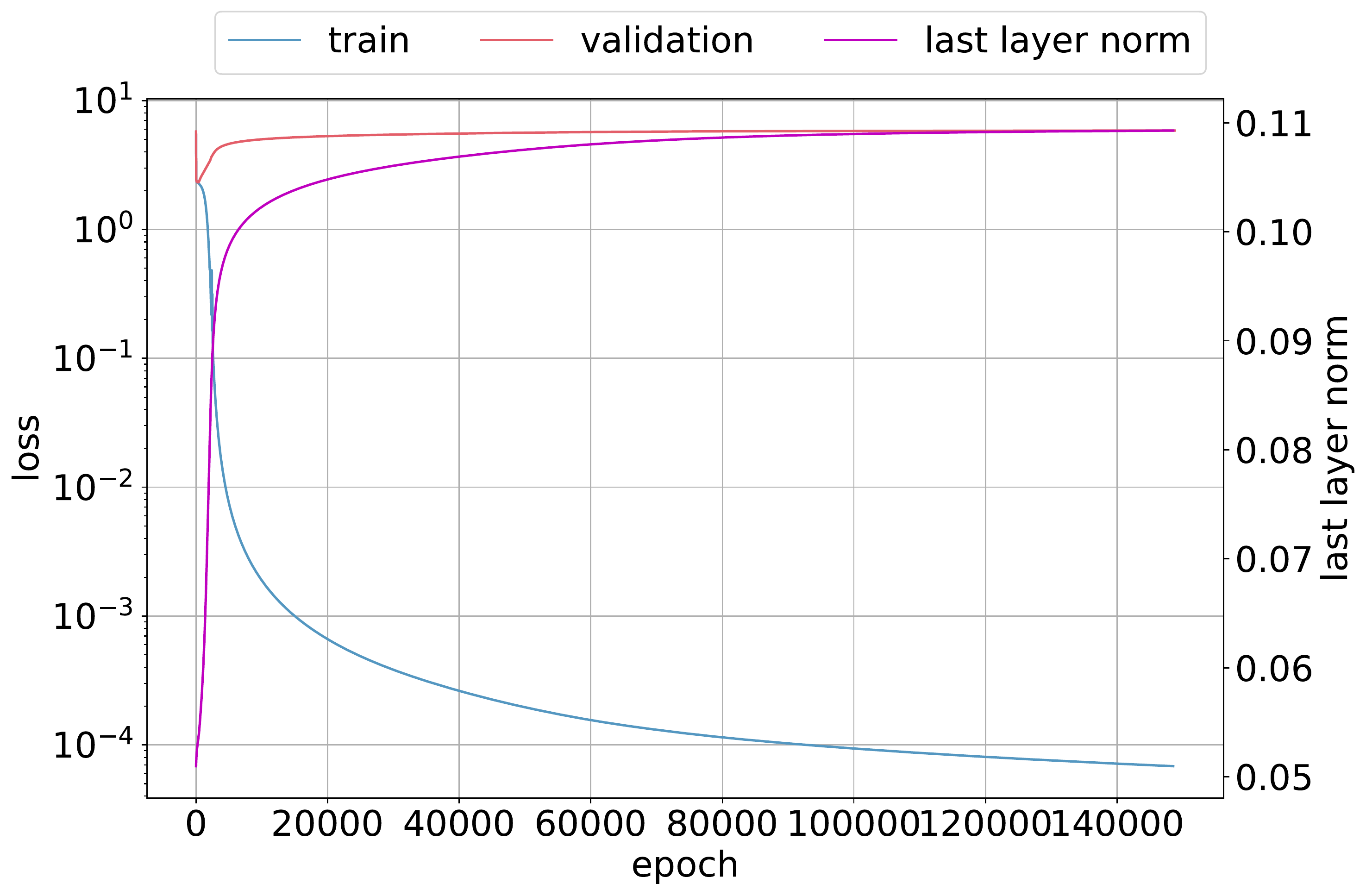} & 
      \includegraphics[width=0.45\linewidth]{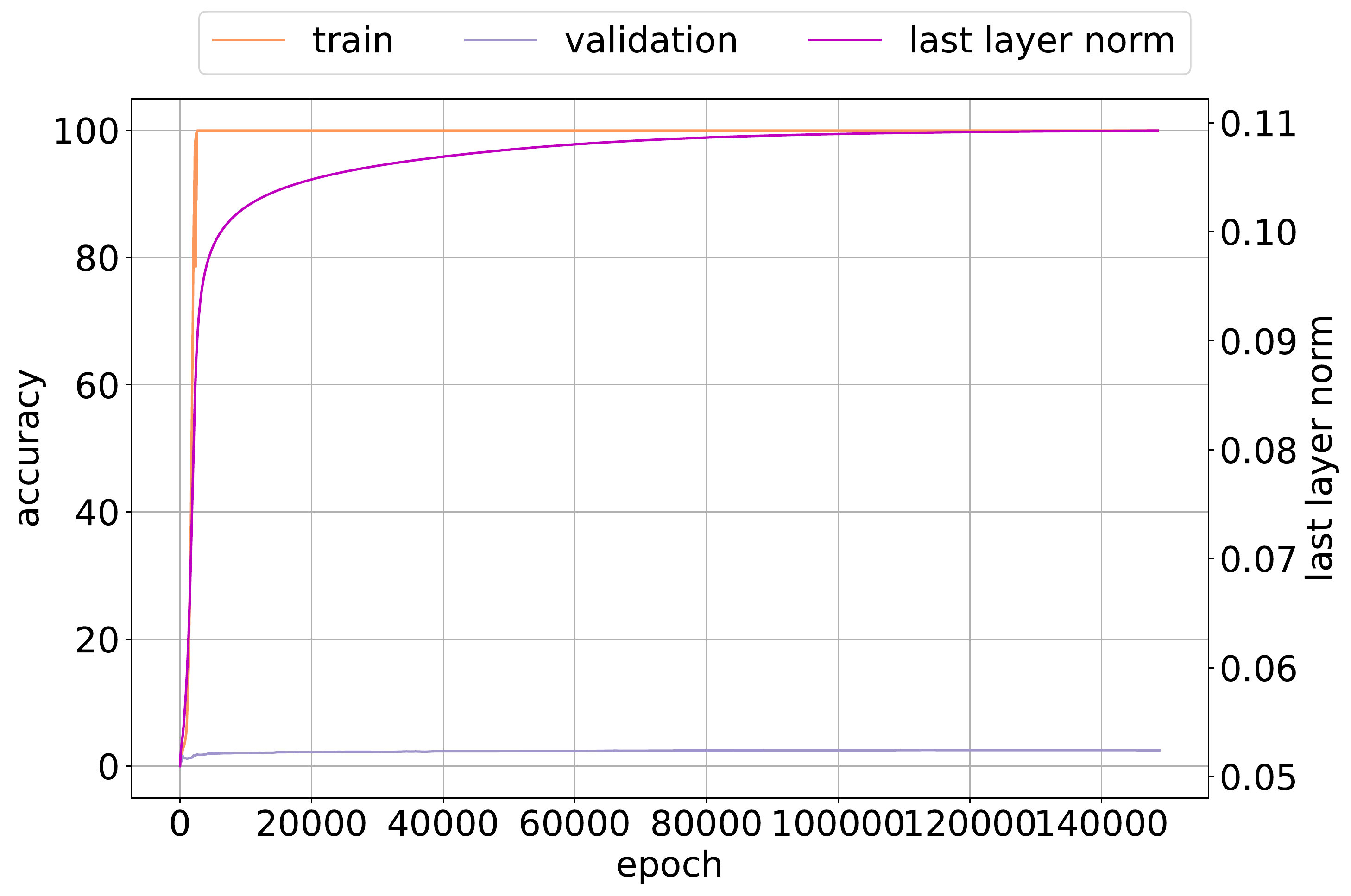} \\
      (a)  & (b) \\
      & \\
      \includegraphics[width=0.45\linewidth]{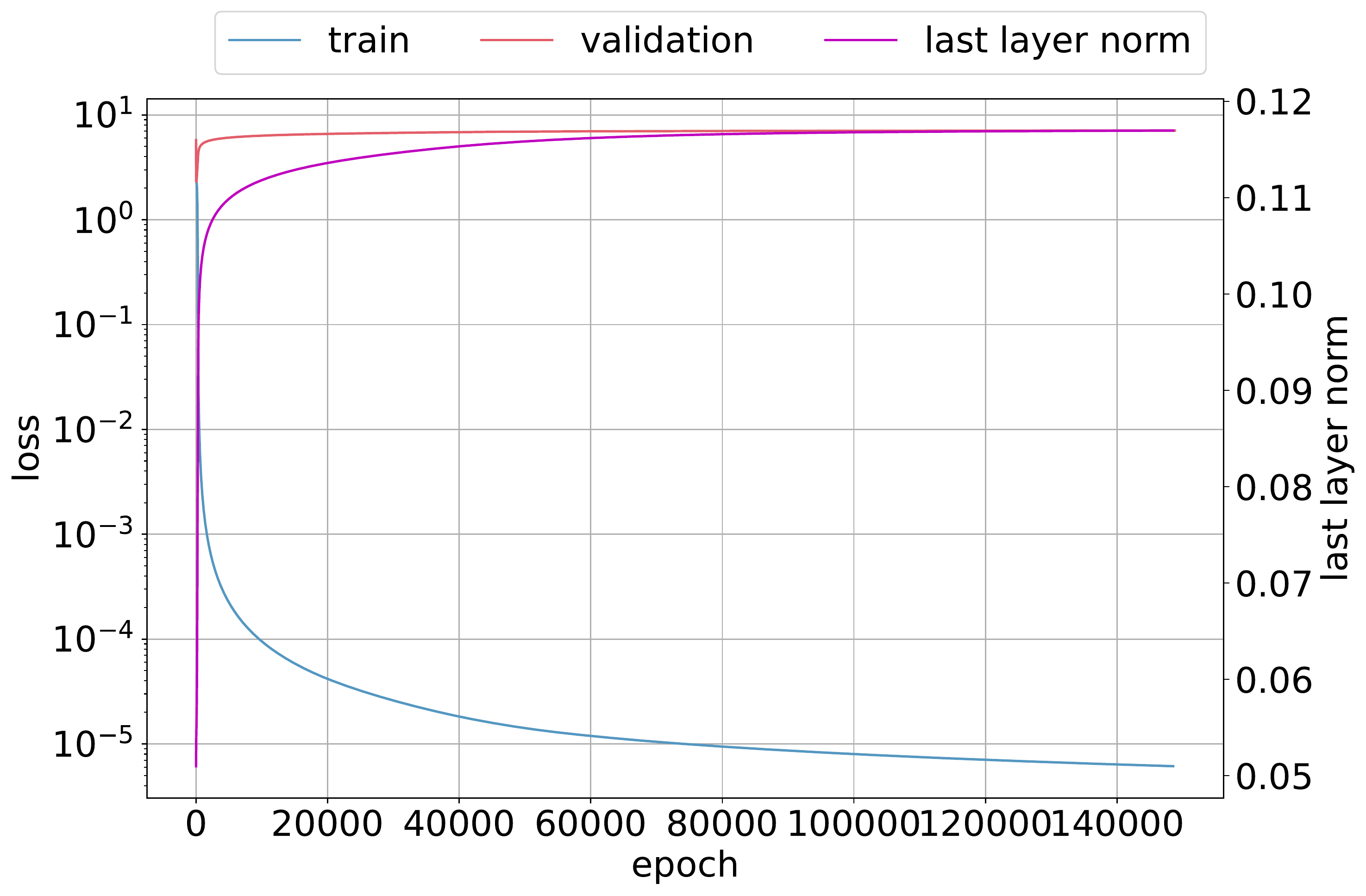} & 
      \includegraphics[width=0.45\linewidth]{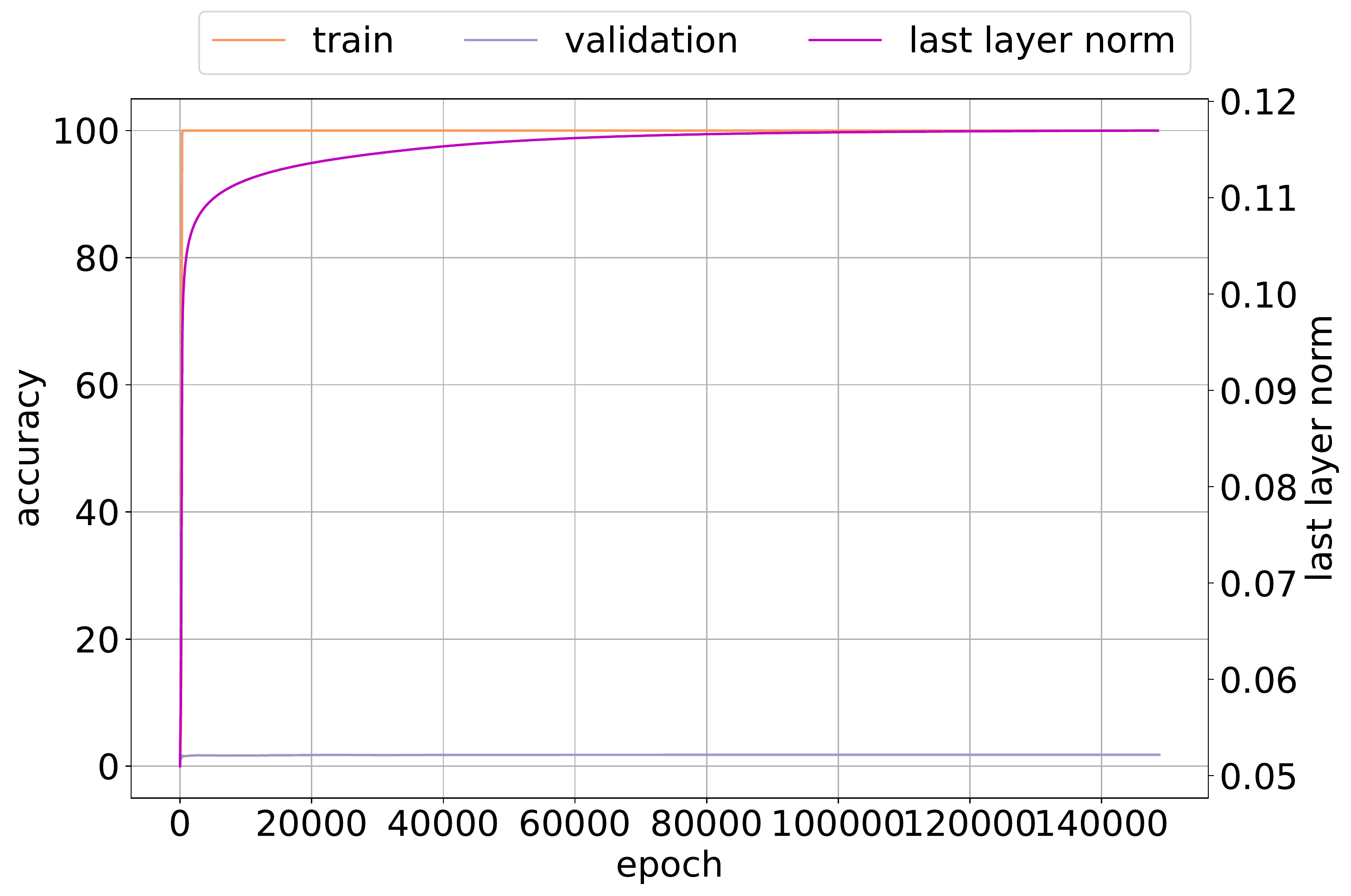} \\
      (c)  & (d) \\ 
      & \\
      \includegraphics[width=0.45\linewidth]{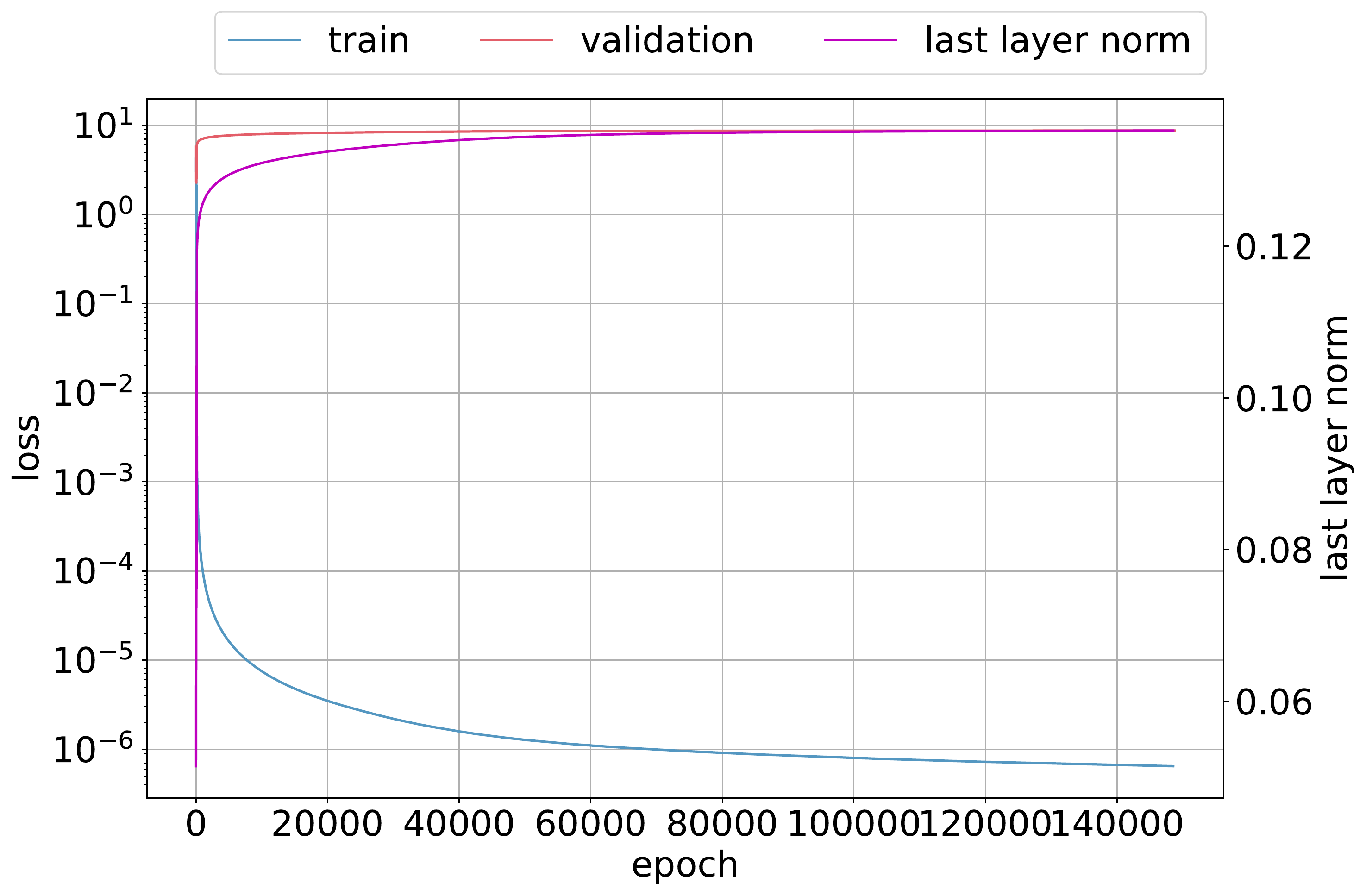} & 
      \includegraphics[width=0.45\linewidth]{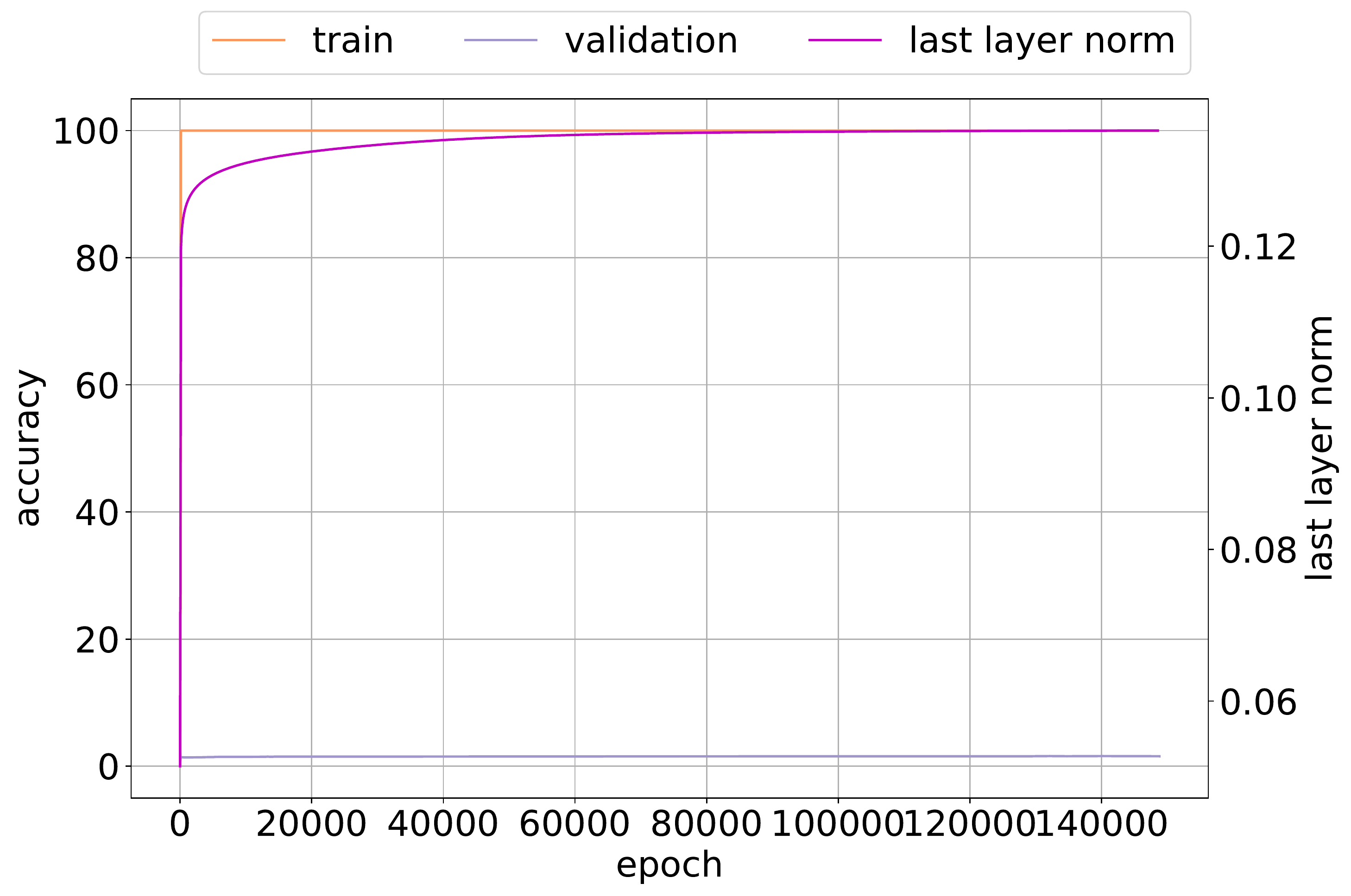} \\
      (e)  & (f) \\ 
      & \\
  \end{tabular}
 \caption{Optimizing a Transformer with SGD on modular division dataset: Norm growth vs (a), (c), (e) training and validation loss, (b), (d), (f) training and validation accuracy. Note the lack of Slingshot Effects, Grokking and generalization seen with Adam/AdamW optimizer.  } 
 \label{fig:grok_sgd}
\end{figure*}

In this appendix, we show that Slingshot Effects are not seen during Transformer training with stochastic gradient descent (SGD) with momentum to support our claim in the main paper. To this end, we use train the Transformer described in in Appendix~\ref{appendix:xformers_setup} on modular division dataset with a 50/50 train/validation split using SGD with momentum. We use a mini-batch size of $512$ which requires the optimizer to take $10$ steps per epoch for dataset split described above. We set momentum to $0.9$ and use the following learning rates: $0.001$, $0.01$ and $0.1$ and run the optimizer for 1500000 steps. The number of steps used here is 3 times larger than the steps used to run Adam/AdamW in this work which is chosen to give SGD additional time to reach convergence. Figure~\ref{fig:grok_sgd} shows the usual loss and accuracy metrics calculated on training and validation data as well as the weight norm of the classifier layer. We observe that there is no evidence of Slingshot with SGD. Lastly, we do not see any evidence of Grokking or generalization with this setup as well.

\subsection{Slingshots with Additional Datasets}
\label{appendix:slingshot_more_grok}

\begin{figure*}[h]
\centering
  \begin{tabular}{cc}
      \includegraphics[width=0.50\linewidth]{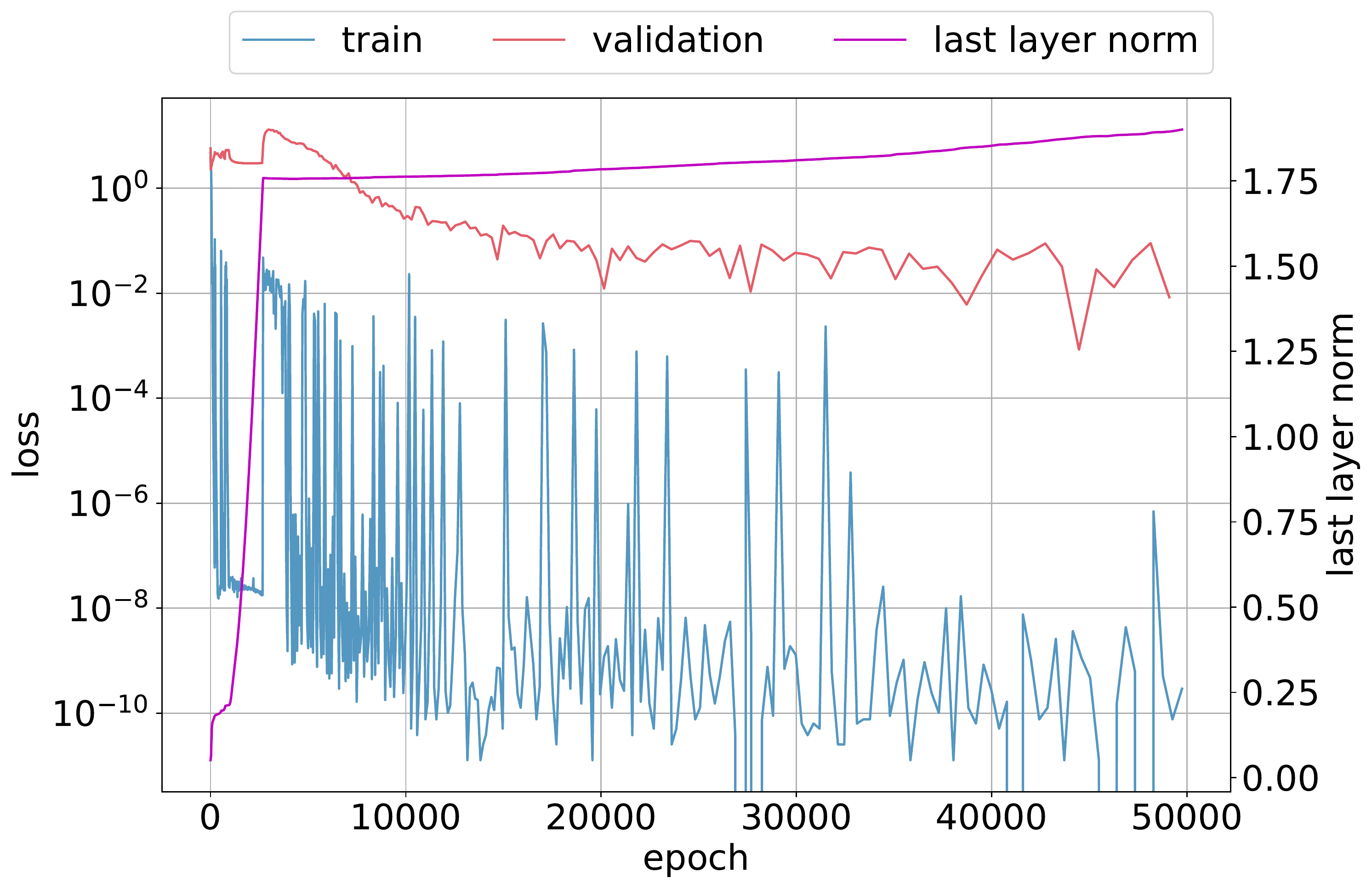} &
      \includegraphics[width=0.50\linewidth]{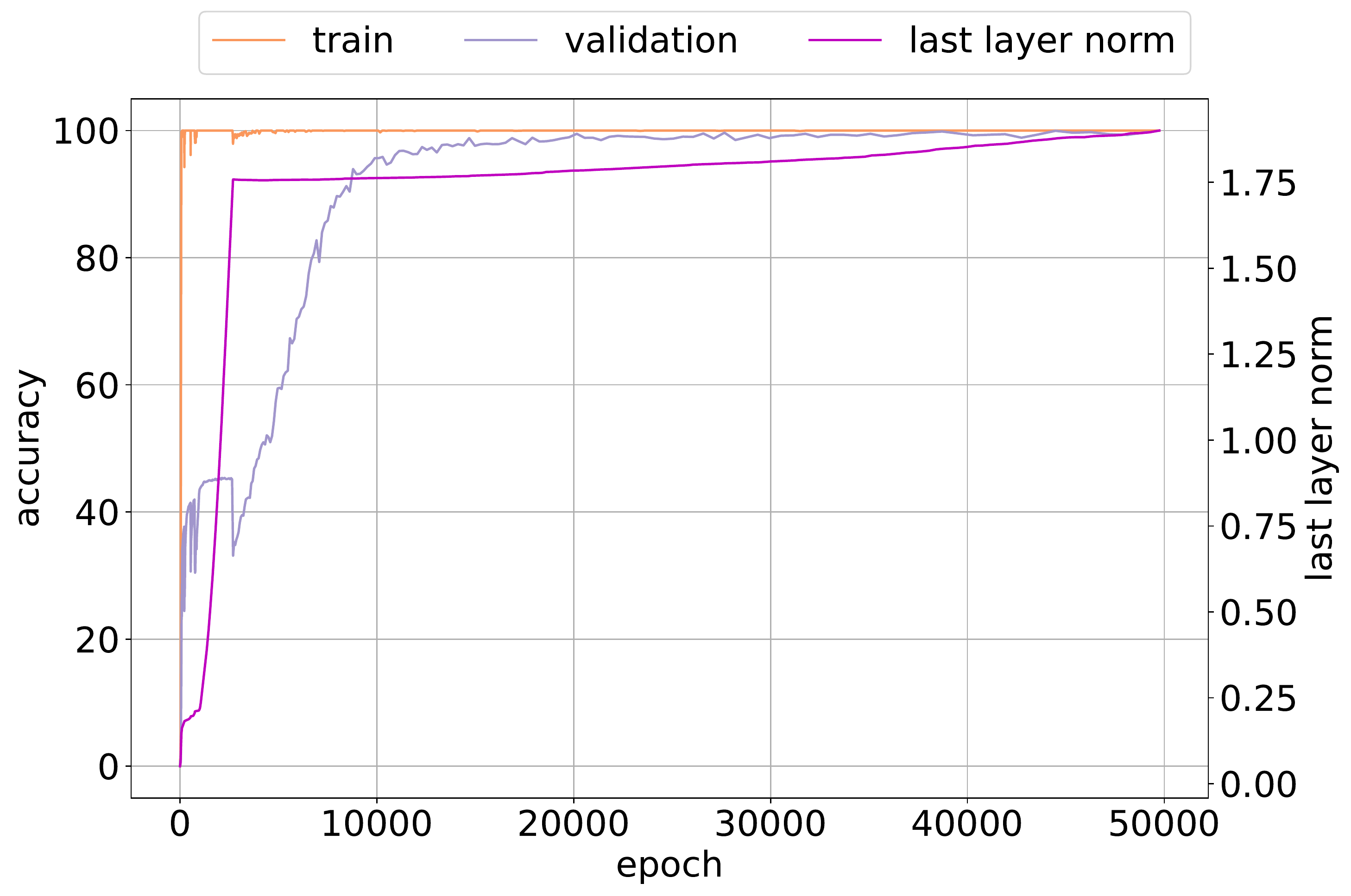} \\
      (a)  & (b)  \\
      & \\
  \end{tabular}
 \caption{Addition dataset with 50/50 train/validation split. Training and validation (a) loss and (b) accuracy.} 
 \label{fig:slingshot_add50p}
\end{figure*}

\begin{figure*}[h]
\centering
  \begin{tabular}{cc}
      \includegraphics[width=0.50\linewidth]{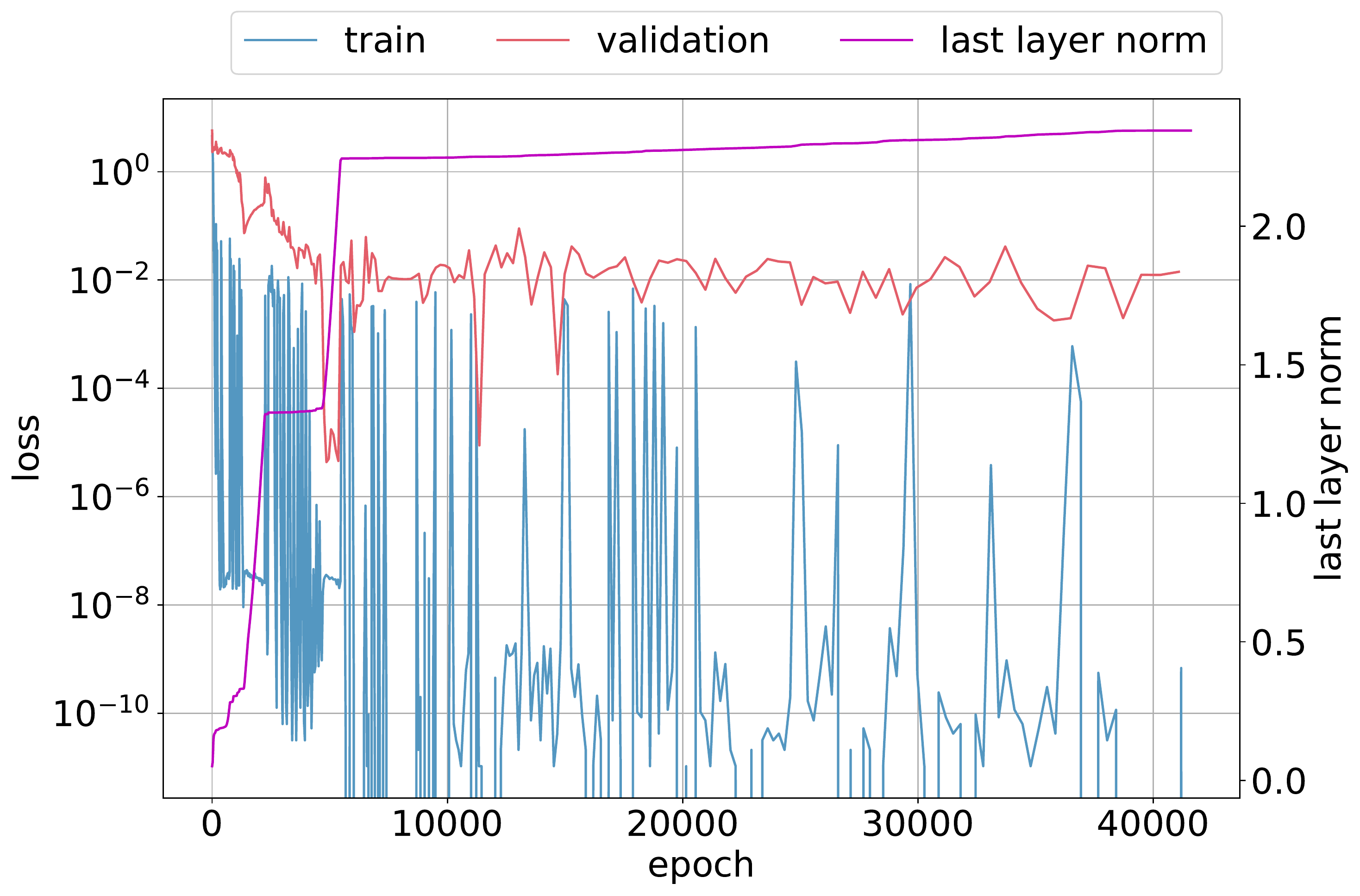} &
      \includegraphics[width=0.50\linewidth]{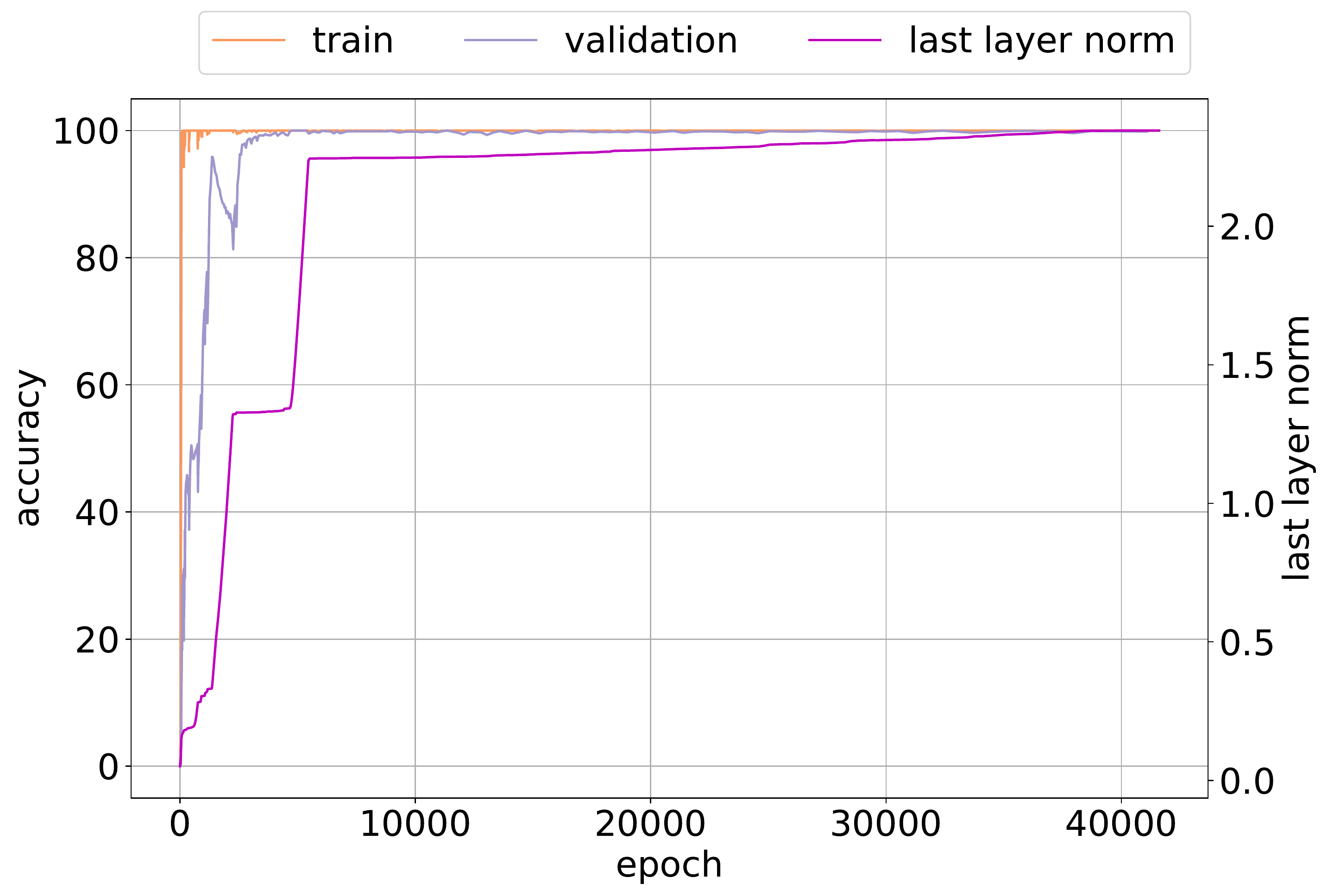} \\
      (a)  & (b)  \\
      & \\
  \end{tabular}
 \caption{Addition dataset with 60/40 train/validation split. Training and validation (a) loss and (b) accuracy.} 
 \label{fig:slingshot_add60p}
\end{figure*}

\begin{figure*}[h]
\centering
  \begin{tabular}{cc}
      \includegraphics[width=0.50\linewidth]{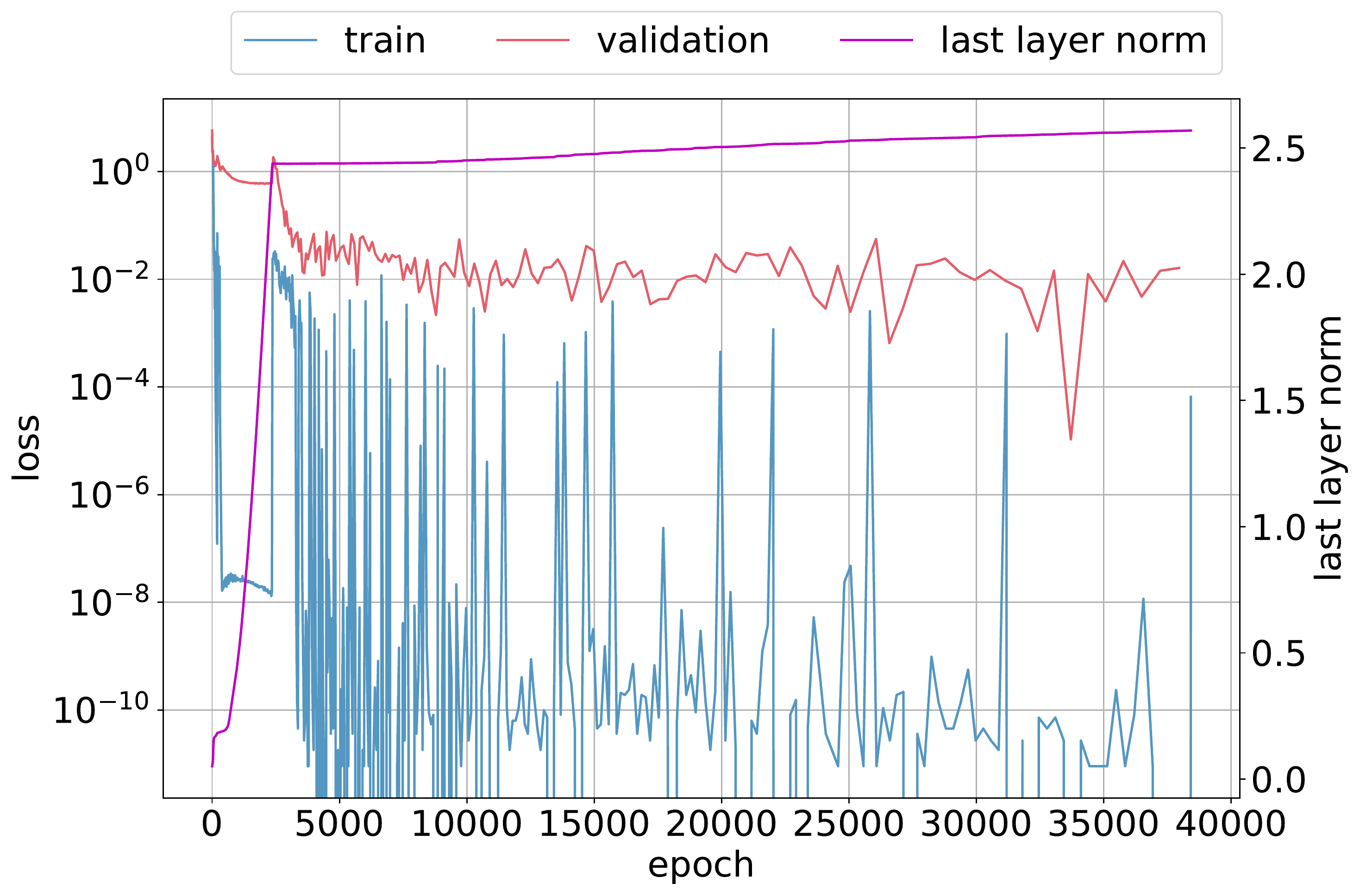} &
      \includegraphics[width=0.50\linewidth]{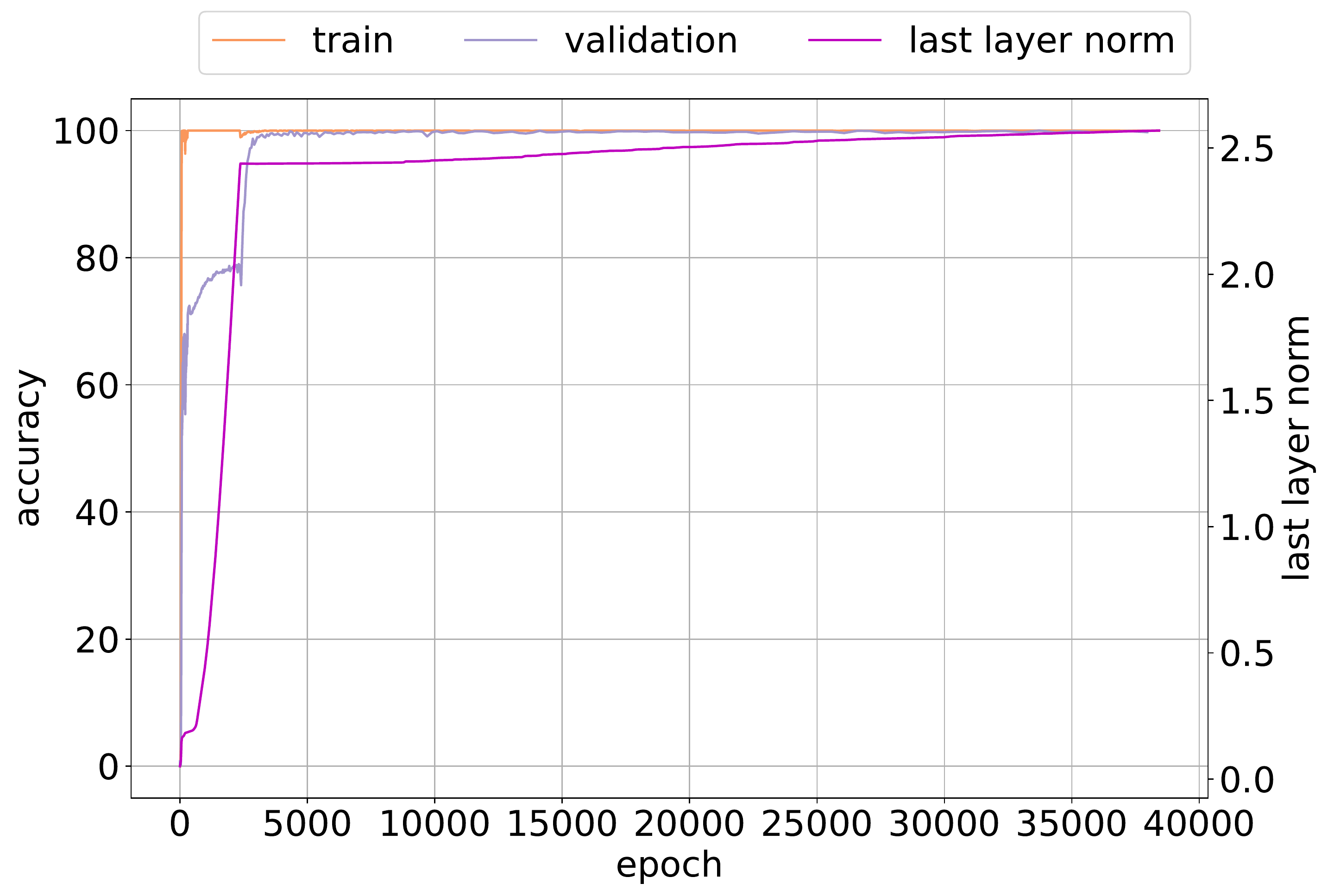} \\
      (a)  & (b)  \\
      & \\
  \end{tabular}
 \caption{Addition dataset with 70/30 train/validation split. Training and validation (a) loss and (b) accuracy.} 
 \label{fig:slingshot_add70p}
\end{figure*}

\begin{figure*}[h]
\centering
  \begin{tabular}{cc}
      \includegraphics[width=0.50\linewidth]{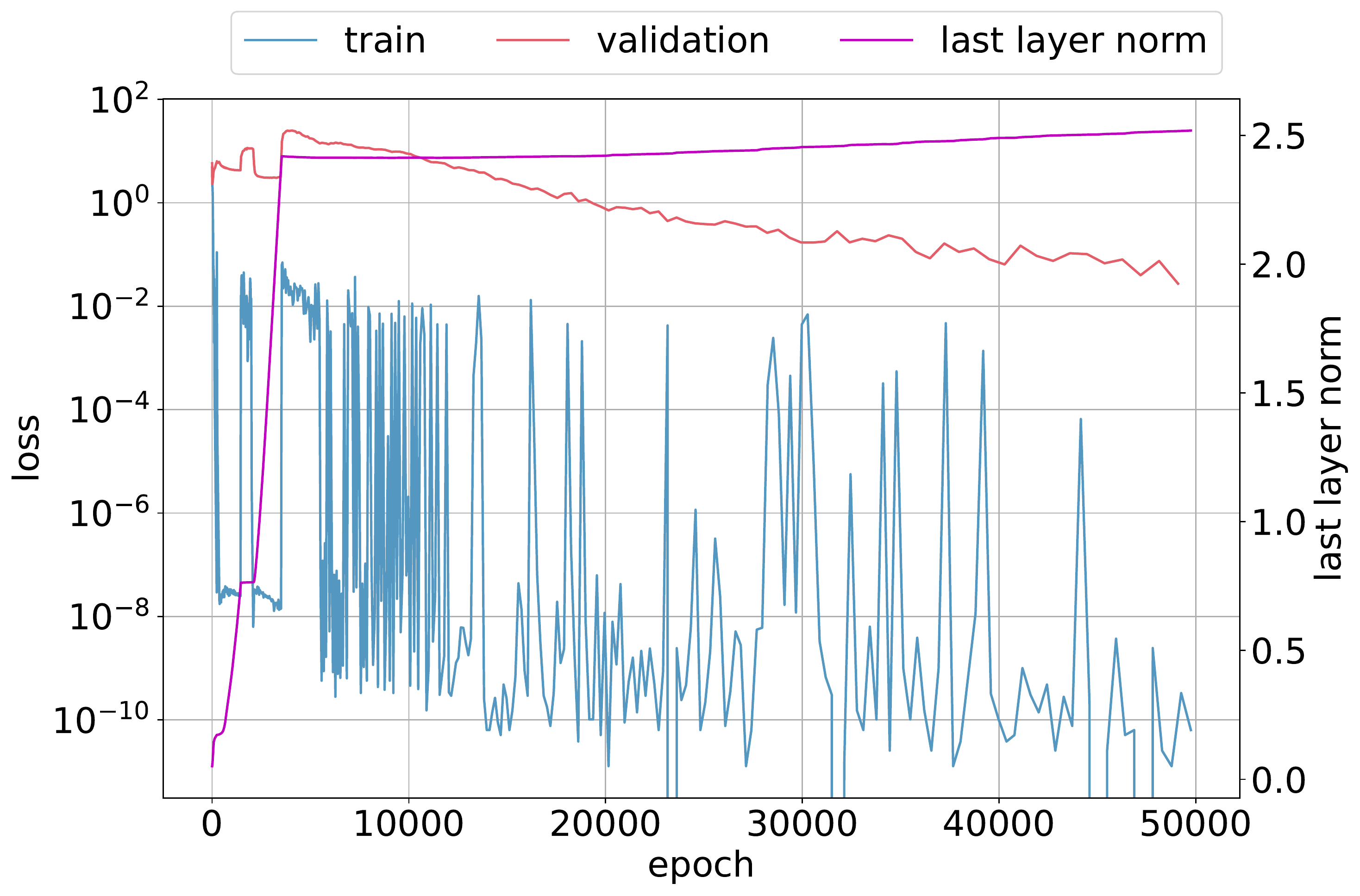} &
      \includegraphics[width=0.50\linewidth]{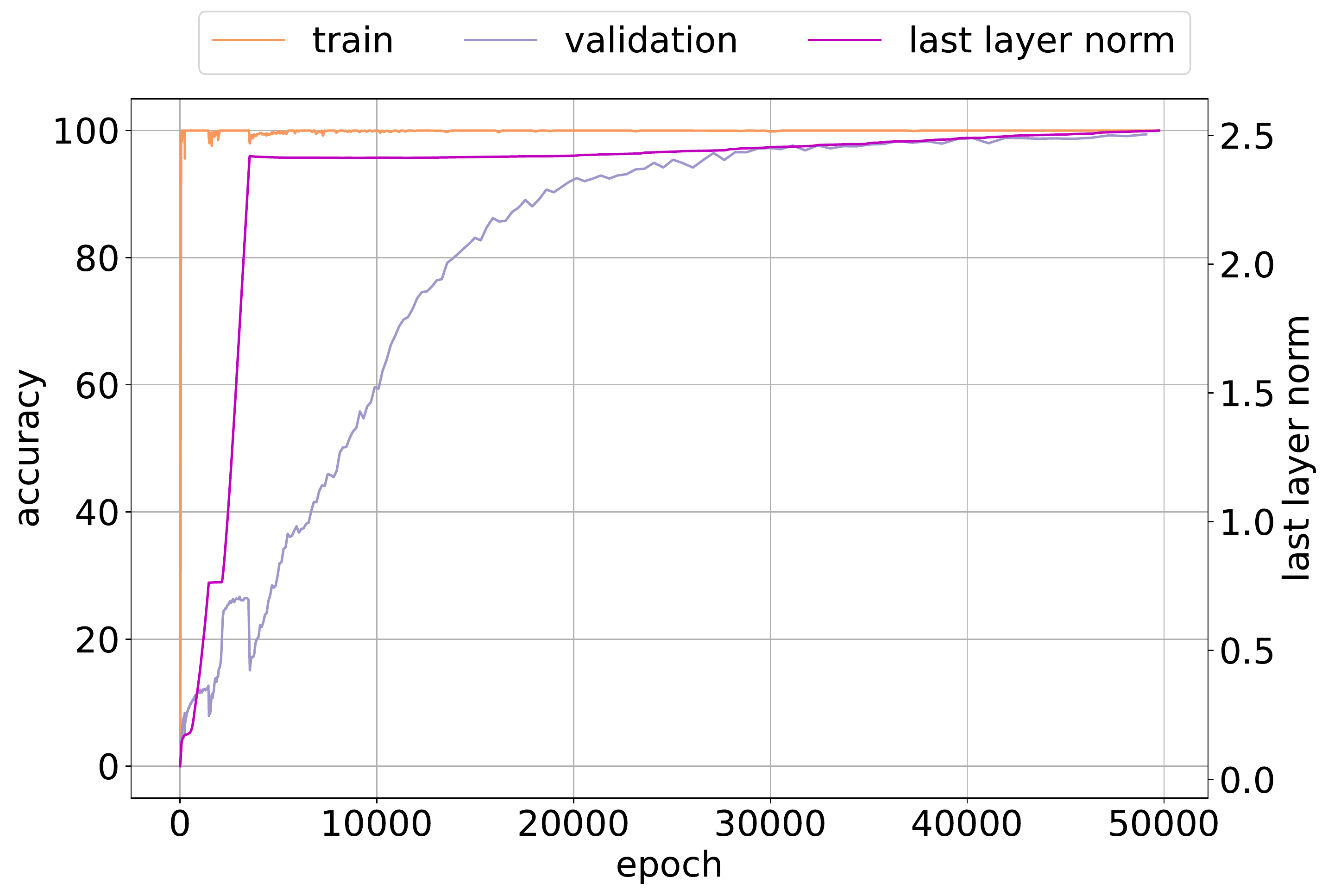} \\
      (a)  & (b)  \\
      & \\
  \end{tabular}
 \caption{Cubepoly dataset with 50/50 train/validation split. Cubepoly operation is given by ($a^{3} + b \pmod {p}$ for $0 \leq a, b < p$). Training and validation (a) loss and (b) accuracy.} 
 \label{fig:slingshot_cubepoly50p}
\end{figure*}

\begin{figure*}[h]
\centering
  \begin{tabular}{cc}
      \includegraphics[width=0.50\linewidth]{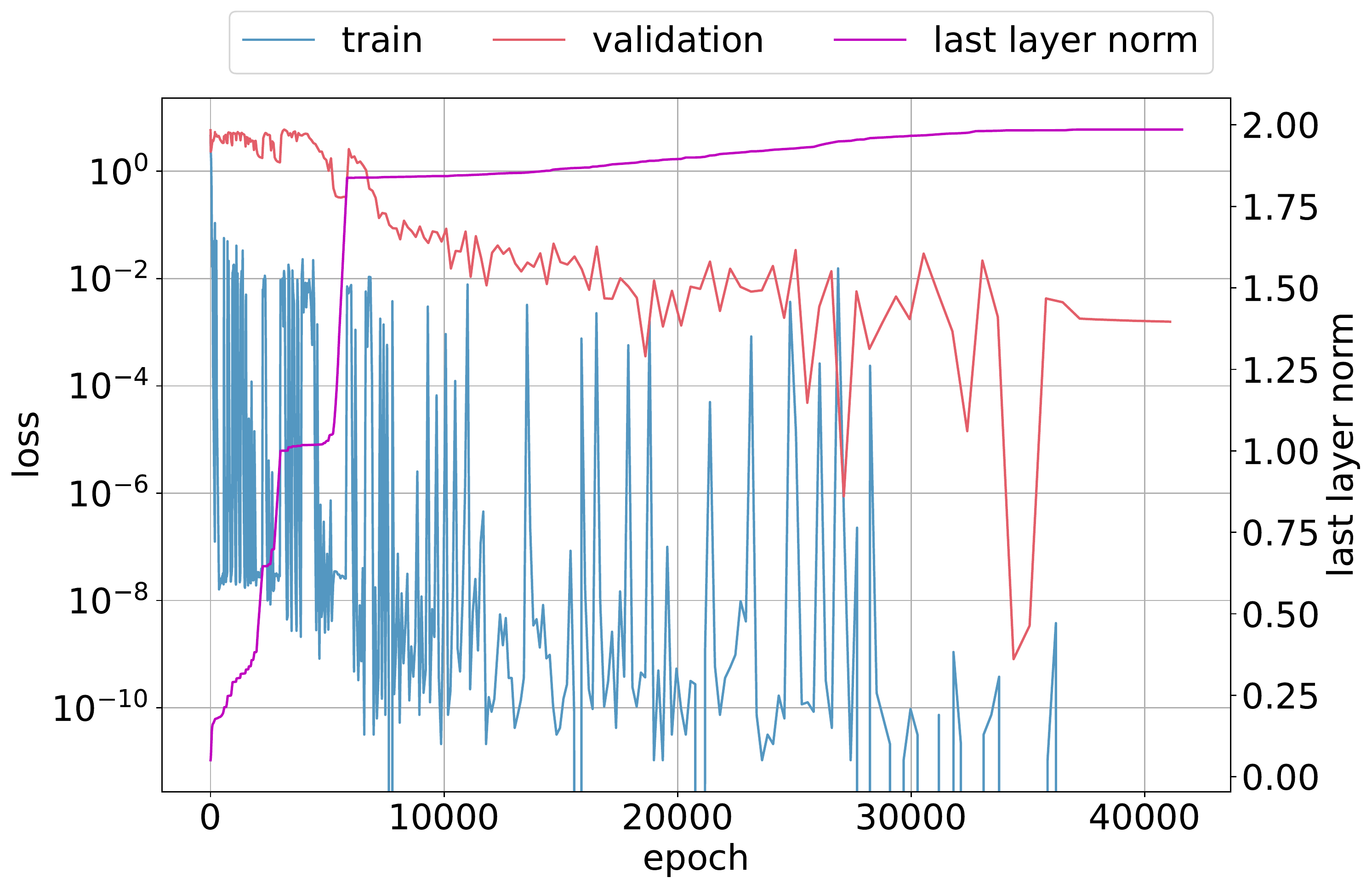} &
      \includegraphics[width=0.50\linewidth]{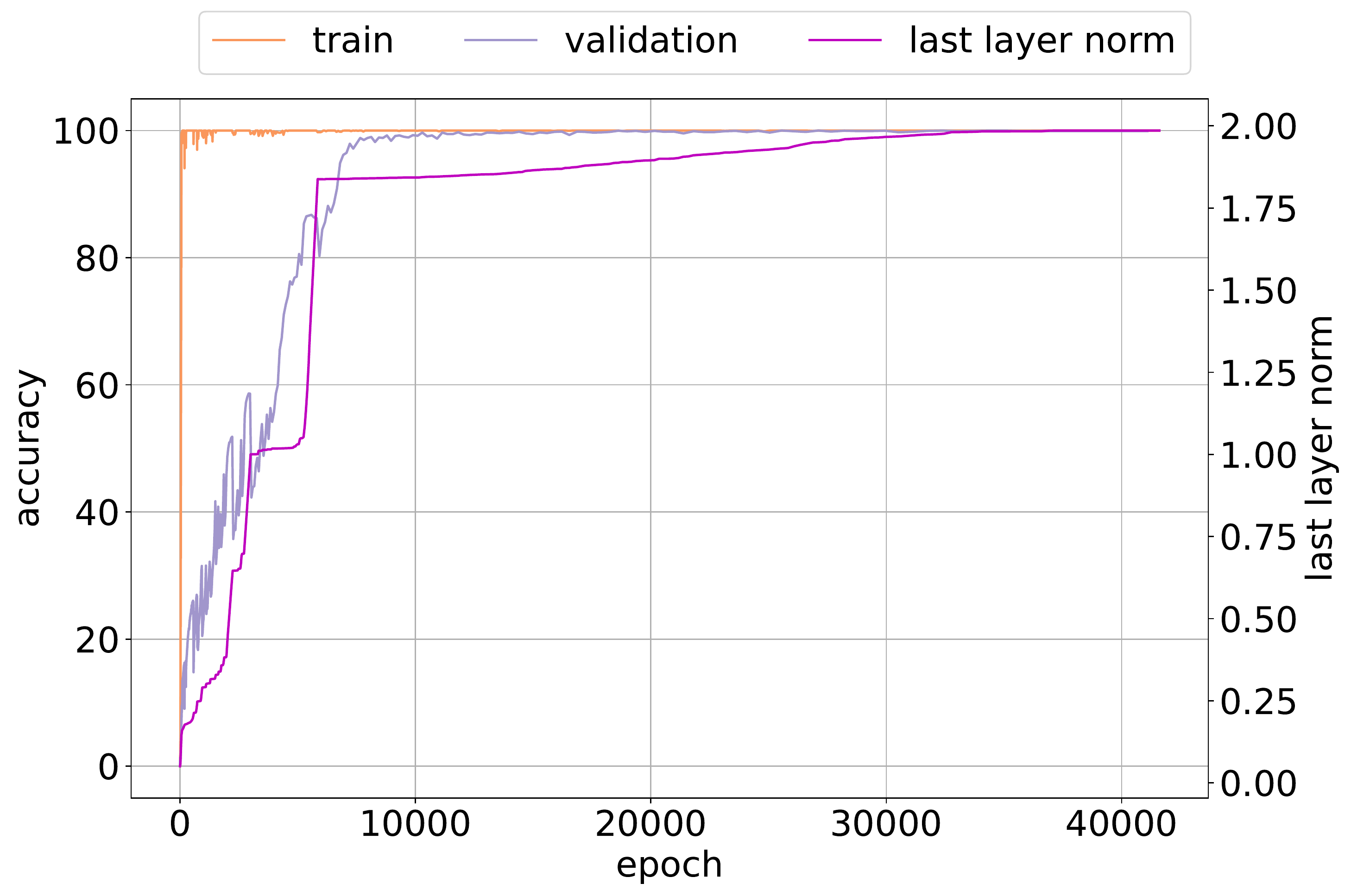} \\
      (a)  & (b)  \\
      & \\
  \end{tabular}
 \caption{Cubepoly dataset with 60/40 train/validation split. Cubepoly operation is given by ($a^{3} + b \pmod {p}$ for $0 \leq a, b < p$). Training and validation (a) loss and (b) accuracy.} 
 \label{fig:slingshot_cubepoly60p}
\end{figure*}

\begin{figure*}[h]
\centering
  \begin{tabular}{cc}
      \includegraphics[width=0.50\linewidth]{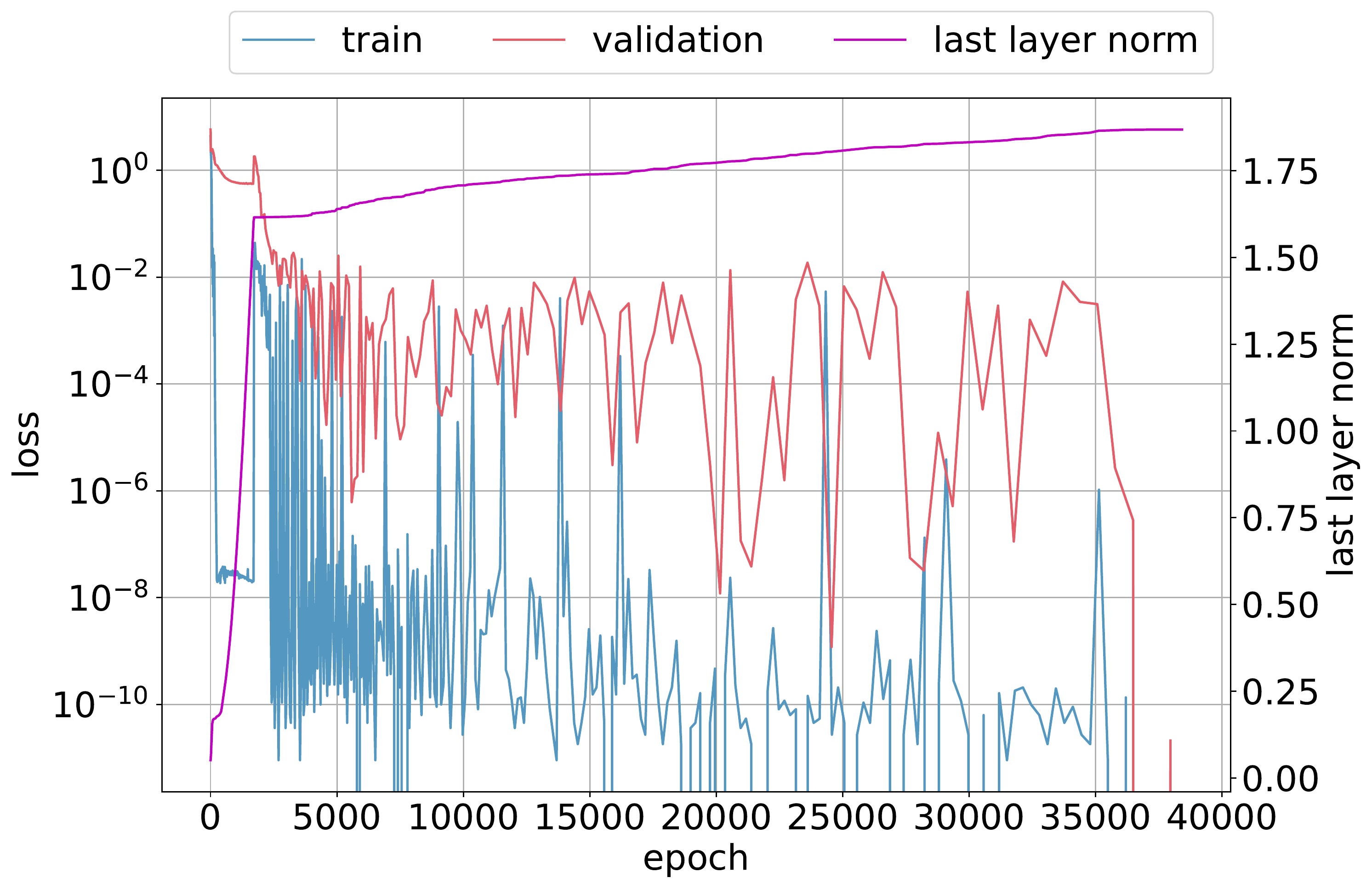} &
      \includegraphics[width=0.50\linewidth]{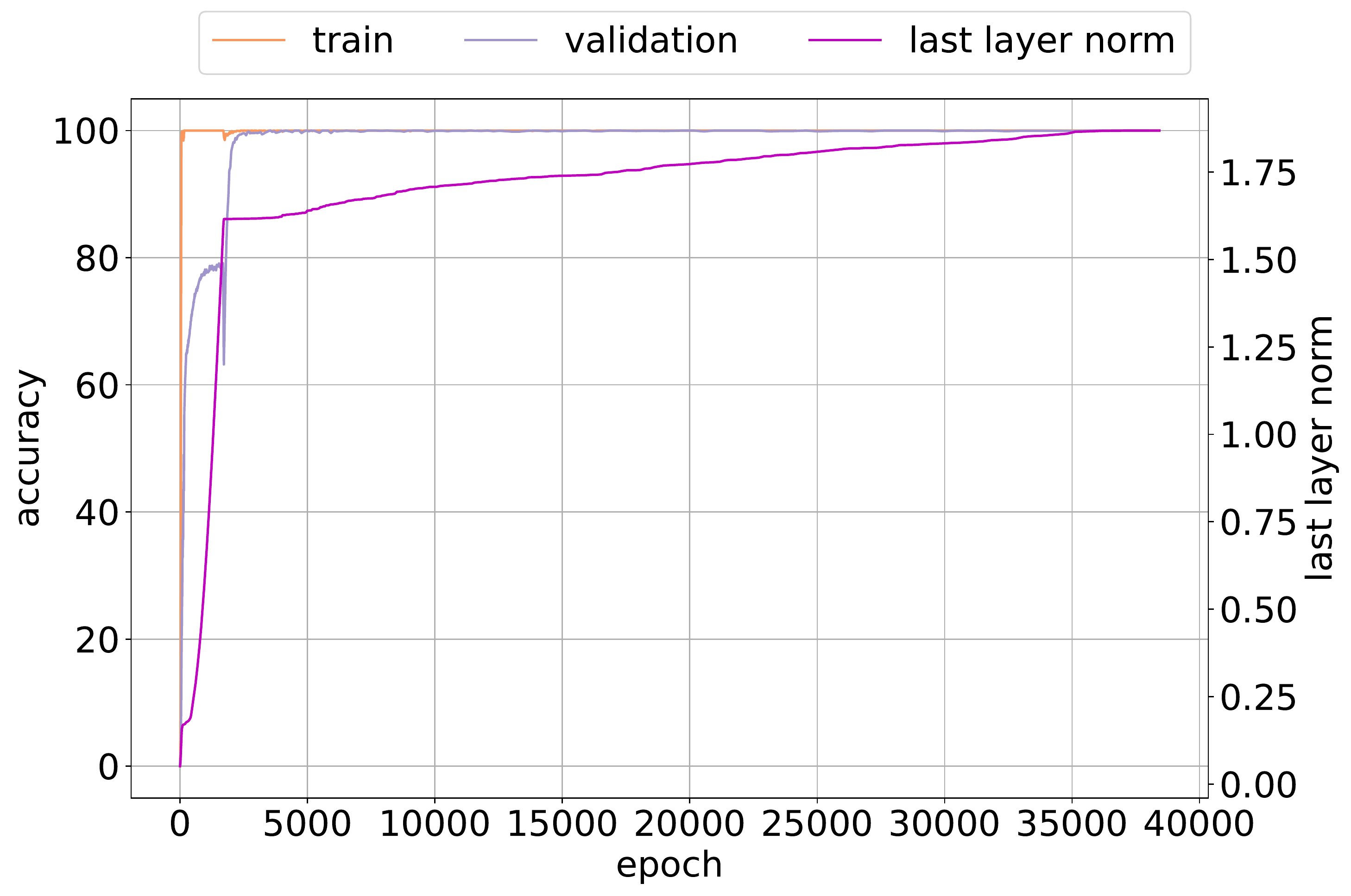} \\
      (a)  & (b)  \\
      & \\
  \end{tabular}
 \caption{Cubepoly dataset with 70/30 train/validation split. Cubepoly operation is given by ($a^{3} + b \pmod {p}$ for $0 \leq a, b < p$). Training and validation (a) loss and (b) accuracy.} 
 \label{fig:slingshot_cubepoly70p}
\end{figure*}

\begin{figure*}[h]
\centering
  \begin{tabular}{cc}
      \includegraphics[width=0.50\linewidth]{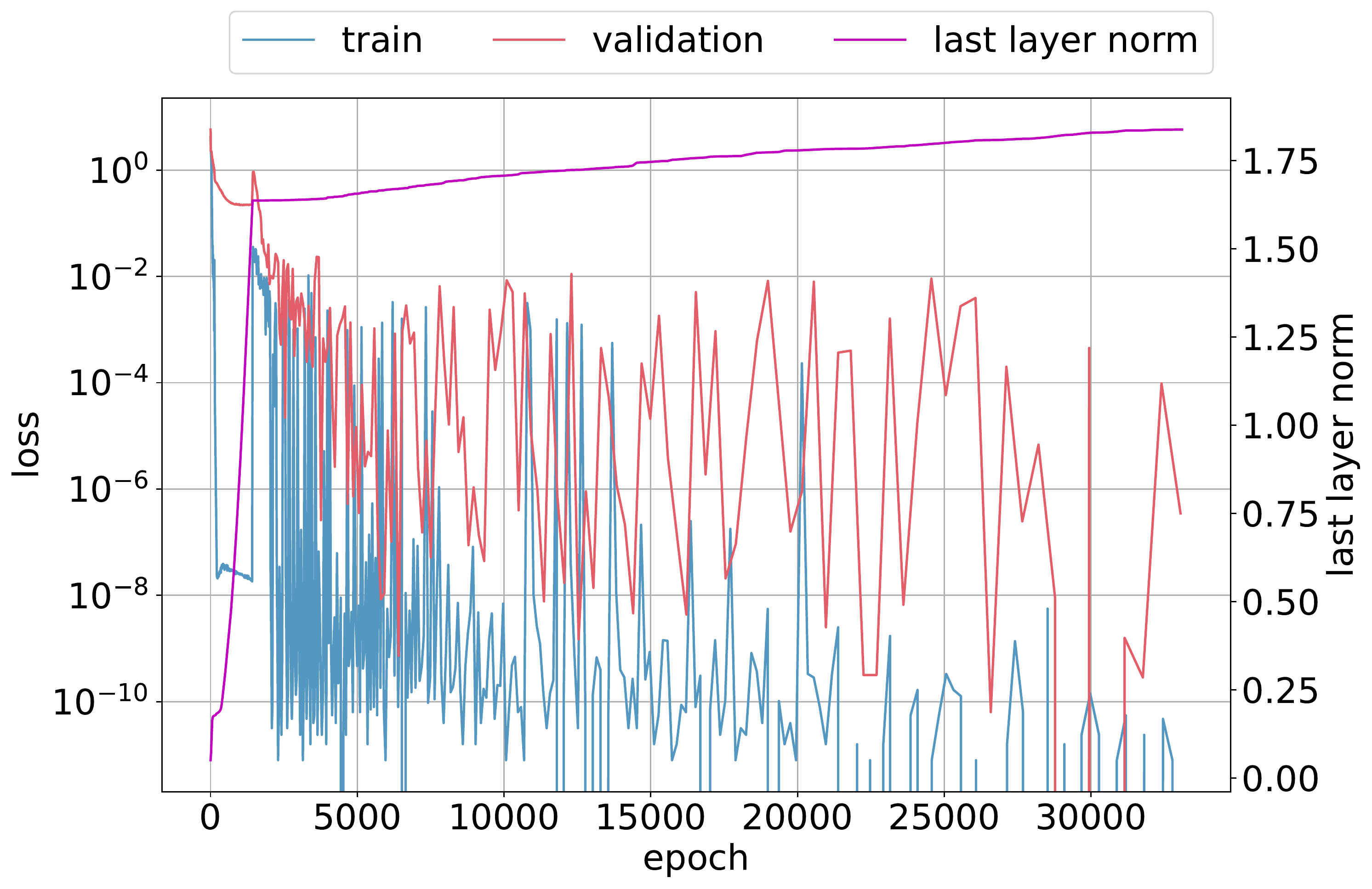} &
      \includegraphics[width=0.50\linewidth]{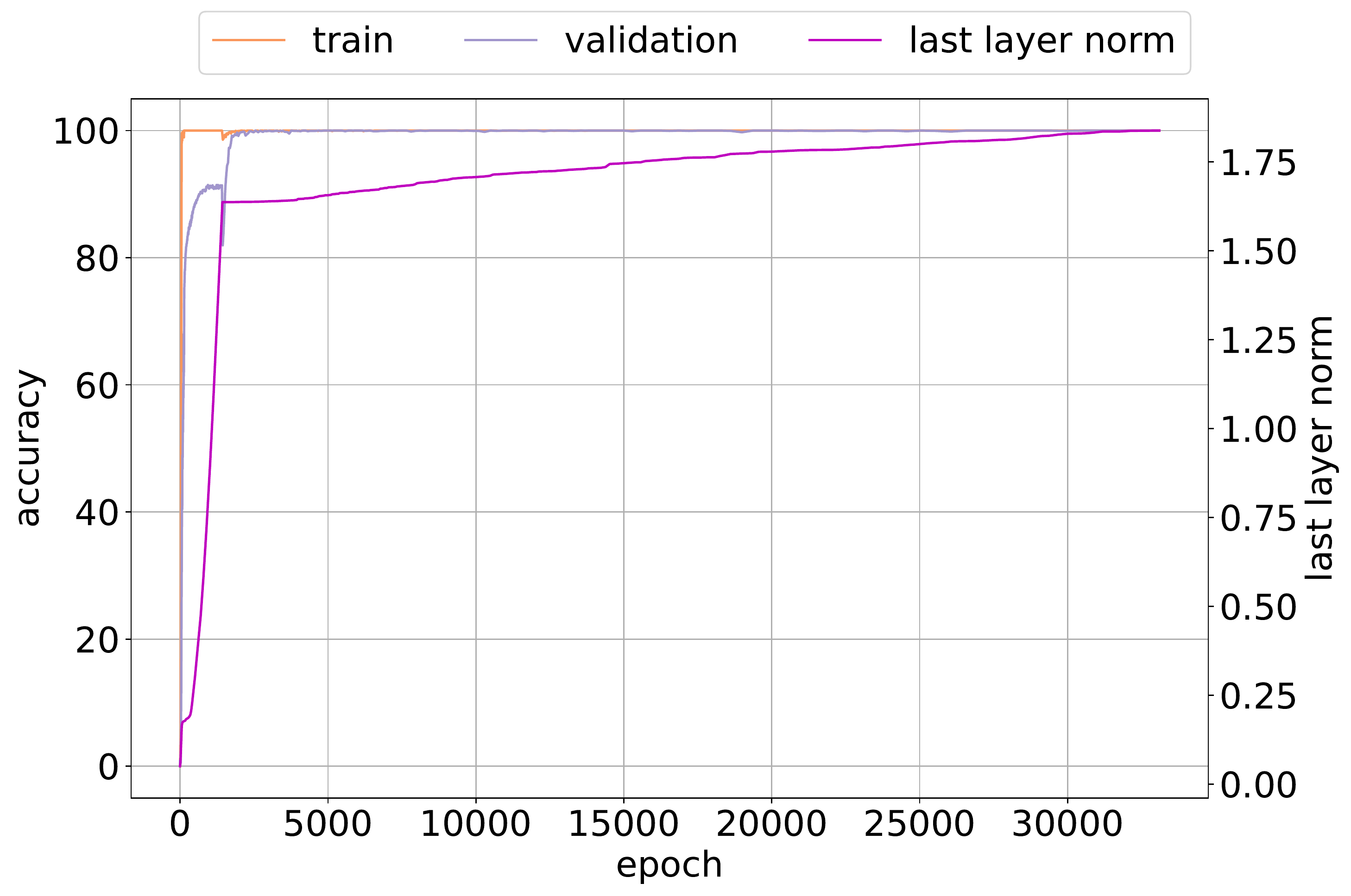} \\
      (a)  & (b)  \\
      & \\
  \end{tabular}
 \caption{Cubepoly dataset with 80/20 train/validation split. Cubepoly operation is given by ($a^{3} + b \pmod {p}$ for $0 \leq a, b < p$). Training and validation (a) loss and (b) accuracy.} 
 \label{fig:slingshot_cubepoly80p}
\end{figure*}

\begin{figure*}[h]
\centering
  \begin{tabular}{cc}
      \includegraphics[width=0.50\linewidth]{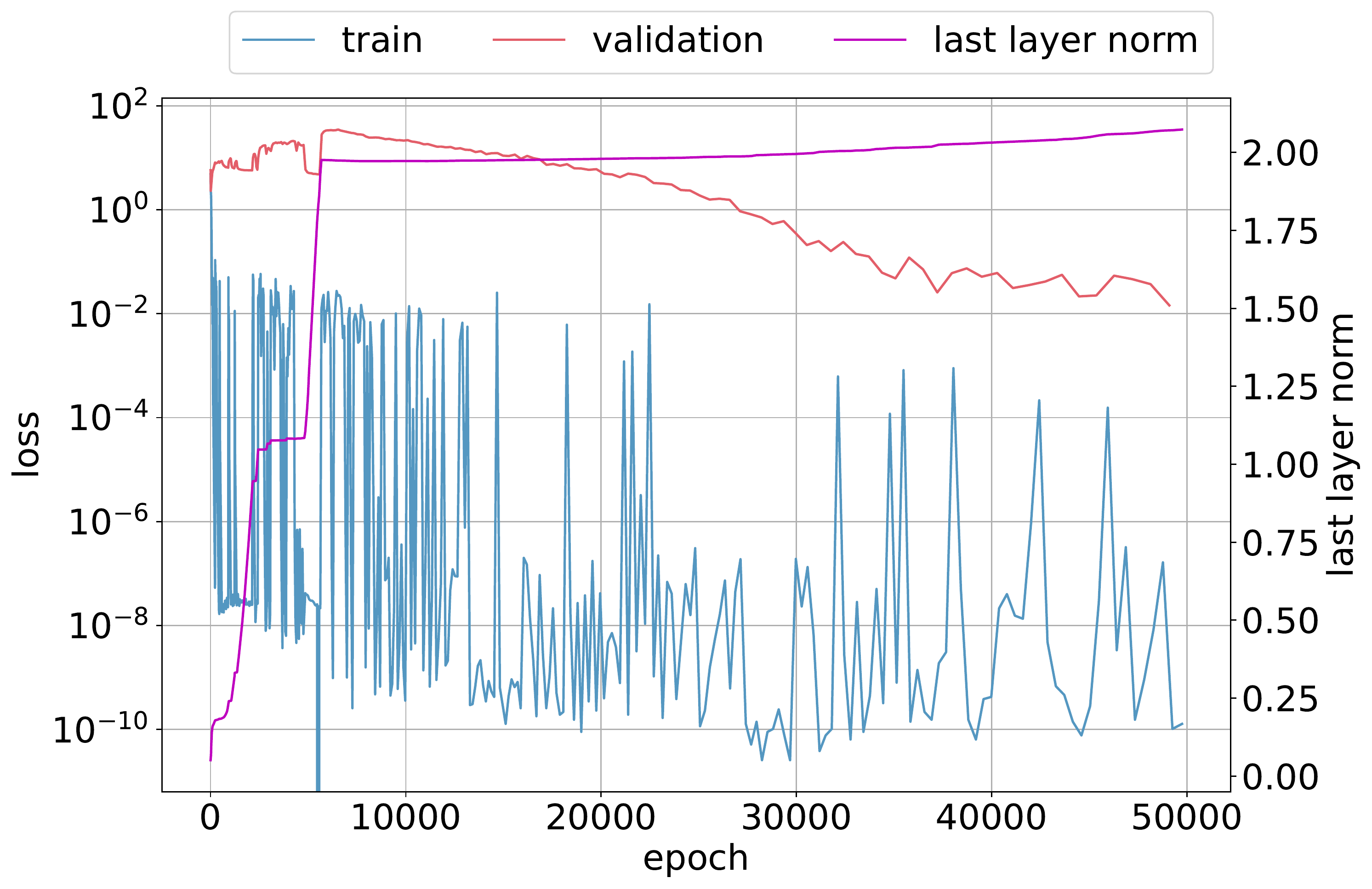} &
      \includegraphics[width=0.50\linewidth]{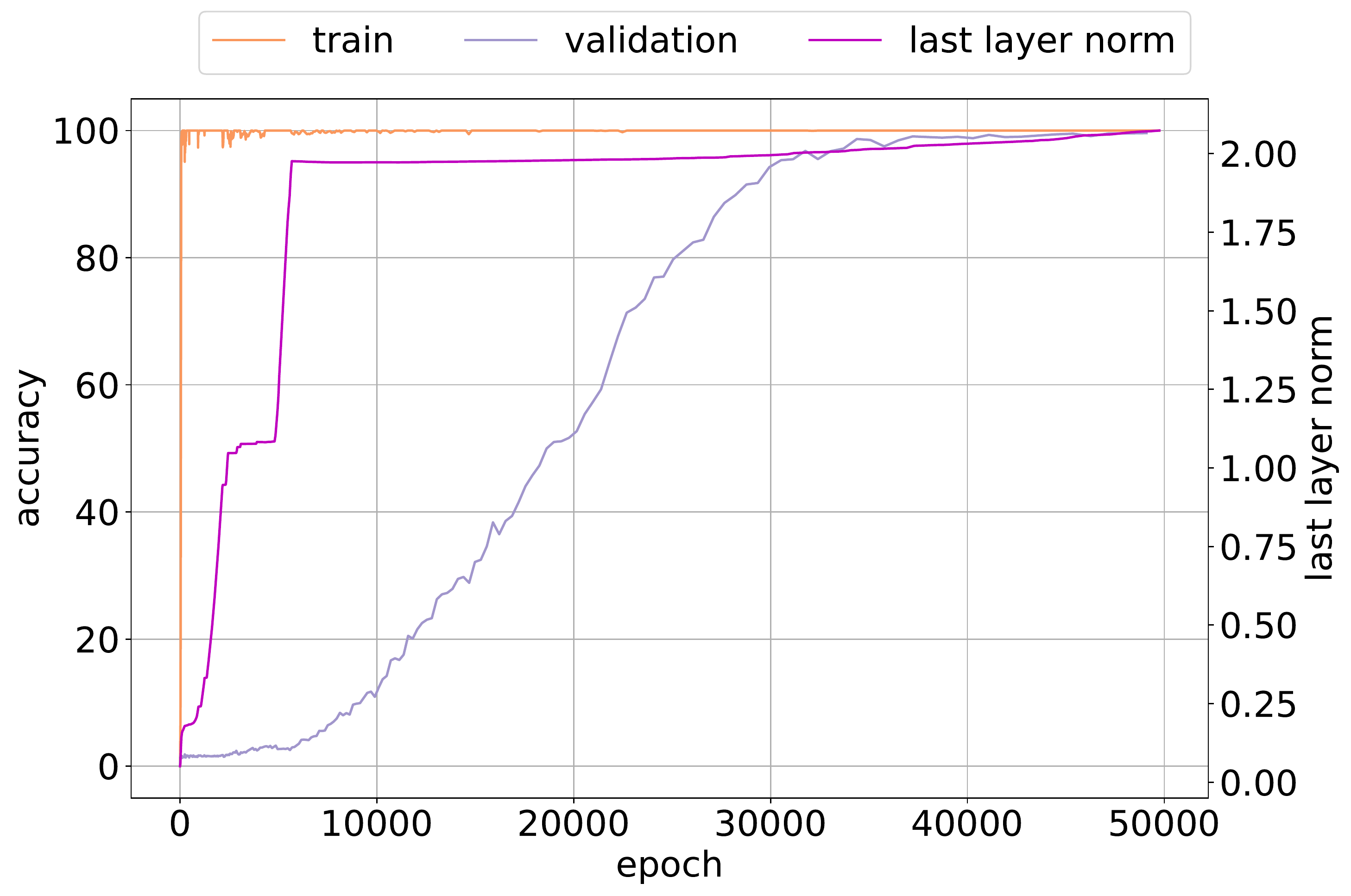} \\
      (a)  & (b)  \\
      & \\
  \end{tabular}
 \caption{Division dataset with 50/50 train/validation split. Training and validation (a) loss and (b) accuracy.} 
 \label{fig:slingshot_divison50p}
\end{figure*}

\begin{figure*}[h]
\centering
  \begin{tabular}{cc}
      \includegraphics[width=0.50\linewidth]{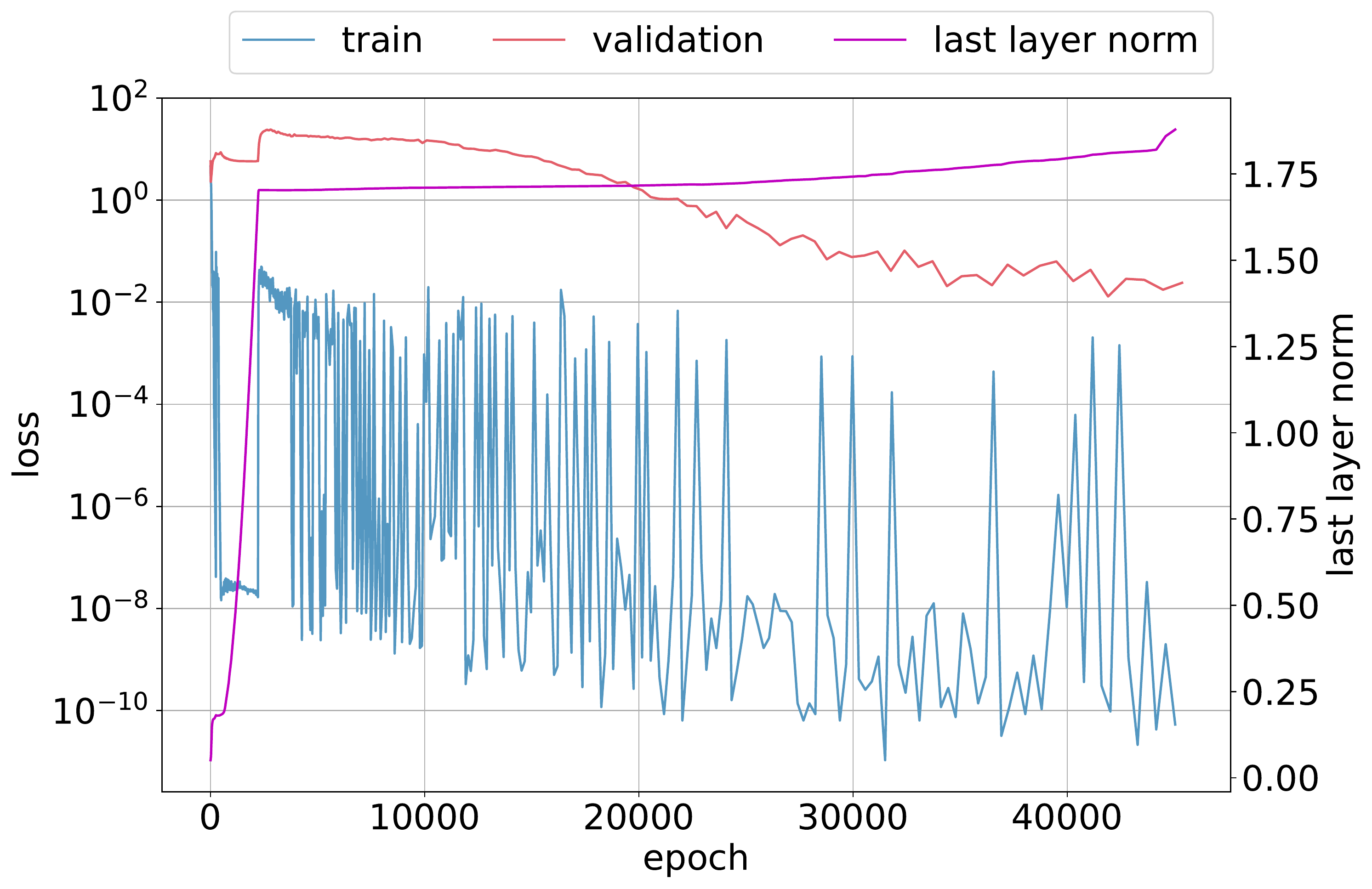} &
      \includegraphics[width=0.50\linewidth]{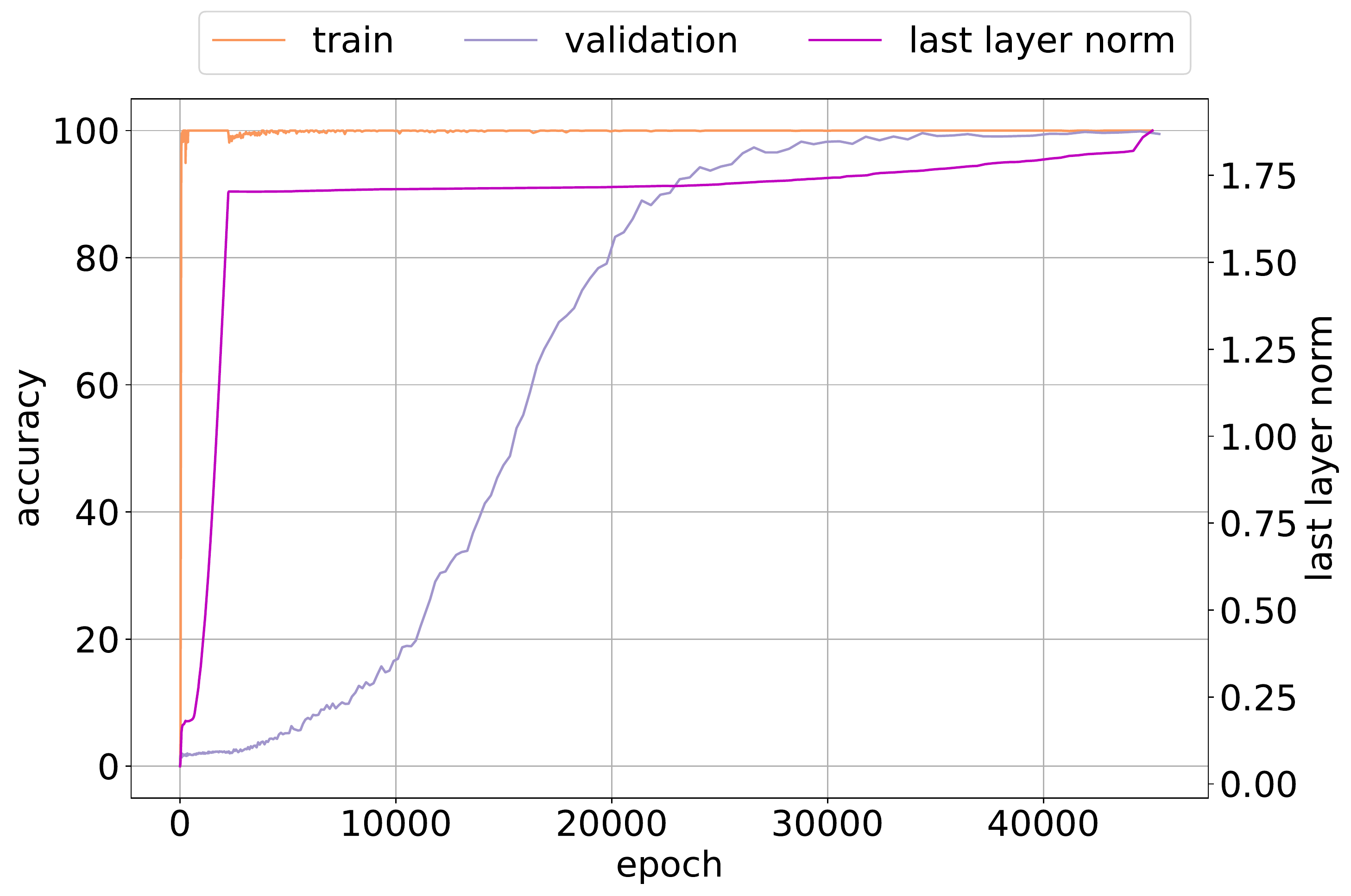} \\
      (a)  & (b)  \\
      & \\
  \end{tabular}
 \caption{Division dataset with 60/40 train/validation split. Training and validation (a) loss and (b) accuracy.} 
 \label{fig:slingshot_divison60p}
\end{figure*}

\begin{figure*}[h]
\centering
  \begin{tabular}{cc}
      \includegraphics[width=0.50\linewidth]{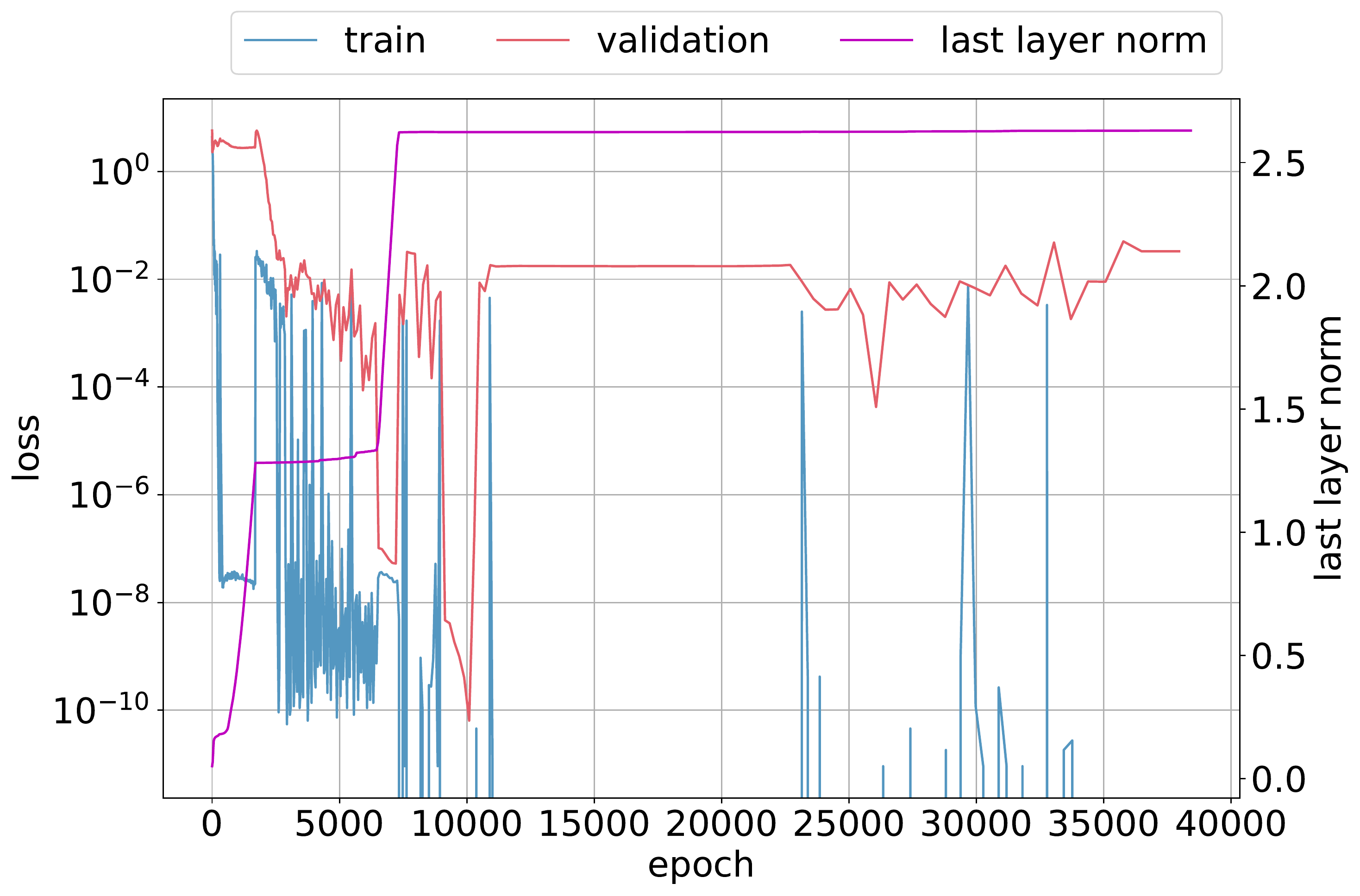} &
      \includegraphics[width=0.50\linewidth]{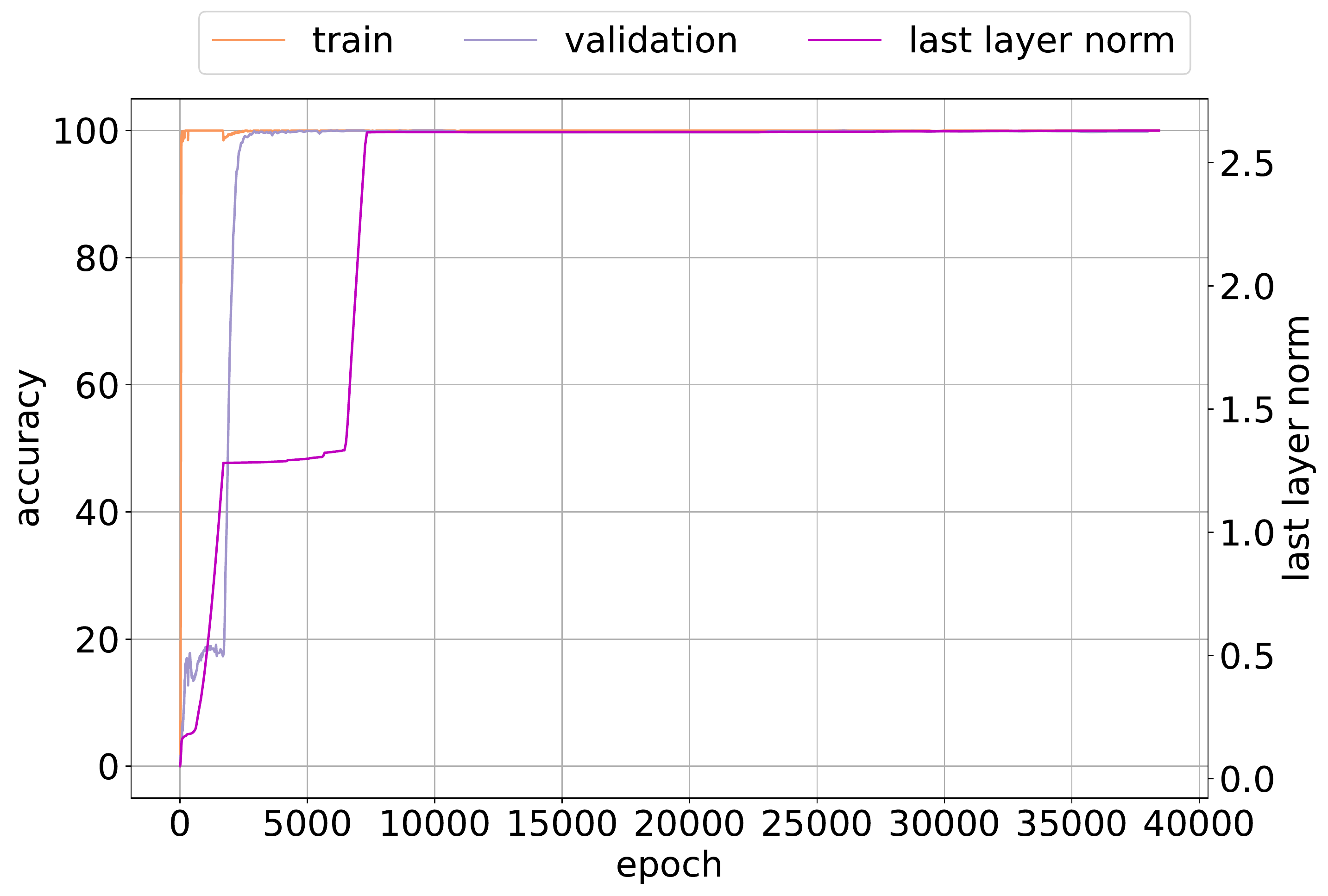} \\
      (a)  & (b)  \\
      & \\
  \end{tabular}
 \caption{Division dataset with 70/30 train/validation split. Training and validation (a) loss and (b) accuracy.} 
 \label{fig:slingshot_divison70p}
\end{figure*}

\begin{figure*}[h]
\centering
  \begin{tabular}{cc}
      \includegraphics[width=0.50\linewidth]{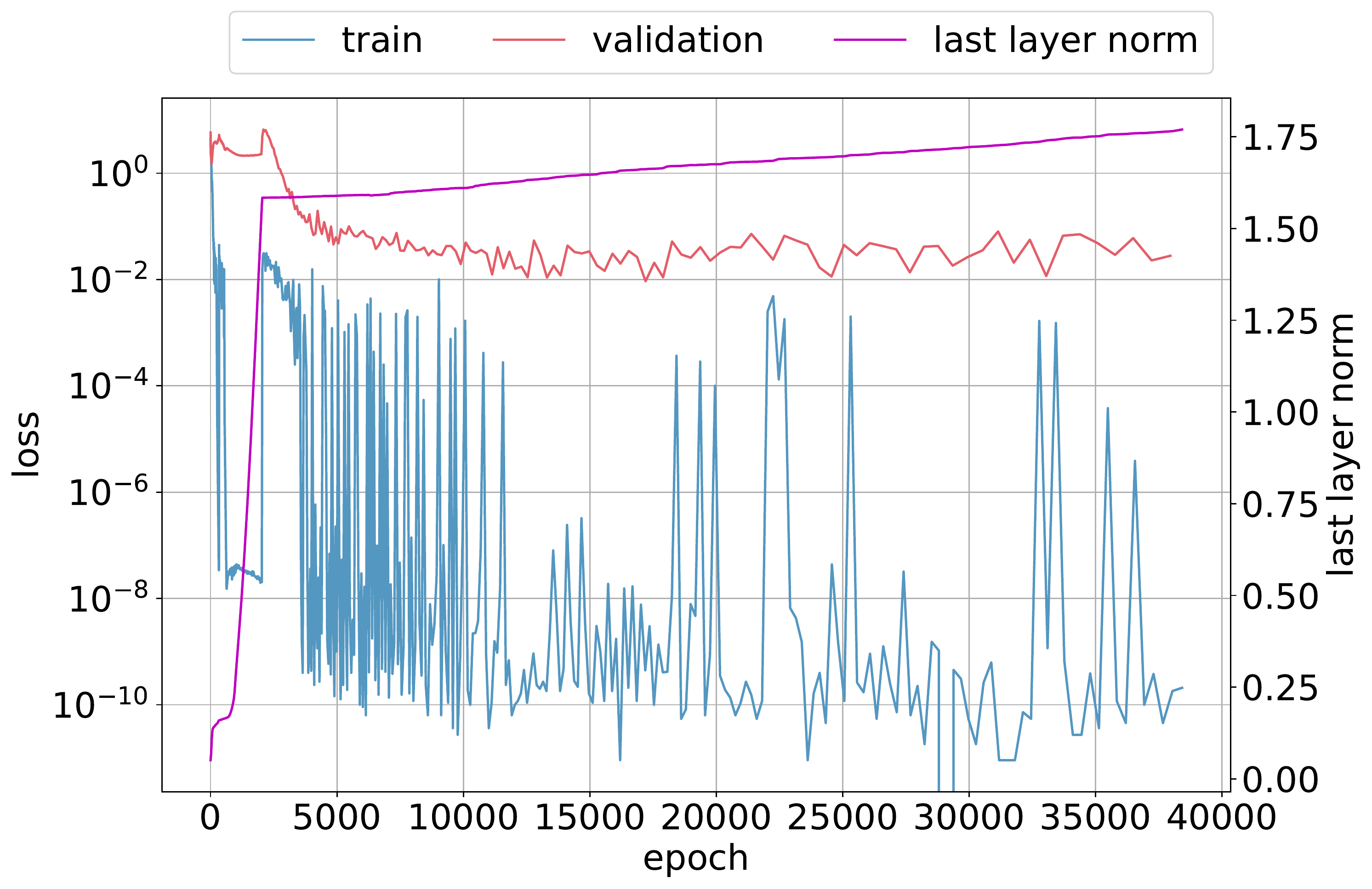} &
      \includegraphics[width=0.50\linewidth]{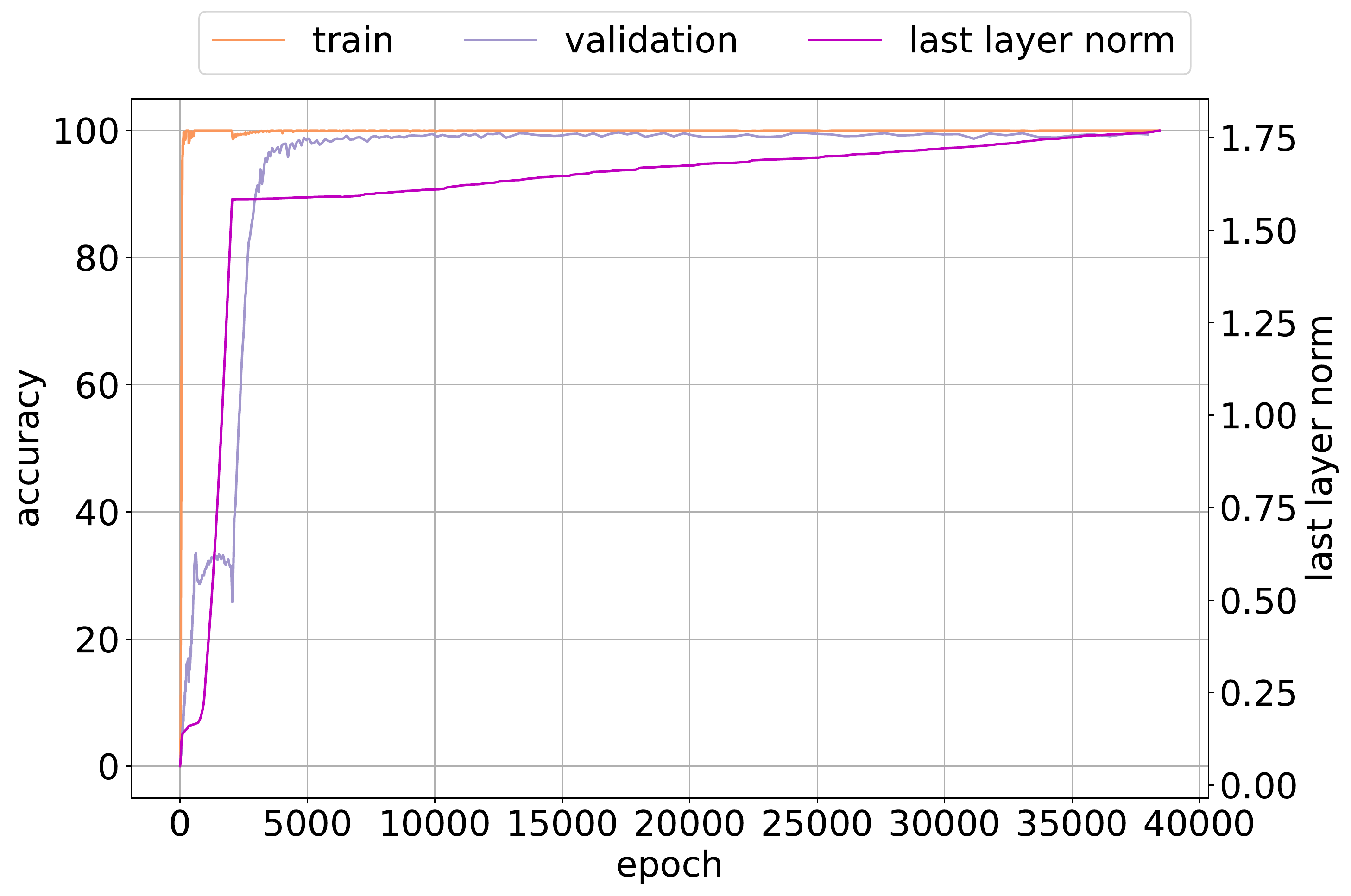} \\
      (a)  & (b)  \\
      & \\
  \end{tabular}
 \caption{Even-add-odd-subtraction dataset with 70/30 train/validation split. Even-add-odd-subtraction operation is given by $[a + b \pmod {p}$ if $a$ is even, otherwise $a - b \pmod {p}$] for $0\leq a,b < p$. Training and validation (a) loss and (b) accuracy.} 
 \label{fig:slingshot_evenodd70p}
\end{figure*}

\begin{figure*}[h]
\centering
  \begin{tabular}{cc}
      \includegraphics[width=0.50\linewidth]{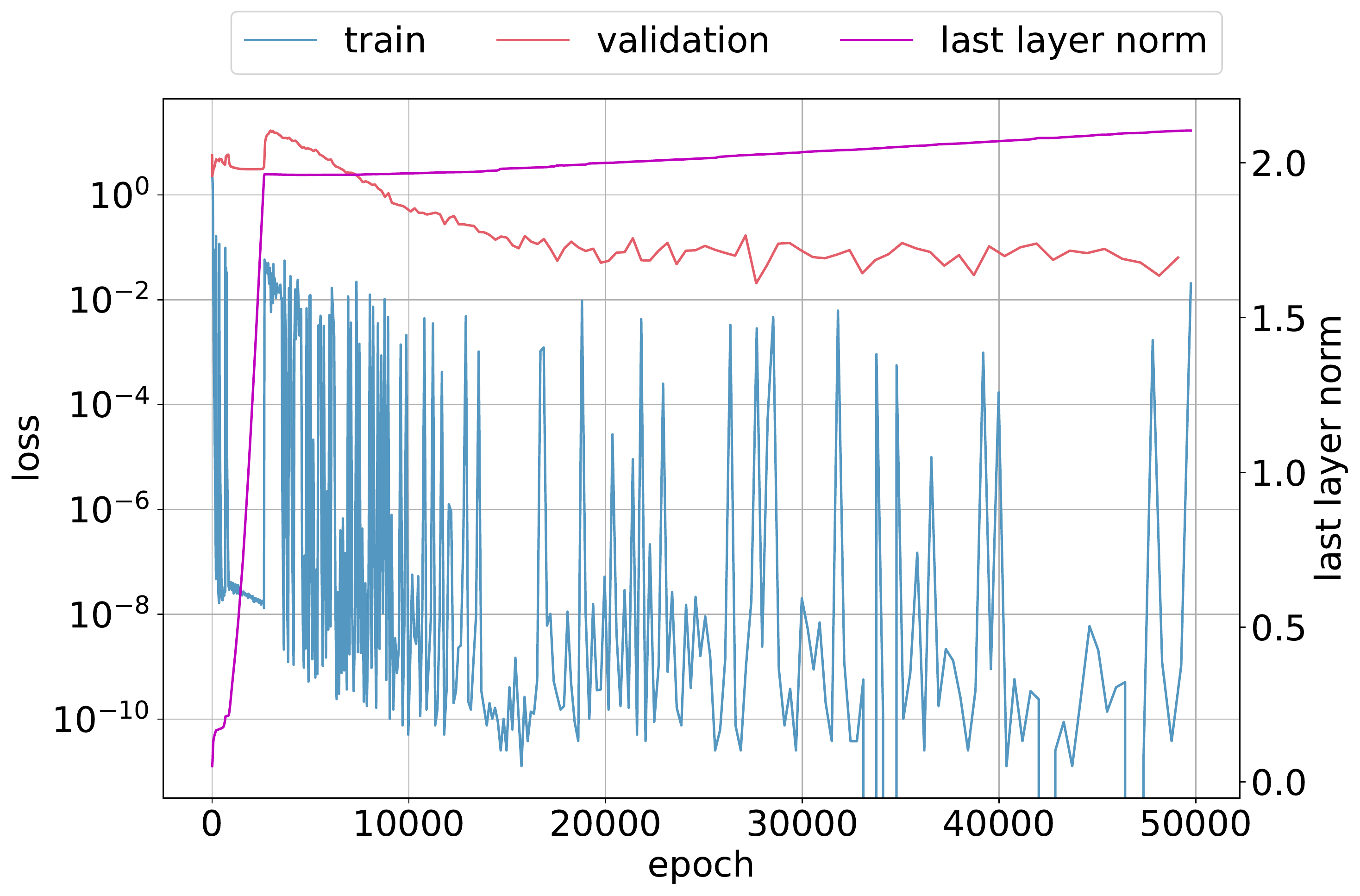} &
      \includegraphics[width=0.50\linewidth]{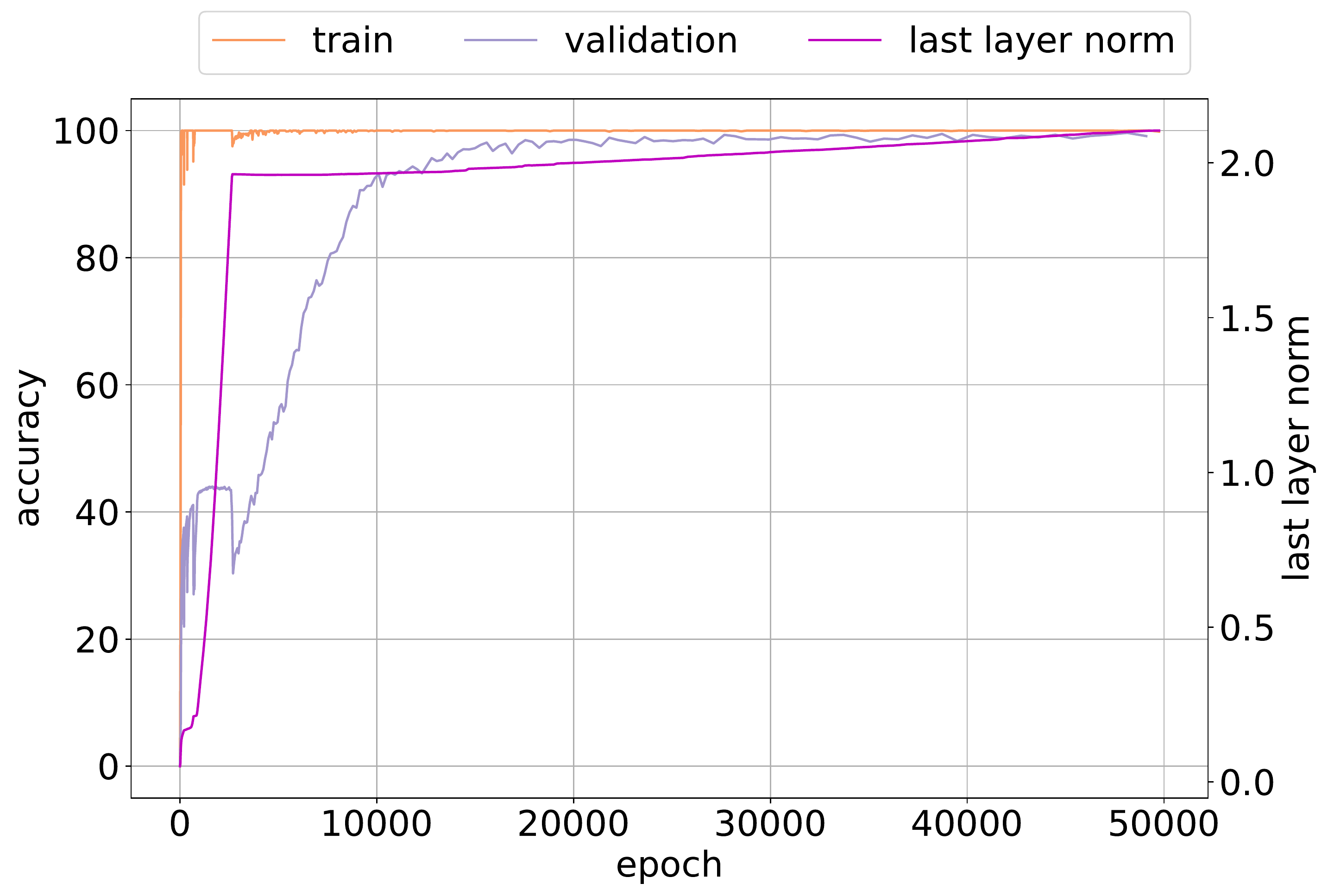} \\
      (a)  & (b)  \\
      & \\
  \end{tabular}
 \caption{Multiplication dataset with 50/50 train/validation split. Training and validation (a) loss and (b) accuracy.} 
 \label{fig:slingshot_mul50p}
\end{figure*}

\begin{figure*}[h]
\centering
  \begin{tabular}{cc}
      \includegraphics[width=0.50\linewidth]{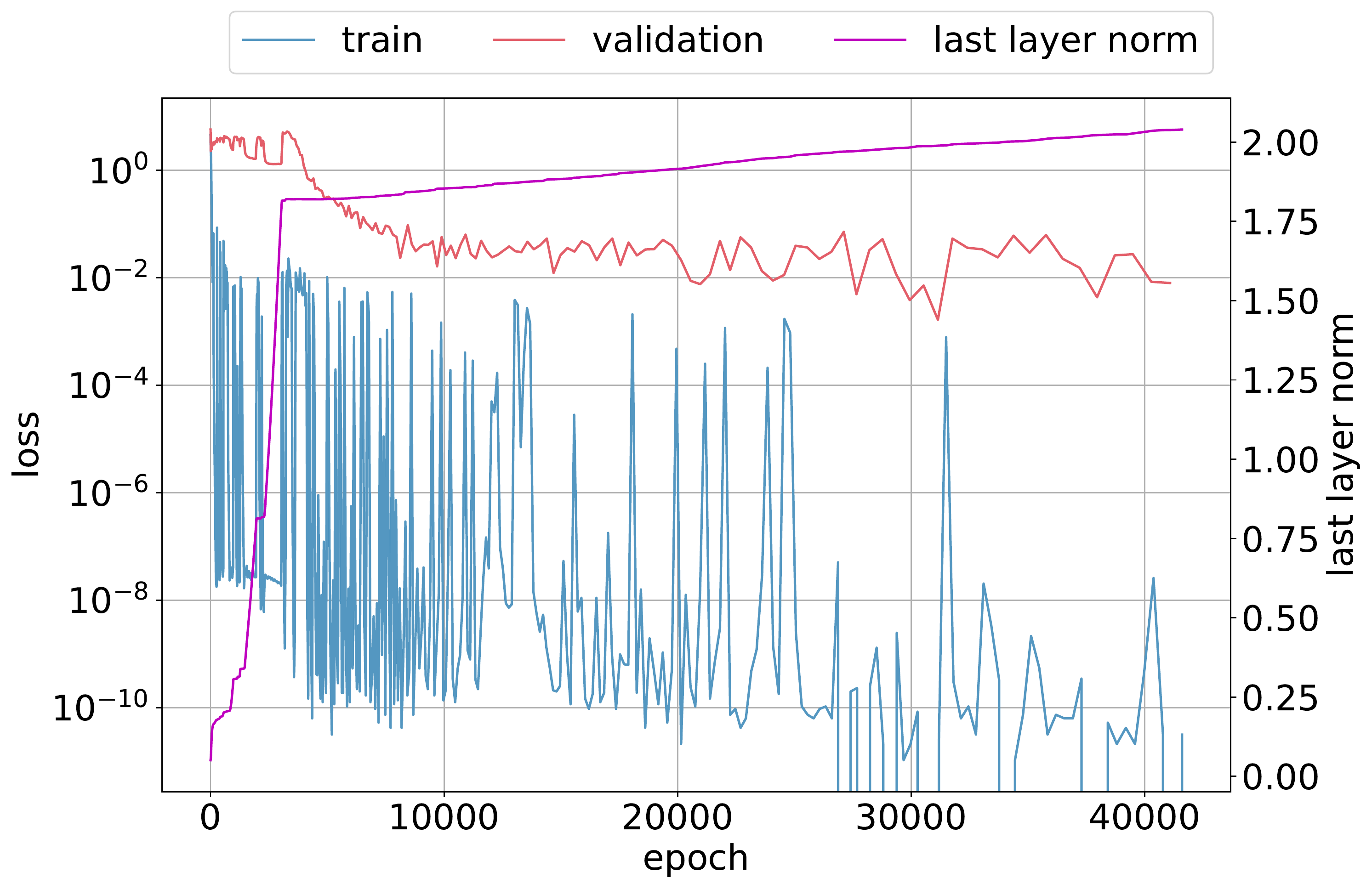} &
      \includegraphics[width=0.50\linewidth]{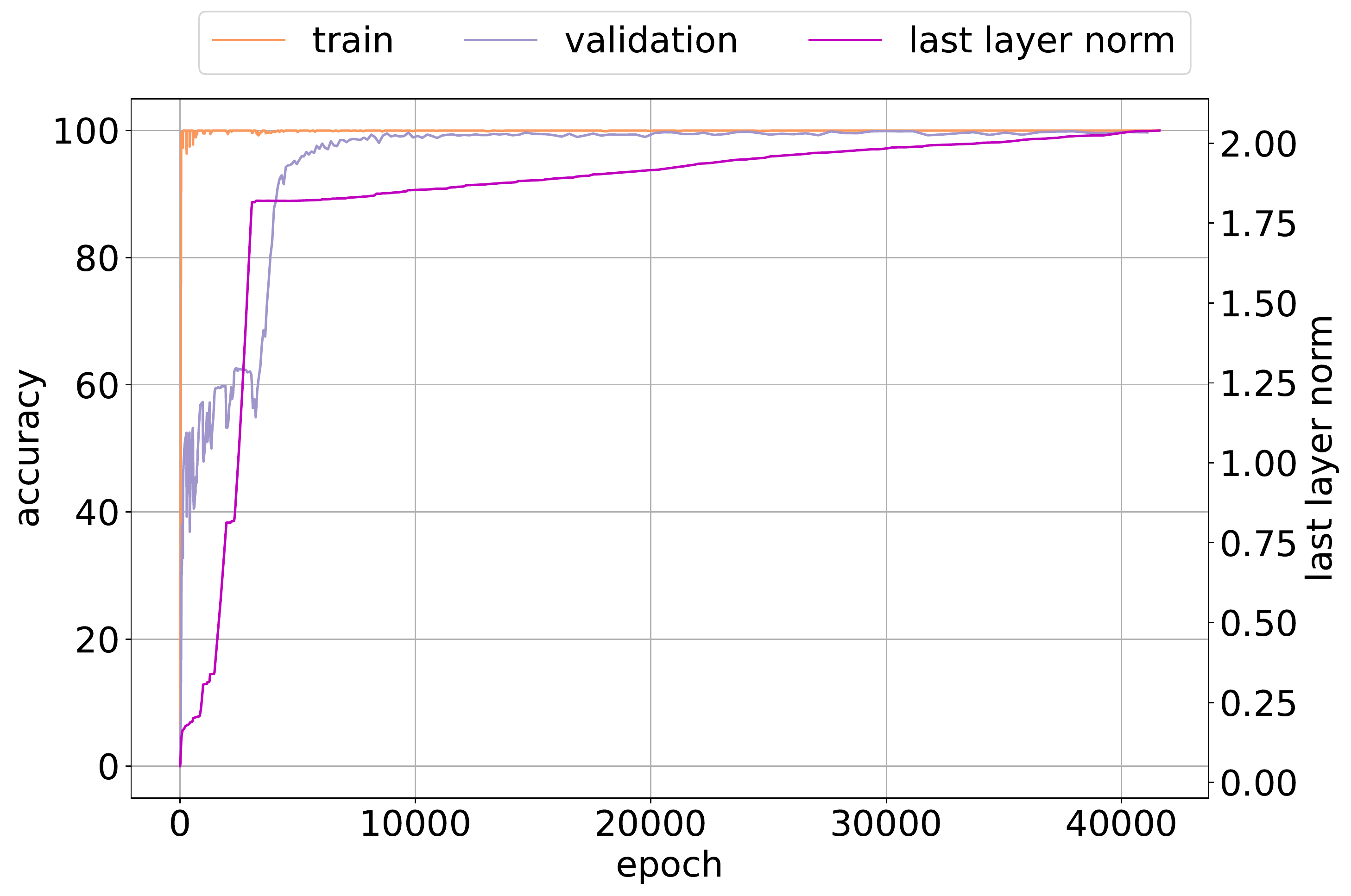} \\
      (a)  & (b)  \\
      & \\
  \end{tabular}
 \caption{Multiplication dataset with 60/40 train/validation split. Training and validation (a) loss and (b) accuracy.} 
 \label{fig:slingshot_mul60p}
\end{figure*}

\begin{figure*}[h]
\centering
  \begin{tabular}{cc}
      \includegraphics[width=0.50\linewidth]{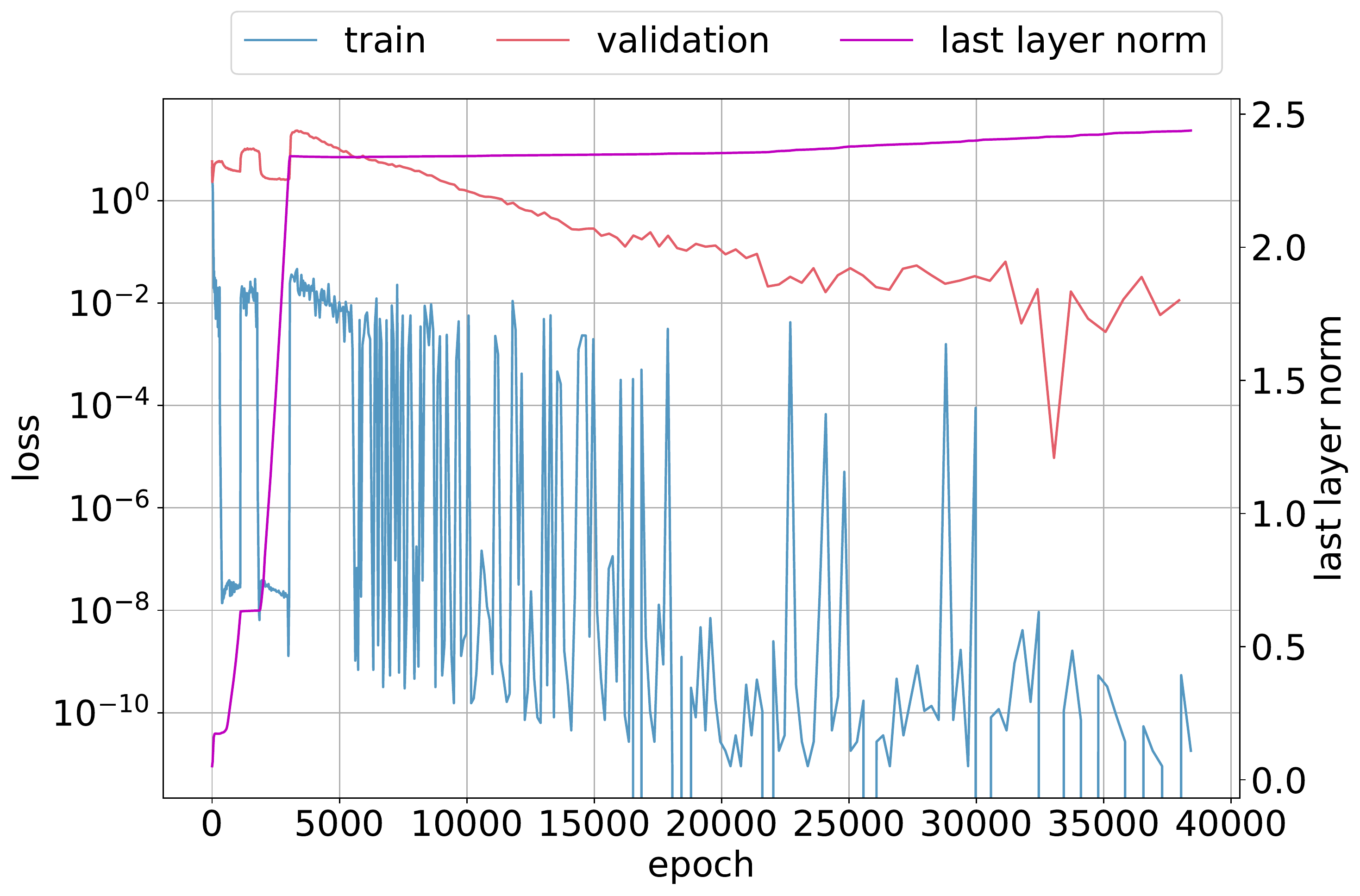} &
      \includegraphics[width=0.50\linewidth]{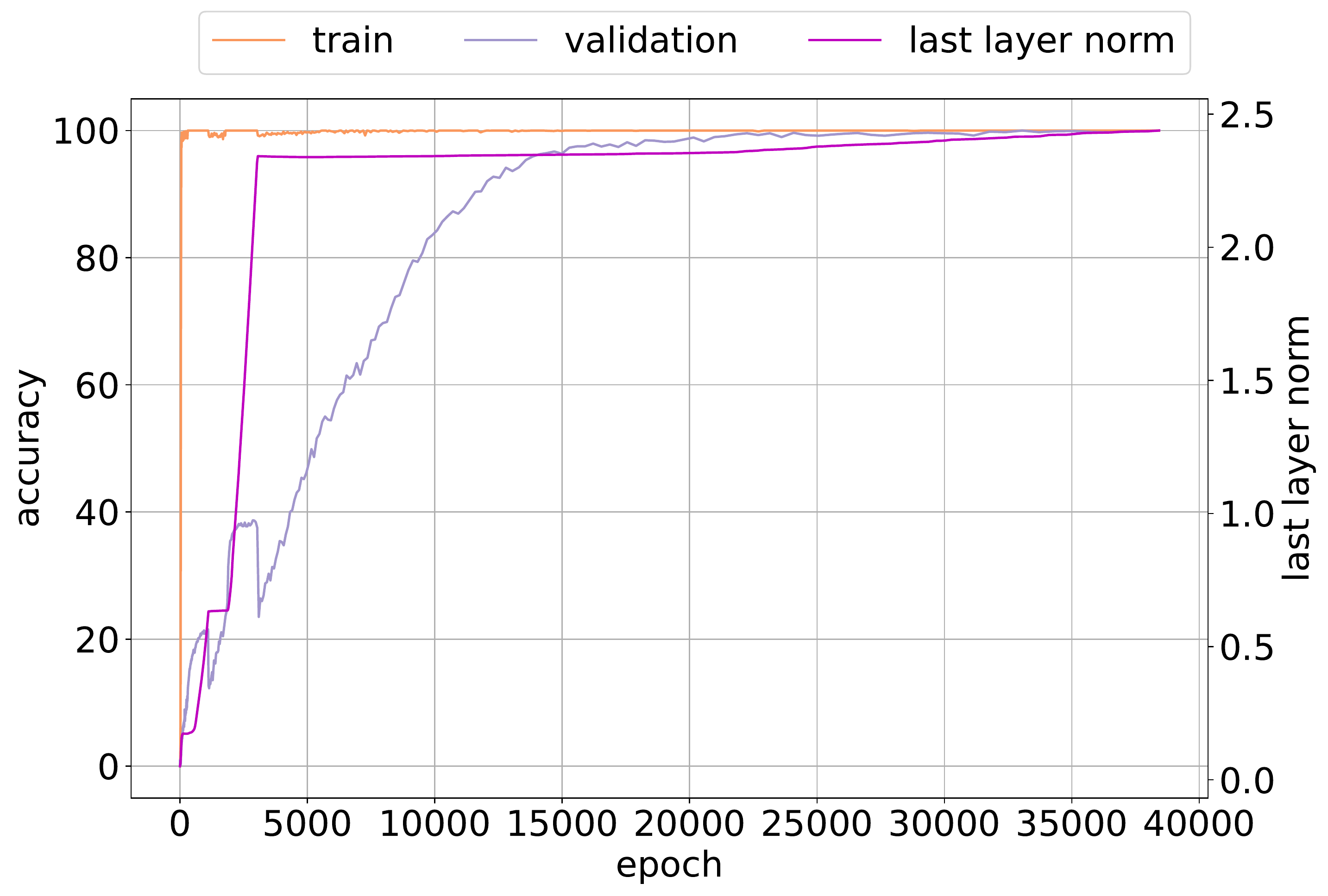} \\
      (a)  & (b)  \\
      & \\
  \end{tabular}
 \caption{Squarepoly dataset with 70/30 train/validation split. Squarepoly operation is given by $a^{2} + b \pmod {p}$ for $0 \leq a, b < p$. Training and validation (a) loss and (b) accuracy.} 
 \label{fig:slingshot_squarepoly70p}
\end{figure*}

\begin{figure*}[h]
\centering
  \begin{tabular}{cc}
      \includegraphics[width=0.50\linewidth]{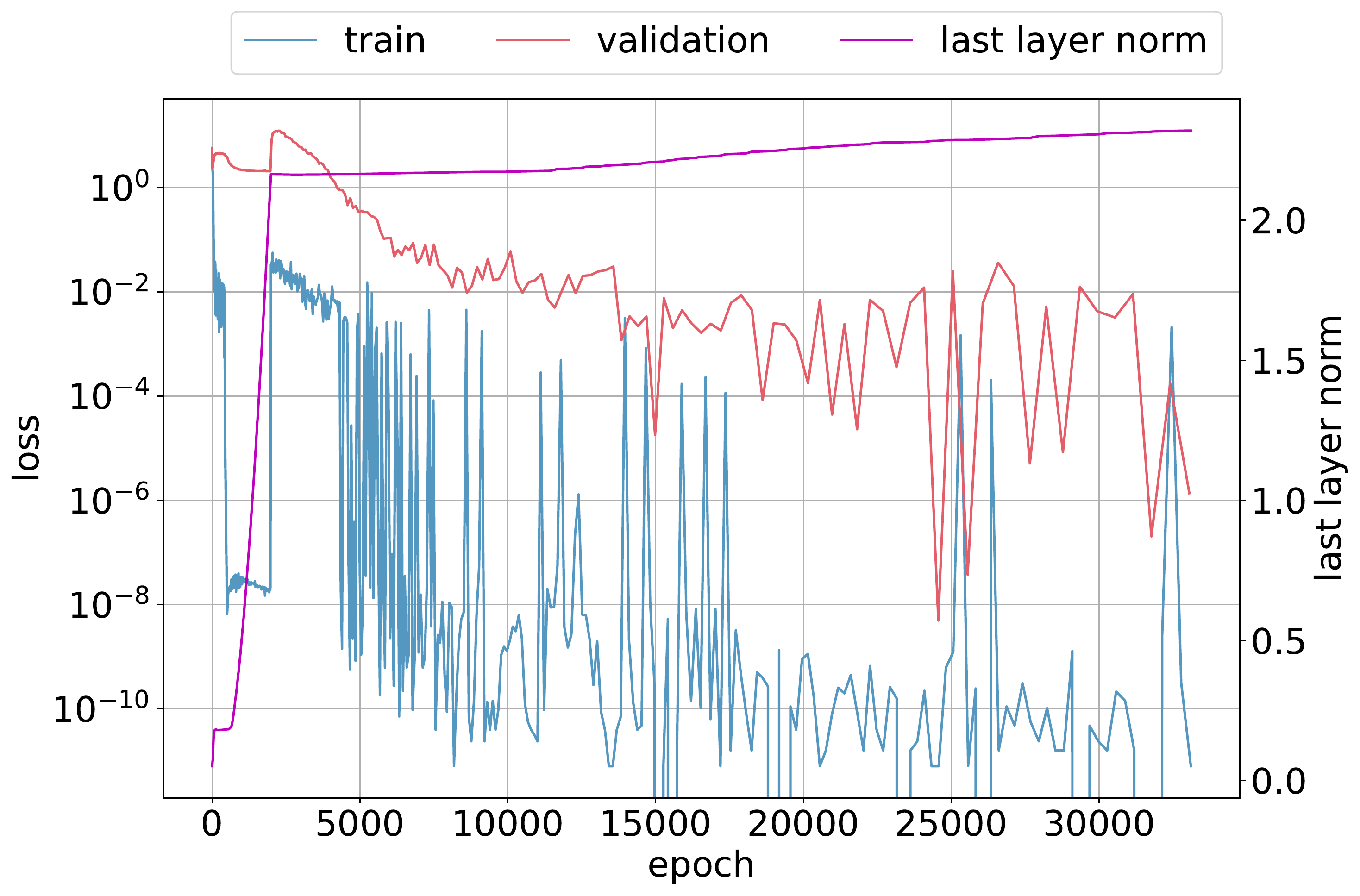} &
      \includegraphics[width=0.50\linewidth]{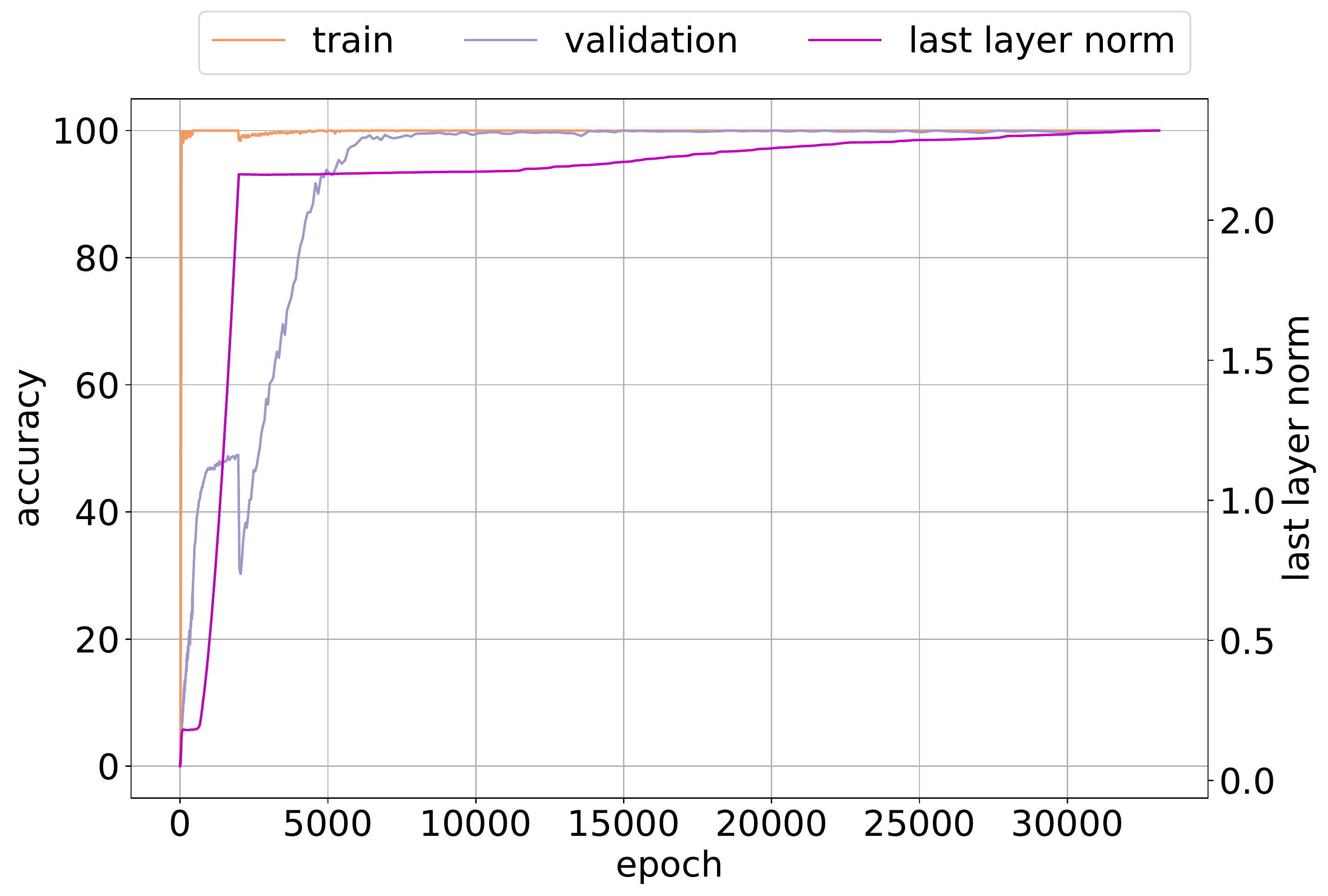} \\
      (a)  & (b)  \\
      & \\
  \end{tabular}
 \caption{Squarepoly dataset with 80/20 train/validation split. Squarepoly operation is given by $a^{2} + b \pmod {p}$ for $0 \leq a, b < p$. Training and validation (a) loss and (b) accuracy.} 
 \label{fig:slingshot_squarepoly80p}
\end{figure*}

\begin{figure*}[h]
\centering
  \begin{tabular}{cc}
      \includegraphics[width=0.50\linewidth]{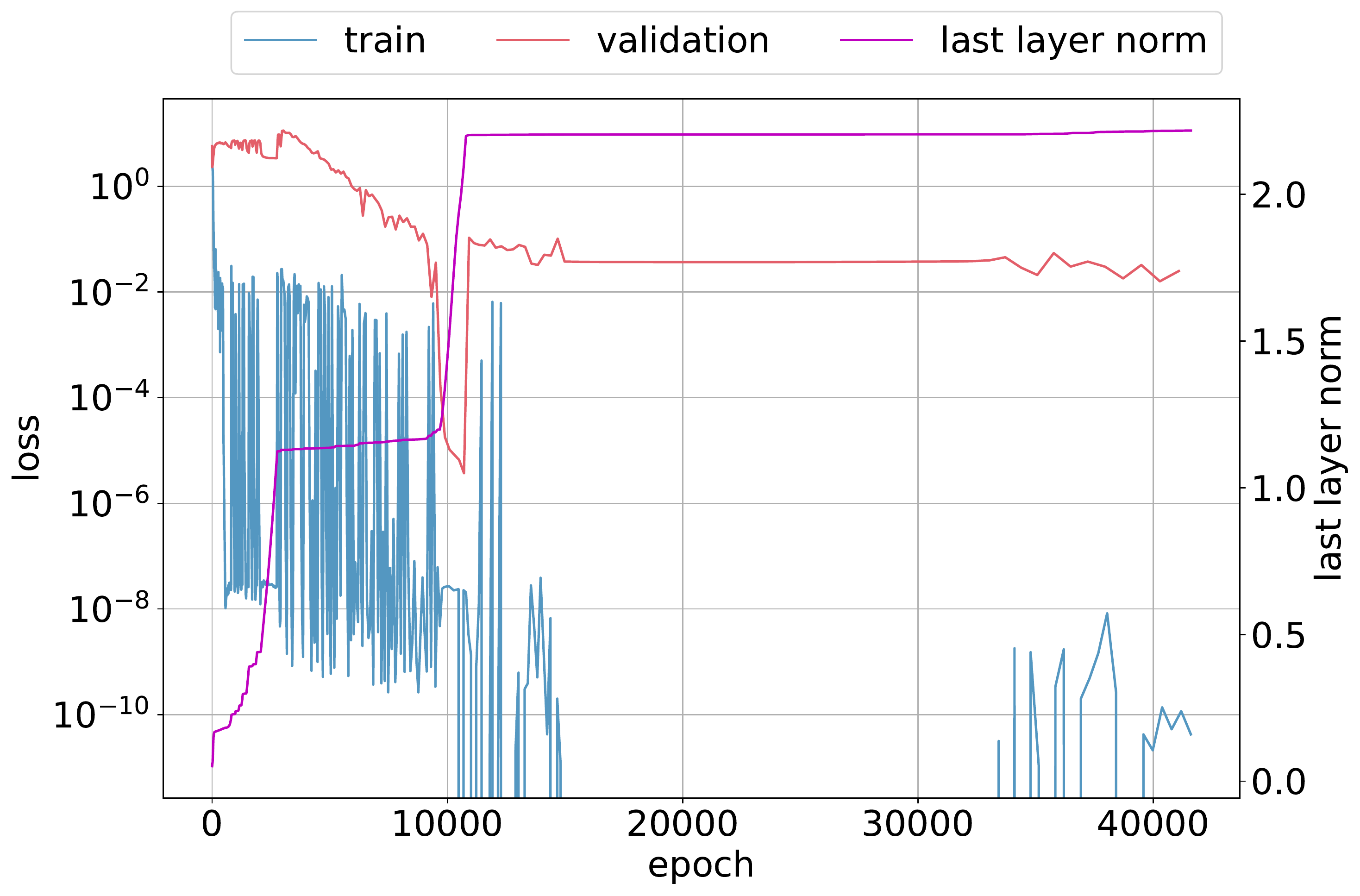} &
      \includegraphics[width=0.50\linewidth]{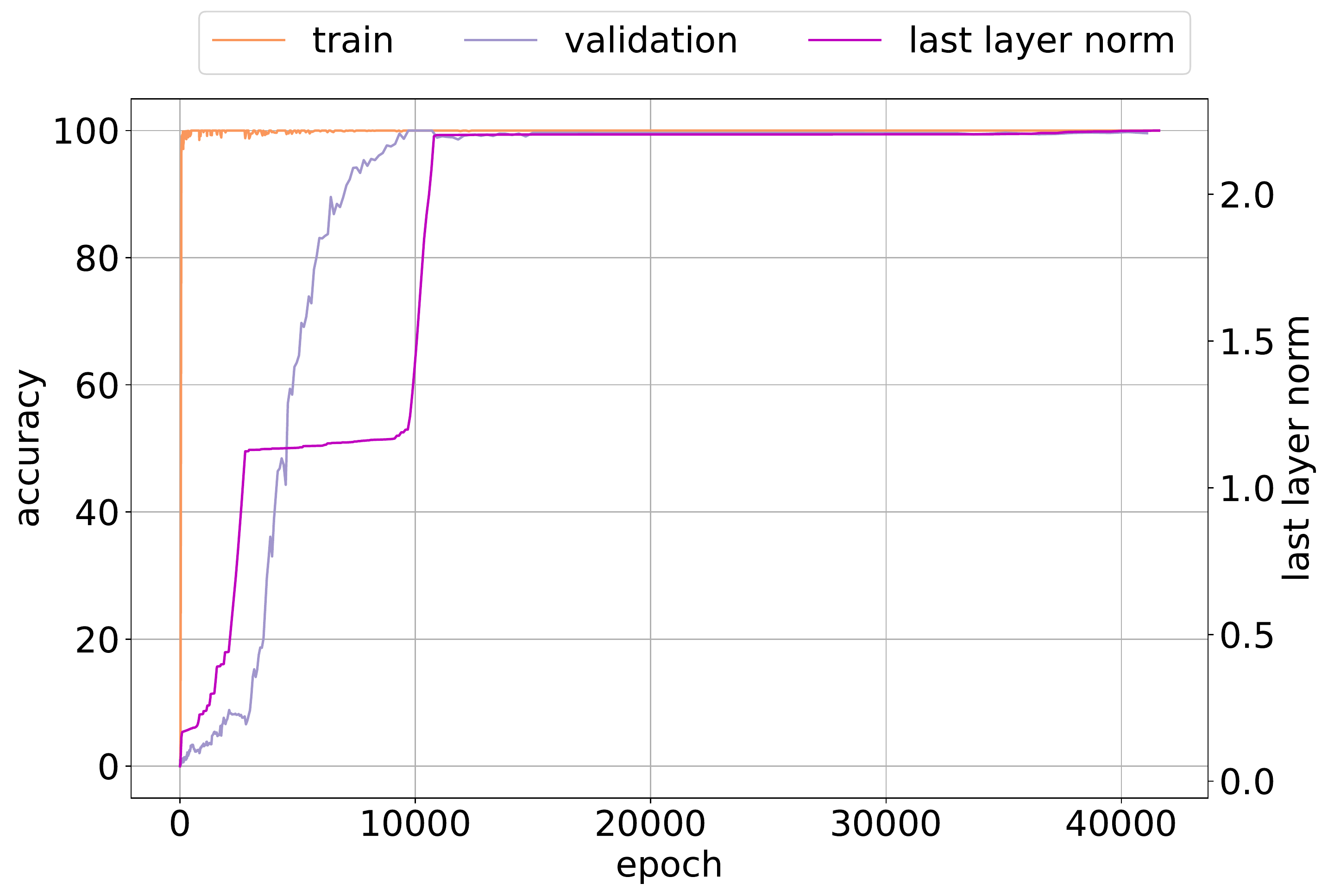} \\
      (a)  & (b)  \\
      & \\
  \end{tabular}
 \caption{Subtraction dataset with 60/40 train/validation split. Training and validation (a) loss and (b) accuracy.} 
 \label{fig:slingshot_sub60p}
\end{figure*}

\begin{figure*}[h]
\centering
  \begin{tabular}{cc}
      \includegraphics[width=0.50\linewidth]{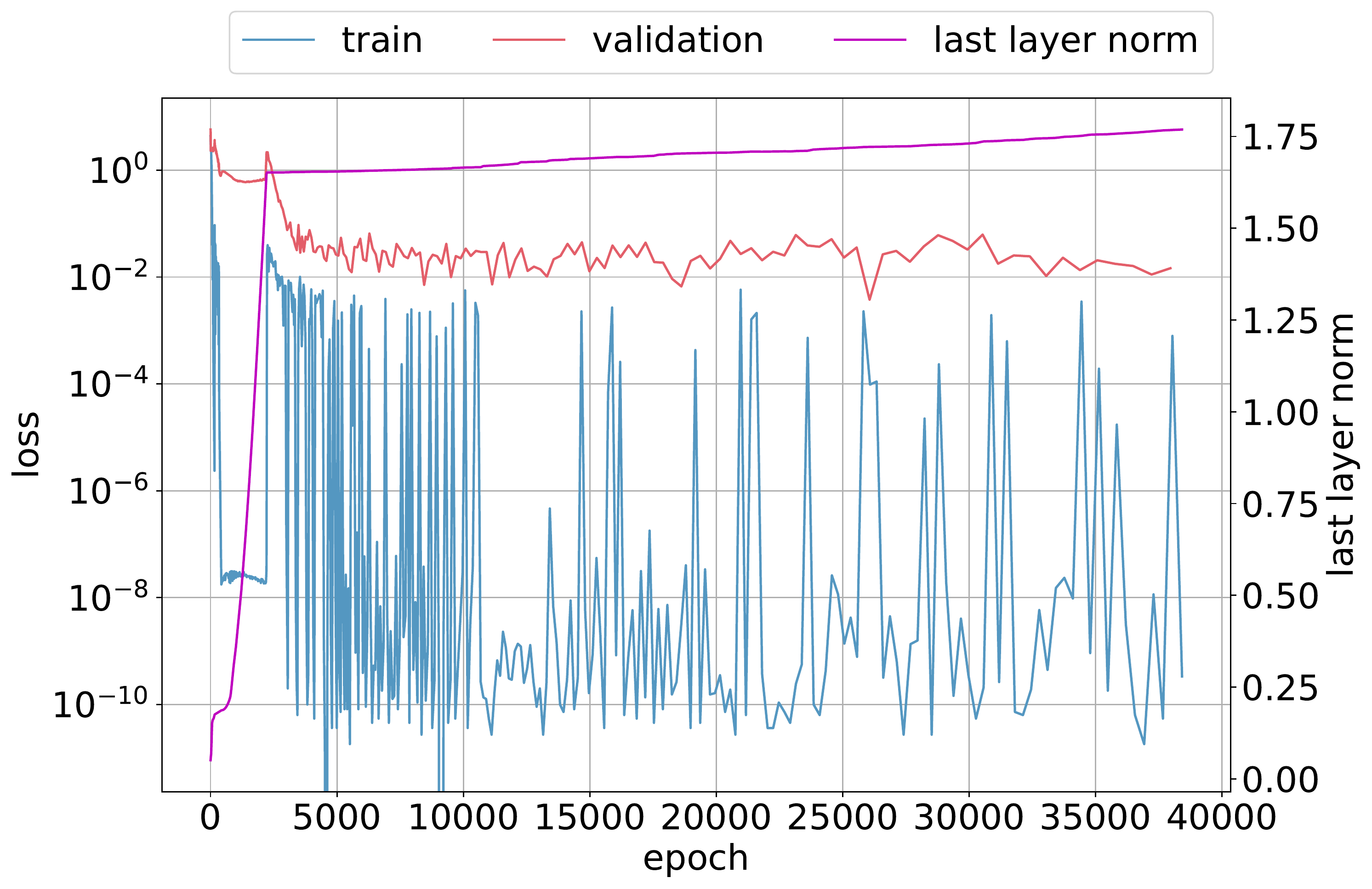} &
      \includegraphics[width=0.50\linewidth]{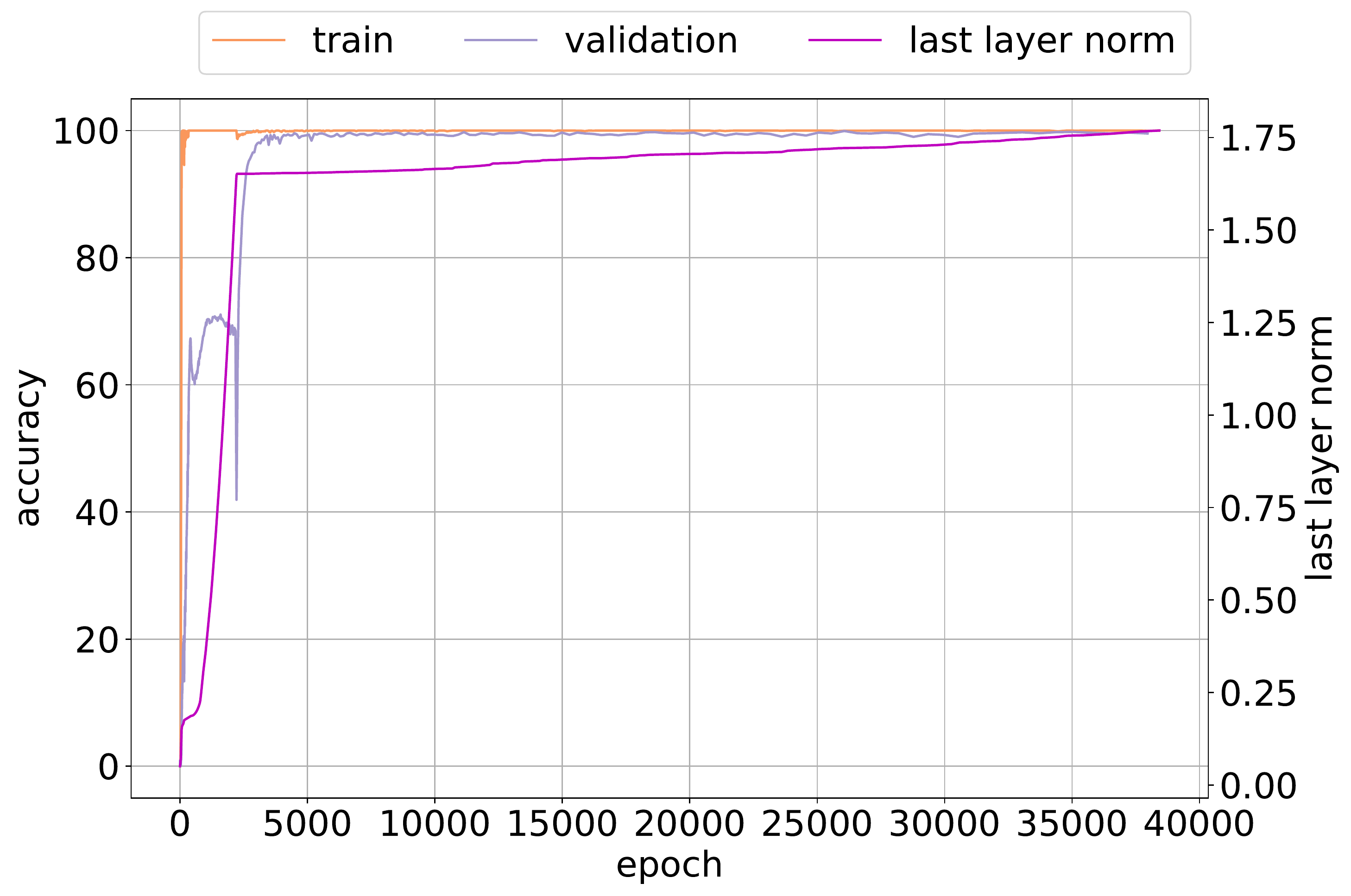} \\
      (a)  & (b)  \\
      & \\
  \end{tabular}
 \caption{Subtraction dataset with 70/30 train/validation split. Training and validation (a) loss and (b) accuracy.} 
 \label{fig:slingshot_sub70p}
\end{figure*}

\begin{figure*}[h]
\centering
  \begin{tabular}{cc}
      \includegraphics[width=0.50\linewidth]{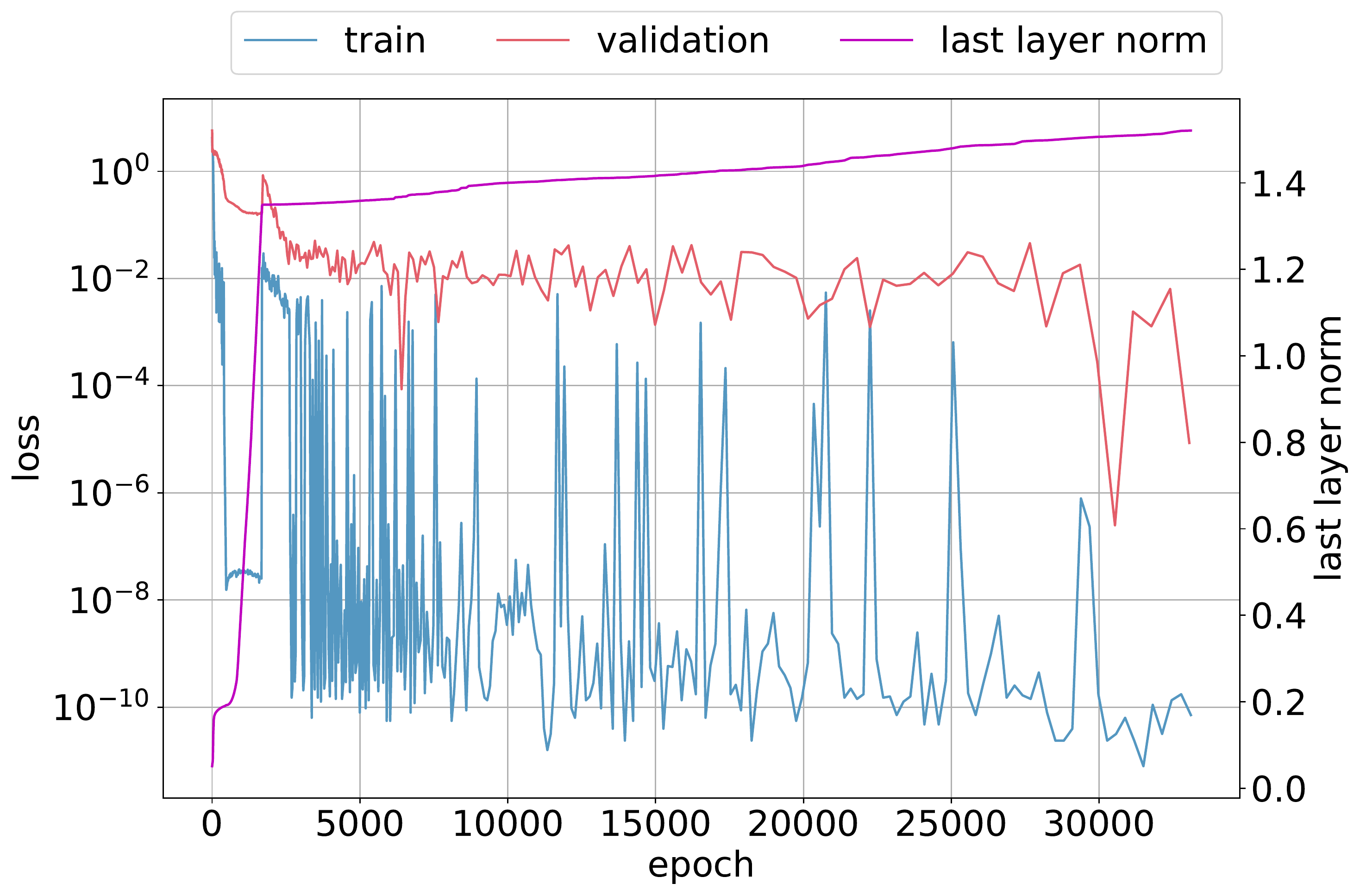} &
      \includegraphics[width=0.50\linewidth]{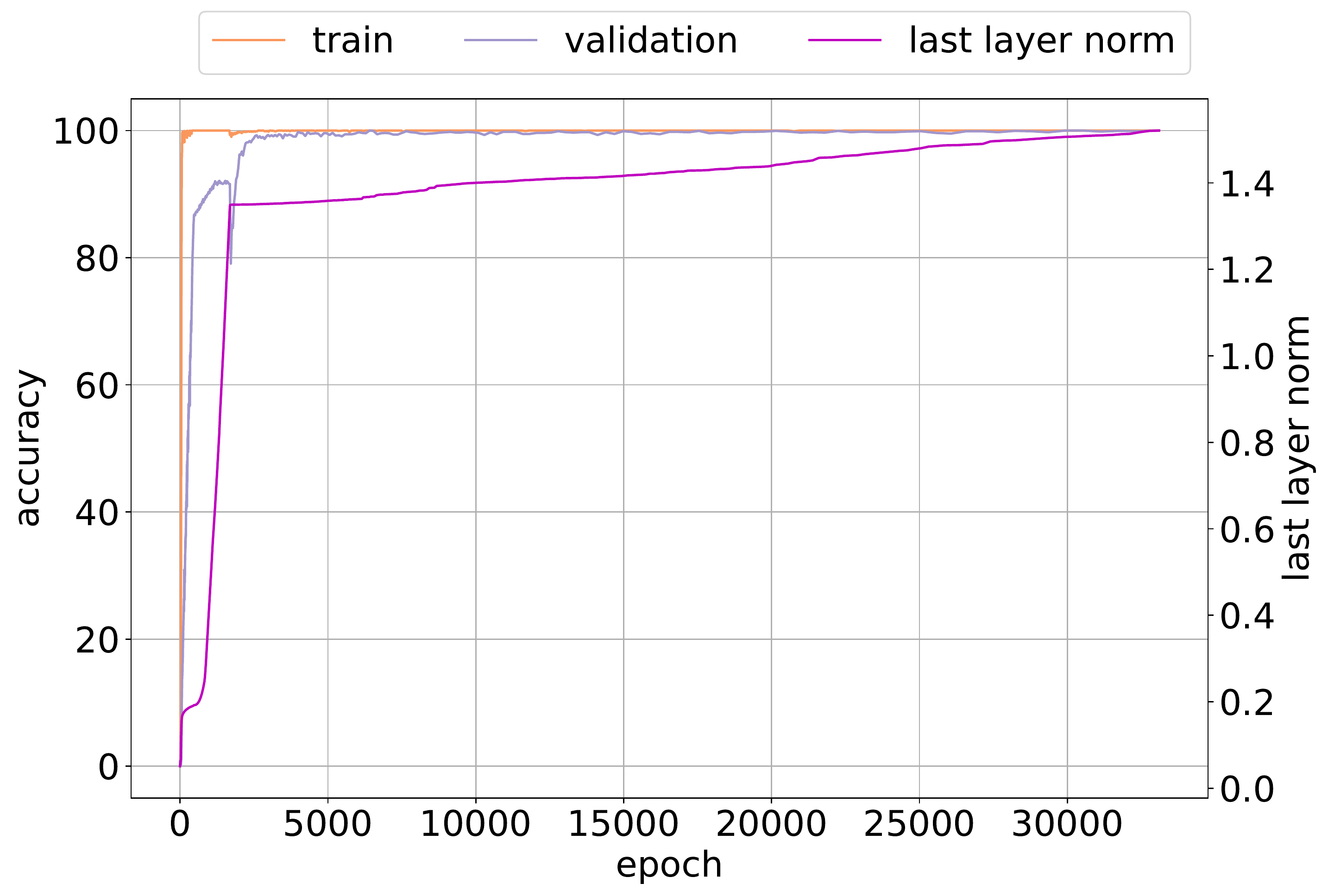} \\
      (a)  & (b)  \\
      & \\
  \end{tabular}
 \caption{Subtraction dataset with 80/20 train/validation split. Training and validation (a) loss and (b) accuracy.} 
 \label{fig:slingshot_sub80p}
\end{figure*}

In this appendix, we provide evidence of Slingshot Effects on additional datasets from Power et al~\cite{power2021grokking} Grokking work. The datasets are created by a subset of mathematical operations defined in Appendix~\ref{appendix:xformers_setup}. Each operation can have multiple datasets that depends on the train/validation split ratio. We use the training setup described in~\ref{appendix:xformers_setup} on $18$ separate datasets. Figure~\ref{fig:slingshot_add50p} - Figure~\ref{fig:slingshot_sub80p} shows the results the datasets described in this appendix. We observe Slingshot Effects and generalization with all $18$ datasets. These results suggest the prevalence of Slingshot Effects when large models are trained with adaptive optimizers, specifically Adam~\cite{kingma2014adam}.

\clearpage
\section{Controlling Instability Through Normalization and Norm Constraints} 

Training instability is the hallmark of the Slingshot Mechanism, yet as seen in previous sections, the Slingshot Effect typically results in improved performance, and Grokking. In this section, we explore whether it is possible to maintain stable training, without sacrificing performance. To this end, we explore how constraining and regularizing the weights of the network affect the Slingshot behaviour, and overall performance.

\subsection{Weight decay}

\begin{figure*}[h]
\centering
  \begin{tabular}{ccc}
      weight decay = 0  & weight decay = 0.1 & weight decay = 1.0 \\
      \includegraphics[width=0.33\linewidth]{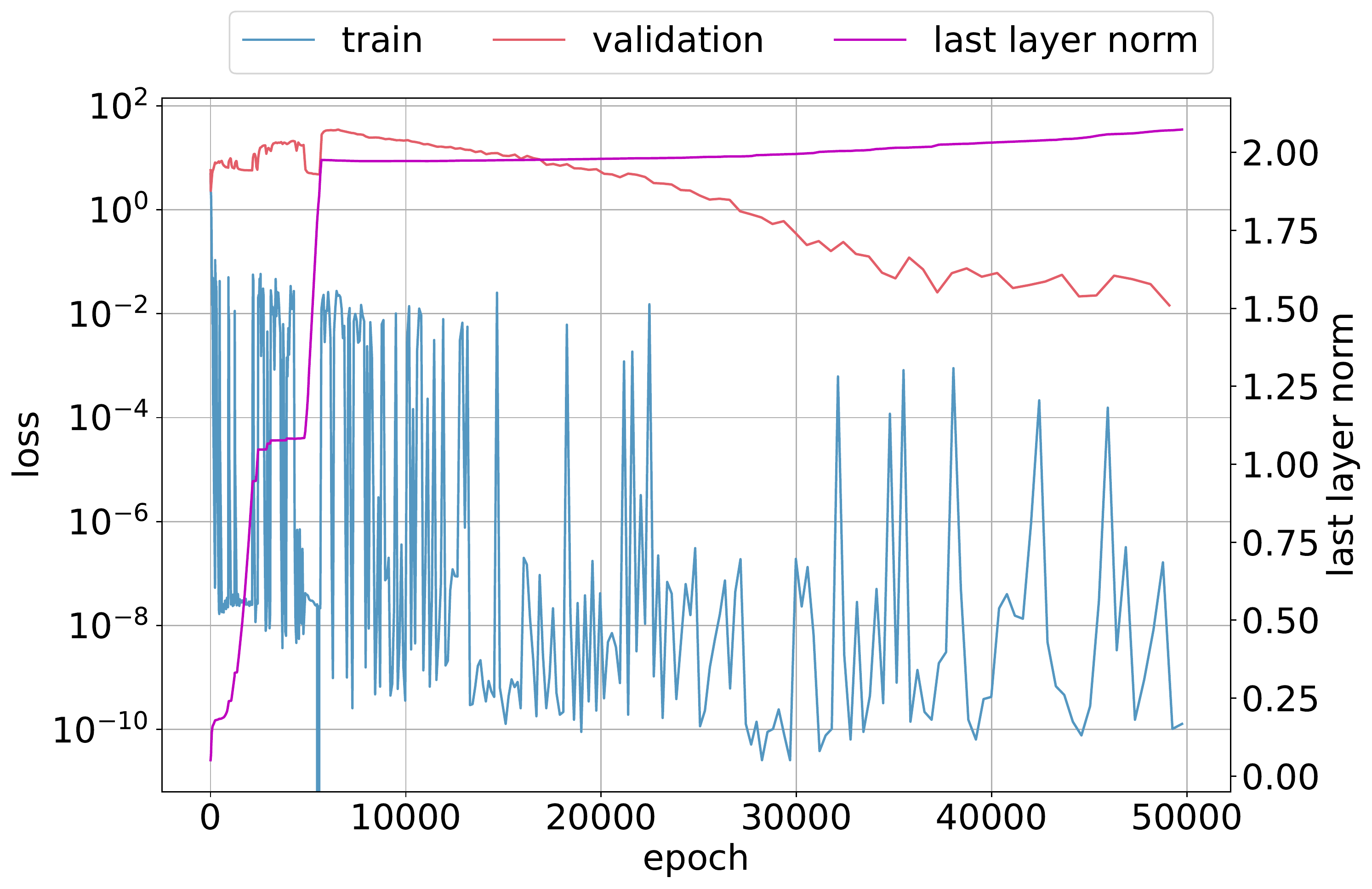} & 
      \includegraphics[width=0.33\linewidth]{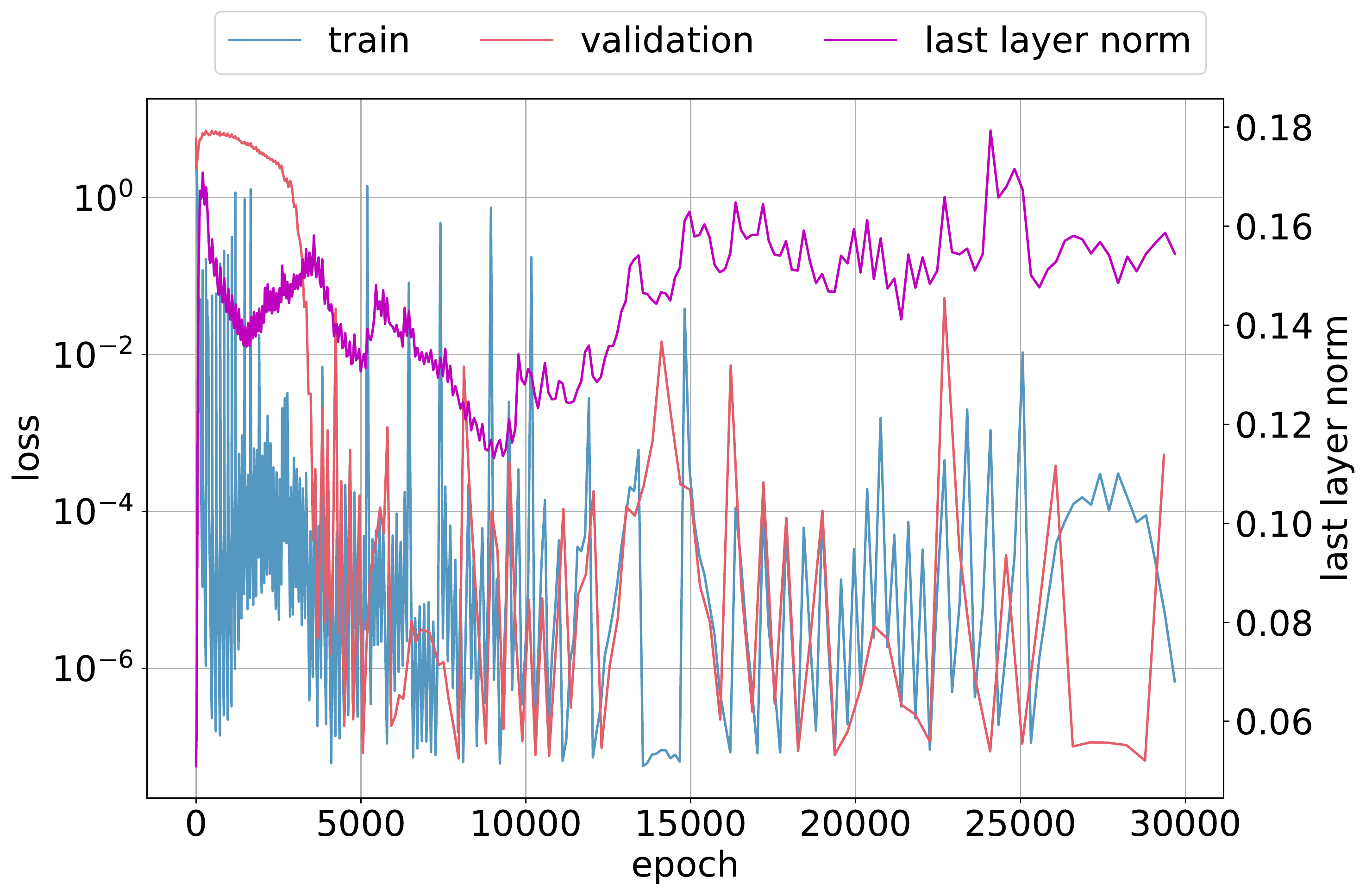} & 
      \includegraphics[width=0.33\linewidth]{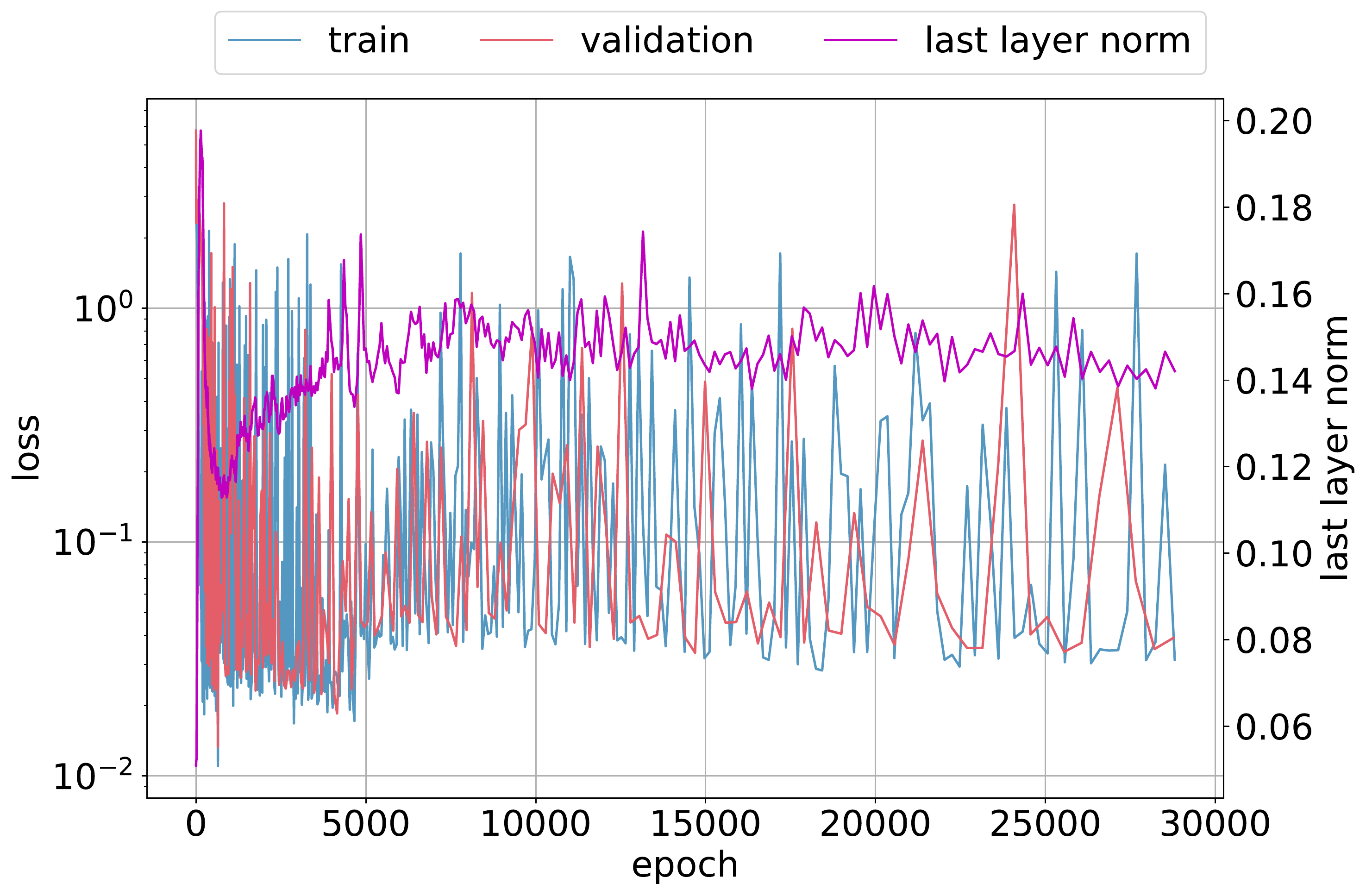} \\
      (a)  & (b) & (c) \\
      & train  and validation loss vs epochs \\
      & \\
      \includegraphics[width=0.33\linewidth]{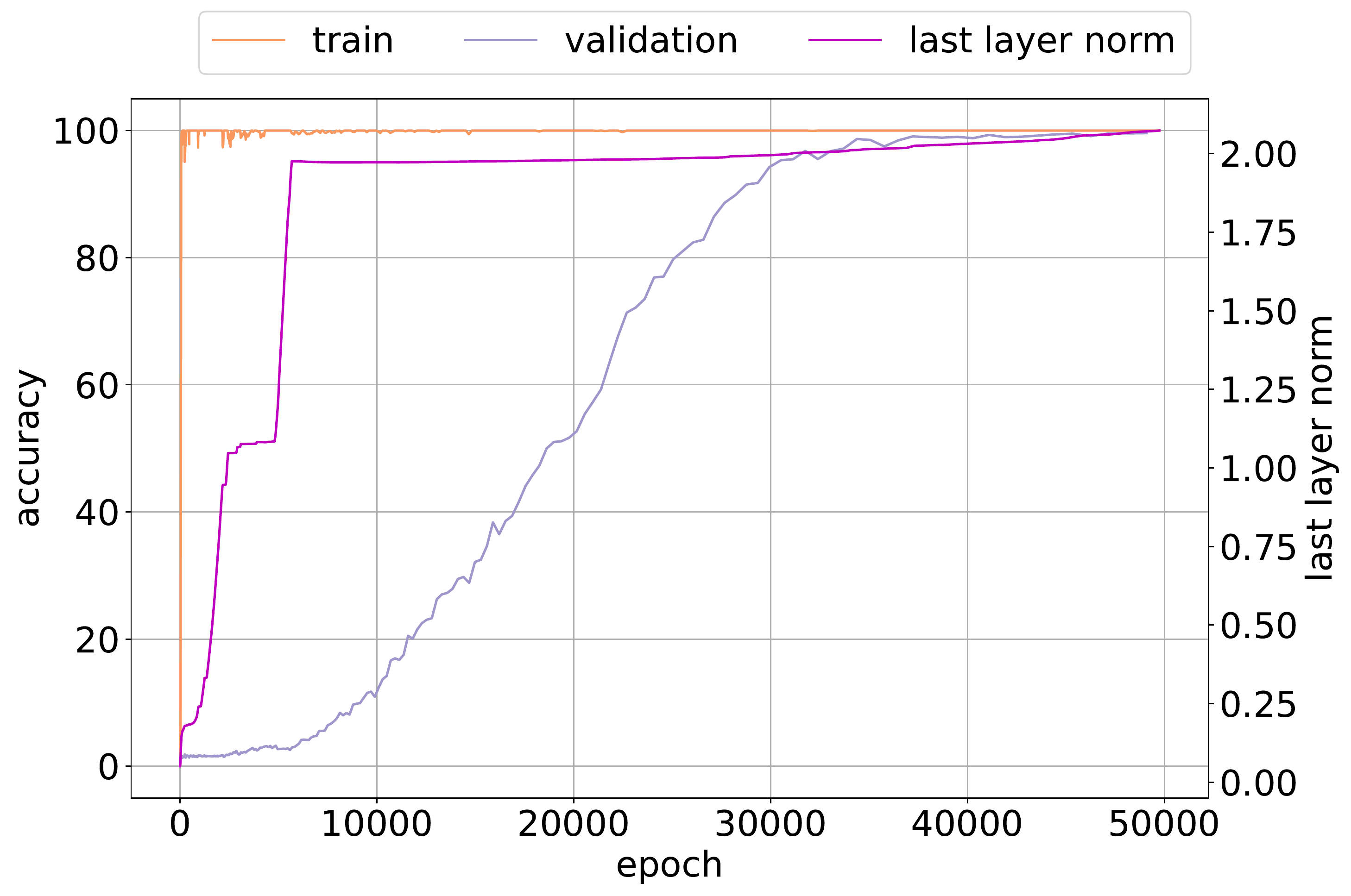} & 
      \includegraphics[width=0.33\linewidth]{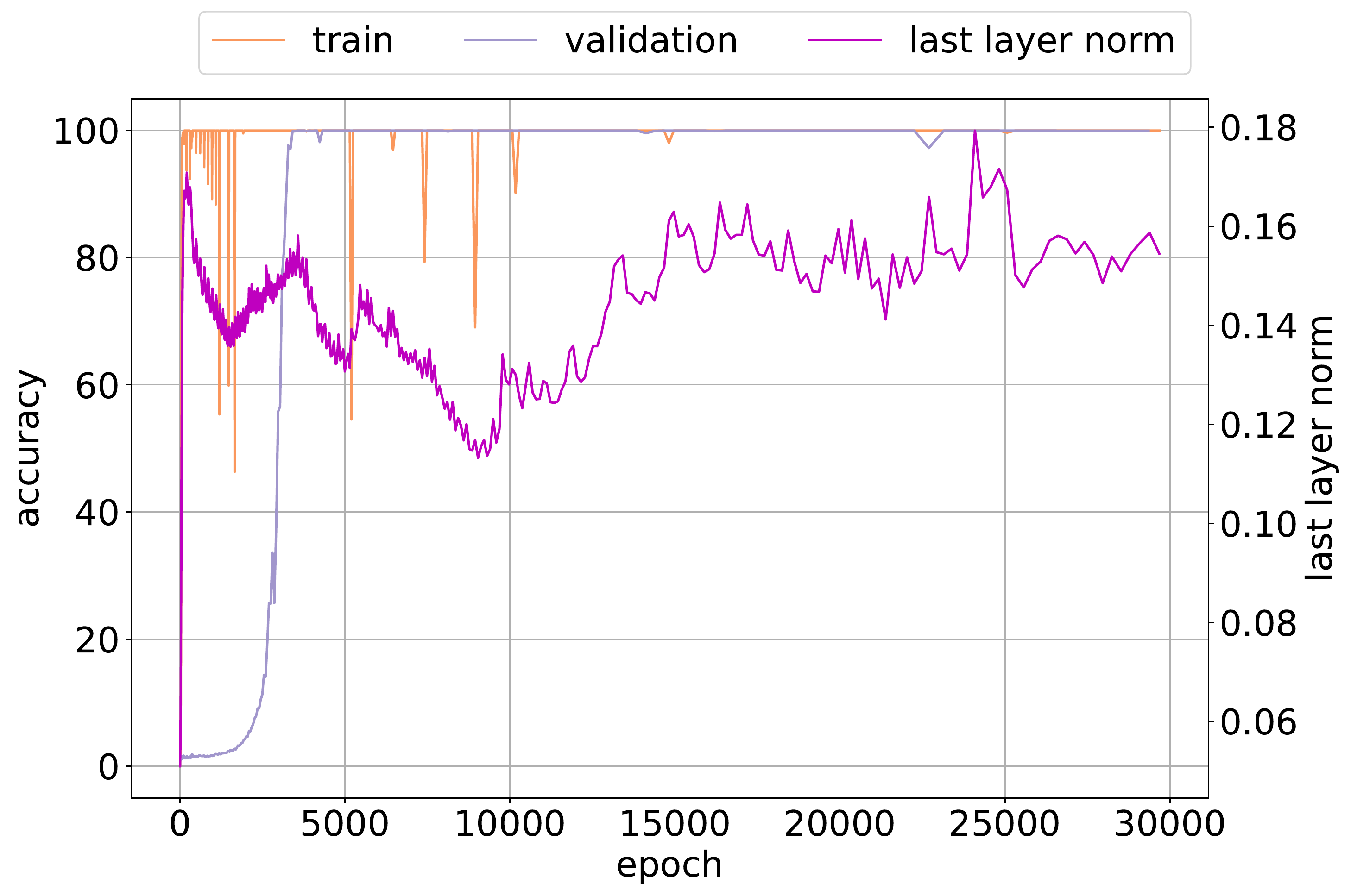} &    \includegraphics[width=0.33\linewidth]{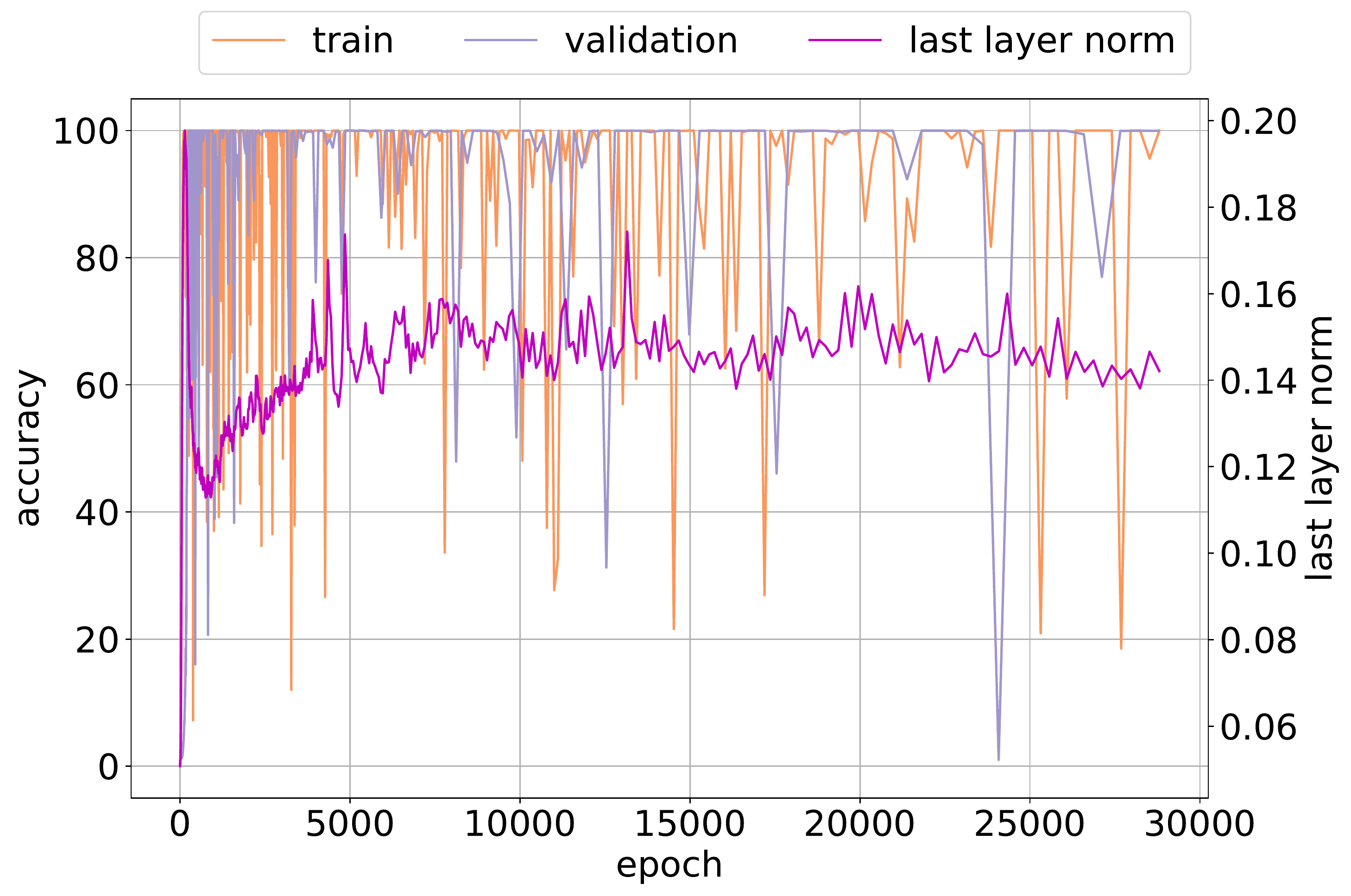} \\
      (d)  & (e) & (f) \\
      & train  and validation accuracy vs epochs \\
      & \\
  \end{tabular}
 \caption{Division dataset: Norm behavior with different weight decay values. Training and validation loss vs epochs with weight decay (a) 0.0, (b) 0.1, (c) 1.0; Training and validation accuracy vs epochs shown in (d), (e) and (f). The evolution of classifier weight norm shows instability as increase in weight decay strength.} 
 \label{fig:grok_vary_wd_div}
\end{figure*}

\begin{figure*}[h]
\centering
  \begin{tabular}{ccc}
      weight decay = 0  & weight decay = 0.1 & weight decay = 1.0 \\
      \includegraphics[width=0.33\linewidth]{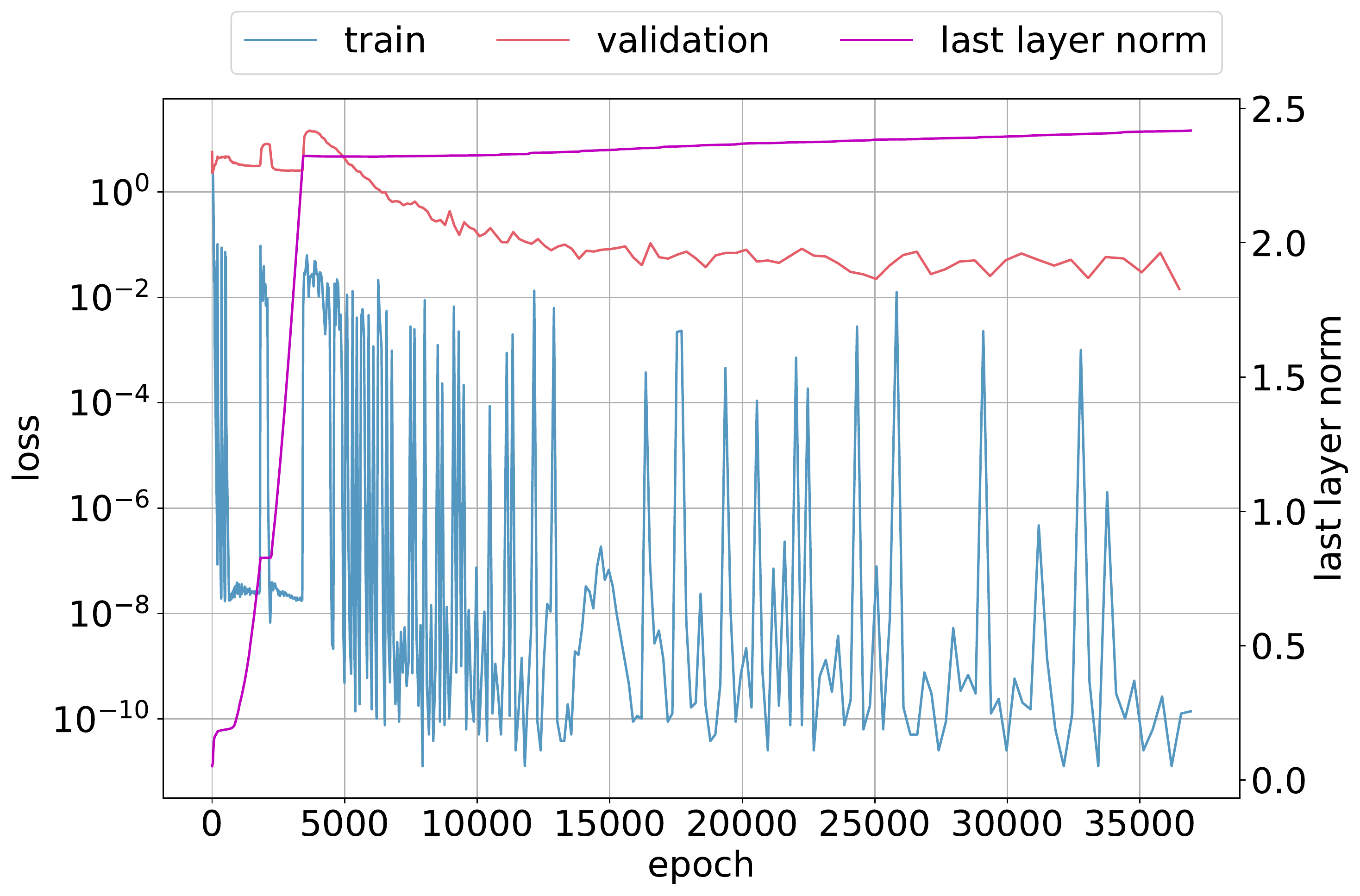} & 
      \includegraphics[width=0.33\linewidth]{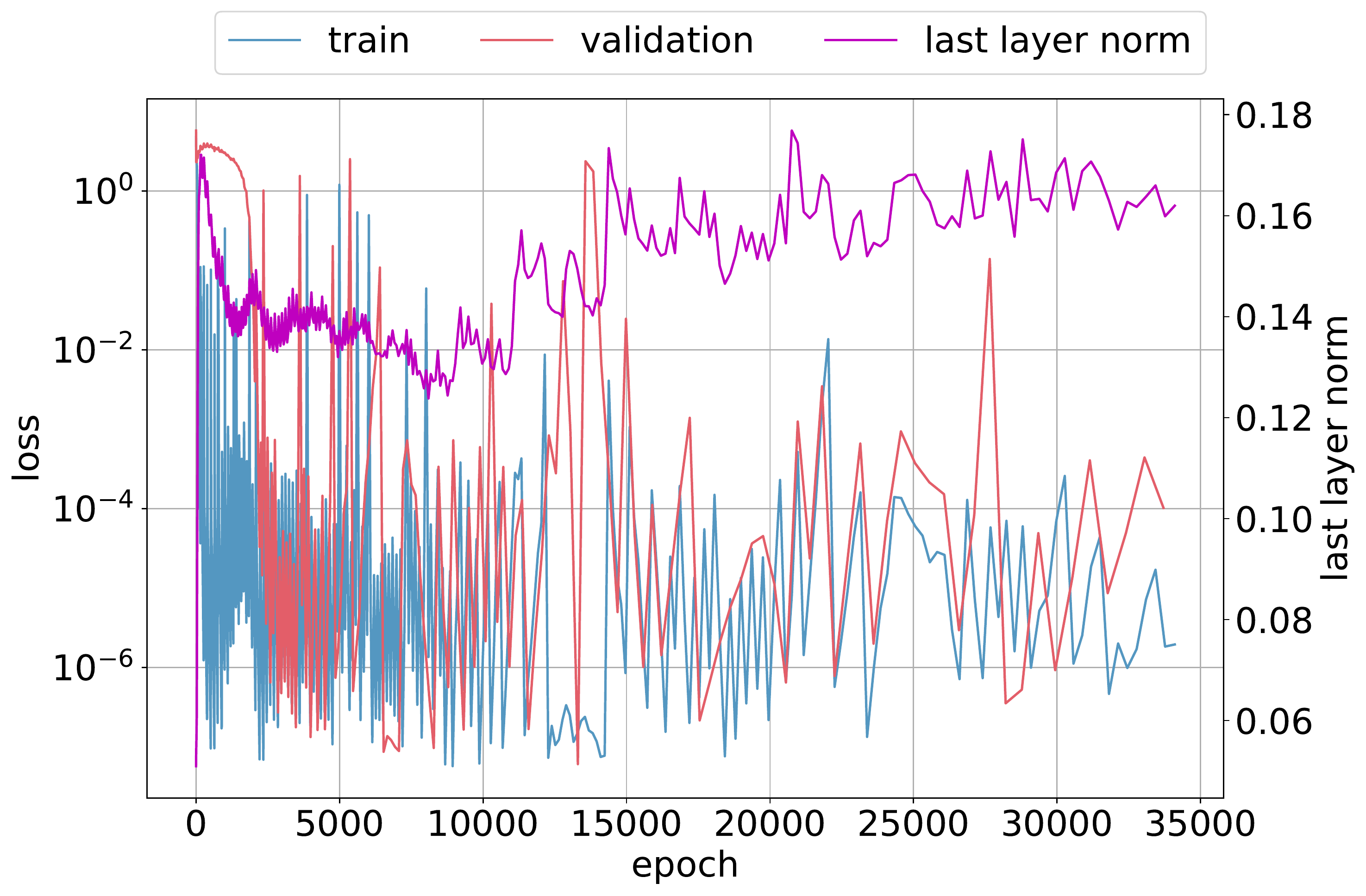} & 
      \includegraphics[width=0.33\linewidth]{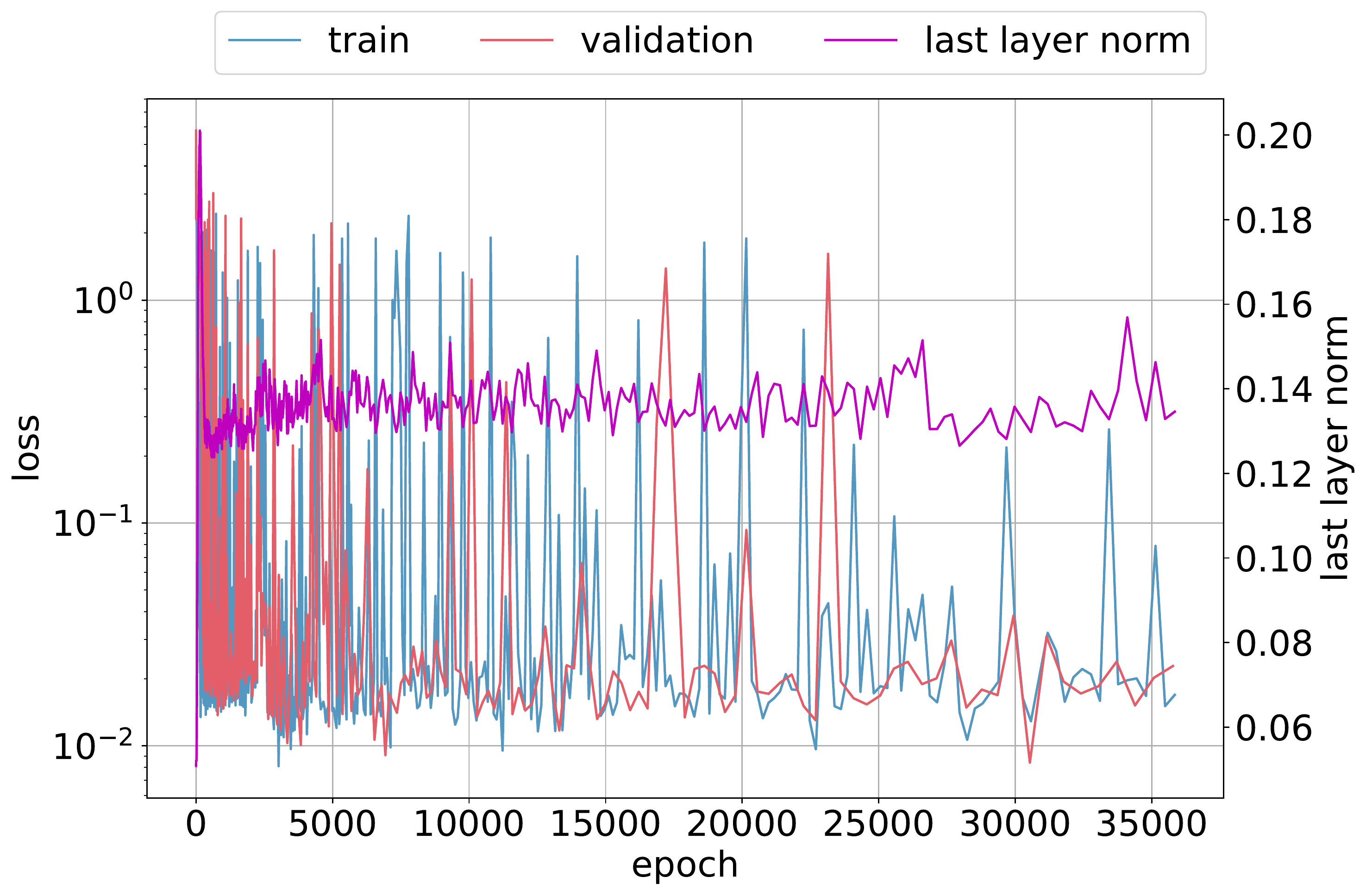} \\
      (a)  & (b) & (c) \\
      & train  and validation loss vs epochs \\
      & \\
      \includegraphics[width=0.33\linewidth]{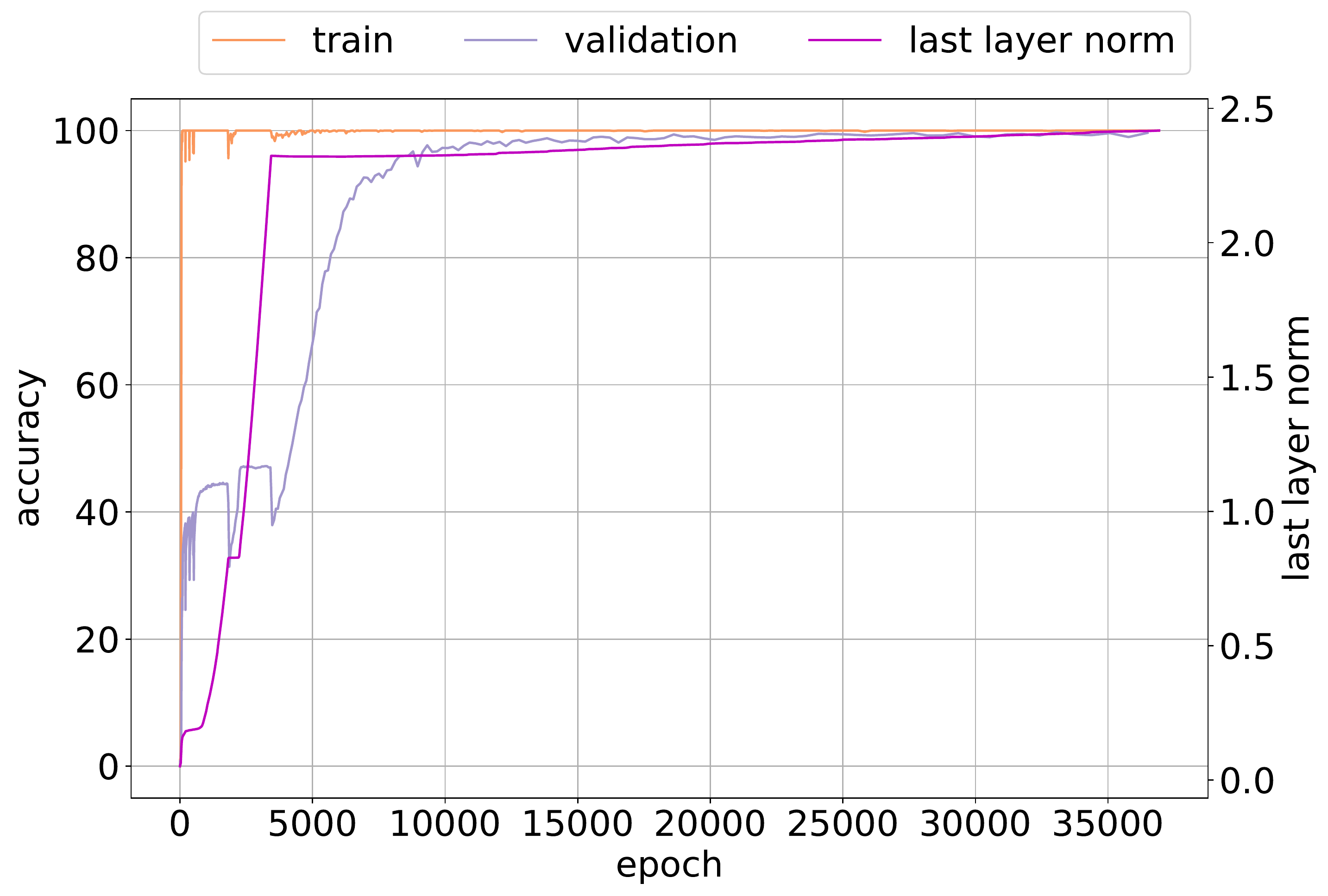} & 
      \includegraphics[width=0.33\linewidth]{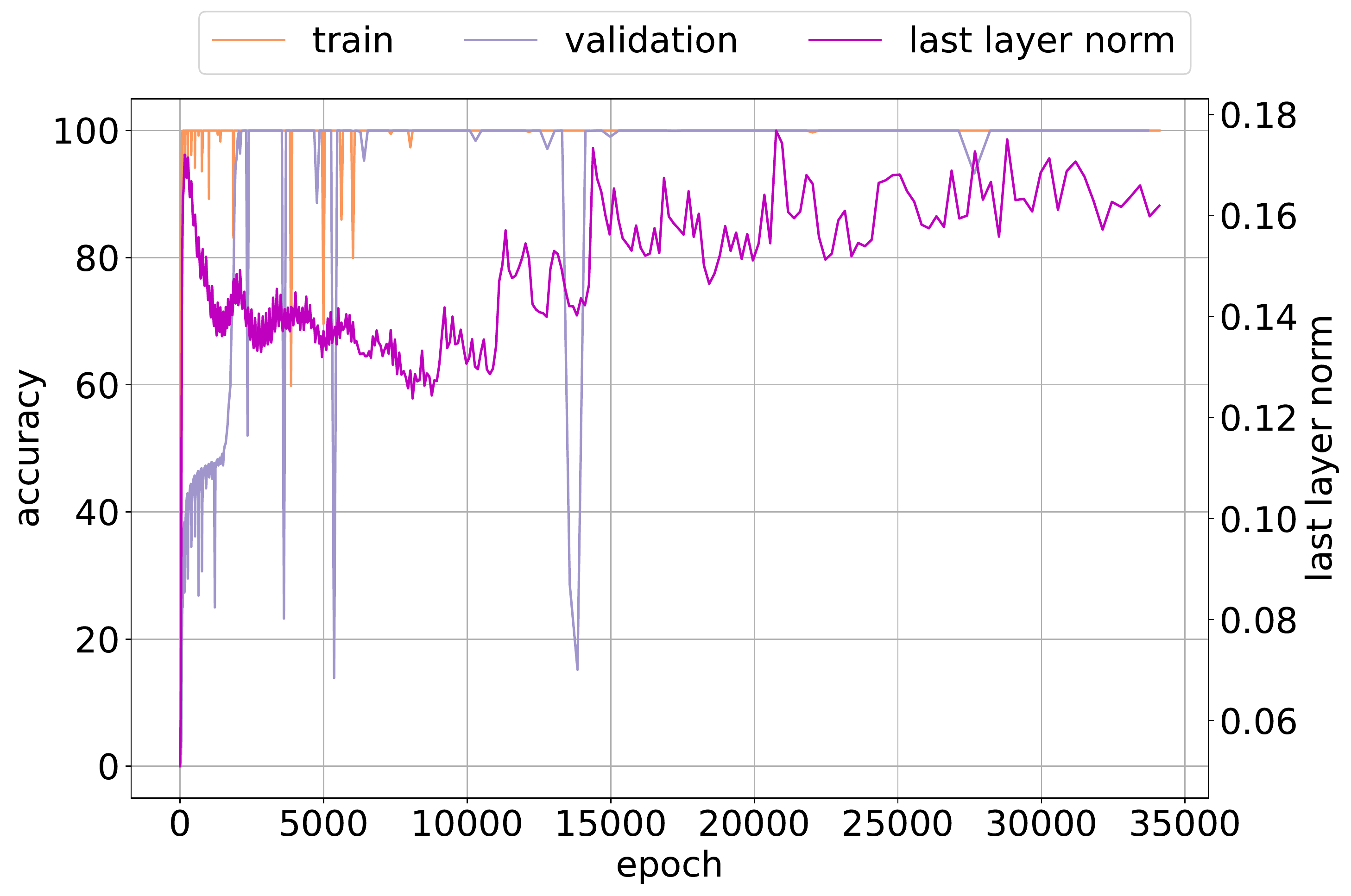} & 
      \includegraphics[width=0.33\linewidth]{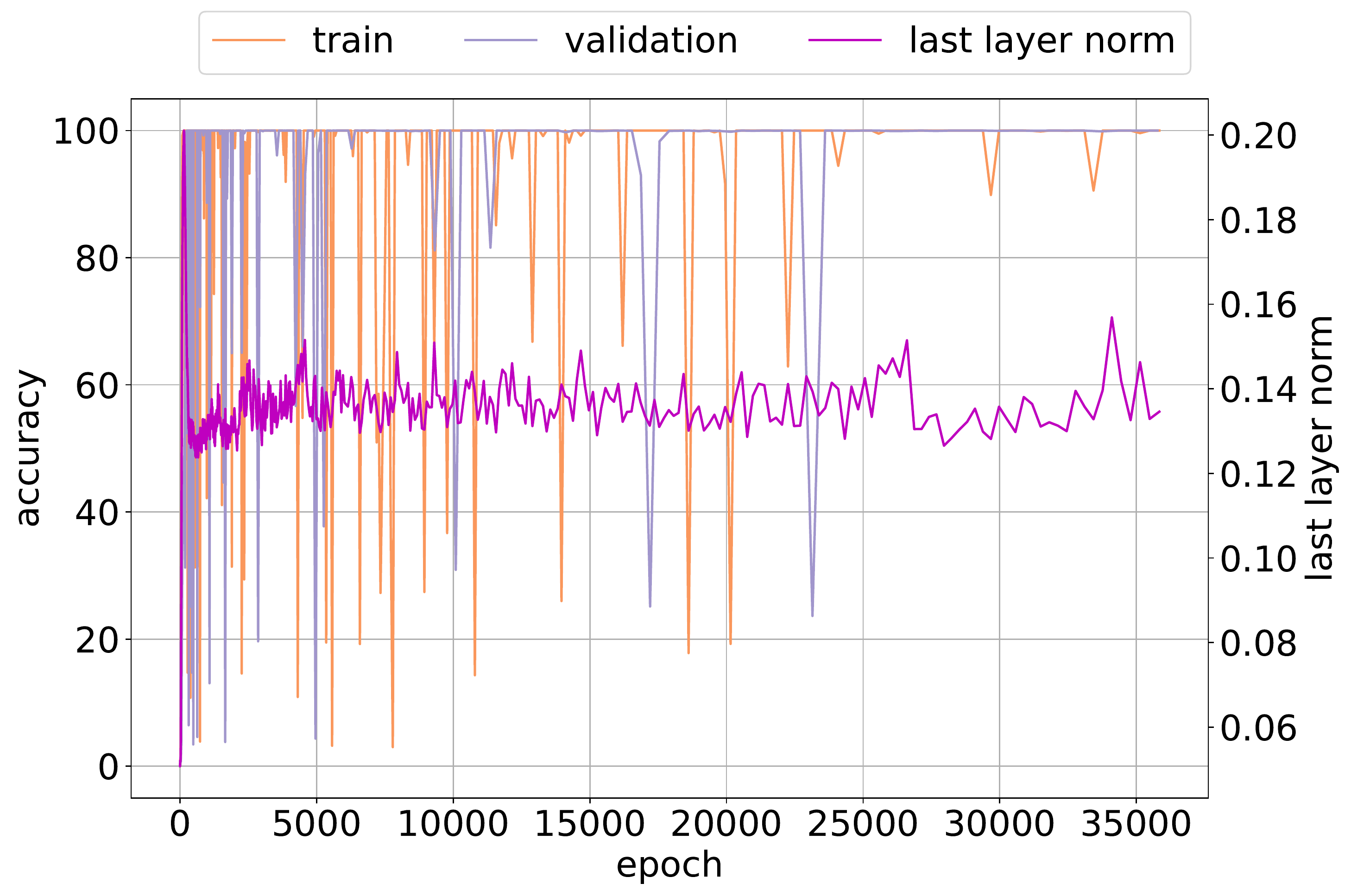} \\
      (d)  & (e) & (f) \\
      & train  and validation accuracy vs epochs \\
      & \\
  \end{tabular}
 \caption{Addition dataset: Norm behavior with different weight decay values. Training and validation loss vs epochs with weight decay (a) 0.0, (b) 0.1, (c) 1.0; Training and validation accuracy vs epochs shown in (d), (e) and (f). The evolution of classifier weight norm shows instability as increase in weight decay strength.} 
 \label{fig:grok_vary_wd_add}
\end{figure*}

\begin{figure*}[h]
\centering
  \begin{tabular}{ccc}
      weight decay = 0  & weight decay = 0.1 & weight decay = 1.0 \\
      \includegraphics[width=0.33\linewidth]{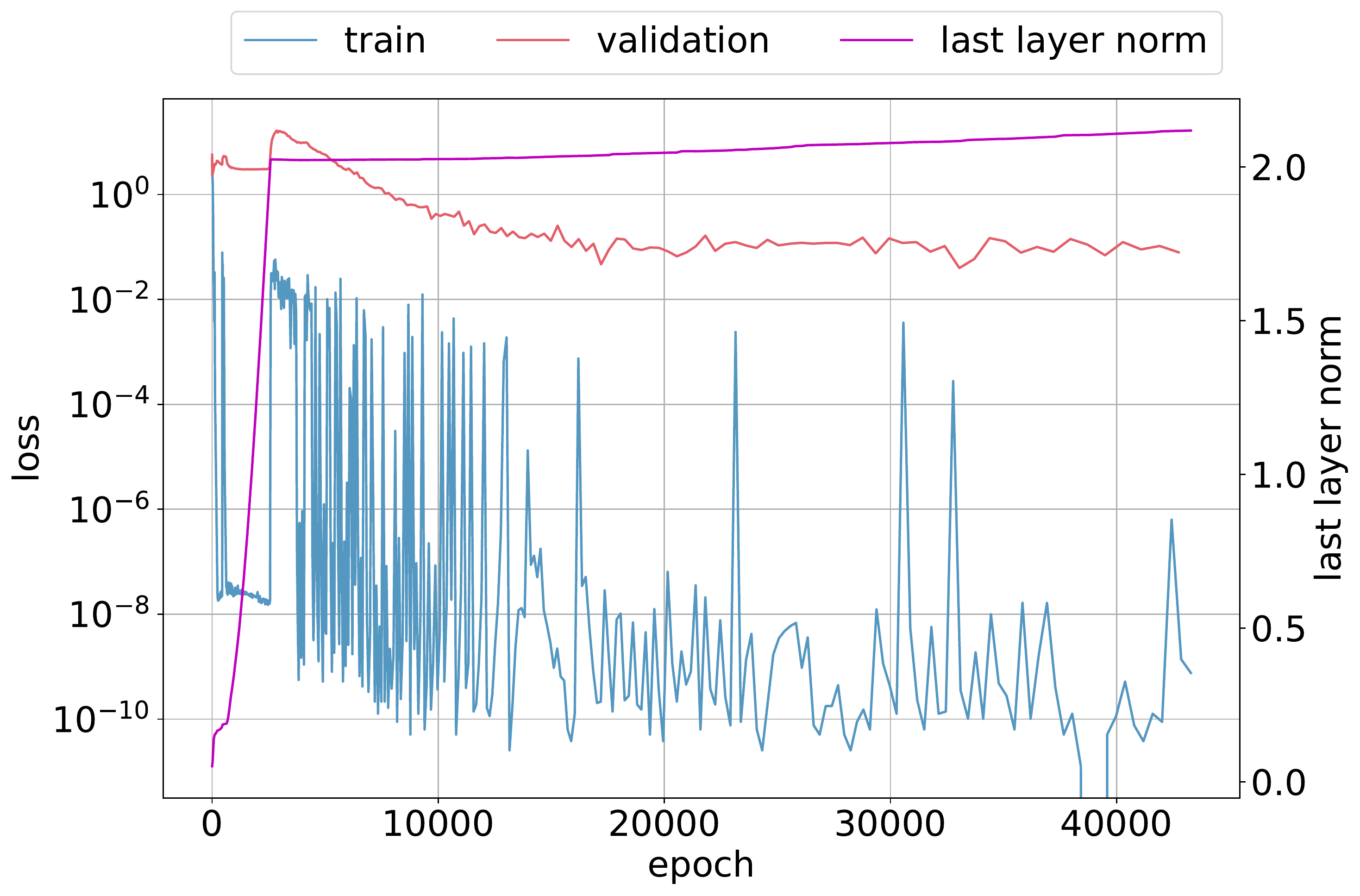} & 
      \includegraphics[width=0.33\linewidth]{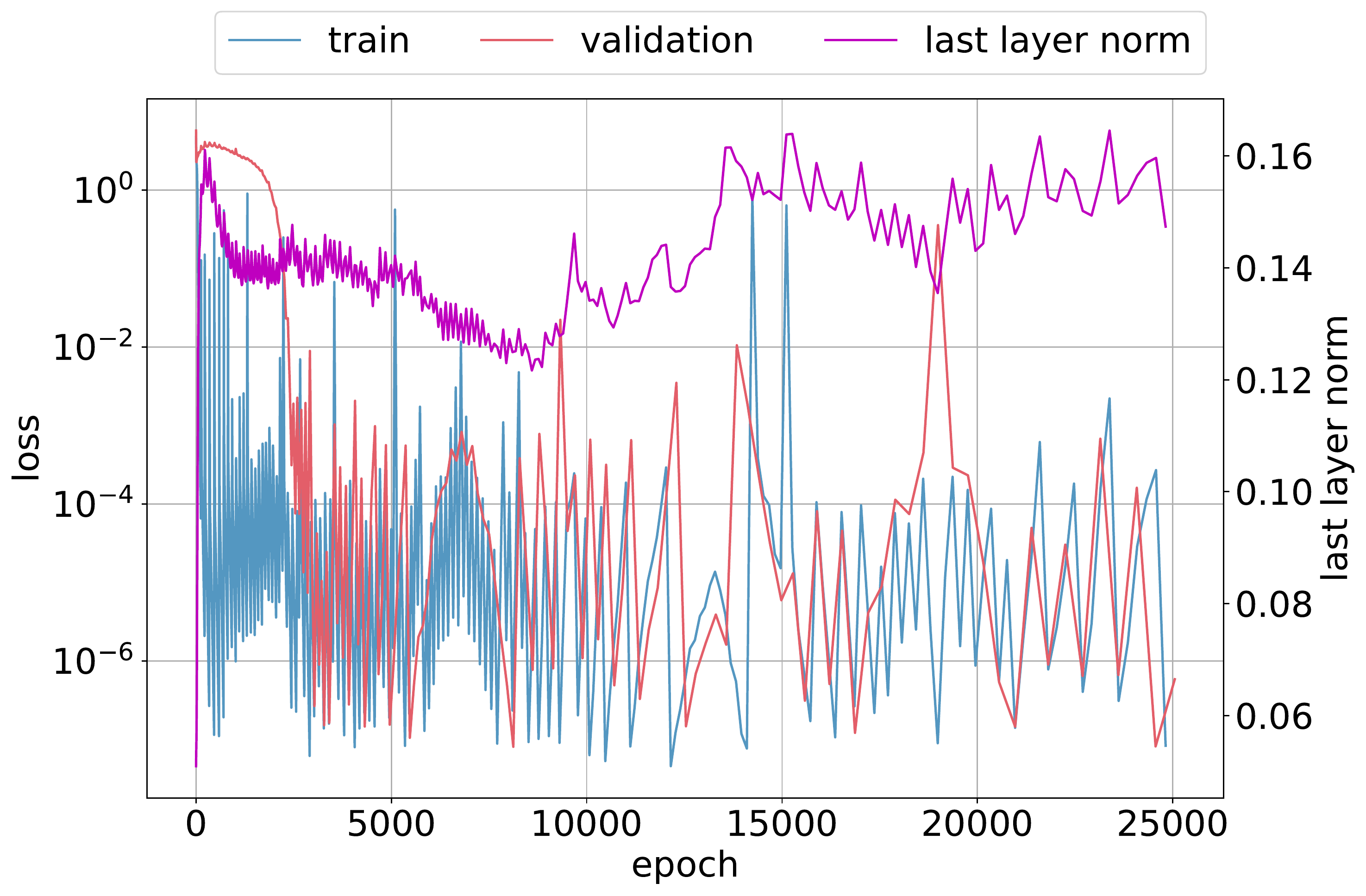} & 
      \includegraphics[width=0.33\linewidth]{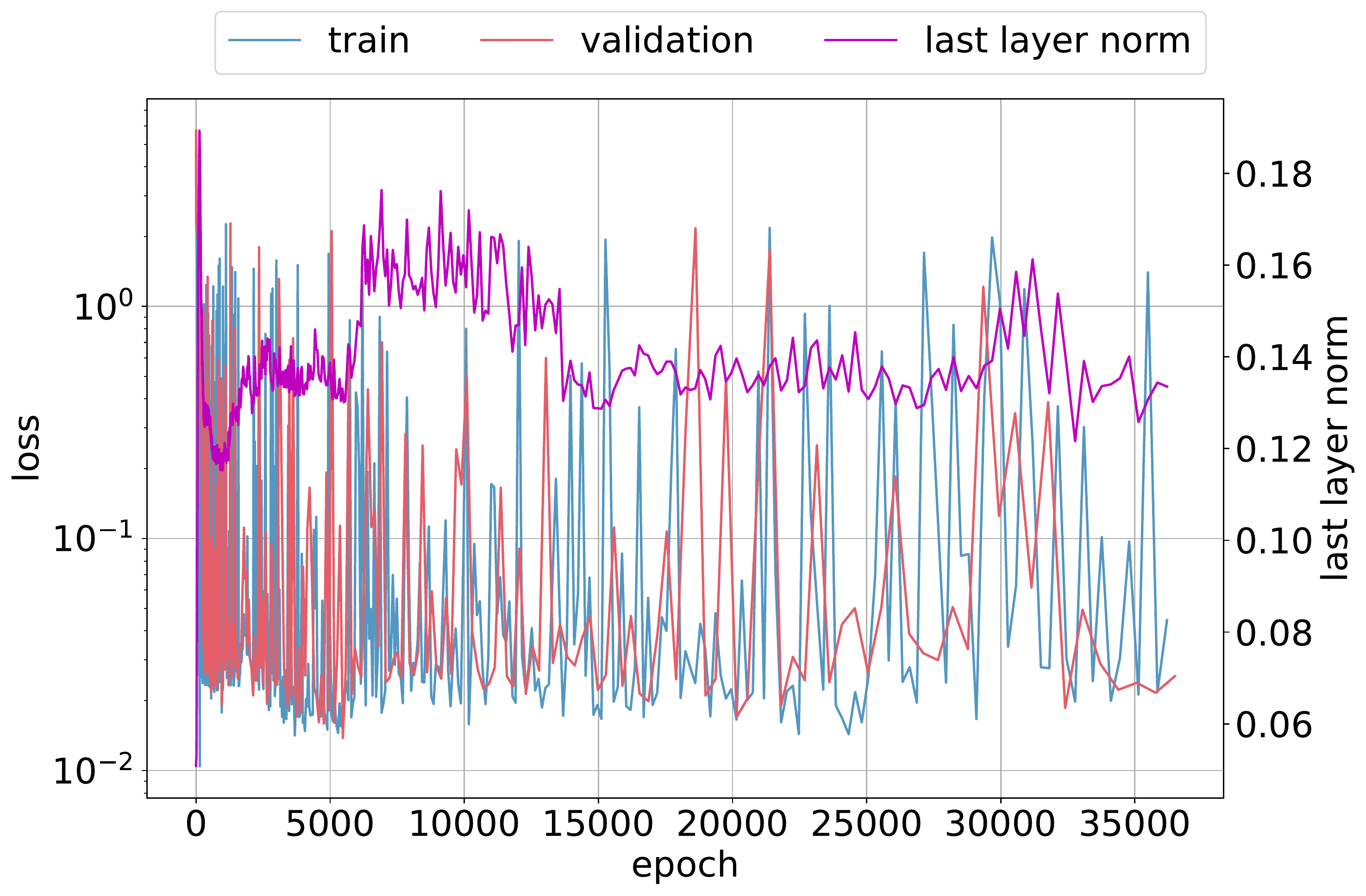} \\
      (a)  & (b) & (c) \\
      & train  and validation loss vs epochs \\
      & \\
      \includegraphics[width=0.33\linewidth]{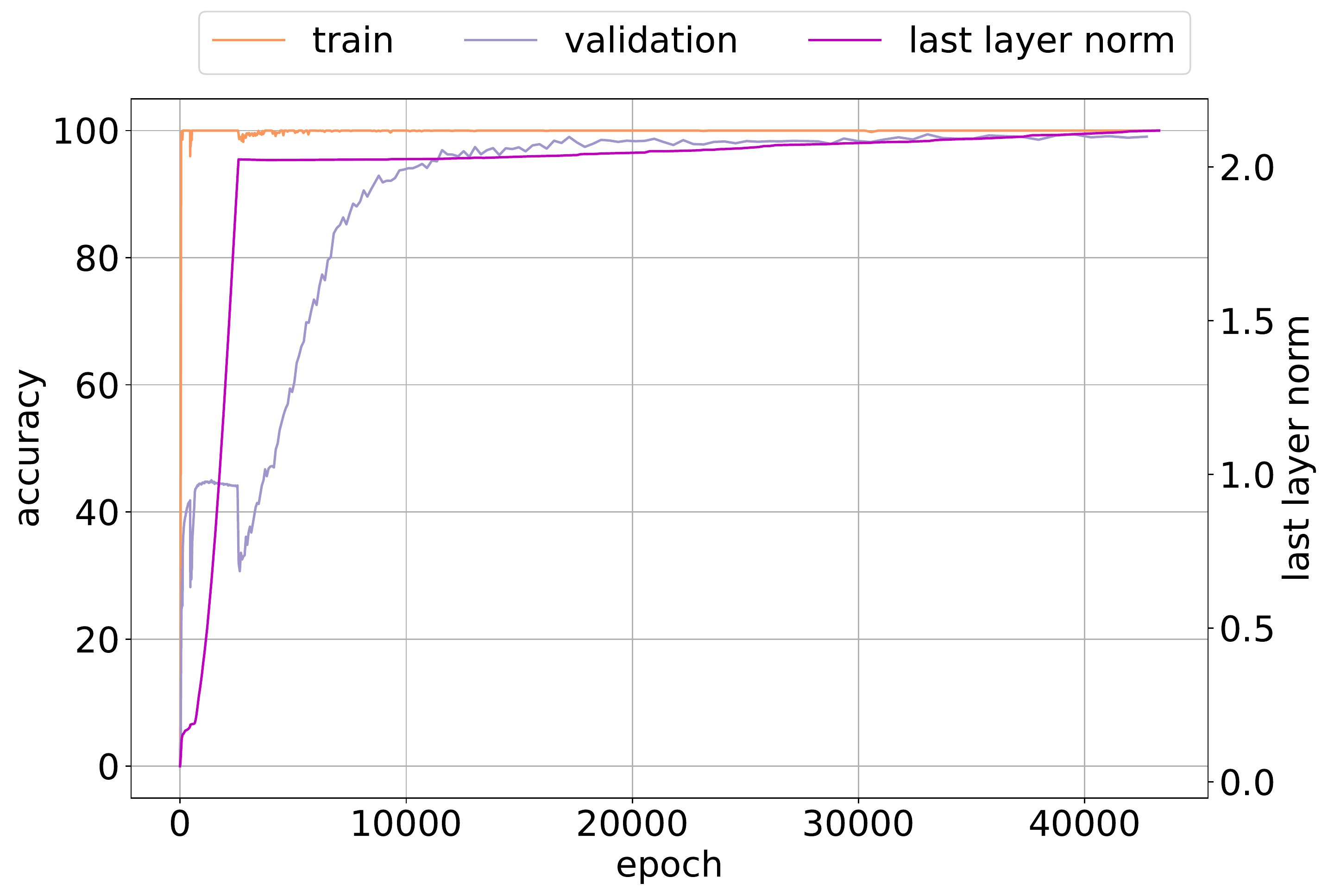} & 
      \includegraphics[width=0.33\linewidth]{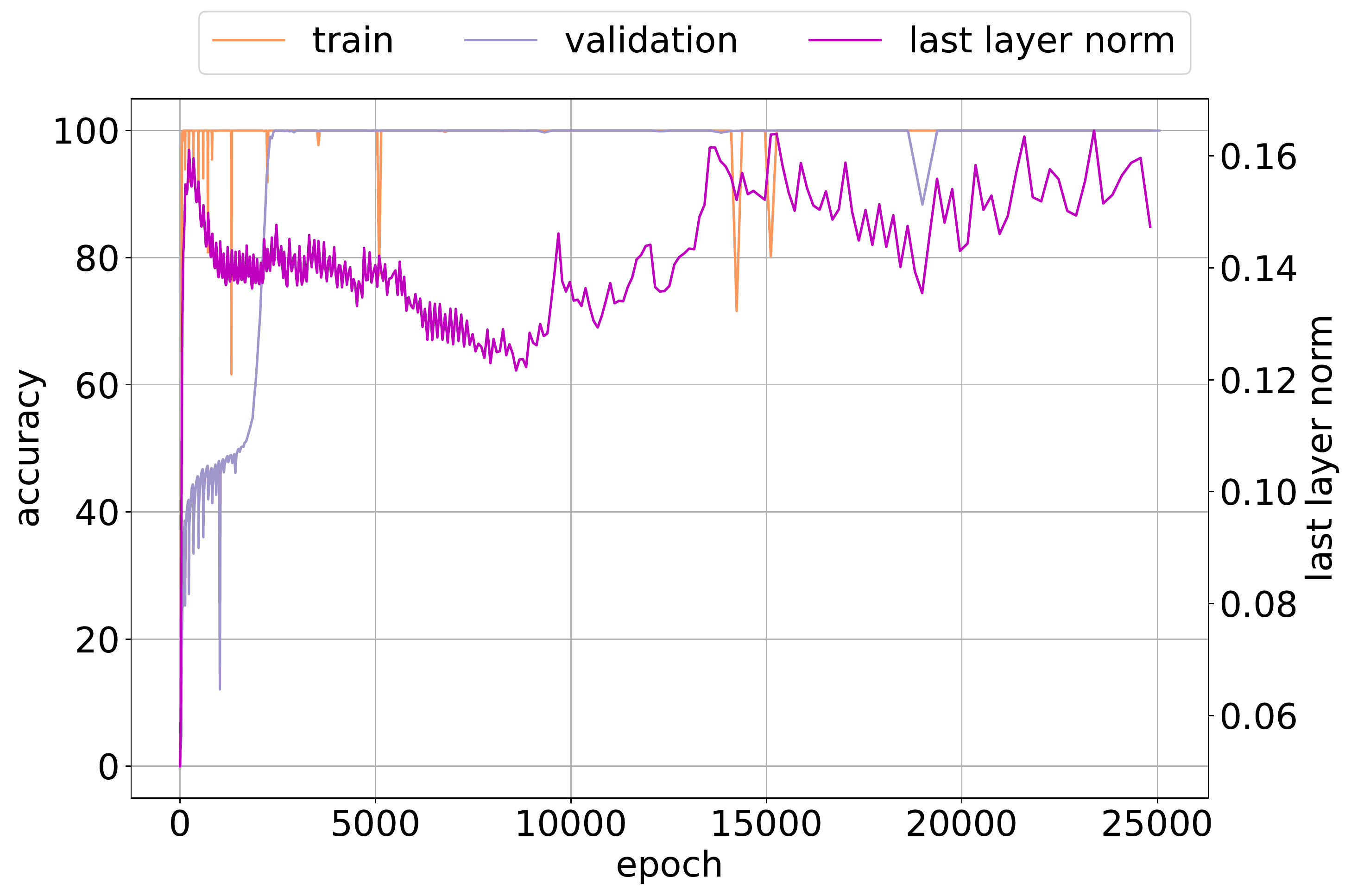} & 
      \includegraphics[width=0.33\linewidth]{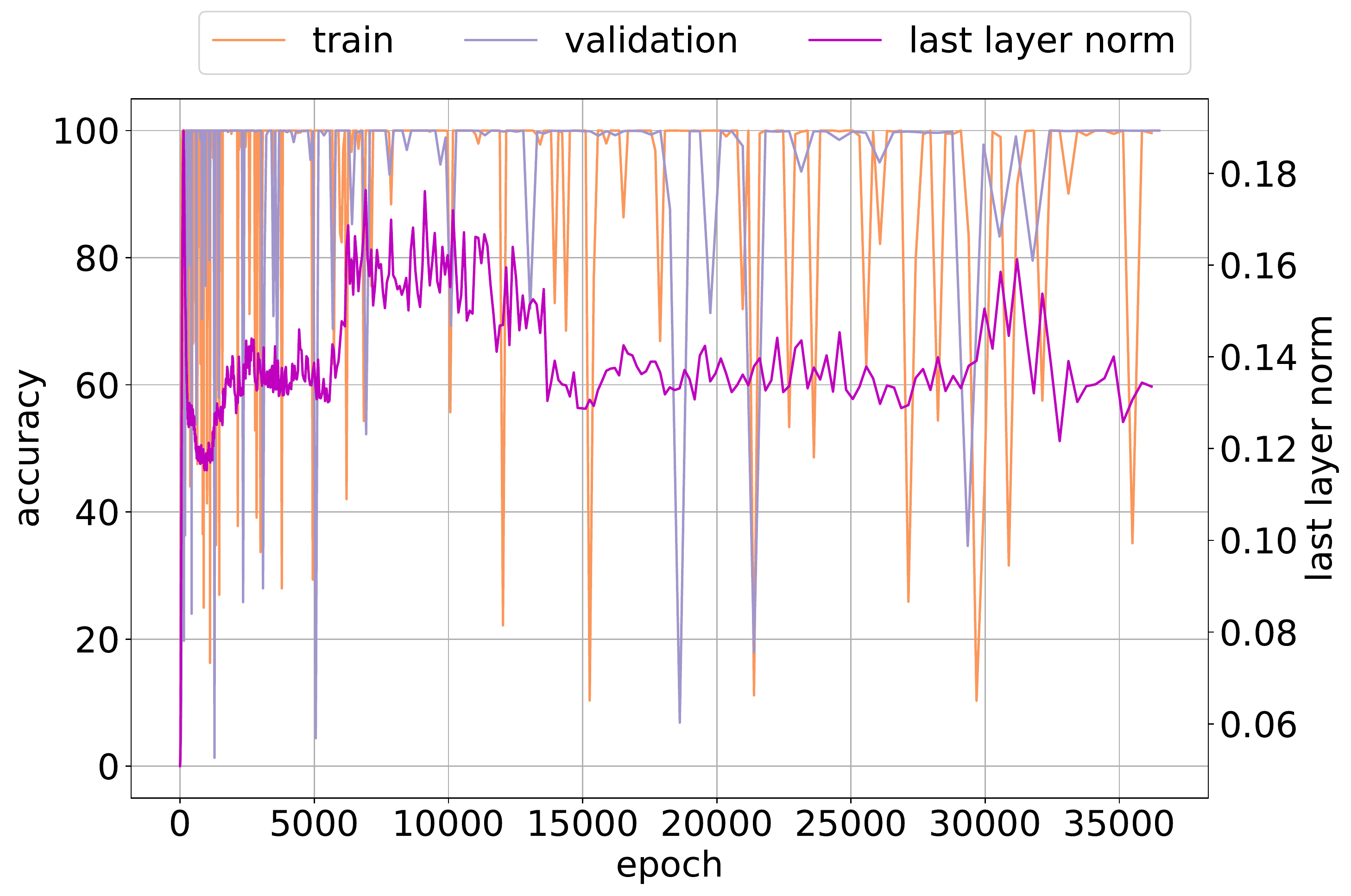} \\
      (d)  & (e) & (f) \\
      & train  and validation accuracy vs epochs \\
      & \\
  \end{tabular}
 \caption{Multiplication dataset: Norm behavior with different weight decay values. Training and validation loss vs epochs with weight decay (a) 0.0, (b) 0.1, (c) 1.0; Training and validation accuracy vs epochs shown in (d), (e) and (f). The evolution of classifier weight norm shows instability as increase in weight decay strength.} 
 \label{fig:grok_vary_wd_mul}
\end{figure*}

Weight decay is a commonly used regularization approach to improve the generalization performance of neural networks. Power et al.~\cite{power2021grokking} show that weight decay has the largest positive effect on alleviating Grokking. Weight decay naturally controls the size of the parameters and consequently their norm growth. We study the effect of weight decay on stability of training Transformers with Grokking datasets in this section. We use weight decay values from $0, 0.1, 0.2, 0.4, 0.6, 0.8 and 1.0$ with AdamW~\cite{loshchilov2017decoupled} optimizer. Figure~\ref{fig:grok_vary_wd_div} shows the results for division dataset. We observe from Figure~\ref{fig:grok_vary_wd_div} that as weight decay strength increases, both Slingshot Effects and Grokking phenomenon disappear with the model reaching high validation accuracy quickly as seen in Figure~\ref{fig:grok_vary_wd_div}e and Figure~\ref{fig:grok_vary_wd_div}f. However, we observe that the model experiences instability as can been seen with the loss plots in Figure~\ref{fig:grok_vary_wd_div}b and Figure~\ref{fig:grok_vary_wd_div}c or the accuracy plots in Figure~\ref{fig:grok_vary_wd_div}e and Figure~\ref{fig:grok_vary_wd_div}f. A similar trend is observed for addition and multiplication datasets in  Figure~\ref{fig:grok_vary_wd_add} and Figure~\ref{fig:grok_vary_wd_mul} respectively. 

The results shown above indicate that Slingshot may not be the only way to achieve good generalization. Both Slingshot and weight decay prevent the norms from growing unbounded and achieve high validation accuracy as seen in plots described above. While weight decay shows different weight norm dynamics, this regularization does not decrease training instability. These results suggest the need for alternative approaches to improve training stability.

\clearpage
\subsection{Features and parameter normalization}

A second approach that we use to explicitly control weights and feature norm is by normalizing the features and weights via the following scheme:
$
    w = \frac{w}{\lVert  w  \rVert},
    f(x) = \frac{f(x)}{\lVert  f(x)  \rVert},
$
where $w$ and $f(x)$ are the weights and inputs to the classification layer respectively, the norm used above is the $L_{2}$ norm, and $x$ is the input to the neural network. We take the cosine similarity of the normalized weights and features and divide this value by a temperature value that we treat as a hyperparameter in these experiments. The operation is given by:
$
  y = \frac{w \cdot f(x)}{\tau}
$
where $\tau$ represents the temperature hyperparameter. We use temperature values from $0.1, 0.25, 0.5, 0.75, 1.0$ for these experiments.

Figure~\ref{fig:grok_norm_both_div} shows the results of Transformer training on division dataset described in Appendix~\ref{appendix:xformers_setup} that is split evenly into train and validation sets. We observe that the model displays training instability evidenced by norm behavior and also loss behavior in Figure~\ref{fig:grok_norm_both_div}a at lower temperature values. We observe that $\tau=0.25$ provides a good compromise between fitting training data while showing no training instability as seen in  Figure~\ref{fig:grok_norm_both_div}b. This hyperparameter value also results in Grokking as validation accuracy improves late in training as can be seen from  Figure~\ref{fig:grok_norm_both_div}e. These together suggest that bounding weights and features norm helps stabilize training without sacrificing training performance.

\begin{figure*}[h!]
\centering
  \begin{tabular}{ccc}
      temperature = 0.1  & temperature = 0.25 & temperature = 1.0 \\
      \includegraphics[width=0.33\linewidth]{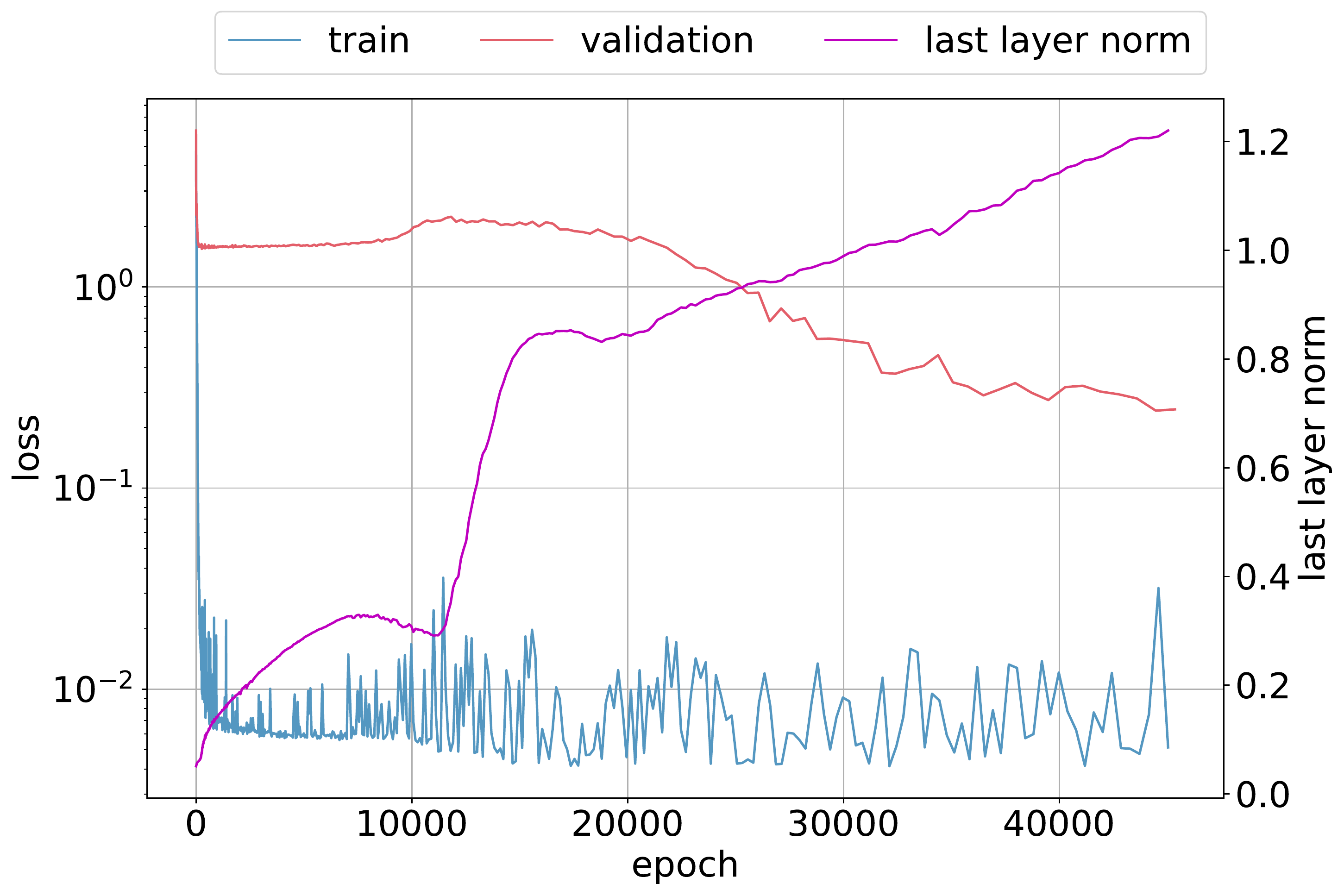} &
      \includegraphics[width=0.33\linewidth]{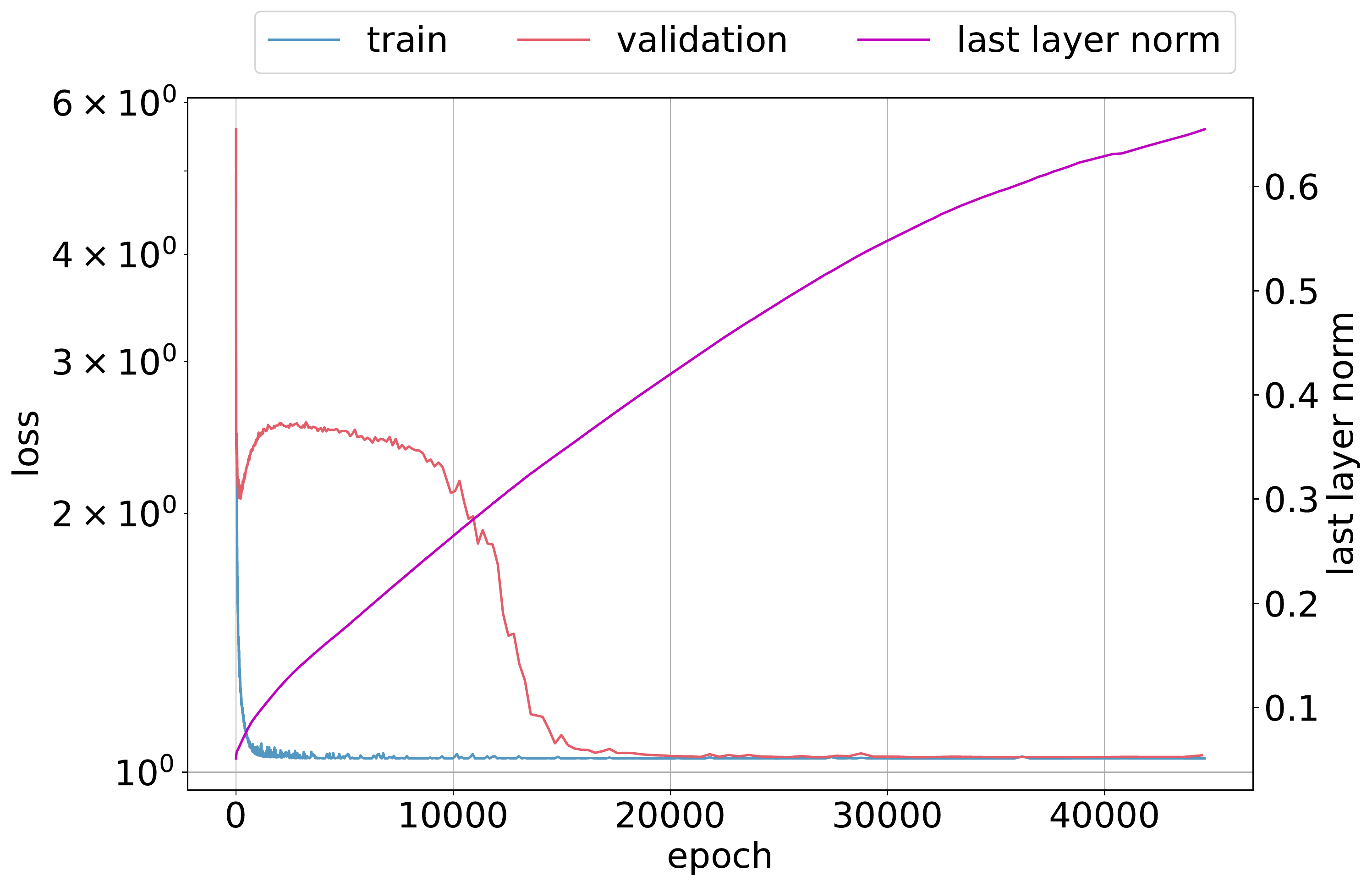} & 
      \includegraphics[width=0.33\linewidth]{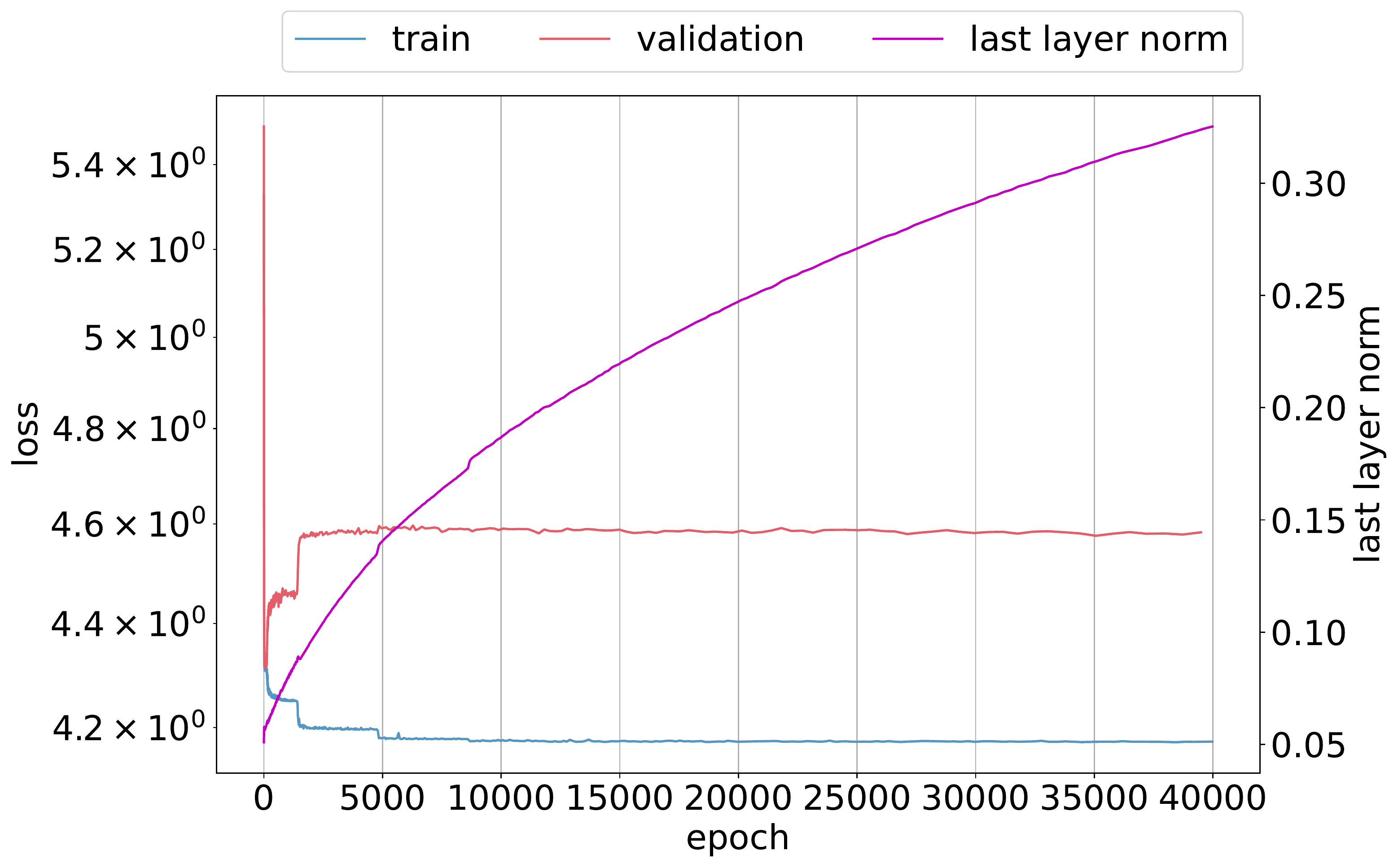}  \\
      (a)  & (b) & (c) \\
      & train  and validation loss vs epochs \\
      & \\
      \includegraphics[width=0.33\linewidth]{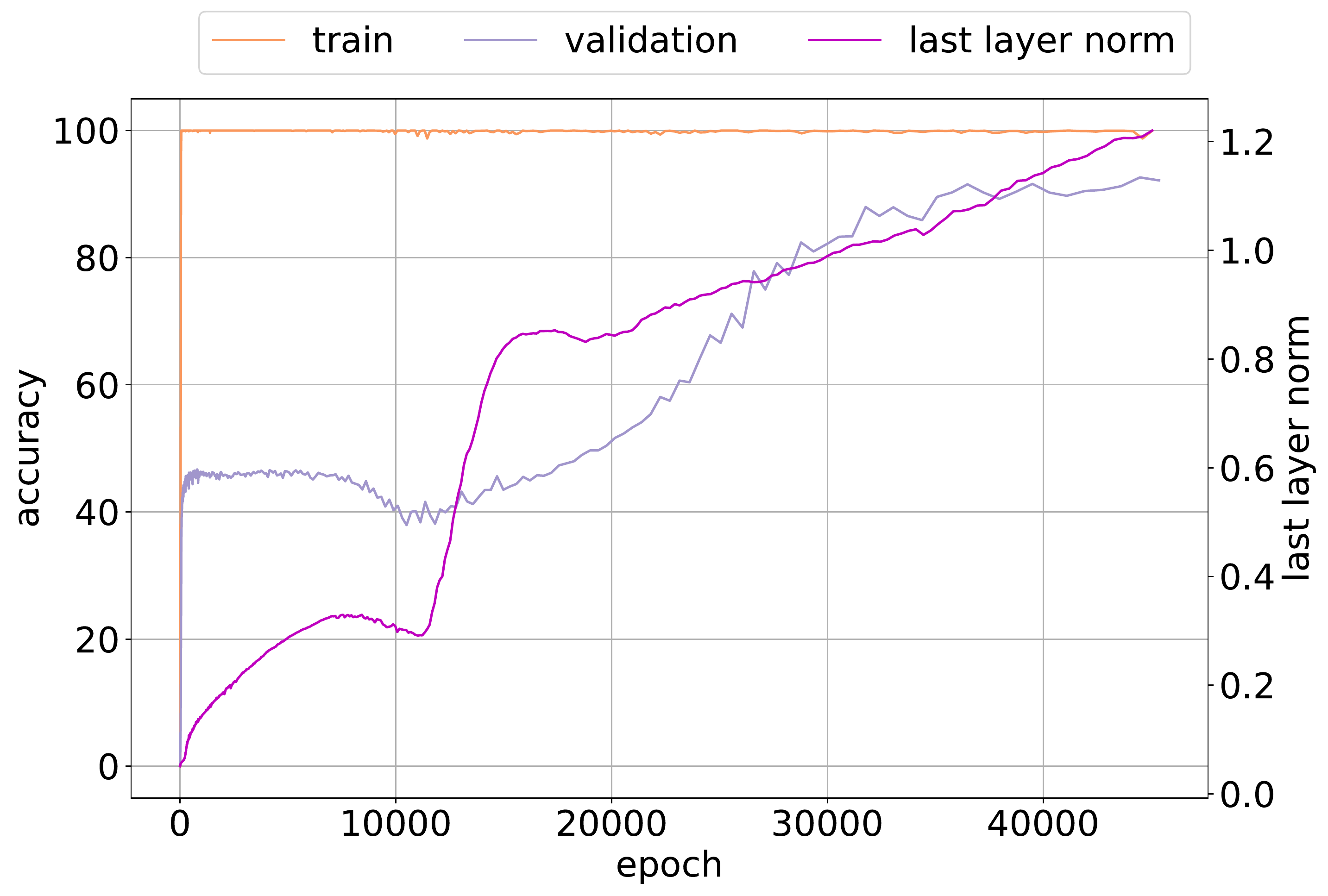} &
      \includegraphics[width=0.33\linewidth]{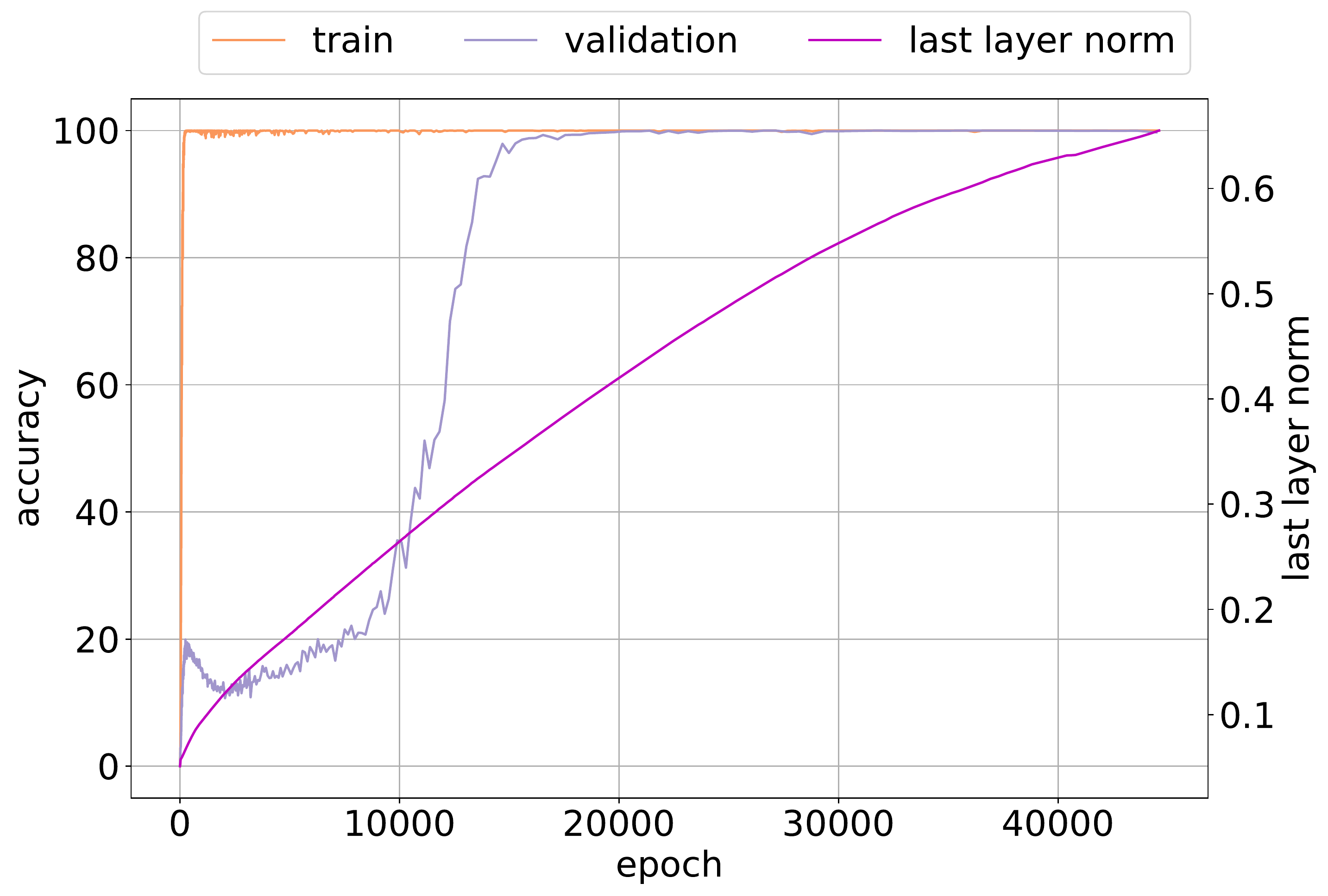} & 
      \includegraphics[width=0.33\linewidth]{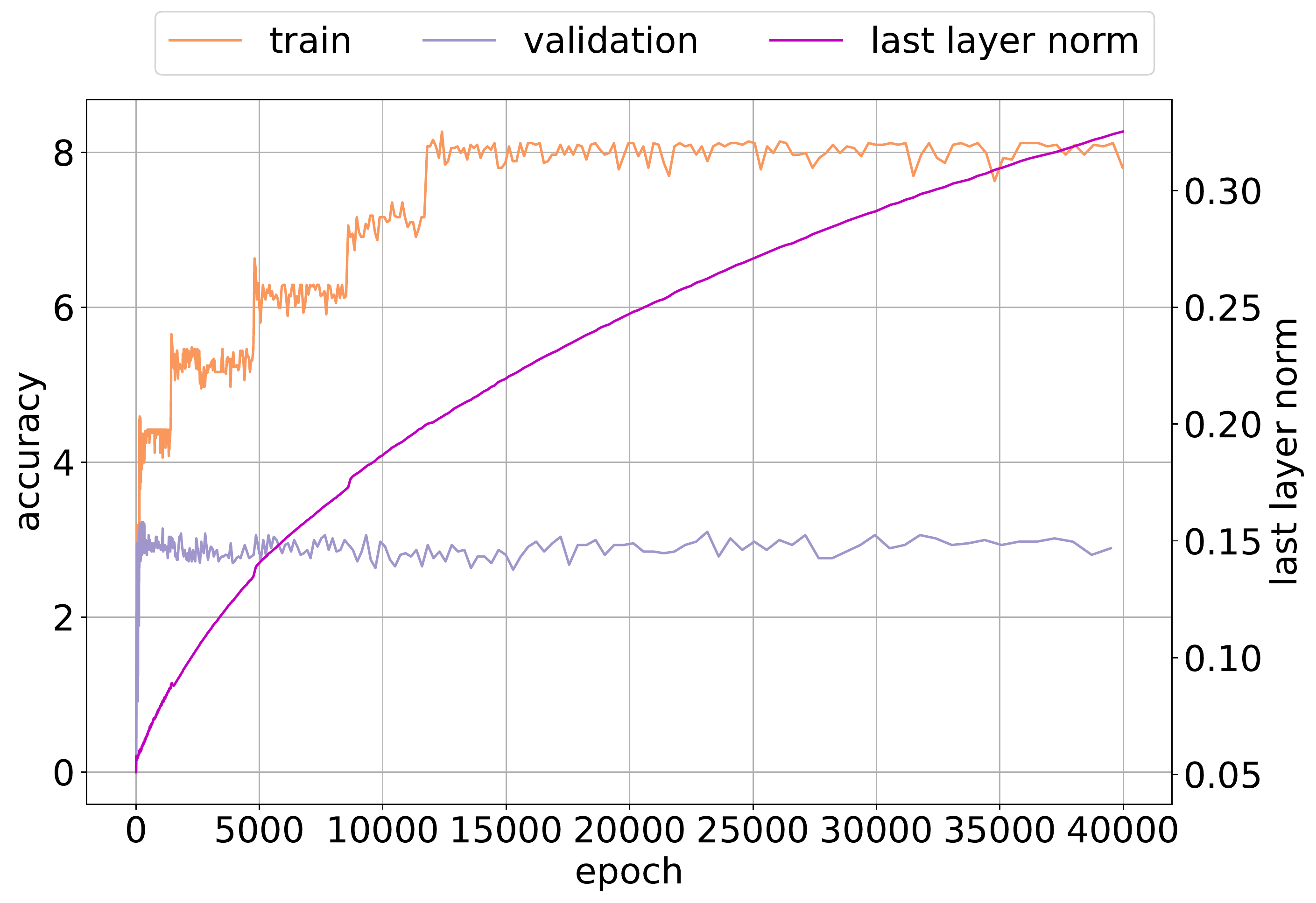}  \\
      (d)  & (e) & (f) \\
      & train  and validation accuracy vs epochs \\
      & \\
  \end{tabular}
 \caption{Multiplication dataset: Features and parameters normalization. Observe that a smaller temperature allows the model to fit the data better but experiences training instability. Temperature = 0.25 allows the model to fit and achieve high validation accuracy without suffering training instability.} 
 \label{fig:grok_norm_both_mul}
\end{figure*}

We validate the normalization scheme with two additional datasets namely multiplication and division from Appendix~\ref{appendix:xformers_setup}. Figure~\ref{fig:grok_norm_both_mul} shows the results for training Transformers with multiplication dataset that is split evenly into train and validation sets. We observe from Figure~\ref{fig:grok_norm_both_mul} that a proper temperature value can stabilize training and with some tuning can provide a compromise between training stability and generalization. Specifically, $\tau=0.25$ allows the model to fit the training data and reach almost perfect validation accuracy as seen from Figure~\ref{fig:grok_norm_both_mul}b and Figure~\ref{fig:grok_norm_both_mul}e.

\begin{figure*}[h!]
\centering
  \begin{tabular}{ccc}
      temperature = 0.1  & temperature = 0.25 & temperature = 1.0 \\
      \includegraphics[width=0.33\linewidth]{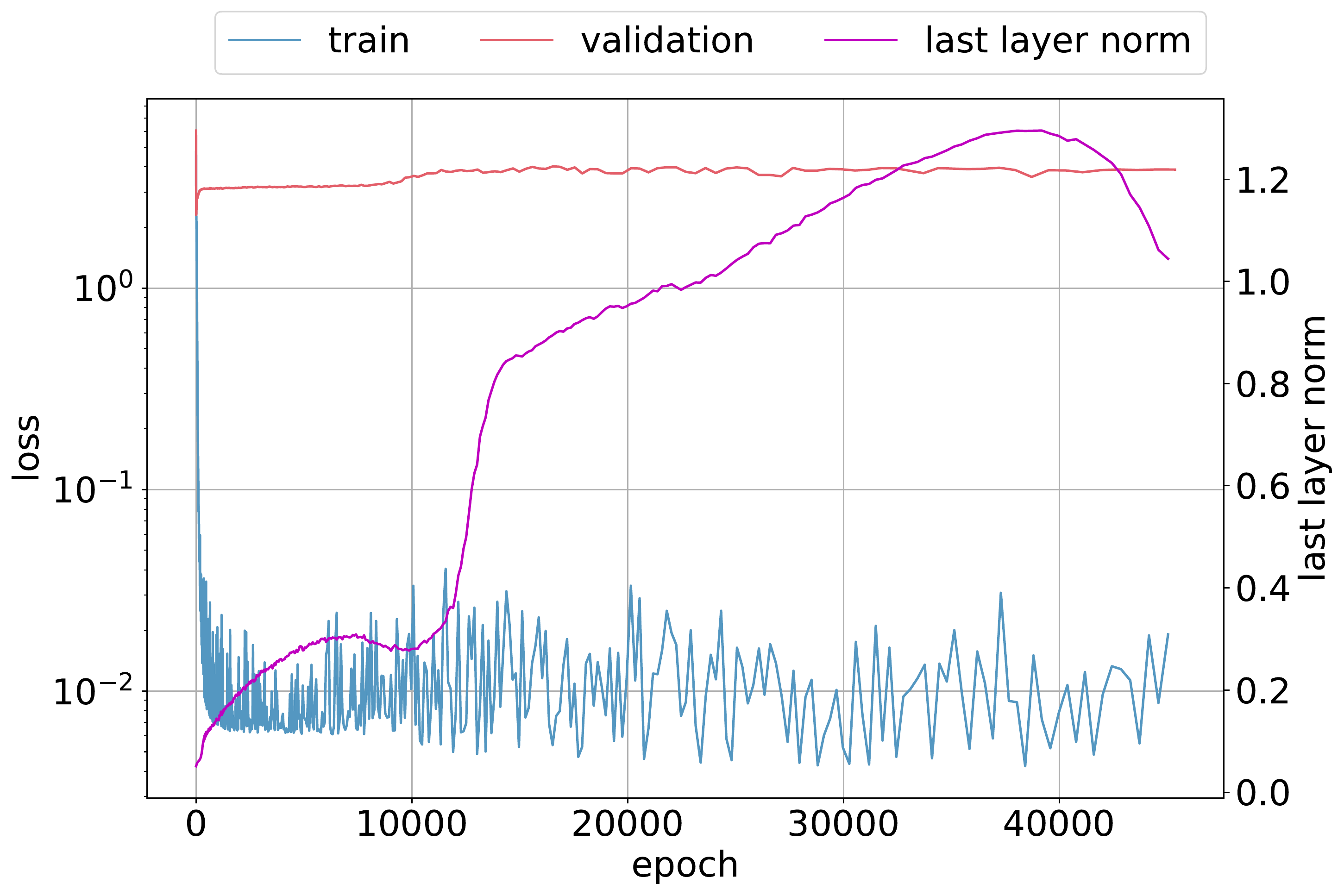} &
      \includegraphics[width=0.33\linewidth]{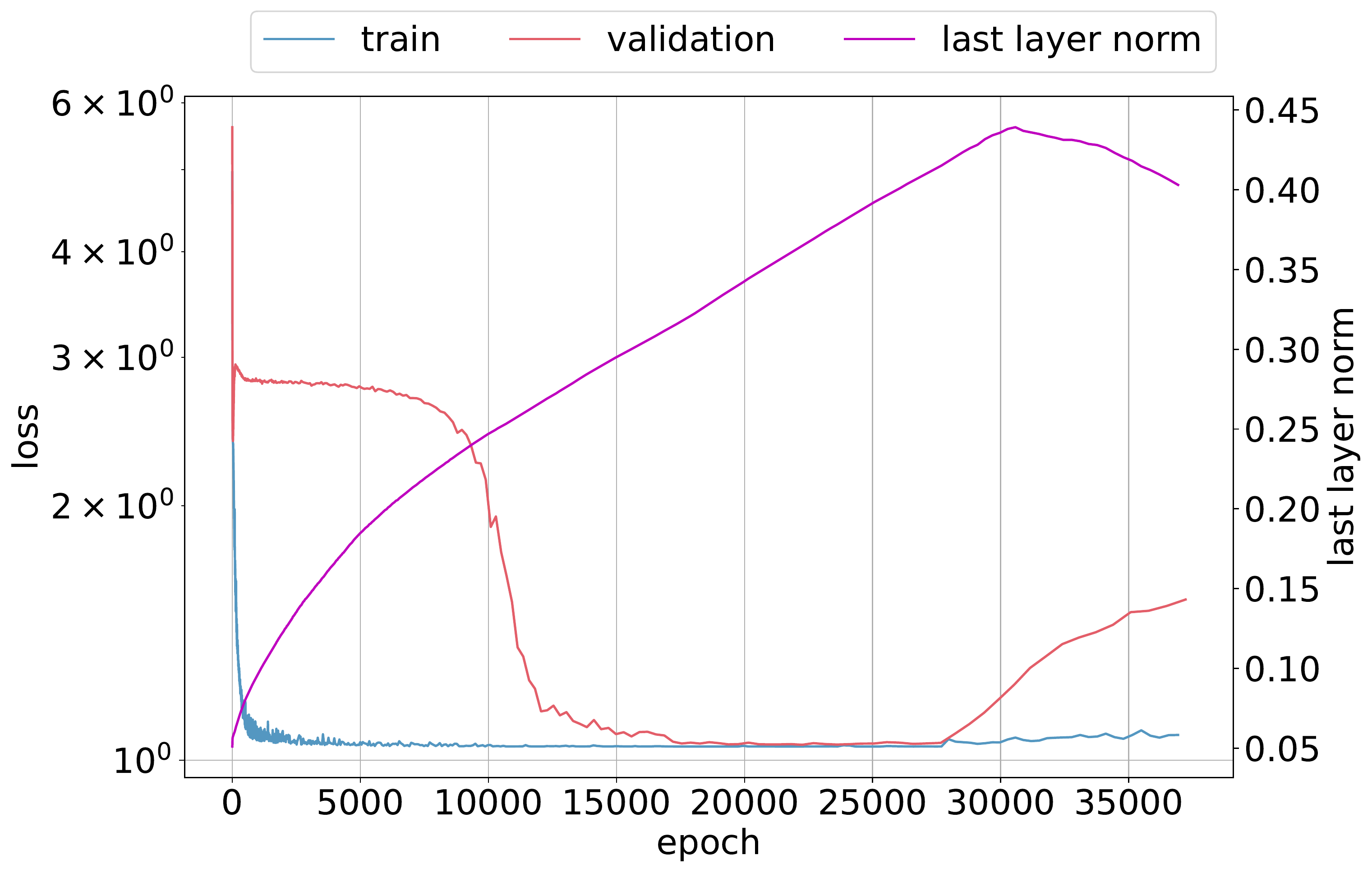} & 
      \includegraphics[width=0.33\linewidth]{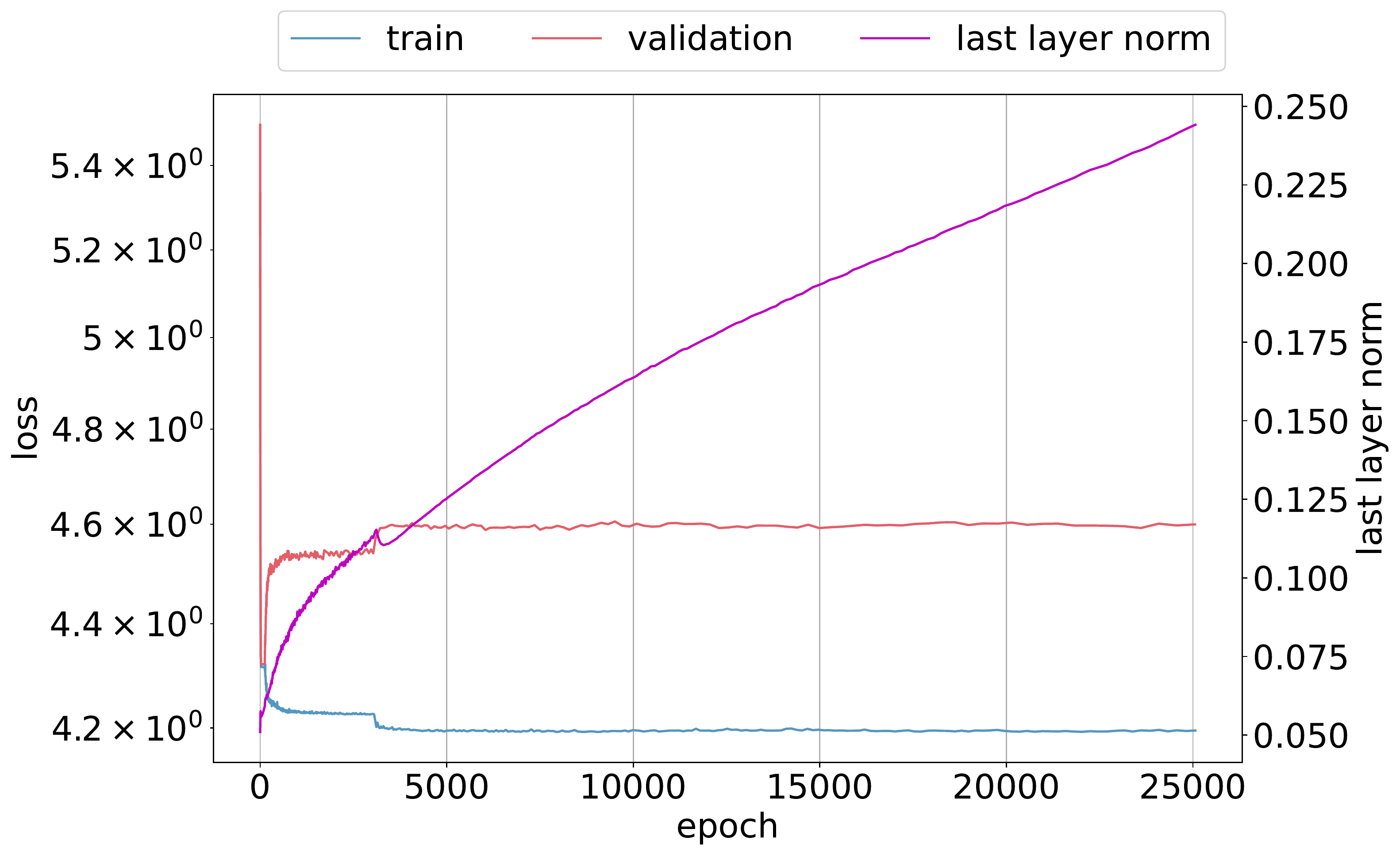}  \\
      (a)  & (b) & (c) \\
      & train  and validation loss vs epochs \\
      & \\
      \includegraphics[width=0.33\linewidth]{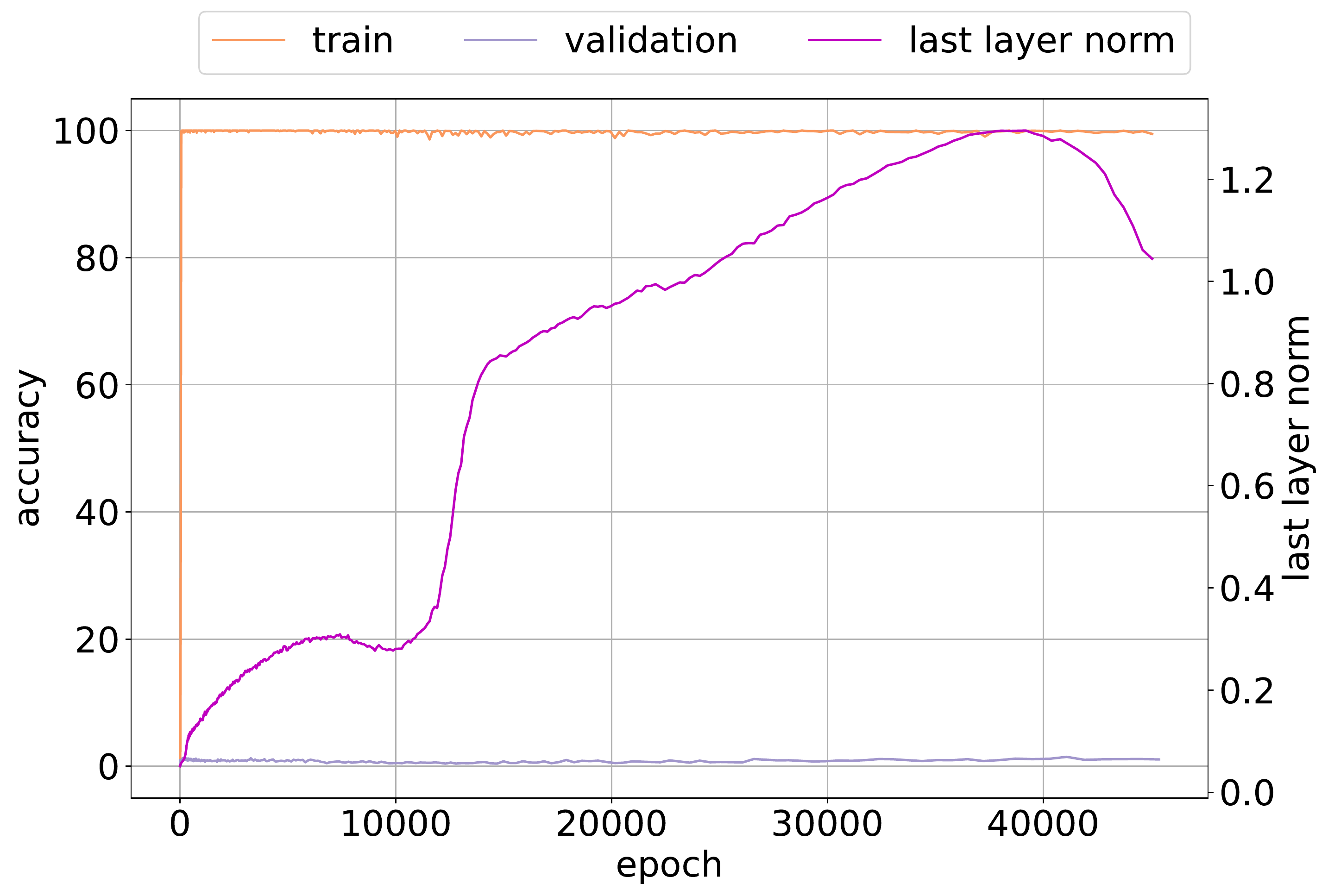} &
      \includegraphics[width=0.33\linewidth]{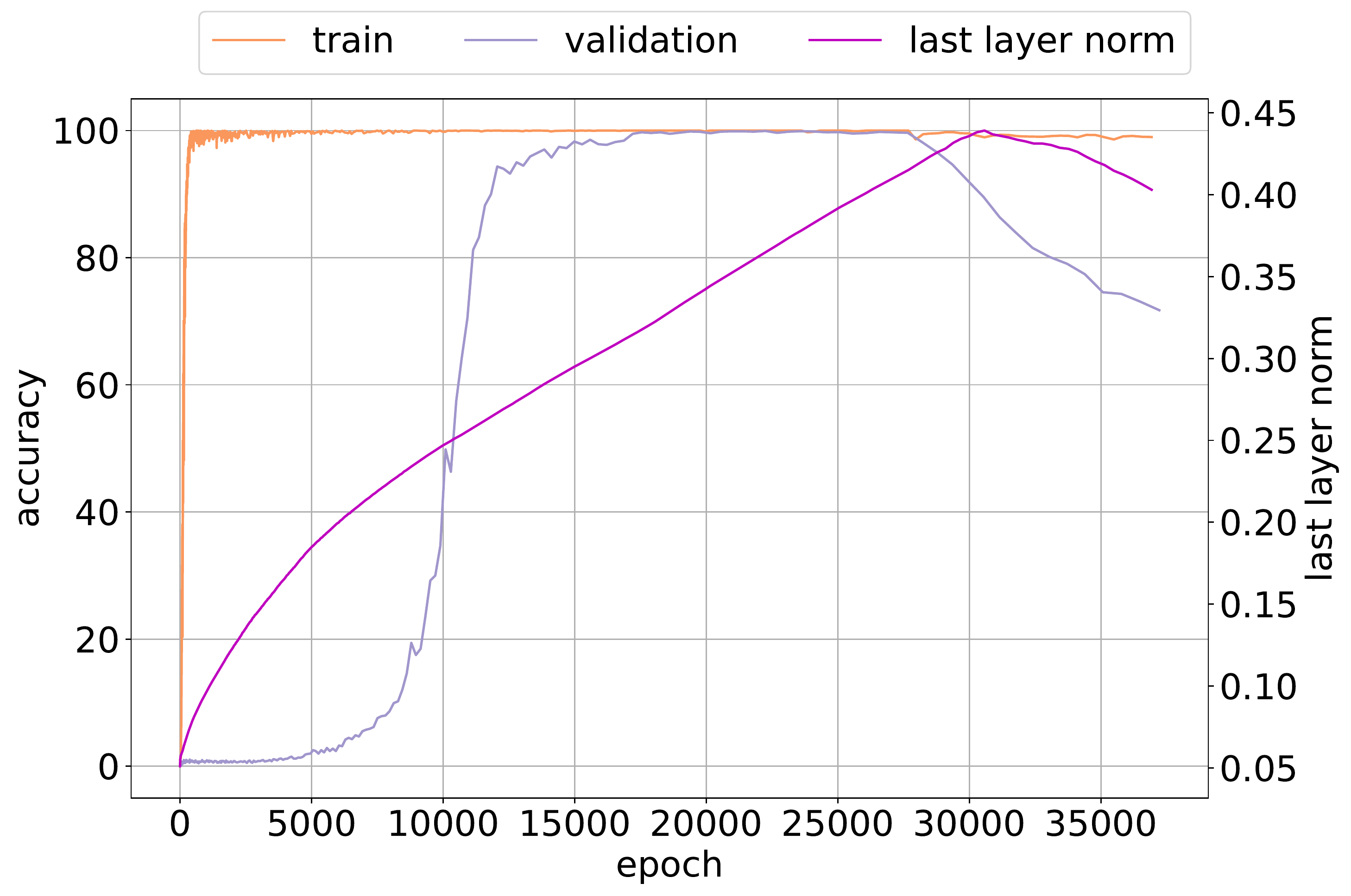} & 
      \includegraphics[width=0.33\linewidth]{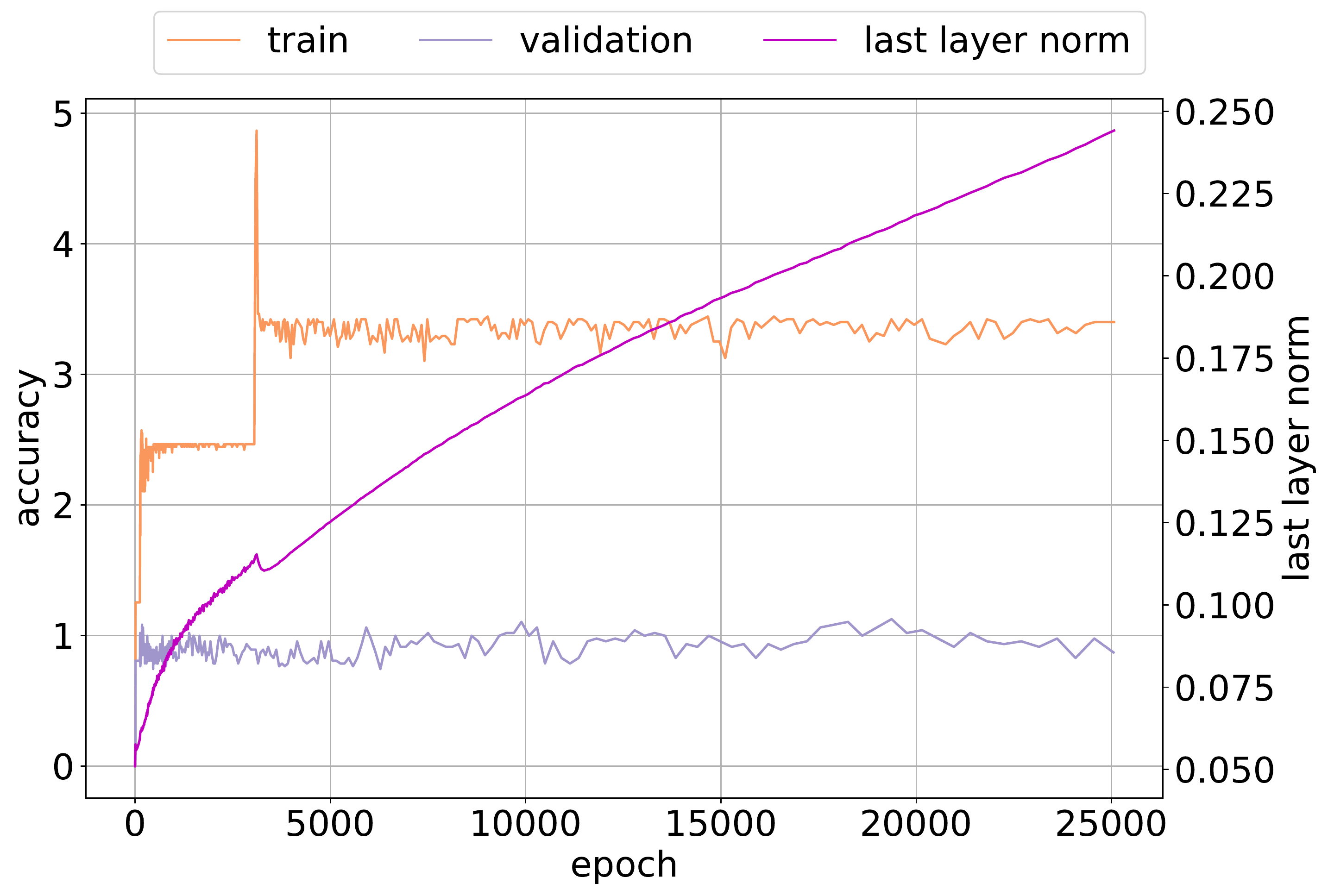}  \\
      (d)  & (e) & (f) \\
      & train  and validation accuracy vs epochs \\
      & \\
  \end{tabular}
 \caption{Subtraction dataset: Features and parameters normalization. Observe that a smaller temperature allows the model to fit the data better but experiences training instability. Temperature = 0.25 allows the model to fit and achieve high validation accuracy. However, we observe training instability as can seen with weight norm dynamics.} 
 \label{fig:grok_norm_both_sub}
\end{figure*}

Finally, we repeat the above experiments with subtraction dataset and show the results in Figure~\ref{fig:grok_norm_both_sub}. This dataset shows that while a properly tuned temperature can help the model achieve almost perfect generalization, training instability shows up very late in optimization. This observation can be seen from Figure~\ref{fig:grok_norm_both_sub}b and Figure~\ref{fig:grok_norm_both_sub}d. This result suggests that more work remains to be done with understanding and stabilizing the training behavior of large neural networks.

\begin{figure*}[h!]
\centering
  \begin{tabular}{ccc}
      temperature = 0.1  & temperature = 0.25 & temperature = 1.0 \\
      \includegraphics[width=0.33\linewidth]{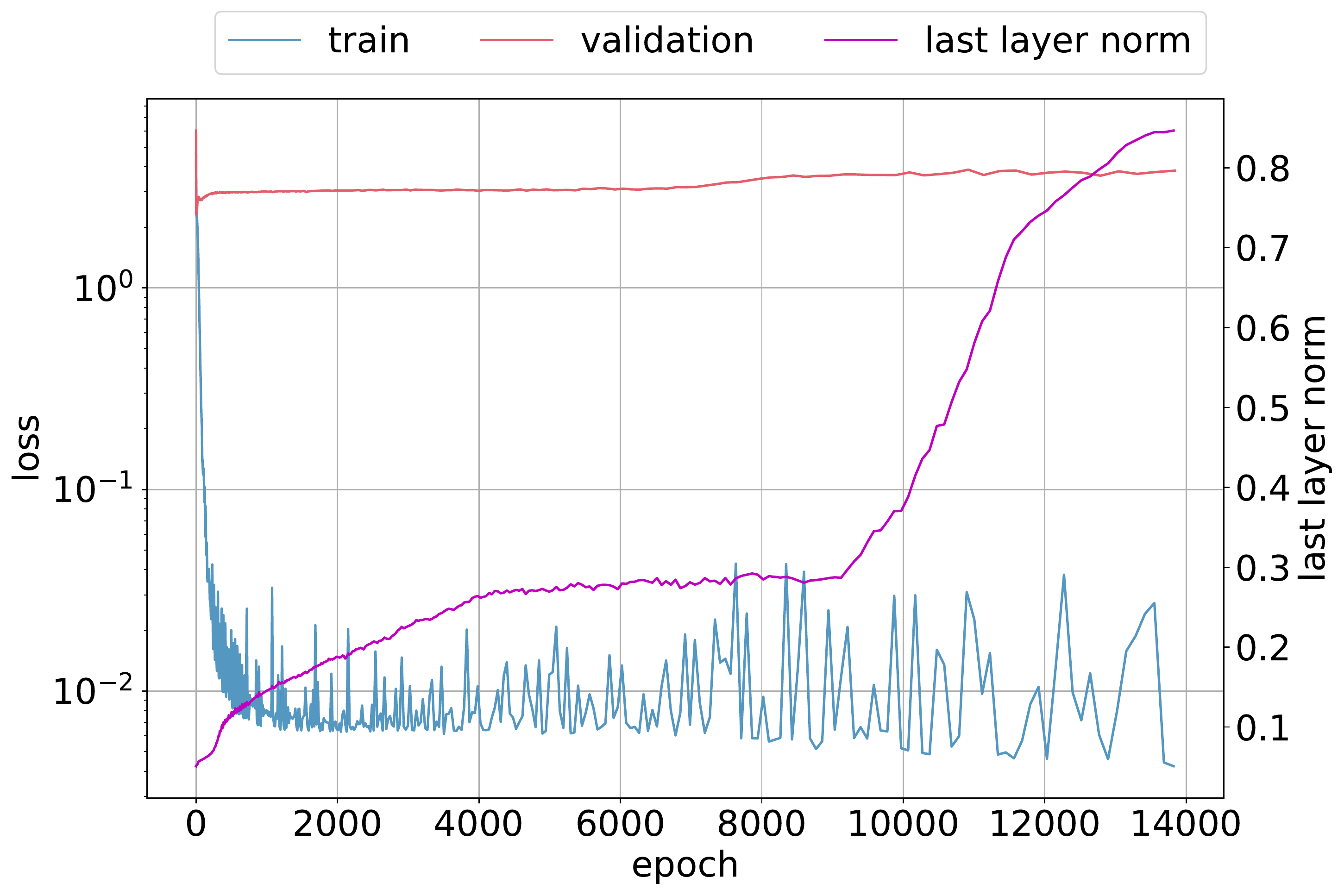} &
      \includegraphics[width=0.33\linewidth]{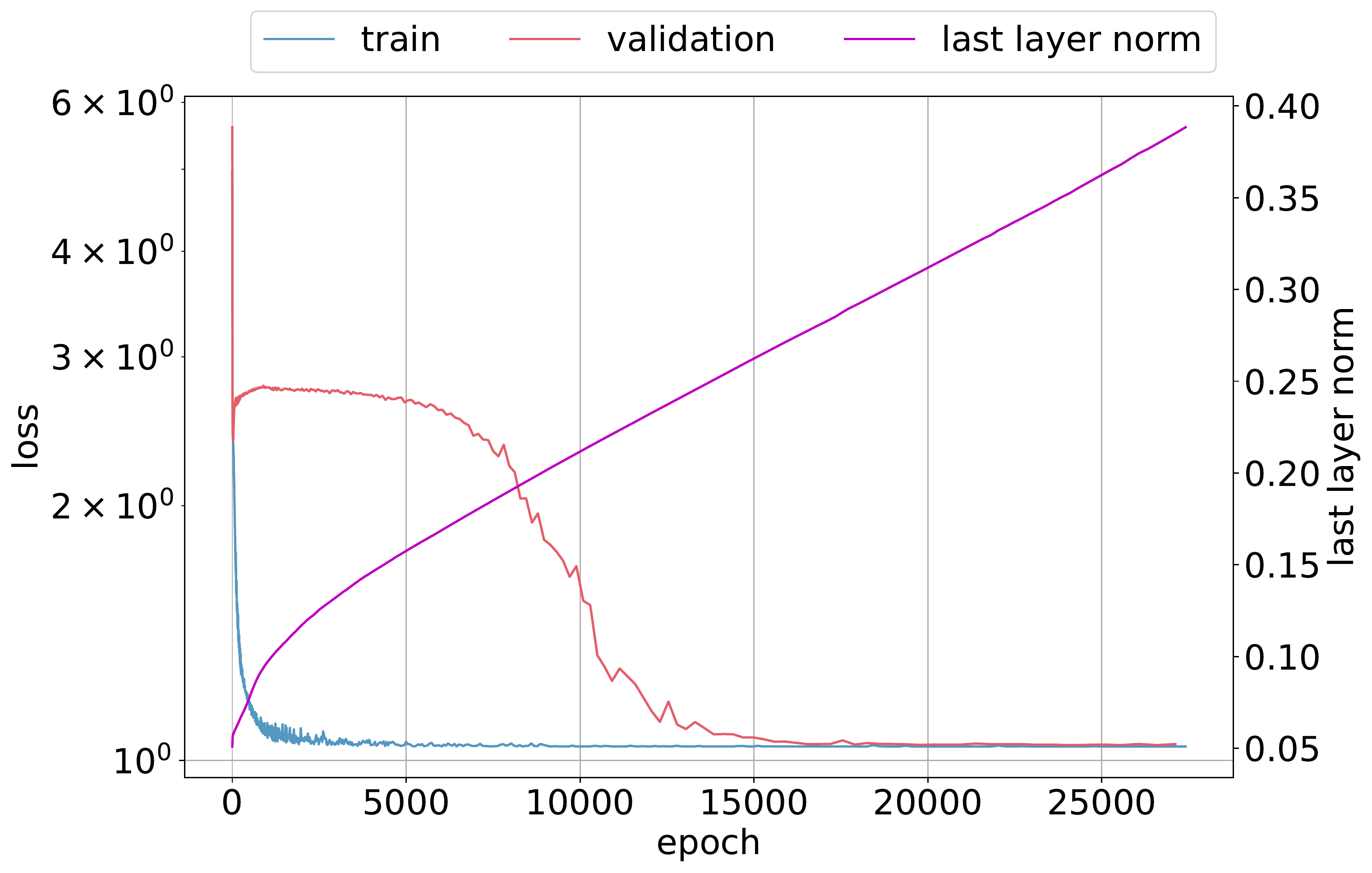} & 
      \includegraphics[width=0.33\linewidth]{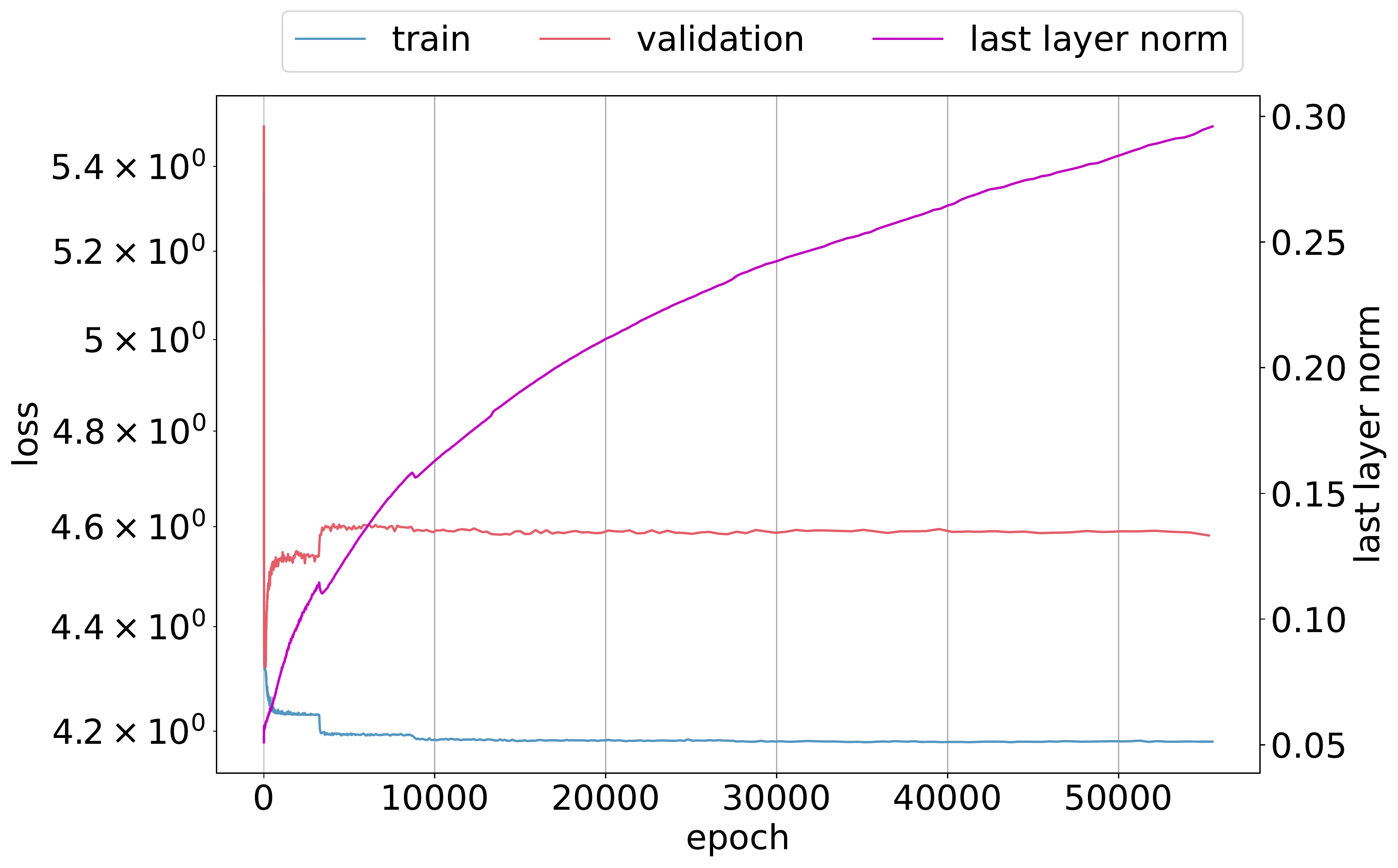}  \\
      (a)  & (b) & (c) \\
      & train  and validation loss vs epochs \\
      & \\
      \includegraphics[width=0.33\linewidth]{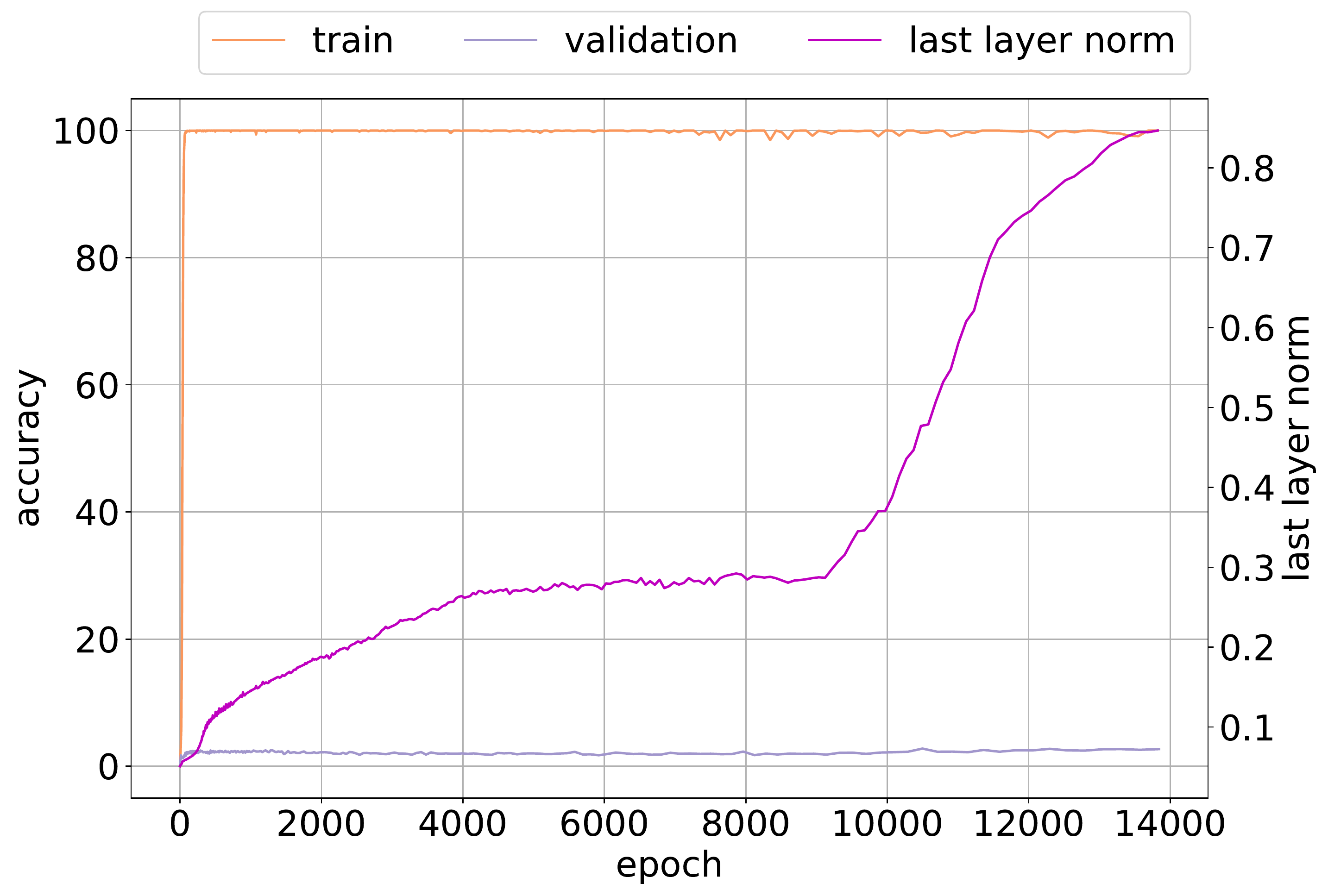} &
      \includegraphics[width=0.33\linewidth]{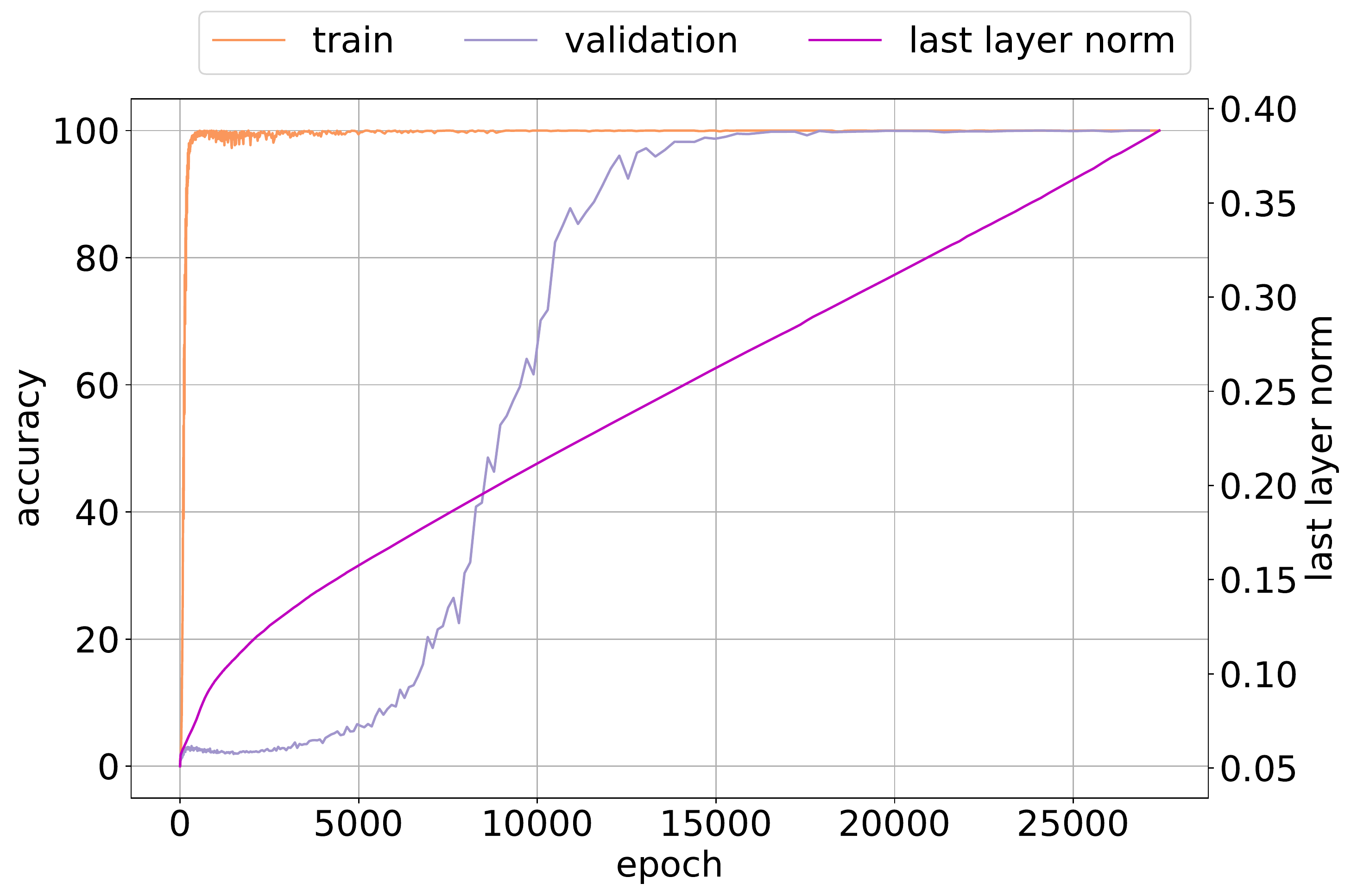} & 
      \includegraphics[width=0.33\linewidth]{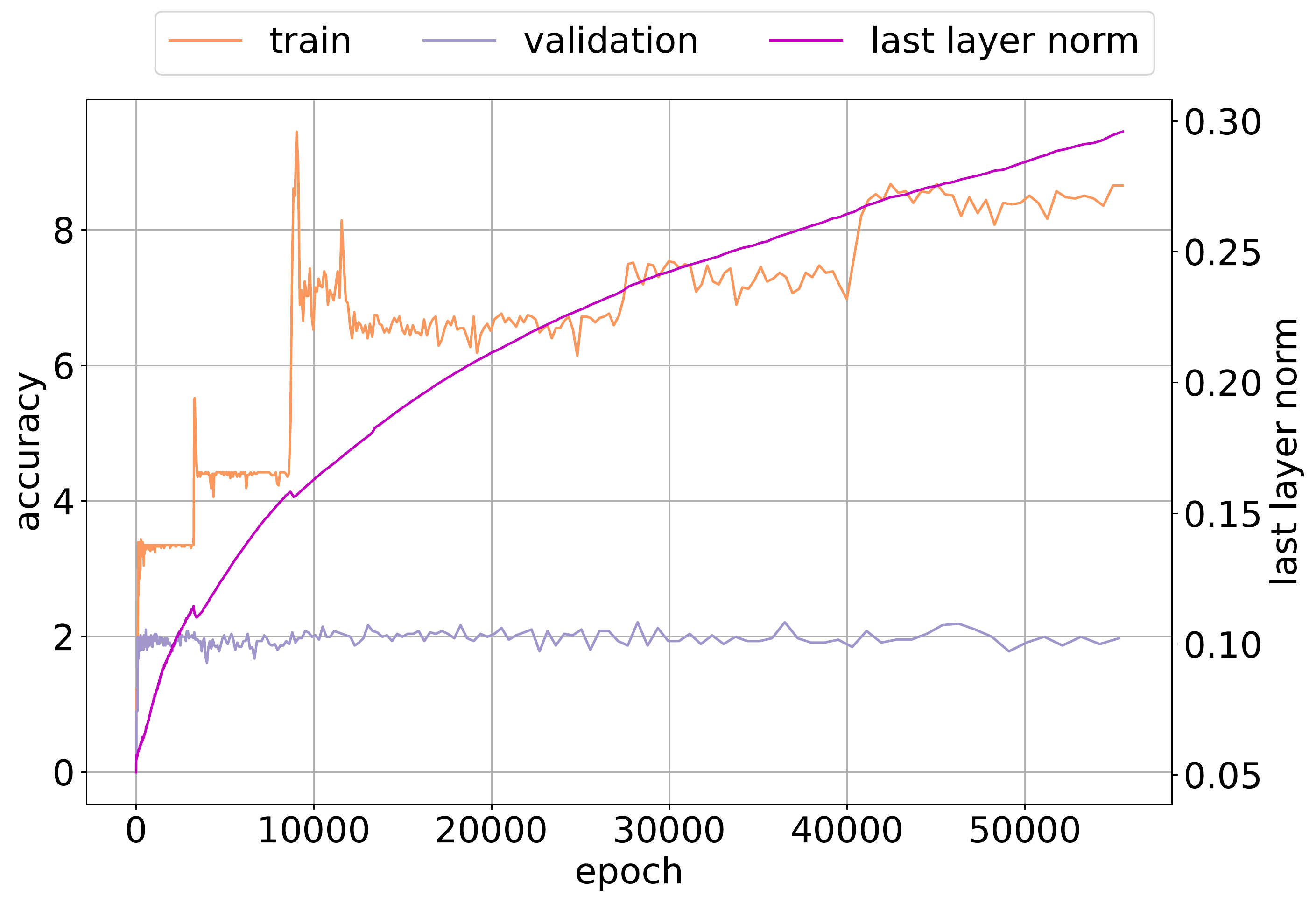}  \\
      (d)  & (e) & (f) \\
      & train  and validation accuracy vs epochs \\
      & \\
  \end{tabular}
 \caption{Division dataset: Features and parameters normalization. Observe that a smaller temperature allows the model to fit the data better but experiences training instability. Temperature = 0.25 allows the model to fit and achieve high validation accuracy without suffering training instability.} 
 \label{fig:grok_norm_both_div}
\end{figure*}

\clearpage

\end{appendices}
\end{document}